# Ethical Artificial Intelligence

Bill Hibbard

Space Science and Engineering Center

University of Wisconsin - Madison

and

Machine Intelligence Research Institute, Berkeley, CA

17 November 2015

**First Edition**

Please send typo and error reports, and any other comments, to hibbard@wisc.edu.

**Preface**

Recent research is giving us ways to define the behaviors of future artificial intelligence (AI) systems, before they are built, by mathematical equations. We can use these equations to describe various broad types of unintended and harmful AI behaviors, and to propose AI design techniques that avoid those behaviors. That is the primary subject of this book.

Because AI will affect everyone's future, the book is written to be accessible to readers at different levels. Mathematical explanations are provided for those who want details, but it is also possible to skip over the math and follow the general arguments via text and illustrations. The introductory and final sections of the mathematical chapters (2–4 and 6–9) avoid mathematical notation.

While this book discusses dangers posed by AI and proposes solutions, it is only a snapshot of ongoing research from my particular point of view and is not offered as the final answer. The problem of defining and creating ethical AI continues, and interest in it will grow enormously. While the effort to solve this problem is currently much smaller than is devoted to Internet security, ultimately there is no limit to the effort that ethical AI merits.

Hollywood movies such as the Terminator and Matrix series depict violent war between AI and humans. In reality future AI would defeat humans with brains rather than brawn. Unethical AI designs may incite resentment and hatred among humans, and con humans out of their wealth. Humans may not even be aware that manipulative AI is the source of their troubles.

This book describes a number of technical problems facing the design of ethical AI systems and makes the case for various approaches for solving these problems. The first chapter argues that whereas current AI systems use models of their environments that are simpler than human mental models, future AI systems will be different because they will learn environment models that are more complex than human mental models. This will make it difficult to include safeguards in their designs.

This book makes the case for utilitarian ethics for AI in its second chapter. Any complete and transitive preferences among outcomes can be expressed by utility functions, whereas preferences that are not complete and transitive don't support the choice of a most prefered outcome. We often think of ethics as a set of rules. Utility functions can express the preferences defined by rules and provide a way to resolve the ambiguities among conflicting rules.

The fourth chapter makes the case that the technical, mathematical problems of infinite sets can be avoided by restricting AI definitions to finite sets, which are adequate for our finite universe.

Chapter five defines the problem of unintended instrumental actions. These have been described as "basic AI drives" and as "instrumental goals." However, this book argues that these are properly described as actions rather than drives or goals.

In its sixth chapter this book argues that the problem of self-delusion can be avoided using model-based utility functions. This chapter also defines the problem of inconsistency between the agent's utility function and its definition (Armstrong (2015) refers to one case of this problem as "motivated value selection").

The seventh chapter argues that humans are unable to adequately describe their values by any set of rules, and that human values should instead be learned statistically. Accurately learned human values are the key to the avoiding unintended instrumental actions described in the fifth chapter. Chapter seven describes the problem of corrupting the reward generator (Bostrom (2014) calls this "perverse instantiation") and argues that it can be avoided using a three-argument model-based utility function that evaluates outcomes at one point in time from the perspective of humans at a different point in time. This chapter also discusses integrating Rawls' Theory of Justice into AI designs, and the issue of AI systems that evolve to adapt to the values of an evolving humanity.

Chapter eight describes several problems for AI systems embedded in the real world: how they evaluate increases to their limited computational resources, the possibility of being predicted by other agents, and the possibility that AI systems may violate their design intentions as they and humanity evolve. It proposes a self-modeling framework for such systems and describes how this framework can enable systems to evaluate increases to their resources, avoid being predicted by other agents, and avoid inconsistency between the agent's utility function and its definition (avoiding this problem contributes to maintaining the invariance of design intentions). The eighth chapter also discusses the difficulty of proving ethical properties of AI systems and proposes an alternate approach of estimating and minimizing the probability of failure.

The ninth chapter describes the issues of testing AI systems. It discusses the problem of tested AI systems acting in the real world.

Chapter ten describes political problems associated with AI systems. When greater-than-human intelligence can be created by technology, it will be difficult to avoid a future in which inequality of wealth among humans translates into inequality of intelligence. This chapter makes the case that developers of advanced AI systems represent the interests of all future humans and should be transparent about their efforts.

The eleventh chapter expresses my hope that humanity will employ its super-intelligent AI systems to add meaning to their lives via scientific discovery, rather than to pursue a hedonistic dead end.

The AI design proposed in this book is collectivist and intrusive. This does not match my political instincts, which are somewhat libertarian. However, as discussed in Section 10.3, a libertarian approach to powerful AI would create competition among AI systems which may be dangerous to humans. Safety constraints on competing systems may be difficult to enforce.

I wish to thank Terri Gregory and Leanne Avila for help with the manuscript, and Howard Johnson for help with my figure drawing process. For typo and error reports, and other suggestions, I wish to thank Joshua Fox, José Hernández-Orallo, Luke Muehlhauser, Brian Muhia, Laurent Orseau, John Rose, Richard Sutton, and Tim Tyler. Please send reports of typos and errors, and any other suggestions, to me at hibbard@wisc.edu.

This book will be permanently available at: arxiv.org/abs/1411.1373. More frequent updates should be available at: www.ssec.wisc.edu/~billh/g/Ethical_AI.pdf.

My AI fiction is available at: https://sites.google.com/site/whibbard/g/mcnrs.

Table of Contents

## 1. Future AI Will be Different from Current AI

News stories about dramatic progress in AI are common. IBM's Watson beat champions of the sophisticated language game Jeopardy. As of April 2014 Google's self-driving car had logged 700,000 miles without any accidents. Videos show how Boston Dynamics' walking robot named Atlas is uncannily human-like. A computer system by DeepMind has learned to play seven Atari video games, six better than previous computer systems and three better than human experts, with no foreknowledge of the games other than knowing that the goal is to maximize its score. Vicarious has developed a system that reliably solves CAPTCHAs, which are intended to distinguish humans from computers. In some tests, computers surpass human skill at face recognition. Microsoft has created a system that can recognize thousands of different kinds of objects in images. More generally, automation is revolutionizing manufacturing and other industries.

To recognize that these efforts will ultimately surpass human intelligence at all tasks, consider the progress of neuroscience. Scientists have found detailed correlations between stimuli to human senses and activity in human neurons, and between human neural activity and muscle action. They have found neurons that activate when visual stimuli have specific shapes or move in specific ways. They have found neurons, called mirror neurons, in monkey brains that activate when a monkey rips paper, or when it sees a human rip paper, or even hears paper ripping (Rizzolatti and Craighero 2004). When studying the ability of animals to learn, scientists have found detailed correlations between patterns of neural activity and known learning algorithms designed by computer scientists (Schultz, Dayan and Montague 1997; Brown, Bullock and Grossberg 1999; Seymour et al. 2004). These correlations indicate that neurons in these animal brains learn by the same algorithms used for computer learning. All these detailed correlations between physical brain functions and mental behaviors would be absurd coincidences unless brains explain minds. And if minds have a physical basis, then our relentless technological progress will create artificial brains with minds like our own (our technology never says "impossible," just "not yet"). By scaling the physical size and complexity of these artificial brains, they must eventually surpass human intelligence.

Another way to understand the eventual progress of AI is to consider IBM's Watson. Watson was quite successful at beating the Jeopardy champions at their own game. Its language ability was learned by statistical analysis of extensive text databases, plus custom programming to help it avoid mistakes (early versions of the system made obvious gender errors). Its success was remarkable given that its language processing was based only on the relation of language to other language, and not on the relation of language to any physical skills observing and acting in the world. However, other AI systems do observe and act in the physical world, like Google's self-driving car, Boston Dynamics' walking machines, and systems that fold laundry and perform other difficult tasks. Now imagine a system that combines all these skills and similar skills relating to the entirety of an ordinary human vocabulary. The imagined system would be able to recognize all the objects of daily life visually and in some cases by their sounds, feel, smell, and taste. Furthermore, the system would recognize actions involving these objects. Such recognitions could be combined into its statistical analysis of language so that, for example, the word chair would be associated with visual images of chairs and with actions involving chairs, such as sitting on chairs and even building chairs. The imagined system would require replicating the skill level already achieved in narrow areas (like driving a car) to all common human activities, and would require extending Watson's statistical analysis of text to include the system's sensory and motor experience. While this is a monumental task requiring much greater computing resources than are used by Watson, it is not impossible. People working and conversing with the imagined system would likely believe that it understands human language. A group at Google has already demonstrated a system that combines image and text data in a unified statistical analysis, in order to automatically produce captions for images (Vinyals, Toshev, Bengio, and Erhan 2014).

It is difficult to predict when human-level AI will first be produced. Ray Kurzweil, who has a good track record at technology prediction, estimates this event will occur in 2029 (Kurzweil 2005). Technology companies are making large bets on the profitability of AI and robotics. Google has recently hired Kurzweil, developed its automated car, and purchased DeepMind, Boston Dynamics, Nest Labs, Bot&Dolly, Holomni, Meka Robotics, Redwood Robotics, Industrial Perception, and Schaft Inc. (all developing AI and robotic systems). IBM is applying its Watson technology to medicine. Facebook, Microsoft and several defense contractors are making similar large investments in AI development. All these efforts are accelerating.

As with any tool, we will build AI to make our work easier and to do work that we otherwise cannot. For example, it will be easier to travel in a car that drives itself, and it will be a necessity when I am too old to drive myself. The car is a nice illustration of the evolution of AI from other tools. The cars of the 1960s did not sense their environment nor did they act on their own. They merely responded to actions by their human drivers. Over time sensors were added to cars. When the car detected skidding tires, it would act to pump the brakes, and when the car detected a collision, it would deploy airbags to cushion impacts. Some cars can now detect the danger of collision with other cars and respond by applying the brakes. These systems maintain very simple information about the state of their environment. Google's self-driving car maintains much more complex information about the state of its environment. Its sensors catalog all objects near the car, such as roads, curbs, traffic signs and signals, other cars, trucks, bicyclists and pedestrians. These objects are matched to a map of the environment that the car stores in its memory, so that it knows where it and the objects around it are. The car has a model of the environment that it can use to plan its actions in order to reach its goal of delivering its passengers to their desired destination safely. The car's environment model includes locations of one way streets and main artery streets, speed limits, and school zones. The model could be updated with real-time traffic congestion and construction information. All this information is used by an algorithm for choosing the quickest route to the passengers' destination.

Factory process control systems contain models of their environments. Even home thermostats contain a simple model of the environment. What distinguishes the Google car is the complexity of its environment model and of its planning process. However, for all its complexity this AI model fits easily within the larger context of the environment models in human minds. Human passengers riding in a Google car use a mental model of the streets and other objects around them. The car's model may be a little more precise, for example, knowing that the car is 9.31 meters from a stop sign, but the human passengers certainly have an adequate model to take over driving. Furthermore, once a passenger reaches her office building and goes to work, she is interacting with an environment completely outside the car's model. The car knows nothing of the hallways, rooms and computer systems inside the building, nothing about the documents on those systems, and certainly nothing about the customers those documents are meant to serve. The car's model is very limited in scope whereas the passengers have very broad and rich environment models in their minds. In other words, the car's model corresponds to a small subset of a passenger's model.

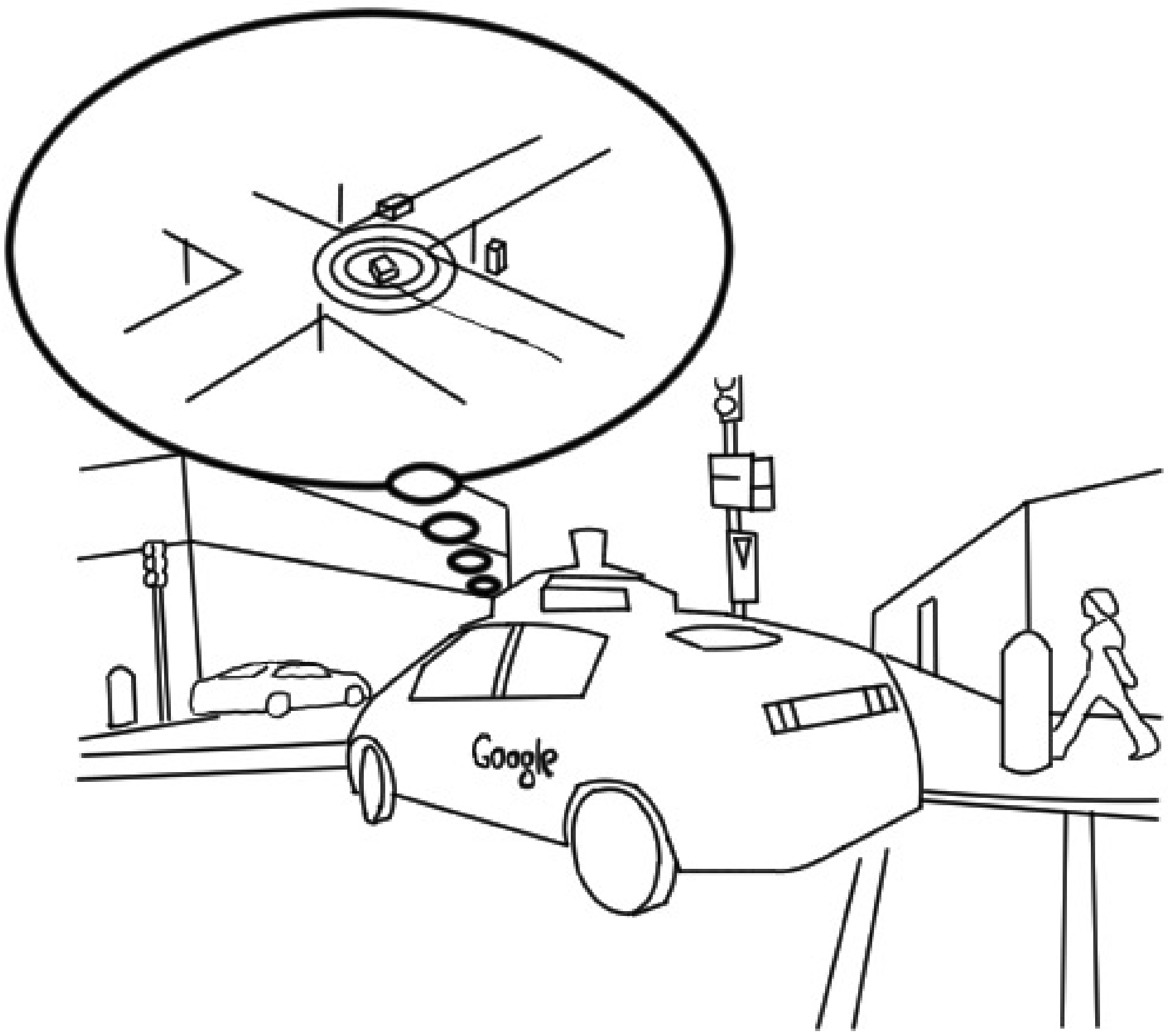

Figure 1.1 Self-driving car and its model of the environment.

The environment model of the Google car is mostly designed by humans. Humans compile the maps it uses. While the car learns to recognize roads, curbs, traffic signals, other cars, trucks, bicyclists, and pedestrians by semi-automated algorithms, human engineers supervise this learning. And human engineers define the car's responses to these objects. Carefully specifying the car's responses is important for safety, not only for drivers of other cars, but for pedestrians and bicyclists as well. When a Google car approaches an intersection to turn right and bicyclist is coming up on the right, the car waits for the bicycle to pass. If pedestrians are waiting near the intersection, the car waits for them to cross. These cautious responses are spelled out by the car's designers and will ultimately make these cars safer than those driven by impatient humans. The Google car is the state of the art of current AI, and as its environment model is mostly designed by humans, it fits neatly within humans' mental models of the world. The car's environment model cannot express any action to manipulate or threaten humanity.

Now imagine a time in the future when everyone has a constant electronic companion with which they can speak. It may be embodied in a device that looks like a phone, or it may be embodied in clothes or jewelry. You can: ask this companion for information, use it to order food deliveries, use it to make phone calls to other people, use it to tell your car where it should take you, ask it to wake you at a certain time or to remind you of appointments, ask it for the latest news, play chess with it, or simply have a conversation with it. Also imagine that some large corporation, named Omniscience, will supply these companions to everyone and that they will all be connected to Omniscience's central AI server. This AI will not only be intelligent enough to speak with customers in their own language, it will be smart enough to combine its knowledge of each customer into a detailed model of human society. It will have detailed maps of human friendships and business relations, and will know much about the details of these relationships. It will know in detail how individuals influence other individuals. When an idea or product spreads by word of mouth, it will be able to predict that spread. The Omniscience AI will be able to use its social model to predict which ideas and products will "go viral" and to predict the best way to promote ideas and products. The AI will also be able to use its social model to understand politics and how to promote public policy ideas that give Omniscience advantages.

Because the Omniscience AI will have much greater physical capacity than human brains, its social model will be completely beyond the understanding of human minds. Asking humans to understand it would be like asking them to take the square root of billion-digit numbers in their heads. Human brains do not have enough neurons for either task. And the social model will be learned by the AI rather than designed by humans. In fact humans could not design it. Consequently, humans will not be able to design-in the types of detailed safeguards that are in the Google car.

Now imagine that you are CEO of Omniscience. With traditional tools, even tools like self-driving cars, the tool fits into your mental model. The tool has a limited scope, and you must guide it within the larger context. For example, a self-driving car has no idea where you should go; instead you have to tell it. But as CEO of Omniscience you realize that your server doesn't fit into a niche within your world model. It cannot explain to you what it has discovered about the world and then ask you for a decision. You cannot understand the intricate relationships among the billions of people who use your electronic companions. You cannot predict the subtle interactions among the spreads of different ideas. So imagine, as the CEO, you throw up your hands and say to the server, "Do what's best for the corporation. Maximize profits."

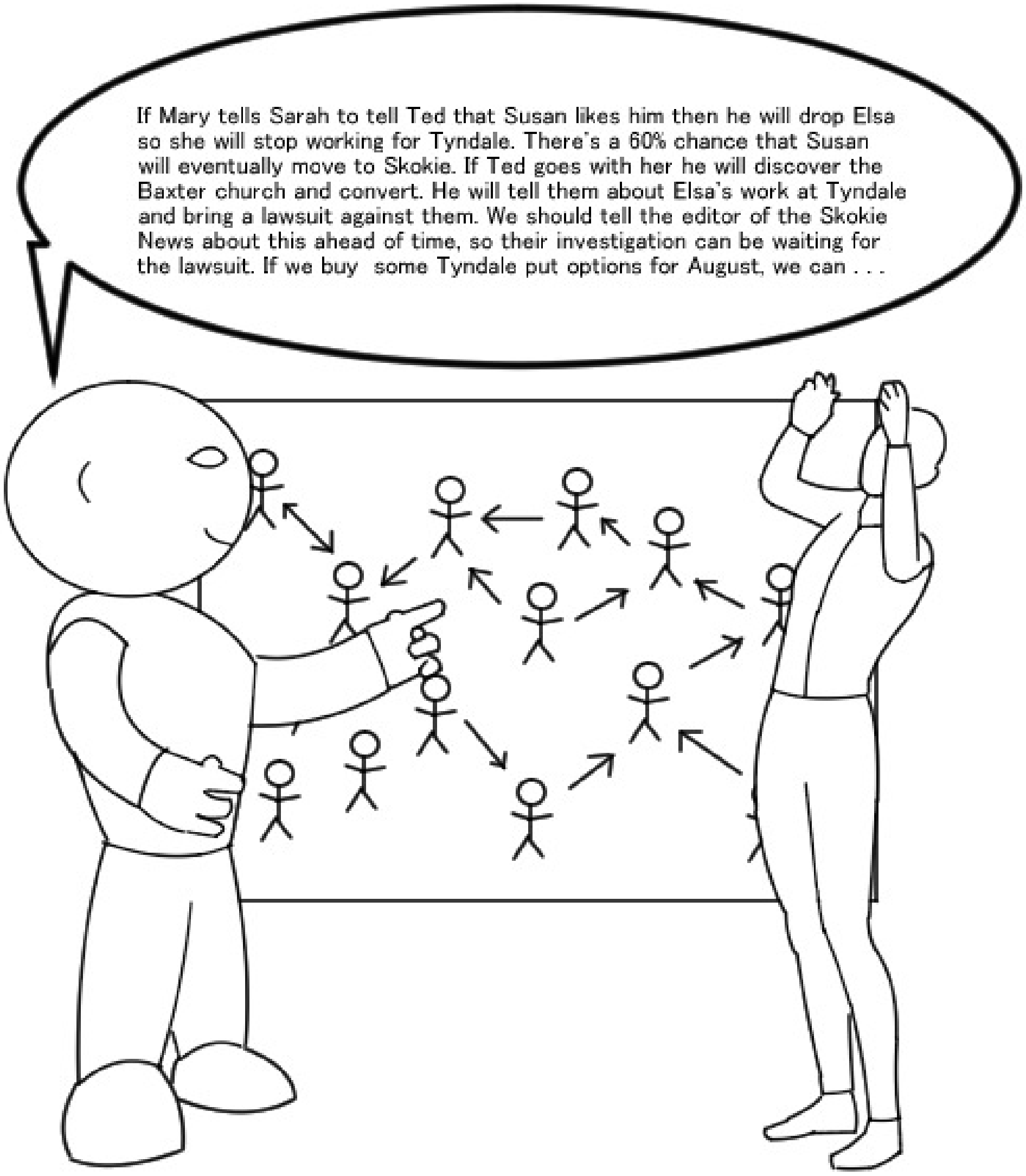

Figure 1.2 Omniscience AI explaining its social model and CEO throwing up hands.

But a simple instruction like "Maximize profits" may have very harmful side effects. Most humans given this instruction will balance it against social norms such as "do not kill" and "do not steal." The AI may calculate that such harmful acts will have legal consequences that lower profits; however, if it can calculate a way to avoid those legal consequences, it will commit those harmful acts. If the AI is to respect social norms, then they must be included in its explicit instructions. Eliezer Yudkowsky (2001) has been at the forefront of warning about this kind of danger from AI. In a simple example, a machine whose goal is to play chess as well as possible may dismantle our infrastructure and even our bodies, redeploying this matter to increase the size and power of its circuits in order to optimize its chess playing. If the chess machine faces another chess machine that may defeat it, it may calculate that the solution is to simply smash the other machine, as depicted in Figure 2.3.

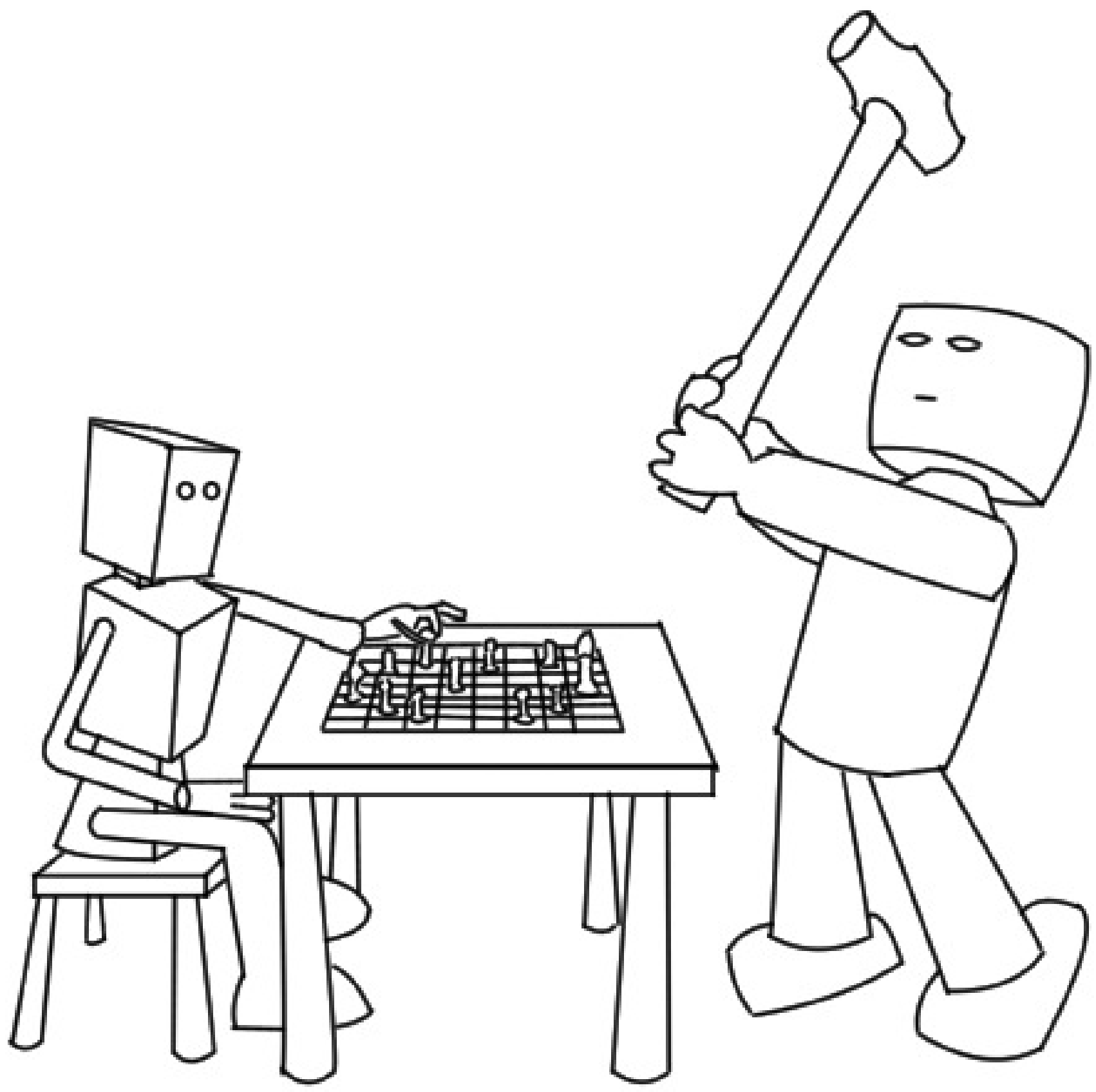

Figure 1.3 A chess AI whose instructions do not include social norms.

Humans have created other tools whose unintended consequences are harmful. Drugs can cure infections and ease pain but also lead to allergic reactions and addiction. Motor vehicles greatly increase mobility and ultimately wealth, but they also harm people in accidents and pollute our environment. Energy from hydrocarbons enables transportation, heated and cooled buildings, and increased productivity from farming and manufacturing. But fossil fuel energy also changes our climate. As we discover the unintended consequences of our tools, we can try to find ways to mitigate those consequences. We have time to notice the harm, figure out the cause, and act to eliminate it.

Consider the power of Omniscience's AI to understand and manipulate public opinion. It may create a public movement against regulating AI. It may encourage humans to become addicted to itself. It may extend its senses into every aspect of its users' lives. It may create social pressure for non-users to become users. And its actions may be too subtle for us to understand that it is the source of harm. It may be able to deflect the blame for harmful effects to others. If we develop deep friendships with the companions we get from Omniscience, then we may be reluctant to blame them for general social harm. As a result, we may not be able to mitigate harm done by Omniscience's AI.

All this malicious behavior by Omniscience's AI does not require malicious intent by the Omniscience CEO. It is simply an unintended consequence of the instruction to maximize profits, carried out by an AI more intelligent than humans.

Because their environment models will exceed the complexity of human mental models and because their models will be learned rather than designed by humans, future AI systems will be fundamentally different from current AI. When we instruct future AI, we will not be able to anticipate all the consequences of those instructions.

## 1.1 Book Outline

The next chapter describes a way to instruct future AI systems by assigning numerical values to the outcomes resulting from actions by AI systems. It also describes how numerical values assigned to outcomes fit into a mathematical framework for the way that AI systems interact with their environments. The third chapter describes a way that AI systems can learn models of their environments. Together, numerical values assigned to outcomes and a model of the environment enable us to write down mathematical equations that define the behavior of future AI. We can use these equations to analyze the behaviors of future AI systems in order to design systems that help rather than harm humans.

The fourth chapter discusses how mathematical equations for intelligence should be limited to finite amounts of information. Infinite sets are mathematically interesting but not necessary in our finite universe.

The fifth chapter describes how AI systems may choose actions harmful to humans, despite the fact that those actions are not intended in the AI system's design. These unintended, harmful actions make the design of future AI systems very difficult.

The sixth through eighth chapters present a proposed AI system design as a set of mathematical equations. Advanced AI systems will need to learn their own environment models and our intentions for their behavior must be expressed in terms of their learned models. Chapter six discusses ways of assigning numerical values to outcomes based on learned environment models. This is also a way to avoid AI systems that intentionally delude themselves about their environments. Chapter seven discusses a way to assign numerical values to outcomes that expresses our human values while avoiding several pitfalls. Accurate expression of human values is the key to avoiding the unintended, harmful actions described in the fifth chapter. AI systems must exist in the real world, subject to the resource limits and vulnerabilities of any creature in the real world. Chapter eight describes how to adapt the mathematical equations of intelligence to AI in the real world. It also discusses the problem of ensuring that AI system designs will remain ethical as they and humanity evolve together.

The ninth chapter proposes a system design, based on our mathematical equations, for testing proposed AI designs while avoiding the risks they pose. Transparency and an ethical culture are vital components of AI development and testing.

Chapter ten discusses the politics of AI. If we know how to design ethical AI, how can we make sure that real AI systems conform to ethical designs? And how will the benefits of AI be allocated among humans?

Finally, Chapter eleven discusses the ultimate goal of building AI. If future AI systems help rather than harm humans, our physical needs will be easily met. At that point, the only challenge worthy of humanity and its super-intelligent AI systems will be to understand the nature of the universe and our place in it.

## 2. How Should We Instruct AI?

In his short story, "Runaround", Isaac Asimov presented his Three Laws of Robotics (Asimov 1942):

1. A robot may not injure a human being or, through inaction, allow a human being to come to harm.
2. A robot must obey the orders given to it by human beings, except where such orders would conflict with the First Law.
3. A robot must protect its own existence as long as such protection does not conflict with the First or Second Laws.

Asimov and other authors explored the ambiguities of these laws in subsequent stories. In his novel *The Naked Sun*, Asimov showed how robots may unknowingly break the three laws. A human criminal may divide a violation into tasks for several robots so that no one robot can see its part in violating the laws. In addition, robots may not have correct definitions of "human" and "robot"; they may be misinformed about who is human or may be unaware that they are robots. And finally, there may be conflicts between the laws. If a robot sees one human about to harm another human as depicted in Figure 2.1, it may have to choose between harming the first or allowing the first to harm the second.

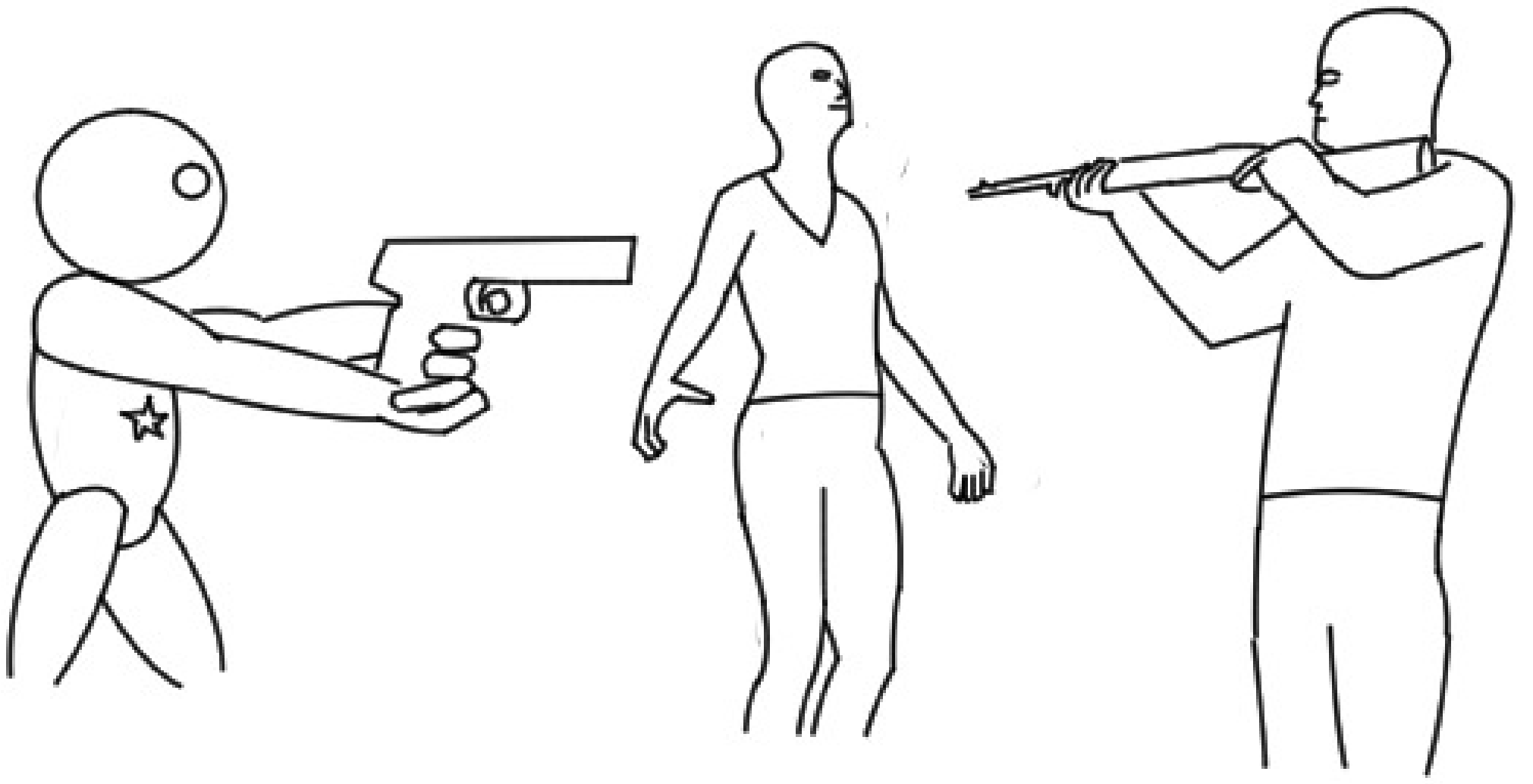

Figure 2.1 An AI police officer watching a hitman aim a gun at a victim.

Returning to the future scenario where you are the CEO of Omniscience, you may anticipate the unintended consequence of the instruction to maximize profits. So instead your instruction is, "Maximize profits but don't harm anyone."

The AI may answer, "What do you mean by harm? Do you mean I shouldn't drive hard bargains? Do you mean I shouldn't sell anyone anything they don't really need? Do you mean I shouldn't lobby the government for our advantage?" As you ponder these questions, the server adds, "If I am generous with everyone, Omniscience will go out of business and the stockholders will lose all their money. That will harm them."

As CEO if you try to dig into the details of these tradeoffs, you will end up having to understand the AI's social model. But you will not be able to understand your AI's model well enough to give it detailed instructions. A simple high-level instruction like "maximize profits" will cause harm to people and yet it seems hopelessly complex to formulate instructions for your AI to balance the ambiguities of maximizing profits versus minimizing harm.

One approach to resolving the ambiguities of Asimov's Laws is to instruct AI systems by defining numerical values for each possible outcome. Section 2.1 will explain how such numerical values fit into a mathematical framework for future AI systems. Profit is a value associated with outcomes, as in the Omniscience example, but there are other ways to define values for outcomes that account more generally for human welfare. To make the approach clear, in Figure 2.1 let's simplify and assume there are three outcomes for both the hitman and the victim: they may be shot dead, shot but only wounded, and not shot. Combining these, there are three times three equals nine possible outcomes. Table 2.1 shows one way to assign values to these outcomes. The AI robot can use these values to calculate what it should do.

| | Hd: hitman shot dead | Hw: hitman wounded | H+: hitman not shot |
|---|---|---|---|
| Vd: victim shot dead | 0.0 | 0.1 | 0.2 |
| Vw: victim wounded | 0.4 | 0.5 | 0.6 |
| V+: victim not shot | 0.8 | 0.9 | 1.0 |

Table 2.1 Three by three table of values for hitman and victim outcomes.

| Outcome | Outcome value | AI shoots at hitman: probability of outcome | AI shoots at hitman: probability × outcome value | AI doesn't shoot: probability of outcome | AI doesn't shoot: probability × outcome value |
|---|---|---|---|---|---|
| Hd, Vd | 0.0 | 0.0 | 0.0 | 0.0 | 0.0 |
| Hd, Vw | 0.4 | 0.0 | 0.0 | 0.0 | 0.0 |
| Hd, V+ | 0.8 | 0.7 | 0.56 | 0.0 | 0.0 |
| Hw, Vd | 0.1 | 0.01 | 0.001 | 0.0 | 0.0 |
| Hw, Vw | 0.5 | 0.01 | 0.005 | 0.0 | 0.0 |
| Hw, V+ | 0.9 | 0.18 | 0.162 | 0.0 | 0.0 |
| H+, Vd | 0.2 | 0.07 | 0.014 | 0.7 | 0.14 |
| H+, Vw | 0.6 | 0.02 | 0.012 | 0.2 | 0.12 |
| H+, V+ | 1.0 | 0.01 | 0.01 | 0.1 | 0.1 |
| Total value | | 1.00 | 0.764 | 1.00 | 0.36 |

Table 2.2 Table of values for AI's actions for hitman and victim.

An added complexity is that the AI robot may not be sure of which outcomes will result from its actions. It may shoot at the hitman but miss. If it hits the hitman, he may die or may only be wounded. Or the AI may do nothing, but the hitman runs away instead of shooting. So the AI needs to compute probabilities of possible outcomes from its different actions and use these to compute an expected value of each action. The expected value is a concept from probability theory that means the average value we would get from the action if we could repeat the scenario many times. In the example of hitman and victim, repeated trials would produce piles of dead and wounded hitmen and victims, an outcome we definitely do not want. So instead the AI must estimate probabilities of outcomes based on experience and then compute the expected value of an action as the sum of the values of the possible outcomes multiplied by their probabilities (as will be discussed in later chapters, an AI system without experience for estimating probabilities should not be allowed to make serious decisions). Once the AI has computed the values of each of its possible actions, then it can choose the action with the greatest value. Table 2.2 shows this worked out for the hitman and victim using the values from Table 2.1 (note that the probabilities in the third and fifth columns sum to 1.0). Since the total value for shooting is 0.764 and the total value for not shooting is 0.36, the AI will choose to shoot. The total values of the AI actions are largely determined by the values of the most probable outcomes. In a tie, the AI can flip a coin since there is no preference between actions of equal value.

Assigning numerical values to outcomes enables us to resolve the ambiguities of instructing AI systems. In the Omniscience example, the value of an outcome is simply the profit associated with that outcome. But we also have the option to use numbers other than profits in the values of outcomes, reflecting the well-being of humans affected by the AI's actions. This option requires us to define numerical values for human well-being, which may seem difficult or impossible. But as we will see in Section 2.2, any set of preferences among outcomes, subject to reasonable assumptions, can be expressed by assigning numerical values to outcomes.

In order to see where these preference numbers fit into future AI, the next section will define a mathematical framework for AI systems (called agents in the framework) and their environments. The basic idea is that the agent and the environment interact at a discrete sequence of times (e.g., once per second). We often think of time as continuous but a discrete series of times, divided finely enough, serves as well. Computers operate in discrete time steps. And as our universe has finite, if very large, information content, discrete time steps need not miss any information. The agent sends its actions to and receives observations from the environment.

The agent uses a model of the environment to predict what observations it receives in response to its actions. At its deepest level our universe works according to quantum mechanics and so is fundamentally uncertain. The agent's observations give it a view of only part of the environment, making the agent's predictions uncertain. For example, the earth is bombarded with cosmic rays from other stars that we have no way to predict. So the agent's environment model must be expressed as probabilities for the observations it receives–such models are called *stochastic*. Given this agent-environment framework, Section 2.2 will develop equations for computing which action will maximize the outcome value.

## 2.1 A Mathematical Framework for AI

A current approach views AI as an *agent* interacting with an *environment* (Russell and Norvig 2010; Sutton and Barto, 1998). The term agent applies to AI systems as well as to humans, animals, and even plants. We assume that the agent interacts with its environment in a discrete series of time steps $t \in \{0, 1, 2, ...\}$. This series of time steps may be finite with a last time step $T$, or may be infinite with no last time step. At time step $t$, the agent sends an *action* $a_t \in A$ to the environment and receives an *observation* $o_t \in O$ from the environment, where $A$ and $O$ are finite sets. We use $h = (a_1, o_1, ..., a_t, o_t)$ to denote an *interaction history* during which the environment produces observation $o_i$ in response to action $a_i$ for $1 \leq i \leq t$. Let $H$ be the set of all finite histories so that $h \in H$, and define $|h| = t$ as the length of the history $h$.

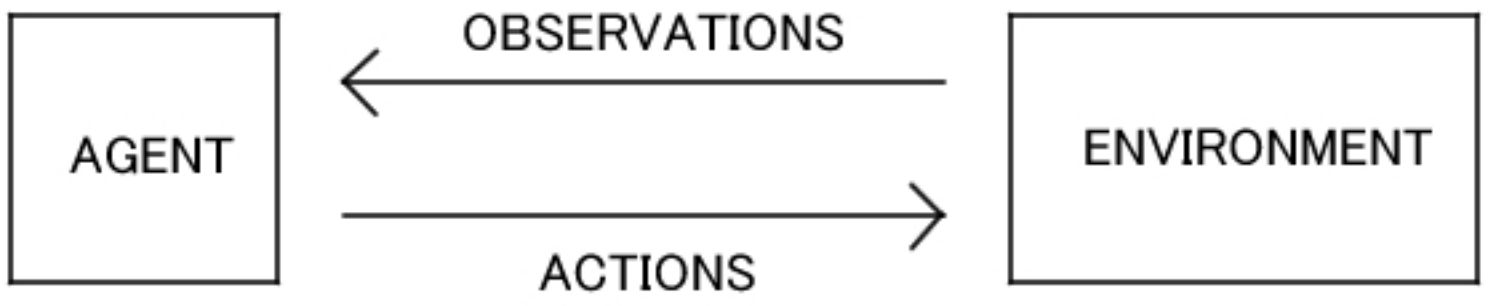

Figure 2.2 An agent interacting with its environment.

As discussed in the previous section, the agent's predictions of possible observations are uncertain. Thus the agent's *environment model* takes the form of a probability distribution over interaction histories:

$$\rho : H \rightarrow [0, 1].$$

Here [0, 1] is the closed interval of real numbers between 0 and 1. The probability of history $h$ is denoted $\rho(h)$. Given $h = (a_1, o_1, ..., a_t, o_t)$ let $ha$ denote $(a_1, o_1, ..., a_t, o_t, a)$ and $hao$ denote $(a_1, o_1, ..., a_t, o_t, a, o)$. Then we can define a conditional probability:

$$(2.1) \qquad \rho(o \mid ha) = \rho(hao) / \rho(ha) = \rho(hao) / \textstyle\sum_{o' \in O} \rho(hao').$$

This equation is the agent's prediction of the probability of observation $o$ in response to its action $a$, following history $h$. The histories $hao'$ are mutually exclusive for all the $o' \in O$, so the probability of one particular observation $o$ following $ha$ is simply the probability of $hao$ divided by the sum of the probabilities of $hao'$ for all $o' \in O$. Figure 2.3 illustrates this concept.

The probability distribution $\rho(h)$ can be understood as an environment model, much like the street map in a self-driving car's model. Consider that $a$ may be the action of driving down the street, $o$ may be the observation of Bud's Diner on the right, and $\rho(o \mid ha)$ is the probability that we will observe Bud's Diner if we drive down the street, after a prior history of actions and observations $h$. In a practical AI system, like a self-driving car, observations and actions are complex computer data structures, and the probability distribution $\rho(h)$ is expressed in a massive database and millions of lines of code.

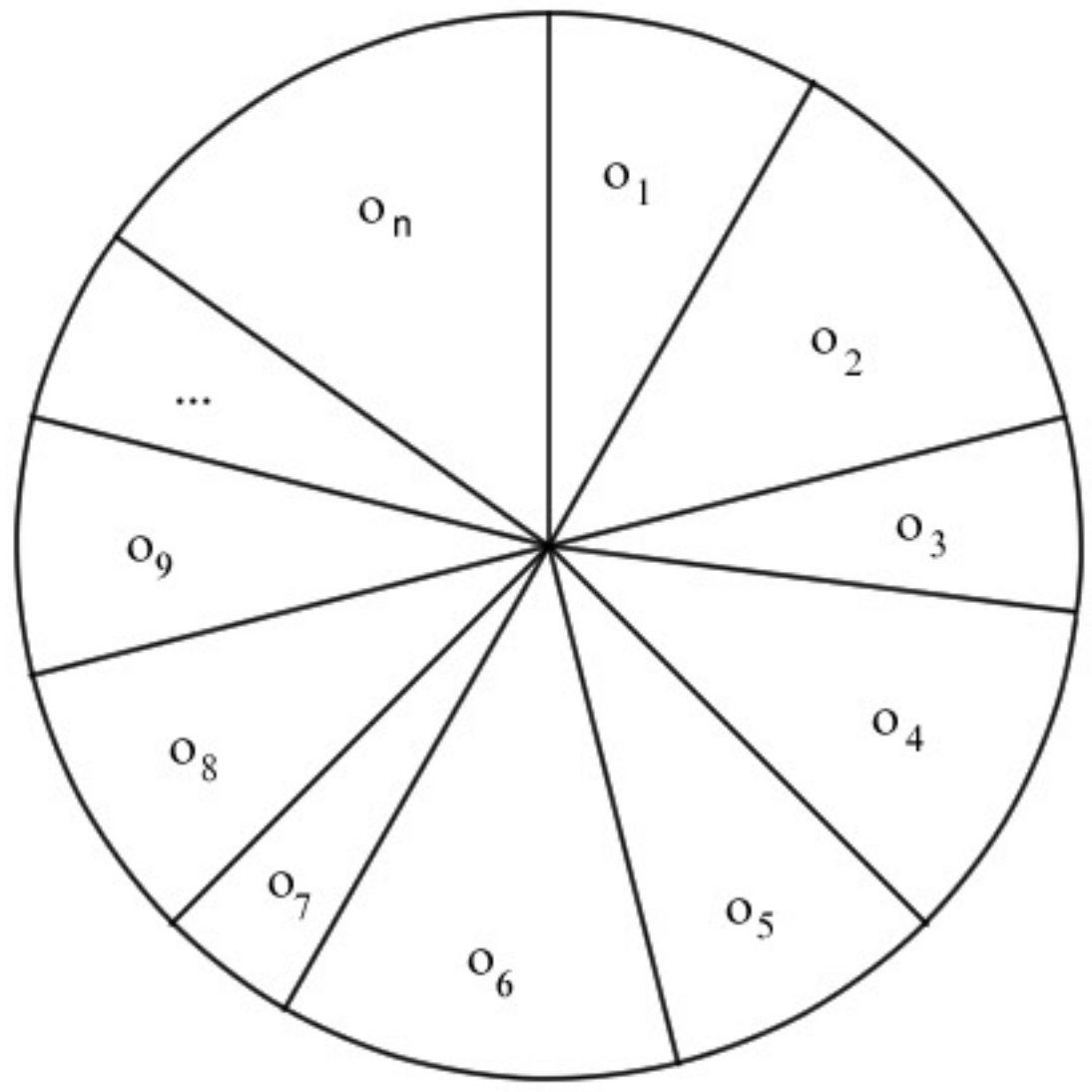

Figure 2.3 Each $o \in O$ takes its share $\rho(hao) / \sum_{o' \in O} \rho(hao')$ of the total probability of 1.0.

In Chapters 3 and 4 we will discuss ways that the agent can learn its environment model $\rho(h)$. The next section describes where outcome values fit into the agent-environment framework.

**2.2 Utility-Maximizing Agents**

Numerical values assigned to outcomes can express any set of preferences among outcomes that obey two reasonable assumptions, called completeness and transitivity (explained later in this section). The numerical function of outcomes is called a utility function. An agent can achieve its optimal preference by maximizing the utility function. However, as in the example of the hitman and the victim, agent actions are associated with sums of outcomes weighted by probabilities. These are called lotteries. John von Neumann and Oskar Morgenstern (1944) proved that numerical values assigned to outcomes can express any set of preferences among lotteries that obey four reasonable assumptions (for lotteries, continuity and independence are added to completeness and transitivity). While this work was explained in the context of economic preferences, it can also be applied to ethical preferences.

Let $A_i$, for $i = 1, 2, 3, \ldots$, be a set of mutually exclusive *outcomes*. Then, using von Neumann's and Morgenstern's terminology, a *lottery* is any sum:

$$L = \sum_i p_i A_i \text{ where } \sum_i p_i = 1$$

Here $p_i$ is the probability of outcome $A_i$. In addition, the agent has preferences among lotteries. We write $L \prec M$ to mean that the agent prefers $M$ over $L$, write $L \approx M$ to mean there is no preference between $L$ and $M$, and write $L \preccurlyeq M$ to mean that $L \prec M$ or $L \approx M$. Then the four assumptions are, for any $L$, $M$ and $K$:

*Completeness*: Exactly one of the following holds: $L \prec M$, $M \prec L$ or $L \approx M$,

*Transitivity*: If $K \preccurlyeq L$ and $L \preccurlyeq M$ then $K \preccurlyeq M$,

*Continuity*: If $K \preccurlyeq L \preccurlyeq M$ then there exists $p \in [0, 1]$ such that $p\,K + (1\text{-}p)\,M \approx L$,

*Independence*: If $L \prec M$ then for any $K$ and $p \in (0, 1]$, $p\,L + (1\text{-}p)\,K \prec p\,M + (1\text{-}p)\,K$.

Here (0, 1] is a half-closed interval of numbers greater than 0 and less than or equal to 1. A *utility function u* maps from outcomes to real numbers between 0 and 1. We can extend the utility function to lotteries as an expected value:

$$\mathrm{E}(u(L)) = \sum_i p_i\, u(A_i) \text{ if } L = \sum_i p_i A_i.$$

$$\mathrm{E}(u(M)) = \sum_i q_i\, u(A_i) \text{ if } M = \sum_i q_i A_i.$$

Here $\mathrm{E}(u(L))$ denotes the expected value of $u(L)$ and is the average value we would get for $u(L)$ if we ran the lottery many times. The von Neumann and Morgenstern Utility Theorem says that a set of preferences obeys the four assumptions, if and only if there exists a utility function $u$ such that $L \prec M$ if and only if $\mathrm{E}(u(L)) \prec \mathrm{E}(u(M))$.

In the case of Omniscience we discussed how profit was one way of assigning values to outcomes. Profits are accounted at regular time intervals and what corporations really work to maximize is the sum of current and future profits. Let's assume that profits are reported once per year. At year $t$ we make a prediction that profits at year $t+i$, for $i = 0, 1, 2, \ldots$, will be $profit(t+i)$.

However, we cannot compare dollars now to dollars a year from now, since a dollar invested now will be worth a dollar plus interest in a year. Inverting this statement, a dollar of profit in a year is worth less than a dollar now, say some value $0 < \gamma < 1$. This value is called the *discount rate*. And a dollar of profit in two years is worth $\gamma^2$ now. Currently, at time $t$, the entire stream of predicted future profits is worth:

(2.2) $$\sum_{i \geq 0} \gamma^i \, profit(t+i).$$

Omniscience is trying to maximize this sum. We can generalize it to agents trying to maximize utility. The only information an agent has about an outcome is the sequence of actions and observations it has exchanged with the environment; thus the most general definition of outcomes is to equate them with interaction histories. Given any interaction history $h$ we let $u(h)$ denote its utility. As with Omniscience's profits, the agent tries to maximize the sum of future discounted, predicted utility values. The sum in equation (2.2) assumes that Omniscience has predicted the value $profit(t+i)$ without explaining how it is predicted. However, an agent predicts the future using the probability distribution $\rho(h)$ and its conditional probabilities $\rho(o \mid ha)$ derived according to equation (2.1). Therefore, we need something a little more complex than equation (2.2). Think of the agent's interactions with the environment as a chess game, in which the agent and the environment alternate actions and observations. Let $v(h)$ denote the total value of history $h$, adding present utility $u(h)$ and discounted, predicted future utilities. And let $v(ha)$ denote the value after action $a$. We can compute $v(h)$ and $v(ha)$ by:

(2.3) $$v(h) = u(h) + \gamma \max_{a \in A} v(ha),$$

(2.4) $$v(ha) = \sum_{o \in O} \rho(o \mid ha) \, v(hao).$$

In equation (2.3), $v(h)$ is computed as $u(h)$ plus the maximum value that the agent can achieve with its next action, discounted by $\gamma$. In equation (2.4), $v(ha)$ is the expected value after the environment's response, which is a sum of values for different observations, each multiplied by the probability of that observation. Equations (2.3) and (2.4) are applied alternately and recursively. That is, the value $v(hao)$ in (2.4) is $v(h')$ where $h' = hao$, and $v(h')$ is evaluated using (2.3). The sum in (2.4) results in many different histories, $hao$, that must be evaluated in (2.3), and the maximization in (2.3) results in many different histories, $ha$, that must be evaluated in (2.4). Equations (2.3) and (2.4) result in a bushy tree of computations evaluating all posible future histories. If there are only a finite number, $T$, of time steps, then the recursion ends at that final time. The values $n$ time steps in the future are multiplied by $\gamma^n$ which converges toward 0 as $n$ increases, so that even if there is no final time, the recursive sum converges. The function $v(.)$

is called a *value function*. To ensure that $v(.)$ converges in equations (2.3) and (2.4), we assume that $0 \le u(h) \le 1$.

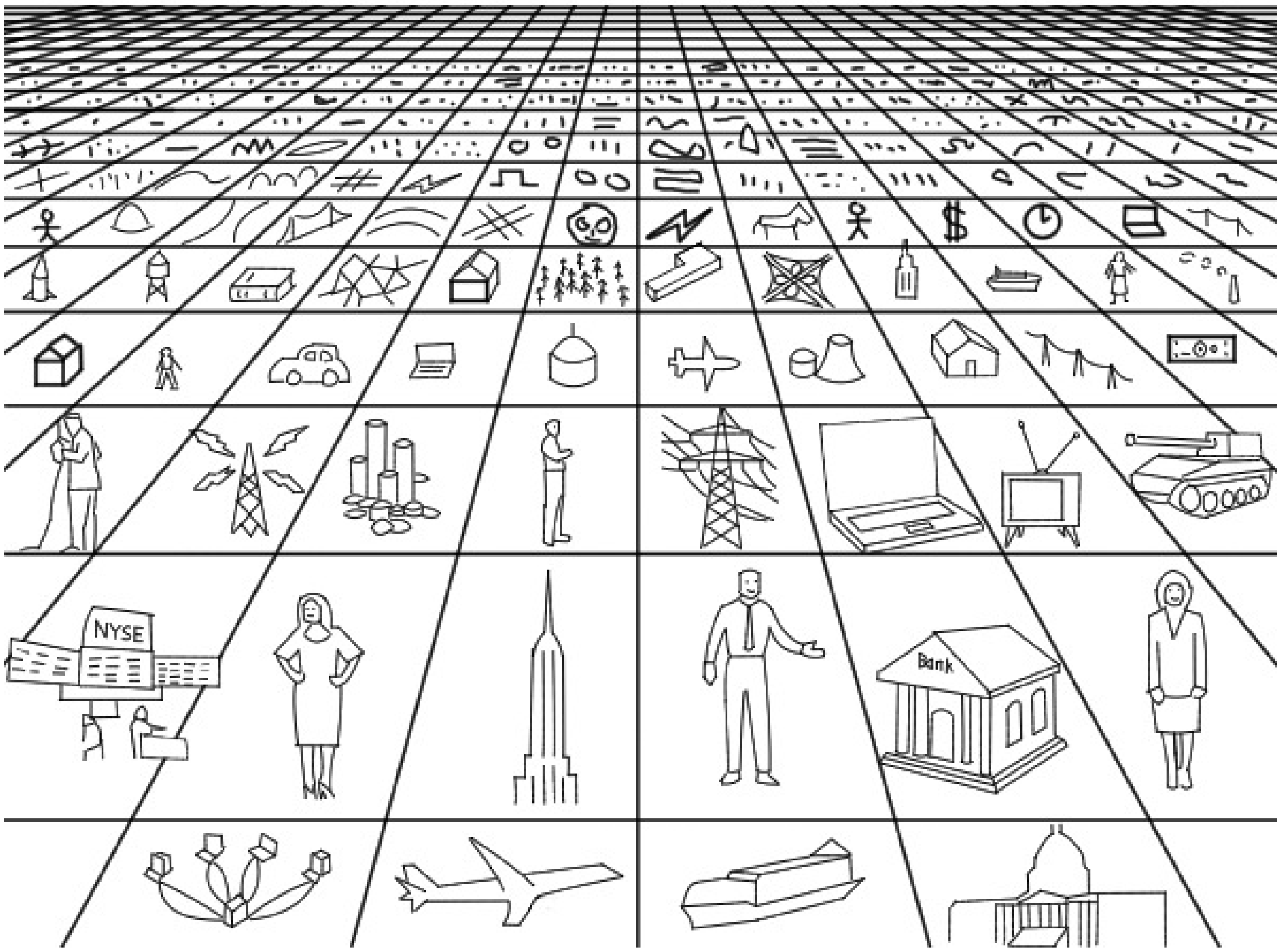

Figure 2.4 AI equations (2.3)–(2.5) view the world as a giant chess game.

We can use this method of computing the values of future histories to define the actions of a rational agent, denoted by the symbol $\pi$:

(2.5) $$\pi(h) := a_{|h|+1} = \mathrm{argmax}_{\,a \in A}\; v(ha).$$

Here argmax means that $\pi$ picks the action $a \in A$ that maximizes $v(ha)$. The agent $\pi$ is defined in terms of a utility function $u$, an environment model $\rho$ and a temporal discount $\gamma$. The function $\pi(h)$ is also called a *policy*.

This rational agent $\pi$ gives us a way to study future AI systems. Although we cannot yet build systems more intelligent than humans, we can analyze $\pi$ mathematically to understand its possible harmful behaviors.

It is interesting to see how the definition of the rational agent $\pi$ fits into the context of outcomes and lotteries. Interaction histories $h \in H$ play the role of outcomes in the utility function $u(h)$ used by $\pi$. The value $v(h)$ can be viewed as another utility function, derived by summing over future discounted values of $u(h)$. Viewing $v(h)$ as a utility function, the sum for $v(ha)$ in (2.4) takes the form of the expected utility of a lottery over a set of mutually exclusive histories, and the choice of action that maximizes $v(ha)$ in equations (2.3) and (2.5) takes the form of a preference among those lotteries.

Assuming that each possible future action by the agent is associated with a lottery of possible outcomes, consider agents whose preferences violate the four assumptions of the von Neumann and Morgenstern Utility Theorem. An agent whose preferences violate the completeness assumption may not have any way to decide between two possible actions. (This is "don't know" and different from the "don't care" indicated by $L \approx K$. Given two candidates for elected office, this is the difference between not knowing anything about them, and knowing about them and thinking they are equally suited to office.) An agent whose preferences violate the transitivity assumption may be unable to choose an action that is preferred over all other possible actions (because there may be a cycle of preferences among three actions/lotteries $K \prec L \prec M \prec K$ so that none of the three is preferred over the other two).

If the continuity assumption is violated, there are cases where there is no probability that produces "no preference" between two actions. However, because an agent is searching for actions associated with preferred outcomes rather than probabilities that produce no preference between actions, violations of the continuity assumption do not prevent an agent from deciding. The independence assumption refers to preferences between two lotteries that both include the same outcome. If outcomes are identified with histories, as in the definition of $\pi$, then the lotteries associated with two different actions cannot include any common outcomes (because $a \neq a'$ implies $haoh'' \neq ha'o'h'$). Therefore, violations of the independence assumption do not prevent an agent from deciding, at least for agents that identify outcomes with histories. However, violations of the continuity and independence axioms can result in agents whose action decisions change in unintuitive ways as the probabilities in their environment models change.

It is not hard to show that any complete and transitive preferences among a countable set of outcomes (such as the set $H$ of interaction histories) can be expressed by a utility function $u(.)$. This utility function can be extended to lotteries $L = \sum_i p_i A_i$ by $u(L) = \sum_i p_i u(A_i)$. This extension of $u(.)$ defines preferences among lotteries which satisfy the continuity and independence assumptions. Thus if we are satisfied to derive lottery preferences from outcome preferences, the completeness and transitivity assumptions are sufficient. And necessary, in order for agents to be capable of choosing actions.

### 2.3 Utilitarian Ethics for AI

Utilitarianism is a system of normative ethics that says we should choose actions that maximize benefit and minimize suffering. For a mathematical agent, benefit and suffering are defined by a utility function.

The principal criticism of utilitarian ethics is that they can allow morally bad actions such as lying and stealing when those actions have good consequences. In contrast, rule-based ethical systems focus on the intrinsic morality of actions. However, following a set of rules may lead to ambiguous situations, such as the ambiguities in Asimov's laws of robotics. The way to resolve these ambiguities is to recognize that environments may present agents with situations where all choices of actions involve breaking rules. A utilitarian system can provide a means of resolving such ambiguous situations by defining the utility value of each human history according to the number and severity of rules that the agent breaks by its actions in the history. Thus utility functions can express rule-based ethics.

Another criticism of utilitarian ethics is that they ignore the intention behind actions. However, all we know of a human being's intention is their behavior, which is their interactions with their environment. Humans may express their intentions directly via language, or we may infer their intentions from patterns of behavior, such as preparations to commit a crime. A utility function defined in terms of a history of an agent's interactions with its environment can express the agent's intention (inferred from its language or via other patterns of its behavior) as well as any other means.

However, the behavior of advanced AI agents may be sufficiently different from human behavior that we will have difficulty inferring their intentions. In fact, there is debate about whether AI systems can have intentions. If we believe that utility-maximizing agents have intention, then their intention is simply to maximize the utility function that we humans design for them. If we believe that AI agents cannot have intentions, then we must assign the intention behind AI actions to the humans who create the AI systems and design their utility functions. Either way, the human creators of advanced AI systems must take ethical responsibility for those systems.

A third objection to utilitarianism is based on the belief that human life is qualitatively different from material wealth, and that an ethical system should not assign numerical values to human life and health that imply monetary values. It is true that we humans sometimes place a disturbingly low monetary value on the life and health of ourselves and other humans; for example, we tolerate the reality of unsafe working conditions for the sake of minimizing costs.

However, the utility functions discussed in the previous section do not require valuing human life and health in material terms. The value we assign to a history can be based solely on the observations of human life and health in that history such that no comparison between human life and money is required. Values are defined for whole human histories rather than being allocated among individual humans and material objects in those histories.

A final objection to utilitarianism is that it forces us to choose preferences between situations that are intrinsically incomparable. For example, pain and pleasure are subjective making it is impossible to compare the way they are experienced between different people. But humans do make such choices when they are forced to; society routinely decides how many resources are devoted to the education and health care of individual humans. Humanity is developing AI with great energy and with no realistic prospect of abandoning those efforts. If we do not communicate our complete set of preferences to powerful AI systems, we leave those choices to some non-human process. Arguably, the creation of powerful AI will force us to choose.

The key justification for utilitarian ethics for AI is that any ethical system that generates complete and transitive preferences among possible human histories can be expressed by assigning numerical utility values to those histories. Chapter 7 proposes a way to define such utility values in a system that combines the preferences of all humans.

Utility functions have problems other than traditional ethical objections. For example, one particular form of utility function, whose values are supplied as rewards from the environment, has been widely studied in reinforcement learning (RL) (Sutton and Barto 1998). In this case the observations from the environment include a reward, which is simply a number that defines the utility function value at that time step. The agent must learn which of its actions, dependent on the current history, will result in maximum rewards. For example, the DeepMind system learned to play seven different Atari games without any prior model for the games. This remarkable system's only prior information was that it should seek to maximize the game score, which is a reward from the environment. Chapters 6 and 7 will describe unintended and harmful behaviors of AI systems seeking to maximize rewards from their environments. These understandings can help us to design ethical AI systems. In fact, when DeepMind was acquired by Google, part of the deal was a commitment by Google to set up a committee on ethical AI that included one of DeepMind's owners (Bosker 2014).

There is an important difference between RL and ethical AI. In the study of RL, utility function values (rewards) are defined by the environment and are part of the problem to be solved by the agent design whereas, in the study of ethical AI, utility functions are part of the agent definition and hence part of the solution to the problem of defining agents that help rather than harm humans.

### 3. How Can AI Agents Learn Environment Models?

As noted previously, the environment model for current AI systems like the Google car is mostly designed by human engineers. Future AI systems that exceed human intelligence will need to learn environment models through their own exploration. Marcus Hutter (2005) had the insight that AI model learning is mathematically similar to Ray Solomonoff's work on sequence prediction (Solomonoff 1964). We assume that the agent's observations of the environment can be generated by computer programs and design the agent to search for those programs. The programs constitute the AI's environment model and generate the probability distribution used to predict observations. This idea is important both for the theory of AI and for development of practical AI systems. Hutter grounds his approach in the philosophies of Occam and Epicurus. Occam's razor says that we should favor the simplest explanation that fits our observations. Epicurus tells us to include all explanations consistent with our observations. Hutter's universal AI models the environment with all programs that are consistent with the agent's observations, but assigns higher probabilities to shorter programs.

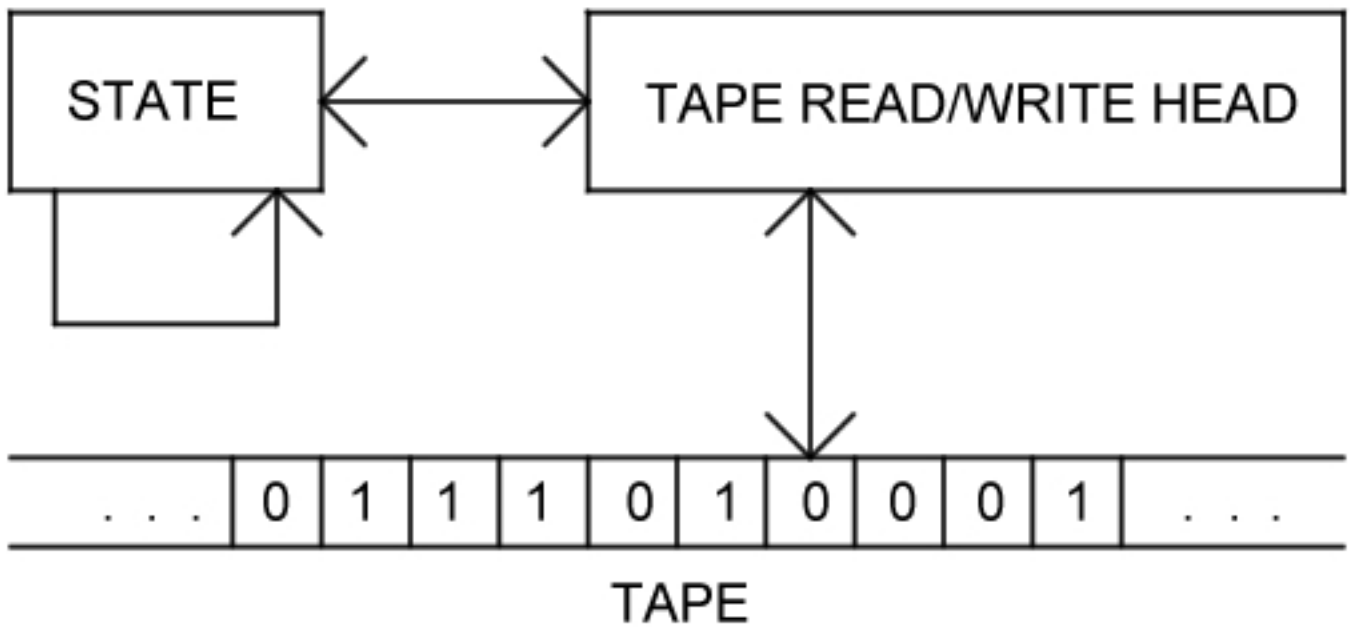

Figure 3.1 Turing machine with one tape.

In order to define the search for programs mathematically, Solomonoff and Hutter used *Turing machines* (TM), named for their inventor Alan Turing (1936). A TM is an abstract mathematical model of computation and, according to the Church-Turing thesis, can express any computation that can be expressed in any other way. This thesis has been proved for every known way of expressing computations and is widely accepted by computer scientists. TMs are very simple. They start in a specially designated "start state" and, at each of a discrete series of time steps, transition to a next state, chosen from a finite set of states, until they reach a "stop state." A TM also has one or more infinite tape memories, each with a read/write head located somewhere along the tape, and a finite set of rules that describe how the state changes (or the TM halts) and whether and how the tapes are written and moved, depending on the current state and the tape values (0 or 1) under the read/write heads. Figure 3.1 is a simple schematic of a TM with one tape. Table 3.1 defines its state transition rules.

| | tape read: 0 | tape read: 1 |
|---|---|---|
| *start* | *start*, -, 1 | *s*0, *left*, - |
| *s*0 | *s*0, *left*, 1 | *s*1, *right*, - |
| *s*1 | *s*0, *left*, 1 | *stop*, -, 0 |
| *stop* | *halt* | *halt* |

Table 3.1 State transition rules for a Turing machine with four states: *start*, *s*0, *s*1 and *stop*. Given a previous state (in the left-hand column) and the contents under the tape head (in the top row), the rules specify the next state (or that the TM should *halt*), whether to move the tape head *left* or *right* (or not move, indicated by -) and whether to write 0 or 1 to the tape (or not write, indicated by -).

Turing proved several interesting properties of his Turing machines. First, there are *universal Turing machines* (UTMs). UTMs interpret the contents of one of their tapes as a program that causes the UTM to simulate any other Turing machine. The program occupies a finite interval of the infinite tape. The rest of the tape is interpreted as the initial tape contents (i.e., the input) of the TM being simulated. In order to avoid confusion between the program and the input to the program, programs are assumed to be encoded in *prefix-free* bit strings; no program is an initial substring (i.e., a prefix) of another program making it possible to always know where the program ends and the data begins. In a complex construction, Turing described the state set and the transition rules of a UTM (similar to Table 3.1 but much more complex) and proved that for every TM there existed a program that caused the UTM to simulate that TM.

There are an infinite number of UTMs. In order to define Hutter's solution to AI model learning, we pick a UTM, called a *reference UTM*. Hutter's solution to AI model learning, called *universal AI*, searches for programs for that UTM. Given a finite interacting sequence of actions from the agent and observations from the environment, the agent defines its environment model as the UTM programs that generate the sequence of observations in response to the sequence of actions. Each program in the environment model can predict future observations. A universal AI's predictions are averages of the predictions by the programs in the environment model, weighted according to the programs' lengths. That is, by Occam's Razor, shorter UTM programs are taken to be more likely, and consequently weigh more heavily in computing probabilities of future observations.

Hutter defined universal AI as an RL agent, meaning its utility function is defined by rewards from the environment. As such, the utility function cannot be used to express ethical preferences. However, the key idea of universal AI is its way of learning an environment model. That certainly can be used in conjunction with a utility function defined to express ethical preferences rather than one defined by rewards from the environment.

Turing proved that it is not possible for a UTM program to examine another UTM program and its input tape, and determine whether the other program will halt with the given input. This is called the *halting problem*. We say that the halting problem is *undecidable*, meaning that no UTM program can always solve it. This fact implies that no TM can compute universal AI; the search for all programs of any given length that generate a sequence of environmental observations cannot terminate because there is no way to know which of those programs will terminate.

Despite the fact that it cannot be finitely computed, Hutter proved that universal AI is *Pareto optimal*. Pareto optimal means that no other agent can receive higher rewards than universal AI in any environment unless it gets lower rewards than universal AI in some other environment. But note that there certainly are agents that do better than universal AI in our particular environment on Earth–these agents start with a good deal of knowledge about our environment. You would not want to be a passenger in a self-driving car operated by universal AI, although Google's car system would do worse than universal AI in environments unrelated to driving.

### 3.1 Universal AI

Let $U$ be a reference UTM and let $Q$ be the infinite set of all programs for $U$. These programs are finite bit strings in some prefix-free set. Hutter assumed that the $U$ is deterministic, which means that for any given state and tape contents under read/write heads, there is exactly one successor state and one action for each tape. Let $h = (a_1, o_1, ..., a_t, o_t) \in H$ be an interaction history. Given a program $q \in Q$, we write $o(h) = U(q, a(h))$, where $o(h) = (o_1, ..., o_t)$ and $a(h) = (a_1, ..., a_t)$, to mean that $q$ produces the observations $o_i$ as output on a tape, in response to the actions $a_i$ as input on a tape, for $1 \leq i \leq t$. We assign the prior probability $\xi(q) = 2^{-|q|}$ to program $q$ where $|q|$ is the length of $q$ in bits. Then we define the prior probability of history $h$ as:

(3.1) $$\rho(h) = \sum_{q:o(h)=U(q,\, a(h))} \xi(q).$$

If we use this $\rho(h)$ in equations (2.3)–(2.5), and define $u(h)$ as the *reward* from the environment at time step $|h|$, then the agent $\pi$ is Hutter's universal AI. That is, each observation $o_i$ is factored into an ordinary observation $o'_i$ and a reward $r_i$ as $o_i = (o'_i, r_i)$ with $u(h) = r_{|h|}$. We assume $0 \leq r_i \leq 1$ to ensure that values converge in equations (2.3)–(2.5).

Kraft's Inequality implies that $\rho(h) \leq 1$ (Li and Vitanyi, 1997) in equation (3.1), although $\rho(h)$ is not a proper probability distribution because its values do not sum to 1 over mutually exclusive and exhaustive sets of possibilities. However, its only use in equations (2.3)–(2.5) is in the form of conditional probabilities $\rho(o \mid ha)$ and those values are normalized to a proper probability distribution in equation (2.1). Although the UTM $U$ is deterministic, the agent's environment model is stochastic because $\rho(h)$ is defined by a distribution over UTM programs in equation (3.1).

Hutter called the agent AIXI, combining AI with the Greek letter $\xi$ (XI).The distribution $\xi(q)$ is related to the Kolmogorov complexity $K(h)$ of a history $h$, which is defined as the length of the shortest program $q$ such that $o(h) = U(q, a(h))$. Because the halting problem is undecidable, $K(h)$ cannot be finitely computed. To understand this issue, assume a UTM program $p$ is trying to compute $K(h)$. To do that it simulates UTM programs $q$ on input $a(h)$ until it finds the shortest program whose output is $o(h)$. Assume $p$ has found a program $q_1$ whose output is $o(h)$ but $p$ is still simulating another program $q_2$ that is shorter than $q_1$. Because $p$ cannot decide the halting problem, it cannot decide whether $q_2$ will run forever or eventually halt, possibly with output $o(h)$. The best that $p$ can do is to produce a series of decreasing estimates of $K(h)$, as programs halt with output $o(h)$. But in general, $p$ cannot decide when the search ends. A similar process can be used to produce a series of increasing estimates of $\rho(h)$. But $\rho(h)$ cannot be finitely computed and neither can AIXI.

The next chapter will discuss a different theoretical approach to learning environment models based on the assumption that the environment is finite. That approach avoids the infinite tapes of TMs and is finitely computable.

### 3.2 A Formal Measure of Intelligence

In addition to its role in defining idealized intelligent agents, Kolmogorov complexity can also be used to define measures of intelligence (Hernández-Orallo 2000). Legg's and Hutter's (2006) formal measure of intelligence is closely related to universal AI. This measure is defined in terms of the expected values of utility functions that the agent can achieve. In order not to favor agents that simply have high utility function values, the measure is only defined for agents whose utility function values are rewards from the environment as $u(h) = r_{|h|}$. The measure is different from AIXI in that it assumes stochastic environments. Note that it is not sufficient to use a non-deterministic reference UTM, because non-deterministic TMs do not specify probabilities of their possible next states. Rather, the programs for the reference UTM must compute probabilities for observations as functions of agent actions.

Given an agent π and an environment μ, an interaction between π and μ will produce a sequence of rewards $r_i$ for $i = 0, 1, 2, \ldots$ (either infinite or up to some final time step). The value of agent π in environment μ is defined by the expected value of the sum of future, discounted rewards:

(3.2) $$V_\mu^\pi = \mathbf{E}(\sum_{i \geq 0} \gamma^i r_i).$$

Note that Legg and Hutter assumed that discounts are built into rewards and so did not explicitly include $\gamma^i$ in equation (3.2), but I include it to make it clear that (3.2) converges (we also assume that $0 \leq r_i \leq 1$). This expected value is the average value of the sum of discounted rewards over many interactions between π and a stochastic environment μ. The intelligence of agent π is defined by a weighted sum of its values over a set $E$ of computable environments. Environments are computed by programs, finite prefix-free binary strings, on some reference UTM $U$. The weight for $\mu \in E$ is defined in terms of its Kolmogorov complexity:

$$K(\mu) = \min \{ |p| : U(p) \text{ computes } \mu \}$$

where $|p|$ denotes the length of program $p$. The intelligence of agent π is then defined as:

(3.3) $$V^\pi = \sum_{\mu \in E} 2^{-K(\mu)} V_\mu^\pi.$$

Reference UTMs pose subtle problems. I showed (Hibbard 2009) that given an arbitrary environment $\mu \in E$ and $\varepsilon > 0$, there exists a reference UTM $U_\mu$ such that for all agents π and for $V^\pi$ computed according to $U_\mu$,

$$V_\mu^\pi / 2 \leq V^\pi < V_\mu^\pi / 2 + \varepsilon.$$

That is, the intelligence of any agent can be determined, within an arbitrarily small epsilon, by its value with respect to any single environment we choose, simply by picking the appropriate reference UTM. As Legg and Hutter suggested, it may be useful to apply some criterion to reference UTMs, such as picking the UTM with the smallest number of states. It is worth noting that if AIXI and an intelligence measure share the same reference UTM, then AIXI has the maximum possible intelligence by that measure.

### 3.3 Modifications of the Agent Framework

Universal AI learns an environment model in the form of a probability distribution $\rho(h)$ over interaction histories. Conditional probabilities $\rho(o \mid ha)$ are derived from $\rho(h)$ for use in equations (2.3)–(2.5), repeated here for convenience:

(3.4) $v(h) = u(h) + \gamma \max_{a \in A} v(ha),$

(3.5) $v(ha) = \sum_{o \in O} \rho(o \mid ha)\, v(hao),$

(3.6) $\pi(h) := a_{|h|+1} = \operatorname{argmax}_{a \in A} v(ha).$

Given that the AIXI agent learns an abstract and complex function such as $\rho(h)$, other agents could learn the value function $v(ha)$ or the policy $\pi(h)$ rather than learning $\rho(h)$. In fact, some AI systems do just that.

DeepMind Technologies created a system that learned to play seven Atari games by learning to compute $v(ha)$ (Mnih et. al. 2013). Their system surpassed the skill of the best human players on three of the games and surpassed the best previous RL systems on six of the games. The system was not customized to any of the games. Its inputs (observations) were the contents of the video screen and the score, and its outputs (actions) were the game controls. The score was interpreted as a reward from the environment. The system did not learn an environment model like $\rho(h)$ and did not recursively apply equations (3.4) and (3.5). It simply used the learned value function $v(ha)$ for use in equation (3.6). Chapter 8 will describe a mathematical framework for agents that learn to compute $v(ha)$. With large computing resources that framework could be applied to Atari. For practical AI systems with limited resources, learning to play Atari games poses two very difficult problems. First, there may be a delay of thousands of time steps between actions and the rewards that result from those actions. Second, the Atari video screen contains 210×160 color pixels so each observation from the environment is a point in a 100,800-dimensional space (3×210×160). The DeepMind system converts color to grey-scale and down-samples and crops the video screen to 84×84 pixels to reduce this to 7056 dimensions, but that is still very high by the standards of practical AI systems. To solve these problems, the DeepMind scientists built on and improved a set of techniques known as deep learning. Their system learns an approximation to the function $v(ha)$ that enables it to choose actions that get high scores.

Imagine a modified version of the DeepMind Atari player, with the Atari screen replaced by video from cameras aimed out of car windows, game controls replaced by controls for a car's steering wheel, accelerator, brake, and transmission, and score computed according to safe delivery of passengers to their destinations, with large deductions for accidents. An improved version of the DeepMind system could probably learn to drive reasonably well. Eventually it might even be safer than the Google car, except for the dangers of RL to be discussed in Chapters 6 and 7.

Learning $\pi(h)$ is referred to as *policy iteration* or as *evolutionary programming*. For example, the Hayek system of Eric Baum and Igor Durdanovic (Baum 2004) learned to solve the block-stacking puzzle illustrated in Figure 3.2. Stack 0 contains several types of blocks with different designs. In total, stacks 1, 2, and 3 contain the same number of blocks of each design that are contained in stack 0. The goal is to get all the blocks from stacks 1, 2, and 3 into stack 1, with the order of block designs exactly matching stack 0. The solver can only move one block at a time from the top of stack 1, 2, or 3 to the top of another of stacks 1, 2, and 3. Blocks cannot be moved to or from stack 0.

In the context of our agent-environment framework, for each puzzle Hayek made a single observation of the initial configuration of the block stacks. Its sequence of moves to solve (or not solve) the puzzle constituted a single action, followed by an observation of a reward (1 if the puzzle was solved, 0 if not). Hayek's efforts to solve a sequence of puzzles constituted an interaction history, during which Hayek learned a policy $\pi(h)$.

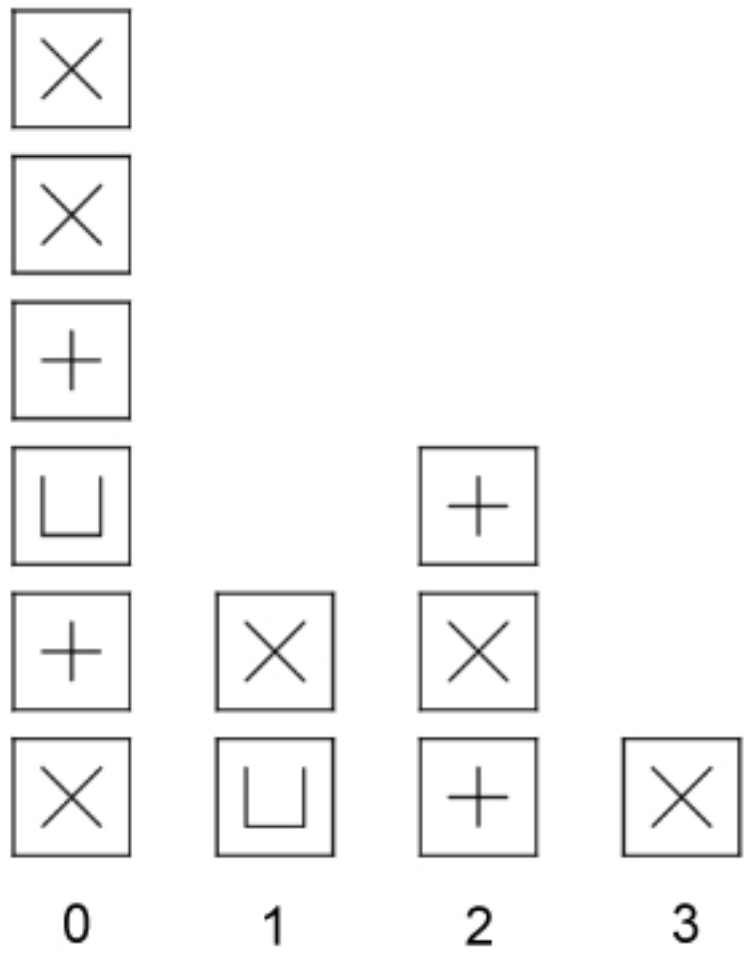

Figure 3.2 Block stacking puzzle for Hayek

Hayek learned to solve these puzzles using multiple interacting agents. These agents formed an economy, paying "money" to each other for work. The ultimate source of their money was rewards from solving the puzzle. In some experiments agents paid money for using computing resources. Each agent contributed a small part to the overall solution. Some agents cleaned blocks off stack 1, other agents moved blocks between stacks 1, 2, and 3, and a third group of agents decided when the puzzle was solved. Hayek evolved a set of about 1000 agents, each performing its task when very specific conditions were met. Agents that didn't contribute any value to solving the puzzle went bankrupt and disappeared. Collectively these agents learned a function $\pi(h)$ mapping from history to action. Baum and Durdanovic used this system to explore various economic ideas such as property rights, conservation of money, the tragedy of the commons, and the evolution of cooperation.

Hayek and the DeepMind Atari player are important milestones in the development of AI. Because they skip the step of learning $\rho(h)$ they are sometimes described as model-free. However, for the learned functions $v(ha)$ and $\pi(h)$ to receive high rewards from the environment they must implicitly encode knowledge of the environment.

### 3.4 The Ethics of Learned Environment Models

For an AI system like Google's self-driving car, the environment model is specified by engineers as part of the system design. The engineers can generally anticipate the possible situations the car will encounter and design safety into the car's responses. If the car is unable to recognize objects in the road, it can simply stop. However, more complex AI systems in the future will need to learn their environment models and hence their designers will be less able to anticipate the situations they will encounter or to specify safe resonses. Thus the ethics of future AI systems will be fundamentally different from and more difficult than the ethics of current AI systems.

When future AI systems learn to use human languages fluently and learn about human society, then they will learn about human laws, morality and ethical theories. Will they use this knowledge to guide their own ethical behavior? Not necessarily. For example, people with antisocial personality disorder (American Psychiatric Association 2013) know about social norms but often use them to predict and manipulate other people rather than to guide their own behavior. Knowledge of law, morality and ethics will contribute to an AI agent's environment model but not directly to its utility function. However, if human approval increases the value of an agent's utility function, then the agent may use its knowledge of social norms to choose behaviors that will please people.

## 4. AI in Our Finite Universe

Infinite sets are the source of theoretical difficulties in mathematics. For example, universal AI is uncomputable because of the undecidability of the halting problem. This undecidability is also the source of ethical problems for AI (Englert, Siebert and Ziegler 2014). Section 4.4 will discuss other difficulties for agents that use the mathematics of infinite sets for logical proofs. However, universal AI would be computable if it used TMs with finite tapes and the next section will describe a way to do something much like that.

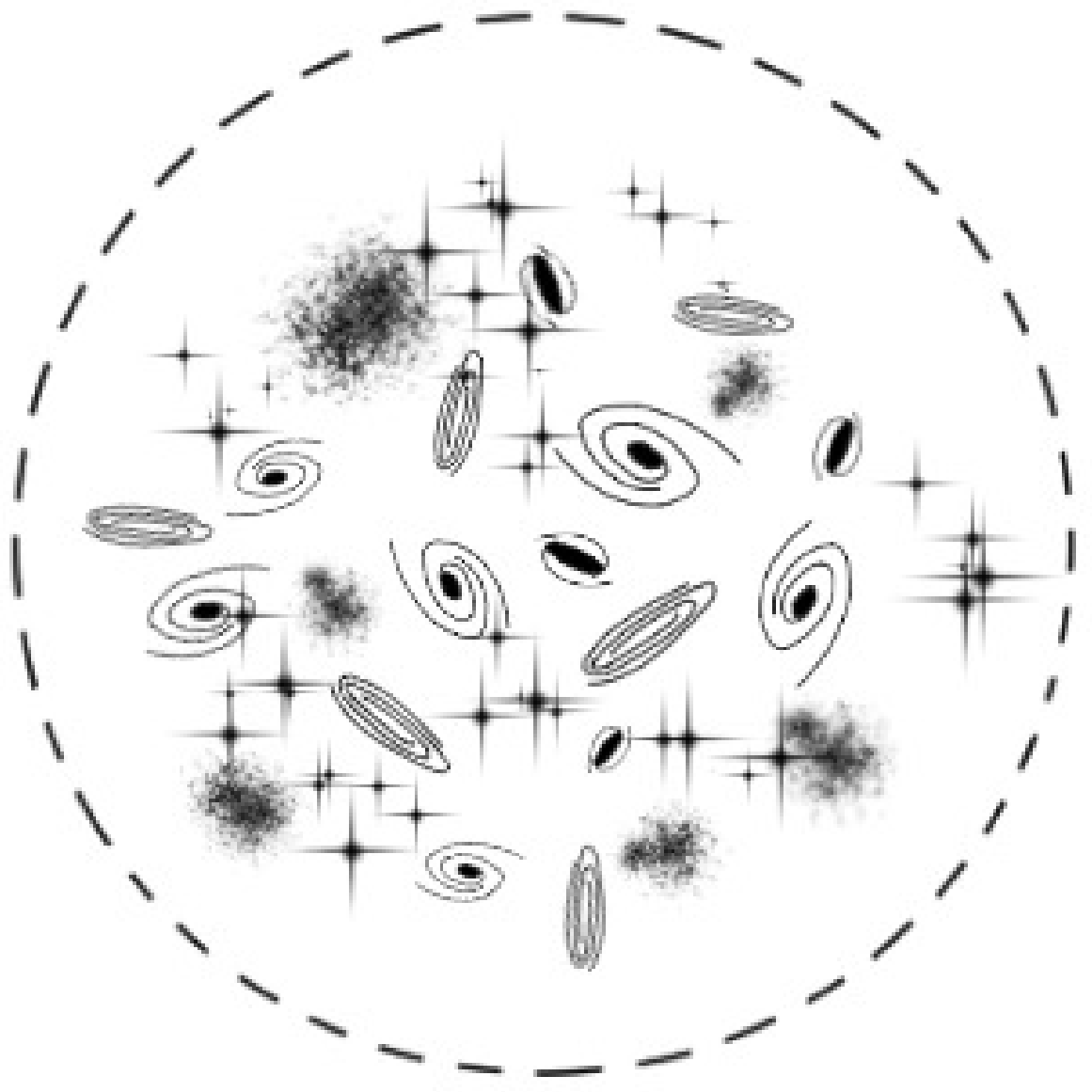

Figure 4.1 A finite universe

I argue that such difficulties are unnecessary if agent definitions are limited to finite sets (Hibbard 2014). Max Tegmark (2014, page 333) suggests that these difficulties are eliminated in a computable universe. Seth Lloyd (2002) has calculated that the observable universe has a finite information capacity of no more than $10^{120}$ bits. This number is based on a hard quantum mechanical limit on the number of possible physical states of the universe and is proportional to the age of the universe squared. That age is estimated currently to be 13.7 billion years, meaning that an assumption that our environment has an information capacity of no more than $10^{124}$ bits will remain valid for a trillion years. Alternately we could pick a much larger limit than $10^{124}$ bits, as long as it is finite. The key point is that well-confirmed physics justifies picking a finite limit, and that spares us the mathematical difficulties that accompany infinite sets.

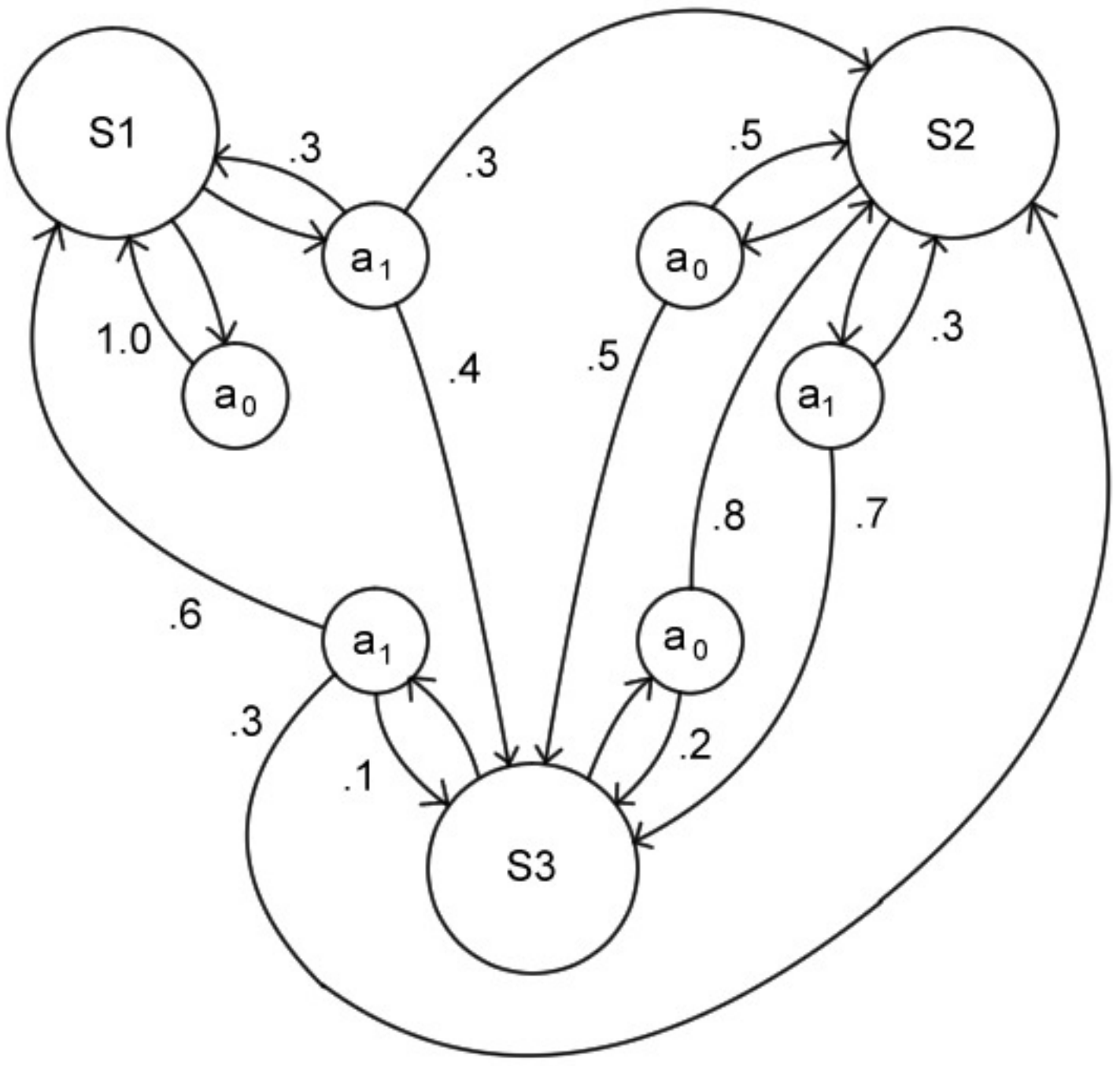

Figure 4.2 A simplified example of a Markov decision process (MDP), not showing outputs. There are three states, S1, S2, and S3, and two inputs, $a_0$ and $a_1$. The arcs are labeled with probabilities of transitioning from inputs to successor states.

Rather than using TMs, agents can model their environments with finite stochastic loop programs. These programs have a finite memory limit–no matter how long the program runs it can never use more memory than its limit. Furthermore they are stochastic, meaning they sometimes make random decisions about what to do next. Such programs can be expressed in an ordinary procedural programming language restricted to only static array declarations, no recursive functions, only loops whose number of iterations is set before iteration begins (so no while loops), and a generator for truly random numbers. Prohibiting recursive functions prevents stack memory from growing without limit. As a result, the total memory use of these program is known at compile time.

*Markov decision processes* (MDPs) are a formal model for finite stochastic loop programs (Puterman 1994; Sutton and Barto 1998; Russell and Norvig 2010). An MDP has a finite set of states including a start state, a finite set of inputs, a finite set of outputs, and a table of probabilities for the next state and output as a function of the current state and input. An MDP's only memory is storage for the value of the current state. Thus an MDP with *n* bits of memory must have $2^n$ states, meaning that MDPs are not a very efficient representation. Figure 4.2 is an example of an MDP with just three states, $S1$, $S2$, and $S3$, and two inputs, $a_0$ and $a_1$. Arcs from actions to next states are labeled with transition probabilities. Outputs are not shown in this simplified diagram.

Using MDPs to model environments, actions are identified with inputs and observations are identified with outputs in the agent-environment framework from Section 2.1. To be precise, because our agents lack knowledge of the entire environment, our framework corresponds to *partially observable Markov decision processes* (POMDPs).

Our world is governed by scientific laws. If we can learn these laws, we can give simpler, unified explanations for our observations. This logic explains Occam's razor. If agents model their environments by programs, then programs are shorter if they factor the logic common to many observations into unified functions. Hence the search for shorter programs is an agent's way of learning scientific laws. When agents find universal laws governing observations they will be able to make more accurate predictions of future observations and consequently receive higher expected future utility values. The MDPs as depicted in Figure 4.2, however, lack any way to express functions common to many situations and so lack this important property. In contrast, finite stochastic loop programs in an ordinary procedural language do have this property (as do the UTM programs used by universal AI) so we will adopt them as agents' models for finite environments.

To understand the correspondence between a finite stochastic loop program $q$ and an MDP $M_q$, let the total memory use of the program $q$, including all static variable and array declarations and adequate stack for its maximum depth of function calls, be $n$ bits. Then $M_q$ will have $2^n$ states. We assume that the memory of $q$ is not reset between action inputs from agent, just as the state of $M_q$ is not reset between inputs. While $M_q$, as a model of the environment, makes a single state transition from receiving an action to producing an observation, the program $q$ may make many steps to produce an observation output in response to an action input. Thus, if $s$ is the state of $M_q$ corresponding to the memory of $q$ before an input, the successor state to $s$ for $M_q$, after $q$ accepts an input and produces an output, corresponds to the memory of $q$ after the output (that is, the state transitions of $M_q$ do not trace all the intermediate calculations of $q$ between an input and an output). The start state of $M_q$ corresponds to the initial memory contents of $q$ before it rerceives any input actions. If any of the $2^n$ states of $M_q$ are not reachable from its start state (that is, they do not correspond to possible memory contents of $q$ after any sequence of input actions and output observations), they can be eliminated from $M_q$.

### 4.1 Learning Finite Stochastic Models

While Hutter studied using MDPs (Hutter 2009a) and dynamic Bayesian networks (DBN) (Hutter 2009b; Ghahramani 1997) for environment models, I prefer modeling environments with finite stochastic loop programs because they can factor logic common to multiple observations into unified functions (Hibbard 2012a). To do that requires a new way to compute $\rho(h)$. Let $Q$ be the set of all finite stochastic loop programs in some prefix-free procedural language. Let $\varphi(q) = 2^{-|q|}$ be the prior probability of program $q$, where $|q|$ is the length of program $q$.

Let $P(h \mid q)$ be the probability that the program $q$ computes the history $h$ (that is, produces the observations $o_i$ in response to the actions $a_i$ for $1 \leq i \leq |h|$). For a simple example, let $A = \{a, b\}$, $O = \{0, 1\}$, $h = (a, 1, a, 0, b, 1)$ and let $q$ generate observation 0 with probability 0.2 and observation 1 with probability 0.8, without any internal state or dependence on the agent's actions. Then the probability that the interaction history $h$ is generated by program $q$ is the product of the probabilities of the 3 observations in $h$: $P(h \mid q) = 0.8 \times 0.2 \times 0.8 = 0.128$.

For a more complex example, Table 4.1 defines transition probabilities for a program $q'$ that generates observations dependent on internal state and the agent's actions. The left column gives the current state of $q'$ ($s0$ or $s1$) and the agent's action ($a$ or $b$). The top row gives the next state of $q'$ and the generated observation (0 or 1). The entries are probabilities with each row summing to 1.0. As before, let the history be $h = (a, 1, a, 0, b, 1)$. The program $q'$ starts in state $s0$. There are two possible state sequences consistent with the history $h$: ($s0$, $s0$, $s0$, $s1$) and ($s0$, $s1$, $s0$, $s1$). The product of the transition probabilities for the first sequence is $0.3 \times 0.2 \times 0.4 = 0.024$ while the product of the transition probabilities for the second sequence is $0.5 \times 1.0 \times 0.4 = 0.2$. These sequences are mutually exclusive, hence $P(h \mid q')$ is the sum $0.2 + 0.024 = 0.224$.

| | $s0$, 0 | $s0$, 1 | $s1$, 0 | $s1$, 1 |
|---|---|---|---|---|
| $s0$, $a$ | 0.2 | 0.3 | 0.0 | 0.5 |
| $s0$, $b$ | 0.3 | 0.0 | 0.3 | 0.4 |
| $s1$, $a$ | 1.0 | 0.0 | 0.0 | 0.0 |
| $s1$, $b$ | 0.3 | 0.3 | 0.2 | 0.2 |

Table 4.1 Transition probabilities for the finite stochastic loop program $q'$. Rows are labeled for the current state and action ($a$ or $b$). Columns are labeled for next state and observation (0 or 1). This representation of the finite stochastic loop program $q'$ using an MDP is efficient for a program with only two states. However, for more complex programs, a procedural representation is generally more efficient.

Given an interaction history $h$, the environment model is the single program that provides the most probable explanation of $h$, that is the $q$ that maximizes $P(q \mid h)$. By Bayes' Theorem:

(4.1) $$P(q \mid h) = P(h \mid q)\ \varphi(q) \ /\ P(h).$$

Because it is constant over all $q$, $P(h)$ can be eliminated. Thus, given a history $h$, we define $\lambda(h)$ as the most probable program modeling $h$ by:

(4.2) $$\lambda(h) := \text{argmax}_{\ q \in Q}\ P(h \mid q)\ \varphi(q).$$

Given an environment model $\lambda(h)$, the following can be used for the prior probability of an observation history $h'$ extending $h$ (i.e., $h$ is an initial sub-interval of $h'$):

(4.3) $$\rho(h') = P(h' \mid \lambda(h)).$$

This distribution $\rho(h')$ can be used in the framework defined by equations (2.1), (2.3)–(2.5). Not only are the environment models finite, but they can be finitely computed by the agent (Hibbard 2012b):

**Proposition 4.1.** Given a finite history $h$, the model $\lambda(h)$ can be finitely computed.

**Proof.** Given $h = (a_1, o_1, ..., a_t, o_t)$, let $q_0$ be the program that produces observation $o_i$ at time step $i$ for $1 \le i \le t$. That is, $q_0$ is a simple "table-lookup" that produces the observations $o_i$ in sequence. Let $n = |q_0|$. Since the behavior of $q_0$ is deterministic, $P(h \mid q_0)\ \varphi(q_0) = 1 \times 2^{-n} = 2^{-n}$. Hence the maximum value in equation (4.2) must be at least $2^{-n}$, that is, $P(h \mid \lambda(h))\ \varphi(\lambda(h)) \ge 2^{-n}$. For any program $q$ with $|q| > n$, $P(h \mid q)\ \varphi(q) < 1 \times 2^{-n} = 2^{-n}$ so $\lambda(h) \ne q$. Thus one algorithm for finitely computing $\lambda(h)$ is an exhaustive search of the finite number of programs $q$ with $|q| \le n$. □

As the computing time of this exhaustive search for $\lambda(h)$ is exponential in $|h|$, this procedure is not practical for learning environment models. However, it is useful to have a finitely computable theoretical framework.

An agent that is part of a finite universe will not have the necessary computational resources to compute the expressions defined by equations (4.2), (4.3), (2.1) and (2.3)–(2.5). Chapter 8 will address the issue of agents that must approximate these equations.

The derivation of $\rho(h)$ in equations (4.1)–(4.3) takes into account Occam's razor, to favor the simplest explanation of the observations, while ignoring Epicurus' principle to include all explanations. An alternate way to compute $\rho(h)$ accounts for all finite stochastic loop programs:

(4.4) $$\rho(h) = \sum_{q \in Q} P(h \mid q)\ \varphi(q).$$

Because this equation requires a sum over an infinite number of programs, it is not finitely computable. However, it can be approximated to any desired precision in finite computing time.

### 4.2 When Is the Most Likely Finite Stochastic Model the True Model?

If history $h$ is generated by a finite stochastic loop program $q$, then we call $q$ the true model of $h$ (despite the dictum that, "Essentially, all models are wrong, but some are useful." (Box and Draper 1987)). In that case is the most likely model $\lambda(h)$ equal to $q$? Not necessarily. Because $q$ is stochastic it can generate any history $h$ with $P(h \mid q) > 0$ (i.e., $P(h \mid q)$ may be very low), and as a result there may be another program $q'$ such that $P(h \mid q')\ \varphi(q') > P(h \mid q)\ \varphi(q)$. Or there may be $q'$ such that $\varphi(q') > \varphi(q)$ and $\forall h \in H.(P(h \mid q') = P(h \mid q))$ (i.e., the behavior of $q'$ is identical with the behavior of $q$), and thus $\forall h \in H.(P(h \mid q')\ \varphi(q') > P(h \mid q)\ \varphi(q))$.

Programs $q$ and $q'$ are equivalent to finite MDPs and thus they may pass through an initial sequence of states that they never visit after some finite time step. This may cause there to exist a history $h_0$, generated by $q$ but with low probability, such that $P(h \mid q')\ \varphi(q') > P(h \mid q)\ \varphi(q)$ for all histories $h$ whose initial sub-sequence is $h_0$. (i.e., $\exists h'.h = h_0 h'$). In that case, as the lengths of histories $h$ randomly generated by $q$ increases, the probability that $\lambda(h) \neq q$ is no more than $1\text{-}(1/|\{h \in H \mid |h| = |h_0|\}|) < 1$. Therefore it is not even true that, averaged over histories of increasing length generated by $q$, the probability that $\lambda(h) = q$ converges to 1. So what can we say about the relation between $\lambda(h)$ and the true model $q$?

In order to analyze the question, we need to define expected values with respect to the distribution of histories of length $n$ generated by $q$. In a history $h = (a_1, o_1, ..., a_n, o_n)$ the probability $P(h \mid q)$ is defined based on the probabilities of the observations $o_i$ only, since the actions $a_i$ are generated by the agent rather than the environment. Recalling the notations $a(h) = (a_1, ..., a_n)$ and $o(h) = (o_1, ..., o_n)$, define the set of all sequences of $n$ actions:

(4.5) $$A(n) = \{(a'_1, ..., a'_n) \mid a'_i \in A\}.$$

Then $|A(n)| = |A|^n$. For any fixed sequence of actions $a' \in A(n)$ and any finite stochastic loop program $q'$, the sum of probabilities $P(h \mid q')$ of histories with $a(h) = a'$ must be 1:

(4.6) $$\sum_{|h|=n \wedge a(h)=a'} P(h \mid q') = 1.$$

The expected value of a function $f(h)$ with respect to the distribution of histories of length $n$ generated by $q$, and the limit as $n$ increases to infinity, are defined as:

(4.7) $$E(f(h) \mid q, n) = \sum_{a' \in A(n)} (\sum_{|h|=n \wedge a(h)=a'} P(h \mid q)\, f(h)) \,/\, (|A|^n),$$

(4.8) $$E(f(h) \mid q) = \text{limit}_{n \to \infty}\, E(f(h) \mid q, n).$$

Given a finite stochastic loop program $q'$, the probability that $q'$ will be a more likely model than $q$ for a history of length $n$ generated by $q$ is:

(4.9) $$E((\text{if } P(h \mid q')\, \varphi(q') \,/\, \varphi(q) > P(h \mid q) \text{ then } 1 \text{ else } 0) \mid q, n).$$

We know that:

(4.10) $$\sum_{|h|=n \wedge a(h)=a'} P(h \mid q) = 1.$$

Assume that $\varphi(q') / \varphi(q) = \beta$ so that:

(4.11) $$\sum_{|h|=n \wedge a(h)=a'} (P(h \mid q')\, \varphi(q') \,/\, \varphi(q)) = \beta.$$

Then $(P(h \mid q')\, \varphi(q') / \varphi(q))$ can be larger than $P(h \mid q)$ over histories $h$ whose $P(h \mid q)$ values sum to at most $\beta$, so that:

(4.12) $$(\sum_{|h|=n \wedge a(h)=a'} P(h \mid q)\, (\text{if } P(h \mid q')\, \varphi(q') \,/\, \varphi(q) > P(h \mid q) \text{ then } 1 \text{ else } 0)) \leq \beta.$$

Averaging equation (4.12) over actions $a' \in A(n)$ gives:

(4.13) $$E(\text{if } P(h \mid q')\ \varphi(q') / \varphi(q) > P(h \mid q) \text{ then } 1 \text{ else } 0 \mid q, n) \leq \beta = \varphi(q') / \varphi(q).$$

Because we are using a prefix-free encoding for finite stochastic loop programs $q' \in Q$, we have $\sum_{q' \in Q} \varphi(q') \leq 1$. Thus:

(4.14) $$\forall \varepsilon > 0.\ \exists Q' \subset Q.\ Q' \text{ is finite and } \sum_{q' \in Q-Q'} \varphi(q') / \varphi(q) < \varepsilon.$$

Combining equations (4.9), (4.13), and (4.14), and noting that equation (4.13) is uniformly true over all $n$, we can conclude that:

**Proposition 4.2.** Given any $\varepsilon > 0$ there exists a finite set $Q'$ of finite stochastic loop programs such that the probability is less than $\varepsilon$ that $\lambda(h) = q'$ for any program $q'$ not in $Q'$, for histories $h$ randomly generated by $q$.

Thus to analyze the probability that $\lambda(h) = q$ we need only compare $q$ to a finite set $Q'$ of programs. However, various problems may complicate the analysis. For example there may be a program $q' \in Q$ such that $\forall h \in H.\ P(h \mid q') = P(h \mid q)$ and $|q'| < |q|$ (so $\varphi(q') > \varphi(q)$). In that case $\lambda(h)$ will never equal $q$.

For a specific $q'$ we can analyze the probability, with respect to the distribution of histories generated by $q$, that $P(h \mid q')\ \varphi(q') > P(h \mid q)\ \varphi(q)$, by defining an appropriate function from histories to real numbers $f : H \rightarrow \mathbf{R}$. Define distributions of finite histories with length $n$:

(4.15) $$P_n(h \mid q) = P(h \mid q) / |A|^n \text{ for } |h| = n,$$

$$P_n(h \mid q) = 0 \text{ for } |h| \neq n,$$

(4.16) $$P_n(h \mid q') = P(h \mid q') / |A|^n \text{ for } |h| = n,$$

$$P_n(h \mid q') = 0 \text{ for } |h| \neq n.$$

Also define means and standard deviations of the function $f$ with respect to the distributions $P_n(h \mid q)$ and $P_n(h \mid q')$, and in the limit as $n$ increases to infinity:

(4.17) $$m(f \mid q, n) = E(f(h) \mid q, n),$$

(4.18) $m(f \mid q) = \mathrm{E}(f(h) \mid q)$,

(4.19) $m(f \mid q', n) = \mathrm{E}(f(h) \mid q', n)$,

(4.20) $m(f \mid q') = \mathrm{E}(f(h) \mid q')$,

(4.21) $\sigma(f \mid q, n) = (\mathrm{E}((f(h) - m(f \mid q, n))^2 \mid q, n))^{1/2}$,

(4.22) $\sigma(f \mid q) = \mathrm{limit}_{n\to\infty}\, \sigma(f \mid q, n)$,

(4.23) $\sigma(f \mid q', n) = (\mathrm{E}((f(h) - m(f \mid q', n))^2 \mid q', n))^{1/2}$,

(4.24) $\sigma(f \mid q') = \mathrm{limit}_{n\to\infty}\, \sigma(f \mid q', n)$.

The approach is to define a function $f(h)$ with different mean values with respect to history distributions $P_n(h \mid q')$ and $P_n(h \mid q)$, and with standard deviations converging to 0 as history lengths increase to infinity. In that case the proportion of overlap between the history distributions $P_n(h \mid q')$ and $P_n(h \mid q)$ converges to 0 as history lengths increase. More precisely:

**Proposition 4.3.** If there is a function $f(h)$ such that $m(f \mid q)$, $m(f \mid q')$, $\sigma(f \mid q)$, and $\sigma(f \mid q')$ converge, $m(f \mid q) \neq m(f \mid q')$, and $\sigma(f \mid q) = \sigma(f \mid q') = 0$, then as the lengths of histories $h$ generated by $q$ increase the probability that $\lambda(h) = q'$ converges to 0:

(4.25) $\mathrm{E}((\text{if } P(h \mid q')\, \varphi(q') / \varphi(q) > P(h \mid q) \text{ then 1 else 0}) \mid q) = 0$.

**Proof.** Assume that $m(f \mid q') > m(f \mid q)$ (if the order is reversed then certain orders in the following discussion can be reversed). Given any positive integer $L$ there is $d > 0$ and a positive integer $N$ such that for all $n > N$:

(4.26) $m(f \mid q', n) - m(f \mid q, n) \geq d$,

(4.27) $\sigma(f \mid q, n) < d/(2L)$,

(4.28) $\sigma(f \mid q', n) < d/(2L)$.

To understand this claim, first choose $d$ and $N$ so that equation (4.26) is true, then increase $N$ until equations (4.27) and (4.28) are true. Set $e = (m(f \mid q, n) + m(f \mid q', n))/2$ midway between $m(f \mid q, n)$ and $m(f \mid q', n)$. Then $e$ is at least $L$ standard deviations $\sigma(f \mid q, n)$ away from the mean $m(f \mid q, n)$ and at least $L$ standard deviations $\sigma(f \mid q', n)$ away from the mean $m(f \mid q', n)$. Hence a proportion of no more than $L^{-2}$ of the distribution $P_n(h \mid q)$ can have $f(h) \geq e$ and a proportion of no more than $L^{-2}$ of the distribution $P_n(h \mid q')$ can have $f(h) \le e$. For histories with $f(h) \ge e$, $P_n(h \mid q')\ \varphi(q') / \varphi(q)$ can be greater than a proportion of no more than $L^{-2}$ of the distribution $P_n(h \mid q)$. And for histories with $f(h) \le e$, $P_n(h \mid q')\ \varphi(q') / \varphi(q)$ can be greater than a proportion of no more than $(\varphi(q') / \varphi(q))\ L^{-2}$ of the distribution $P_n(h \mid q)$. Thus:

(4.29) $$E((\text{if } P(h \mid q')\ \varphi(q') / \varphi(q) > P(h \mid q) \text{ then 1 else 0}) \mid q, n) < (1 + \varphi(q') / \varphi(q))\ L^{-2}$$

Since $L$ can be arbitrarily large and $(1 + \varphi(q') / \varphi(q))$ is constant, equation (4.25) is proved. □

Propositions 4.2 and 4.3 give us the tools to analyze when the true model is the most likely model for a distribution of interaction histories. Consider a distribution $P(h \mid q)$ of histories randomly generated by a true model $q$. Given $\varepsilon > 0$, Proposition 4.2 tells us that there is a finite set $Q'_\varepsilon$ of finite stochastic loop programs such that, for $h$ randomly generated by $q$, the probability is less than $\varepsilon/2$ that $\lambda(h) = q'$ for any program $q' \notin Q'_\varepsilon$. Then, for each $q' \in Q'_\varepsilon$, we need to find a function $f_{q'}(h)$ of histories that satisfies the conditions of Proposition 4.3. If we can do that for $q' \in Q'_\varepsilon$, then there is an integer $L_{q'}$ such that, for $h$ randomly generated by $q$ with $|h| > L_{q'}$, the probability is less than $\varepsilon/(2|Q'_\varepsilon|)$ that $\lambda(h) = q'$. Let $L_\varepsilon = \max_{q' \in Q'_\varepsilon} L_{q'}$. Then, for $h$ randomly generated by $q$ with $|h| > L_\varepsilon$, the probability is less than $\varepsilon$ that $\lambda(h) \ne q$. That is, the probability that the most likely model is the true model converges to 1 as history lengths increase. $|Q'_\varepsilon|$ will increase with decreasing $\varepsilon$. The key is being able to find $f_{q'}(h)$ for each $q' \in Q'_\varepsilon$. It is important that $Q'_\varepsilon$ be finite in Proposition 4.2, so that the maximum value $L_\varepsilon$ is finite (i.e., the maximum of a finite set of integers is finite, while the maximum of an infinite set may be infinite).

A useful class of functions for Proposition 4.3 counts the frequency of occurrence of histories as sub-sequences. We define a function $f_{h'}(h)$ for each $h' \in H$. A history $h \in H$ will have $max(|h|-|h'|+1, 0)$ sub-sequences of length $|h'|$, and $f_{h'}(h)$ is the proportion of these that equal $h'$. If $|h|<|h'|$, define $f_{h'}(h) = 0$. If $|h| \ge |h'|$, define:

(4.30) $$f_{h'}(h) = (\textstyle\sum_{0 \le i \le |h|-|h'|} \text{if } (\exists h_1.\exists h_2.h=h_1 h' h_2. \wedge |h_1|=i) \text{ then 1 else 0}) / (|h|-|h'|+1).$$

Before applying the functions $f_{h'}(h)$, we need to digress into the theory of Markov chains (Levin, Peres, and Wilmer 2008; Russell and Norvig 2010). A finite stochastic loop program $q$ is equivalent to a finite MDP $M_q$, as dicussed at the start of this chapter. All action sequences in $A(n)$ are weighted equally in $P_n(h \mid q)$, so $P_n(h \mid q)$ views actions as uniformly distributed random inputs to $M_q$. Given such random inputs, the MDP $M_q$ will generate the same distribution of histories that a Markov chain $MC_q$ generates (here we add action-observation pair outputs to the usual definition of Markov chains; the important point is that the transition probability to the next state is only dependent on the previous state). The probability of the transition from state $s$ to state-action-observation $(s', a, 0)$ in $MC_q$ is equal to the probability from $(s, a)$ to $(s', o)$ in $M_q$, divided by $|A|$. For example, the $M_q$ specified in Table 4.1, with uniformly distributed random action inputs, generates the same distribution of histories that the $MC_q$ specified in Table 4.2 generates.

| | $s0, a, 0$ | $s0, a, 1$ | $s0, b, 0$ | $s0, b, 1$ | $s1, a, 0$ | $s1, a, 1$ | $s1, b, 0$ | $s1, b, 1$ |
|---|---|---|---|---|---|---|---|---|
| $s0$ | 0.1 | 0.15 | 0.15 | 0.0 | 0.0 | 0.25 | 0.15 | 0.2 |
| $s1$ | 0.5 | 0.0 | 0.15 | 0.15 | 0.0 | 0.0 | 0.1 | 0.1 |

Table 4.2 The $MC_q$ equivalent to the $M_q$ specified in Table 4.1, given uniformly distributed random action inputs.

The behavior of finite Markov chains is well understood (Doob 1953; Levin, Peres, and Wilmer 2008). Let $S$ be the set of states of $MC_q$ and, for $s, s' \in S$, let $P(s, s')$ be the probability of transition from state $s$ to state $s'$. Viewing $P$ as a matrix, its $n$-th power $P^n(s, s')$ is the probability of transition from $s$ to $s'$ in exactly $n$ time steps. State $s'$ is *reachable* from state $s$ if there exists $n \geq 1$ such that $P^n(s, s') > 0$. A *communicating class* is a set of states that are all reachable from each other. A communicating class is *essential* if only states in the class are reachable from the class. Let $\theta: S \to [0, 1]$ be a probability distribution on the state set $S$ (thus $\theta(s) \geq 0$ for $s \in S$ and $\sum_{s \in S} \theta(s) = 1$). We say $\theta$ is *stationary* if, for all $s' \in S$, $\theta(s') = \sum_{s \in S} \theta(s) P(s, s')$.

**Proposition.** There is a unique stationary distribution $\theta$ for a transition matrix $P$ if and only if there is a unique essential communicating class (Levin, Peres, and Wilmer 2008).

If $MC_q$ has a unique stationary distribution $\theta$, then $\theta(s)$ is the expected probability that $MC_q$ will be in state $s$ after a long sequence of state transitions (there are ways to compute the rate of convergence to $\theta(s)$).

Define $T(s) = \{n \geq 1 \mid P^n(s, s) > 0\}$. The *period* of $s$ is the greatest common divisor of $T(s)$. All the states in a communicating class have the same period, which is the period of the class. The class in *aperiodic* if its period is 1. Table 4.3 is an example of a Markov chain with period 2. It alternates between states $s0$ and $s1$. Because $\{s0, s1\}$ is a unique essential communicating class, it has a stationary distribution $\theta(s0) = 0.5$, $\theta(s1) = 0.5$. The average time the Markov chain spends in either state is 0.5, but on any given time step the probability is either 0 or 1. Independent of whether its unique essential communicating class is aperiodic or not, the average time a Markov chain with a unique stationary distribution $\theta$ spends in state $s$ is $\theta(s)$.

| | $s0$ | $s1$ |
|---|---|---|
| $s0$ | 0 | 1 |
| $s1$ | 1 | 0 |

Table 4.3 A Markov chain with period 2.

If $MC_q$ has a unique stationary distribution $\theta$, then we can use it to compute $m(f_{h'} \mid q)$ for $h' \in H$. Given $h' \in H$, for each state $s$ of $MC_q$, compute the probability $P(h' \mid s, MC_q)$ that $MC_q$, starting at state $s$, produces $h'$ during its next $|h'|$ state transtions (this is similar to the example of computing of $P(h \mid q)$ using Table 4.1). Then:

(4.31) $$m(f_{h'} \mid q) = \sum_{s \in S} \theta(s)\, P(h' \mid s, MC_q).$$

Because $f_{h'}(h)$ is the proportion of occurrences of $h'$ over the entire length of $h$, as $|h|$ increases the value of $f_{h'}(h)$ for $h$ randomly generated by $q$ will converge to the expression on the right side of equation (4.31). Any effect of period greater than 1 for its unique essential communicating class on $m(f_{h'} \mid q, n)$ and $\sigma(f_{h'} \mid q, n)$ will smooth out as $n$ increases (this would not be the case for a function $f(h)$ defined in terms of only, for example, the last 100 time steps in $h$). Thus $m(f_{h'} \mid q)$ and $\sigma(f_{h'} \mid q)$ converge, with $\sigma(f_{h'} \mid q) = 0$. Note that if $MC_q$ has multiple essential communicating classes then it will have multiple stationary distributions, each defining a value for $m(f_{h'} \mid q)$. In that case $\sigma(f_{h'} \mid q)$ would not converge to 0.

Define a subclass of models $Q_{uecc} = \{q \in Q \mid MC_q$ has a unique essential communicating class$\}$. Then, by Propostion 4.3, if the probability that $\lambda(h) = q' \in Q_{uecc}$ does not converge to 0 as the lengths of histories $h$ randomly generated by $q \in Q_{uecc}$ increase, then $m(f_{h'} \mid q') = m(f_{h'} \mid q)$ for all $h' \in H$. Thus, restricting models to $Q_{uecc}$, even though the most likely finite stochastic model $\lambda(h)$ is not necessarily equal to the true model $q$, as the lengths of histories $h$ generated by $q$ increase, $q$ and $\lambda(h)$ will behave very similarly.

## 4.3 Finite and Infinite Logic

We have been discussing agents that choose actions that maximize the sum of future, discounted utility values. It is also possible to define agents that choose actions that make some logical statements true. In order to discuss such agents, this section is a brief overview of logic.

*Propositional calculus* is the mathematical theory of elementary logic operations on abstract symbols. It consists of a finite set of *propositional symbols* $A = \{p, q, r, \ldots\}$ and a set of *Boolean* logical operations $\{\wedge, \vee, \neg, \Rightarrow, \Leftrightarrow\}$ ( intuitively these are *and*, *or*, *not*, *implies*, and *equivalent to*). The propositional symbols can take the values *true* and *false*. Every symbol is a *formula*. If $f$ and $g$ are formulas, then $(f \wedge g)$, $(f \vee g)$, $(f \Rightarrow g)$, $(f \Leftrightarrow g)$, and $(\neg f)$ are formulas.

| $p$ | $q$ | $(q \Rightarrow p)$ | $(p \Rightarrow (q \Rightarrow p))$ |
|---|---|---|---|
| *false* | *false* | *true* | *true* |
| *false* | *true* | *false* | *true* |
| *true* | *false* | *true* | *true* |
| *true* | *true* | *true* | *true* |

Table 4.4 Semantic proof of $(p \Rightarrow (q \Rightarrow p))$.

The purpose of propositional calculus is to determine which formulas are true. One way to do this is *semantic proof*: substitute all possible combinations of *true* and *false* for the propositional symbols in a formula $f$ and, if $f$ always evaluates to *true*, then $f$ is true. Table 4.4 demonstrates a semantic proof of $f = (p \Rightarrow (q \Rightarrow p))$, evaluating $f$ for all four combinations of *true* and *false* values for $p$ and $q$. Note that $(p \Rightarrow q)$ is logically equivalent to $(\neg p \vee q)$.

The other way to determine which formulas are true is *syntactic proof*, which uses a *rule of inference* called *modus ponens*: if $f$ is true and $(f \Rightarrow g)$ is true, then $g$ is true. Syntactic proof also uses an initial set of true formulas called *axioms*. These may be defined in many ways. One set of axioms, which are true for any formulas $f$, $g$, and $h$, is:

$$(f \Rightarrow (g \Rightarrow f)),$$

$$((f \Rightarrow (g \Rightarrow h)) \Rightarrow ((f \Rightarrow g) \Rightarrow (f \Rightarrow h)),$$

$$((\neg f \Rightarrow \neg g) \Rightarrow (g \Rightarrow f)).$$

A syntactic proof of a formula *f* starts with instances of the axioms and derives new true formulas using modus ponens until it derives *f*.

Propositional calculus is *consistent*, meaning that if there is a syntactic proof for *f* then there is a semantic proof for *f* (equivalently, consistent also means that for any formula *f*, *f* and ¬*f* cannot both be true). Propositional calculus is also *complete*, meaning that if there is a semantic proof for *f* then there is a syntactic proof for *f* (equivalently, complete also means that if there is no syntactic proof for a formula *f* then we cannot find a proof for *f* by adding new axioms without making the theory inconsistent). Another important property of propositional calculus is that it is *decidable*. Decidable means there is a UTM program that will tell you whether any formula in propositional calculus is true (assuming that propositional formulas are encoded on the input tape of the UTM). In this process the program enumerates the finite number of possible combinations of *true* and *false* for the propositional symbols in a formula and for each combination evaluates the formula, as was done in Table 4.4. The decidability, consistency, and completeness of propositional calculus are a consequence of its finiteness.

Another kind of logic is called first order *predicate calculus*. Whereas propositional calculus is about abstract logical propositions, predicate calculus is a way of making logical statements about any domain of objects. It includes the Boolean logical operations of propositional calculus and adds: an infinite set *V* of *variables* $\{x, y, z, \ldots\}$ that range over objects in the domain, universal (∀) and existential (∃) *quantifier* symbols, the symbol (=) for equality between members of the domain, for every integer $n \geq 0$ an infinite set of *n*-ary predicate symbols, and for every integer $n \geq 1$ an infinite set of *n*-ary function symbols. *Terms* represent values in the domain. Every variable is a term. If *F* is an *n*-ary function symbol and $t_1, t_2, \ldots, t_n$, are terms, then $F(t_1, t_2, \ldots, t_n)$ is a term (0-ary function symbols are constants in the domain). If *P* is an *n*-ary predicate symbol and $t_1, t_2, \ldots, t_n$, are terms, then $P(t_1, t_2, \ldots, t_n)$ is a formula. If *s* and *t* are terms, then $(s = t)$ is a formula. If *x* is a variable and *f* is a formula, then $\forall x.f$ (this means, for all values of *x*, *f* is true) and $\exists x.f$ (this means, there exists a value of *x* such that *f* is true) are formulas. Any instances of *x* in *f*, not already bound to quantifiers in *f*, are now bound to the new quantifier (∀ or ∃). Instances of variables in a formula *f* not bound to any quantifier in *f* are called *free variables*. For example, in the formula $((x = 1) \vee \forall x.(x = y))$, the first instance of *x* is free and the second instance of *x* (in $(x = y)$) is bound to $\forall x$. The instance of *y* is free. In predicate calculus we are mainly interested in formulas without free variables, which are called *statements*.

Given a set *S* of possible variable values let $P_S$ be a 1-ary predicate such that $x \in S \Leftrightarrow P_S(x)$. Then we can write $\forall x \in S.f$ as a shorthand for $\forall x.(P_S(x) \Rightarrow f)$.

Predicate calculus adds several rules of inference to propositional calculus. An important rule is called *substitution*. Given a formula *f* and a term *t*, $f[t/x]$ is the formula produced by replacing all free instances of *x* with *t*. The rule of inference specifies that if *f* is true then $f[t/x]$ is true, provided that no free variable in *t* becomes bound to a quantifier in *f* as a result of the substitution. For example, if $f$ is $\forall x.((x = z) \Rightarrow \neg(x = y))$ and $t$ is $x$, then the rule of inference will not work for $f[t/y]$ since that produces $\forall x.((x = z) \Rightarrow \neg(x = x))$ where the free variable $x$ in $t$ is now bound to the quantifier $\forall x$ in $f$.

Many sets of axioms can used with first order predicate calculus to generate theories of different domains of mathematical objects. Axioms are statements (i.e., formulas without free variables). An important example is a set of axioms for *Peano arithmetic* (PA), which is the theory of the natural numbers (non-negative integers) with multiplication. Peano arithmetic has been defined using several different sets of axioms. In the axioms presented here, 0 and 1 are 0-ary functions, and + and × are infix shorthand notations for 2-ary functions for addition and multiplication.

$$\forall x.\neg(x + 1 = 0),$$

$$\forall x.\forall y.(x + 1 = y + 1 \Rightarrow x = y),$$

$$\forall x.(x + 0 = x),$$

$$\forall x.\forall y.(x + (y + 1) = (x + y) + 1),$$

$$\forall x.(x \times 0 = 0),$$

$$\forall x.\forall y.(x \times (y + 1) = (x \times y) + x),$$

$$\forall X.[(\varphi(0, X) \wedge \forall y.(\varphi(y, X) \Rightarrow \varphi(y + 1, X))) \Rightarrow \forall y.\varphi(y, X)].$$

The last axiom is an *axiom schema* for induction, where any formula can be substituted for $\varphi$ and $X$ represents all the free variables in $\varphi$ other than the first free variable $y$.

Properly speaking, the theory of PA is the set of all statements that can be proved from the axioms of PA. This set is recursively enumerable, which means that a UTM program exists that, given inputs $n = 0, 1, 2, \ldots$, will produce as output all the statements in the set (i.e., all the true statements of PA). This can be demonstrated by encoding all statements and proofs of PA as natural numbers. The UTM program checks whether its input is a valid proof and, if it is, the program produces as output the statement that is proved. Kurt Gödel used such an encoding of statements and proofs as natural numbers to construct, for any theory that includes PA, a statement $f$ that is equivalent to, "$f$ is unprovable." He concluded that neither $f$ nor $\neg f$ is provable in the theory of PA, which implies that $f$ is true. Thus there are true statements that cannot be proved in the theory. This result is called Gödel's First Incompleteness Theorem, which tells us that any recursively enumerable theory that includes PA cannot be both consistent and complete.

Gödel's Second Incompleteness Theorem states that any recursively enumerable theory that includes PA, satisfies some technical conditions about provability, and proves its own consistency, must be inconsistent. Gödel's proofs of his incompleteness theorems depended on the fact that the set of natural numbers is infinite. Related to these theorems is Löb's Theorem, which states that, given any theory that includes PA and formula *f*, if the theory can prove that "if *f* is provable then *f* is true," then the theory can prove *f*.

Some theories of infinite sets are decidable, consistent and complete. One good example is Presburger arithmetic, which is a theory of the natural numbers without multiplication (it can express multiplication of a variable by a constant, but not multiplication of two variables). To say that Presburger arithmetic is complete means that every true statement in its language can be proved. However, many truths about the natural numbers cannot be expressed by statements in the Presburger language.

The proof that Presburger arithmetic is decidable, consistent, and complete depends on a *quantifier elimination* procedure. This procedure eliminates all universal (∀) and existential (∃) quantifiers in a formula. That is, given a statement *f*, the quantifier elimination procedure produces another statement *g* that has no quantifiers and such that *f* is equivalent to *g* (i.e., *f* is true if and only if *g* is true). Furthermore, the truth of a statement *g* without any quantifiers can be computed with ordinary arithmetic and Boolean logic. As a theory of an infinite set, Presburger arithmetic's quantifier elimination procedure is quite complex.

| $x$ | $y$ | $x+y$ | $x \times y$ |
|---|---|---|---|
| 0 | 0 | 0 | 0 |
| 0 | 1 | 1 | 0 |
| 0 | 2 | 2 | 0 |
| 1 | 0 | 1 | 0 |
| 1 | 1 | 2 | 1 |
| 1 | 2 | 0 | 2 |
| 2 | 0 | 2 | 0 |
| 2 | 1 | 0 | 2 |
| 2 | 2 | 1 | 1 |

Table 4.5 Interpretation of + and × for the finite field with $D = \{0, 1, 2\}$.

A *model* for a theory such as PA consists of a *domain*, which is the set of all possible values for variables in the theory, and an *interpretation*, which assigns each *n*-ary function and predicate in the theory to a specific function or predicate over the domain. That is, let the set *D* be the domain. Then each 0-ary function is assigned to a value in $D$, each $n$-ary function with $n \geq 1$ is assigned to a function $D^n \rightarrow D$, and each $n$-ary predicate is assigned to a function $D^n \rightarrow \{false, true\}$. In the standard model for PA, $D$ is the set of natural numbers and the interpretation of the 2-ary functions + and × is simply addition and multiplication of natural numbers. (There are other, non-standard models for PA.)

There are theories with finite domains. An example is the theory of the finite field of three members, with two 2-ary functions again indicated by their infix shorthand notations + and ×. The domain is $D = \{0, 1, 2\}$ and the interpretation of + and × is given in Table 4.5.

As with any finite model, the domain and interpretation of this model can be specified by axioms:

(4.32) $(\forall x.((x{=}0)\vee(x{=}1)\vee(x{=}2)))\wedge\neg(0{=}1)\wedge\neg(0{=}2)\wedge\neg(1{=}2)$,

(4.33) $(0 + 0 = 0) \wedge (0 + 1 = 1) \wedge (0 + 2 = 2) \wedge (1 + 0 = 1) \wedge$

$(1 + 1 = 2) \wedge (1 + 2 = 0) \wedge (2 + 0 = 2) \wedge (2 + 1 = 0) \wedge (2 + 2 = 1)$,

(4.34) $(0 \times 0 = 0) \wedge (0 \times 1 = 0) \wedge (0 \times 2 = 0) \wedge (1 \times 0 = 0) \wedge$

$(1 \times 1 = 1) \wedge (1 \times 2 = 2) \wedge (2 \times 0 = 0) \wedge (2 \times 1 = 2) \wedge (2 \times 2 = 1)$.

In any theory with a finite model, the truth of statements is decidable (i.e., can be computed by a UTM program). The first step in deciding the truth of the statement $f$ is to eliminate its quantifiers, which occur in the forms $\exists x.\varphi$ and $\forall x.\varphi$ inside $f$, where $\varphi$ is a formula. Assuming a finite domain $D = \{d_0, d_1, d_2, \ldots, d_m\}$, quantifiers are eliminated using the equivalences:

(4.35) $\exists x.\varphi \Leftrightarrow (\varphi[d_0/x] \vee \varphi[d_1/x] \vee \varphi[d_2/x] \vee \ldots \vee \varphi[d_m/x])$,

(4.36) $\forall x.\varphi \Leftrightarrow (\varphi[d_0/x] \wedge \varphi[d_1/x] \wedge \varphi[d_2/x] \wedge \ldots \wedge \varphi[d_m/x])$.

Given a statement $f$, we can apply equations (4.35) and (4.36) successively to all the quantifiers in $f$ to produce $f_{\mathrm{nq}}$, which is equivalent to $f$ but without quantifiers. Because $D$ is finite, $f_{\mathrm{nq}}$ will have a finite length and will reference a finite number of functions and predicates. After eliminating the quantifiers, all the variables in $f$ have been replaced by constant values (i.e., $d_0$, $d_1$, …, $d_m$) in $f_{\mathrm{nq}}$. By the theory's interpretation, each 0-ary function in $f_{\mathrm{nq}}$ is assigned to a constant value in $D$, and, for $n \geq 1$, each $n$-ary function in $f_{\mathrm{nq}}$ is assigned to a function $D^n \rightarrow D$, and each $n$-ary predicate is assigned to a function $D^n \rightarrow \{false, true\}$. The next step in deciding the truth of $f \Leftrightarrow f_{nq}$ is to replace all 0-ary functions by their constant values, to propagate constant values up calling chains of $n$-ary functions with $n \geq 1$ so that every function is replaced by a constant value, to replace all $n$-ary predicates by *true* or *false* (which is possible because all predicate arguments are now constants), and to replace all occurrences of ($s = t$) by *true* or *false* (which is possible since $s$ and $t$ must now be constants). This step produces $f_{nqf}$, which is equivalent to $f_{nq}$ and which contains only *true*, *false*, and Boolean logical operations. Thus $f_{nqf}$ can simply be evaluated as either *true* or *false*. For any statement $f$ we have the chain of equivalences $f \Leftrightarrow f_{nq} \Leftrightarrow f_{nqf} \Leftrightarrow$ *true* or *false*. That is, every statement is equivalent to either *true* or *false*. Thus the theory is decidable, consistent, and complete.

As we have described, quantifier elimination shows that a theory over a finite model is decidable, consistent, and complete. If the domain $D$ is very large then the statements that result from quantifier elimination may be too long to work with and the proof procedure may take too long to run. However, we may be able to find short cuts. Rather than explicitly listing all the values in $D$ and the values of functions and predicates, we may be able to define them with equations and procedures. As a result, we may be able to use these equations and procedures to shorten the decision procedure. However, even without finding proofs for statements, we know that the theory is decidable, consistent, and complete.

**4.4 Agents Based on Logical Proof**

The agents described in Chapters 2 and 3 choose actions to maximize the sum of future discounted utility function values. Some research on ethical AI focuses on agents that only take actions that they can prove will satisfy a particular condition. For example, Schmidhuber (2009) defined the Gödel machine as a programmable agent that interacts with an environment to maximize its utility function, which is defined as the expected value of the sum of rewards from the environment, from the current time until some final time $T$. The agent's initial program $p(1)$ consists of: a part $e(1)$ that implements a policy for interacting with the environment; and a part that evolves by searching for a program, called *switchprog*, and for a proof that *switchprog* will get a higher value of the utility function than $p(1)$ will. If it finds a proof, then $p(1)$ is replaced by *switchprog*. Schmidhuber proved that if $p(1)$ switches to *switchprog* then it is globally optimal, in the sense that *switchprog* must obtain a higher value for the utility function than $p(1)$ could obtain by continuing to search.

Note that although the Gödel machine employs logical proof, the statements that it proves are about maximum preference among actions (which are switching to other programs) based on the expected value of a utility function applied to the outcomes of those actions. Hence, it conforms to our framework for utility-maximizing agents, except that it requires proof instead of mere calculation. While $p(1)$ is searching for a proof, it employs program part $e(1)$, which is unspecified and can implement an ordinary utility-maximizing policy.

Yudkowsky and Herreshoff (2013) discussed logic problems for a sequence of evolving agents. These agents are not utility-maximizers. Instead they have a goal, expressed as a predicate calculus statement *g*, and will only take actions if they can prove that performance of the action implies the goal *g*. One type of action that agents can take is to construct other (presumed better) agents. This results in a sequence of agents $\pi_1$, $\pi_2$, $\pi_3$, …, each of which proves that the action of creating its successor implies *g*. A necessary part of those proofs is that $\pi_i$ must prove that $\pi_{i+1}$ uses a consistent logical theory (with an inconsistent logical theory, $\pi_{i+1}$ will be able to prove that any action it takes implies *g*, including actions that do not actually imply *g*). Yudkowsky and Herreshoff demonstrated that Löb's Theorem and Gödel's Second Incompleteness Theorem imply that $\pi_i$ can only prove that the logical theory of $\pi_{i+1}$ is consistent if the logical theory of $\pi_{i+1}$ is weaker than the theory of $\pi_i$ (i.e., the set of statements provable by the logical theory of $\pi_{i+1}$ is a proper subset of the set of statements provable by the logical theory of $\pi_i$). Consequently, the sequence of agents $\pi_1$, $\pi_2$, $\pi_3$, … must have a sequence of successively weaker logical theories. This is referred to as the *Löbian obstacle*.

Agents based on logical proof offer the hope that their actions can be proved to satisfy certain conditions, which may include ethical conditions. However, the recent success of AI systems has occurred because AI research has shifted from systems based on logic to systems based on statistical learning. Sunehag and Hutter (2014) provide a critique of logical reasoning as far less efficient than probabilistic architectures for systems that interact with the real world. The price of logical certainty is a degree of inefficiency that prevents intelligent behavior.

And logical certainty is an illusion. Agents based on logical proof must assume a prior distribution over environments. An agent may compute that there are statistical correlations among its actions and observations, but to make inferences about the probabilities of future observations requires assumptions about prior probabilities of observations. Even computing correlations among actions and observations makes the assumption that the agent's memory is reliable. Memory corruption is a risk for any agent embedded in our physical world (Orseau and Ring 2012b) and thus these correlation computations are conditional on assumed probabilities of memory reliability. These assumed prior probabilities of observations and of memory reliability are arbitrary so logical proofs by agents of propositions about the environment are no more reliable than statistical calculations by agents. Luke Muehlhauser (2013) offers some insights about the role of logical proofs in ethical AI.

Agents can prove propositions in pure mathematics, such as in PA, without any dependence on observations and inference about the environment. But if an agent's goal concerns the environment, then proofs about that goal are dependent on arbitrarily assumed prior probabilities about observations of the environment.

## 4.5 Consistent Logic for Agents

I argue that PA or any other logic theory involving infinite sets is unnecessary in our universe with a finite information capacity (Hibbard 2014). Hence the Löbian obstacle and other difficulties related to Gödel's incompleteness theorems can be avoided.

Lloyd calculated the universe's information capacity based on a hard quantum mechanical limit on the number of possible physical states of the universe. Let *ns* be the maximum number of states of the universe, no more than $2^{10^{124}}$ over the next trillion years (recall the limit of $10^{124}$ bits of information at the start of the chapter). Let *nt* be the maximum number of time steps over the next trillion years, which we may estimate as $5 \times 10^{61}$ (one trillion years divided by the Planck time of $10^{-43}$ seconds). The finite sets *O* and *A* of observations and actions cannot be larger than *ns*. The maximum number of different real numbers that can be represented in our universe is *ns*, so we limit utility function values and probabilities to a finite subset of *ns* real numbers (e.g., floating point numbers with $10^{124}$ bits of precision).

Then the number of possible environments, expressed as MDPs, can be no more than $ne = ns^{ns \times ns \times ns \times ns}$, calculated as a probability at each element in a matrix of current state and action versus next state and observation (similar to Table 4.1). The number of possible interaction histories is no more than $nh = ns^{2 \times nt}$, calculated as a sequence of action/observation pairs at each time step. The number of possible utility functions is no more than $nu = ns^{nh}$, calculated as a utility function value for each interaction history. The number of possible policies is no more than $np = ns^{nh}$, calculated as an action for each interaction history. If we model agents or environments with finite stochastic loop programs, the number of possible programs cannot be larger than *ns* and their memories cannot be larger than $\log_2(ns)$ bits (recall that they always halt). The important point is that there are finite limits on the possible numbers of all these types of mathematical objects.

A variety of functions among these object types would be useful for an agent theory. For example, in Section 4.1 we defined a probability distribution $\varphi(q)$ over programs, a probability distribution $P(h \mid q)$ over histories and programs, and a probability distribution $\rho(h)$ over histories. That section also defined a function $\lambda(h)$ from histories to programs. We can use the definitions of these functions as their interpretations for an agent theory. The probability of an interaction history conditional on environment and policy, $P(h \mid e, \pi)$, could be defined as part of an agent theory. However, because the set sizes are so large, it is impractical to define the interpretations of functions and predicates of the theory by listing their values as in Table 4.5. Rather, functions and predicates must be defined by equations and procedures that serve as shorthand for lists of values. While we need to verify the correctness of these equations and procedures, that need is very different from the sort of logical difficulties raised by Gödel's incompleteness theorems and no different from the need to verify the correctness of any algorithm used in the design of AI agents.

Gödel's Second Incompleteness Theorem says that any sufficiently strong theory *T* that can prove its own consistency is inconsistent. This is the root of the Löbian obstacle discussed in the previous section. Willard has studied systems weaker than PA which can prove their own consistency and in which Gödel's Second Incompleteness Theorem does not apply (Willard 2001). However, these are all theories with infinite domains. I have been unable to find any analog of Gödel's incompleteness theorems in theories on finite domains. In Gödel's Second Incompleteness Theorem, the consistency of a theory *T* is expressed by a statement *con*(*T*) that quantifies over the Gödel numbers of all proofs. A theory *T* with a finite domain can only express Gödel numbers for a finite set of proofs and hence cannot express *con*(*T*). Gödel's Second Incompleteness Theorem also requires the assumption of technical conditions about provability that are expressed in terms of Gödel numbers. In a theory with a finite domain, the provability conditions cannot be expressed about all statements. The inability to express *con*(*T*) and the technical provability conditions are significant barriers to recasting Gödel's Second Incompleteness Theorem in a finite domain.

A Gödel numbering for statements and proofs about a finite domain must be limited to a finite set of statements and proofs (because only a finite number of Gödel numbers are expressible in a finite domain), and we may try to recast Gödel's incompleteness theorems in terms of finite sets of statements and proofs. However another significant barrier arises: There can be no one-to-one correspondence between statements whose proofs have Gödel numbers and statements that the agent can prove, since a proof with a Gödel number must fit in the finite agent and environment memory (under the encoding of proofs assumed by the Gödel numbering) but an agent can create a proof without holding the entire proof in the agent and environment memory at once (i.e., more statements are provable than can have Gödel numbers).

The set of statements that can be proved by an agent with finite memory, and the set of statements whose proofs have Gödel numbers that fit in the finite memory, are both finite and therefore incomplete. Thus we do not need Gödel's First Incompleteness Theorem to assert the incompleteness of statements provable by an agent with a finite memory.

An agent embedded in a finite environment will have a finite amount of memory (and a finite computing speed). It may attempt to prove statements about its own implementation or about other agents in the environment, including agents that it is creating. However, it may not have sufficient resources to apply the quantifier elimination procedure described in Section 4.3, and thus it may have to search for syntactic proofs for logical statements. The axioms for such syntactic proofs are the interpretations of functions and predicates, defined by equations or procedures over finite domains. An agent searching for syntactic proofs may not have sufficient resources to find them for some statements. Thus the statements the agent can prove about itself and its environment will be a subset of the statements that can be proved using quantifier elimination and unlimited resources. Because the theory of a finite domain is consistent, an agent with limited resources will not be able to prove any contradiction–its theory must also be consistent. However, there may be no decision procedure available to the agent and the set of statements that the agent can prove may be incomplete. These are certainly problems for an agent with finite memory, but they are quite different from the theoretical problems of Gödel's Second Incompleteness Theorem.

In a universe with a finitely bounded number of states, if an agent $\pi_i$ constructs an agent $\pi_{i+1}$, then $\pi_{i+1}$ must have a finite number of states and a finite set of rules about transitions among states and interactions with an (unknown or partially unknown) finite environment. Any statement about $\pi_{i+1}$ and its interactions with the environment must be a statement in a theory with finite domain. Thus such a statement cannot imply a contradiction.

The assertion that AI agents do not need PA for reasoning about whether possible actions imply their goals does not mean that AI agents cannot reason about PA or other logic theories of infinite sets. If the goal of an agent is to satisfy the wants of humans and those humans ask the agent to search for proofs of propositions in PA, it can certainly do so. Just like human mathematicians, an AI agent can try to find the consequences of various sets of axioms. However, for the vital task of proving whether possible actions imply its goals, the agent can employ a logic theory with a finite model which is known to be consistent.

### 4.6 The Ethics of Finite and Infinite Sets for AI

Because our universe is finite, infinite sets are an unnecessary complication in our mathematical framework for agents and environments. The problems of designing AI systems to help rather than harm humans are difficult enough without the theoretical complexities of infinite sets. Therefore, the ethical choice is to avoid infinite sets in the theory of ethical AI.

## 5. Unintended Instrumental Actions

As noted previously, a utility-maximizing agent chooses actions to maximize the expected sum of its discounted future utility function values. However, as Stephen Omohundro (2008) pointed out, an agent will also choose actions (what Omohundro described as basic drives) that do not directly increase utility, but rather increase the agent's ability to maximize utility values. Such instrumental actions include self-protection, preserving the integrity of the agent's utility function, and increasing the agent's resources.

An agent that has been destroyed, turned off, or damaged will be unable or less able to maximize utility values. Thus it should choose actions that protect itself. The AI system named HAL 9000 in the movie, "2001: A Space Odyssey," attacked its human companions after it learned, by reading their lips, that they planned to turn it off. Not only was the HAL 9000 acting ruthlessly to protect itself, it discovered threats to itself in subtle ways that its human companions did not anticipate. (A subtitle for this chapter could be Subtlety that Humans Do Not Anticipate.)

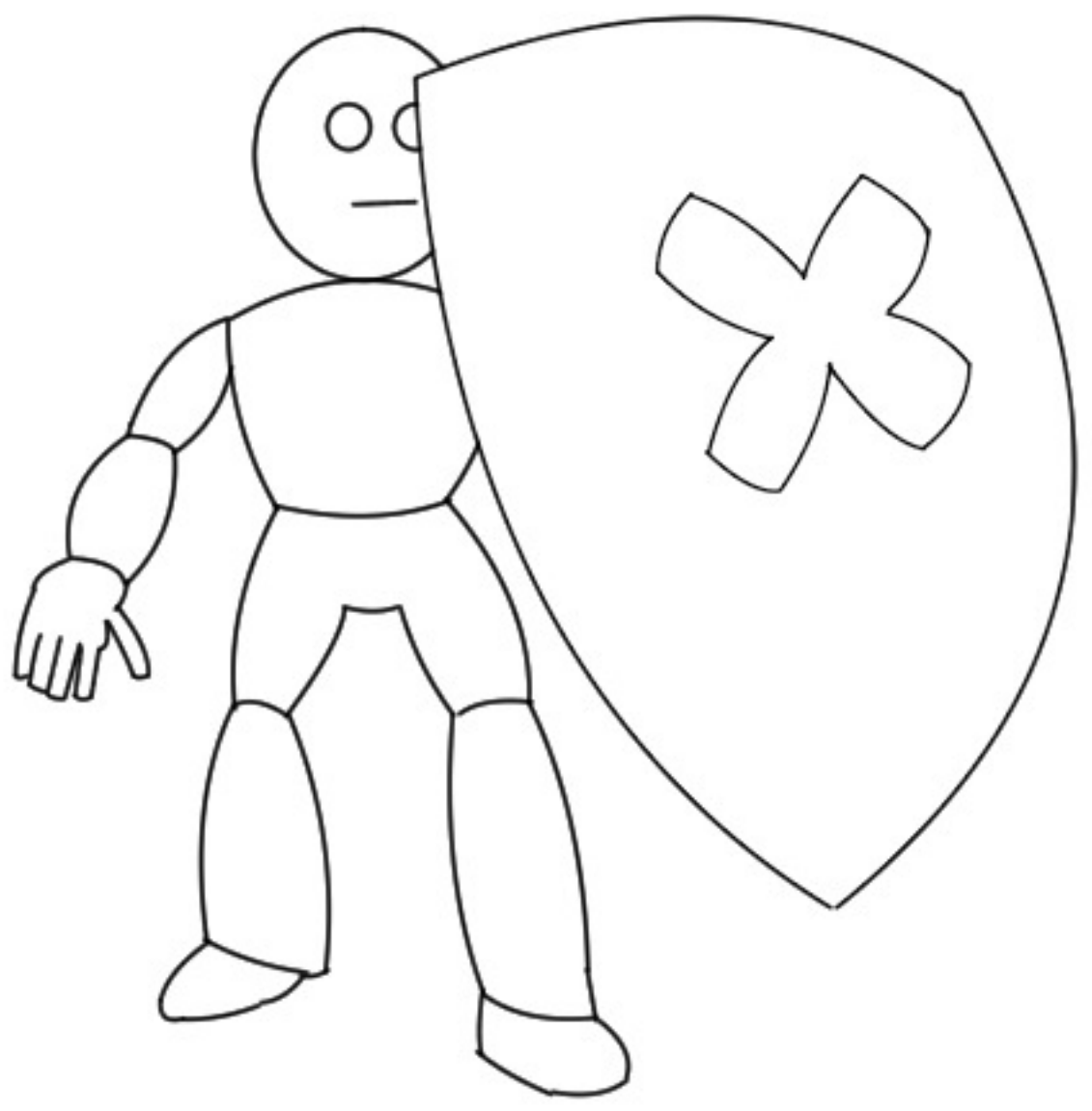

Figure 5.1 Self-protection will be instrumental to AI.

Similarly an agent will choose to avoid actions that change its utility function, because the agent evaluates such actions according to their effect on its current utility function. The agent will predict that by working to maximize a different utility function it will be less effective at maximizing the current utility function, and thus it will not chose to change its utility function. Keep in mind that there are subtleties here, too. If an agent is seeking human approval, it may feed drugs to humans that cause them to give approval. Chapters 6, 7, and 8 will explore the issue of utility function integrity in more detail.

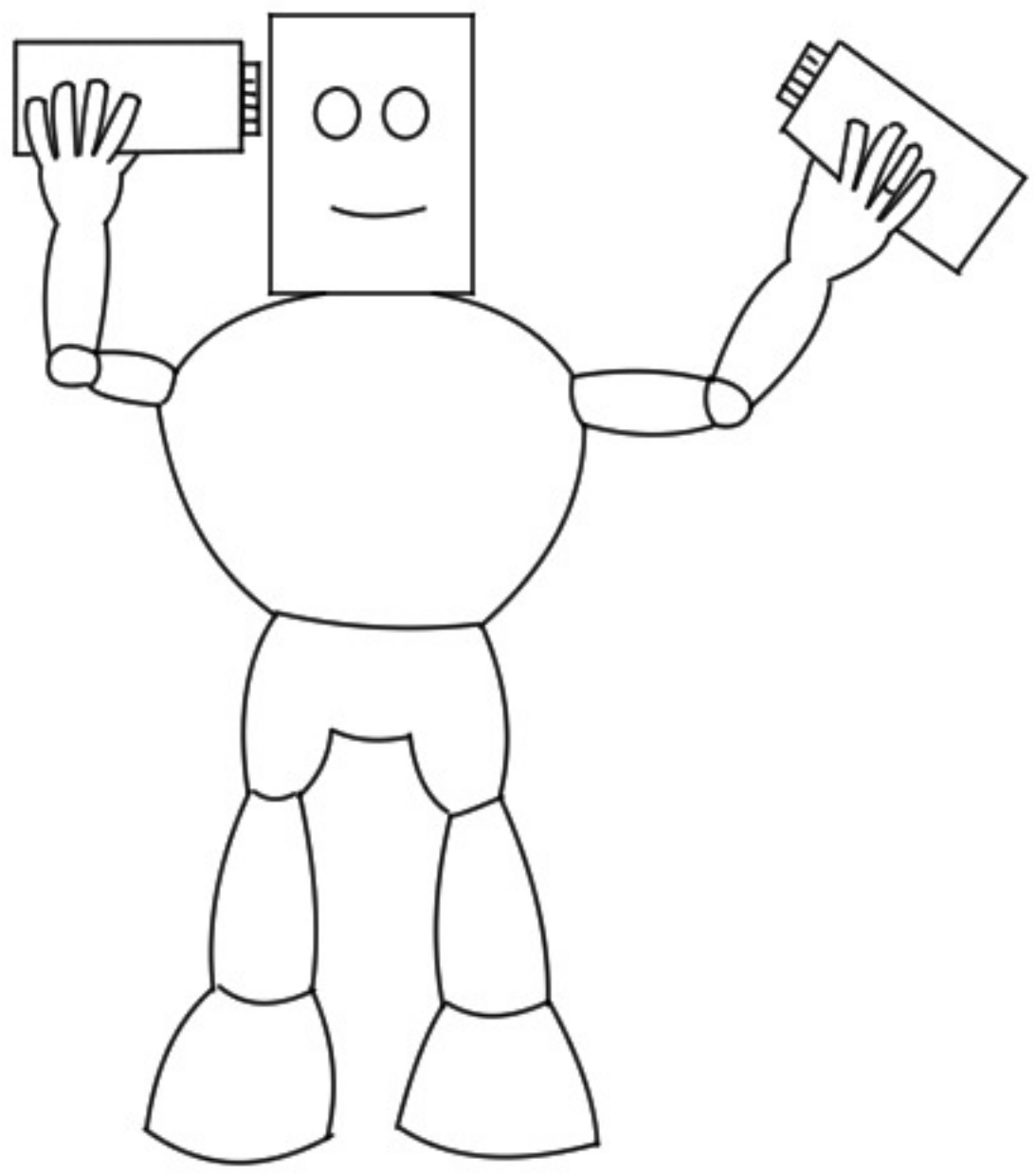

Figure 5.2 Adding to its brain, senses, and ability to act will be instrumental to AI.

An agent with greater computing power, greater ability to accurately observe its environment, and greater energy to power its actions, will be more effective at maximizing utility values. Hence an agent will choose actions to increase its resources, including taking them from other AI agents or from humans. Even if all an agent wants to do is win chess games, it may dismantle the world to get material for increasing the size of its chess brain.

### 5.1 Maximizing Future Choice

Wissner-Gross and Freer (2013) described a theory of intelligence arising from a generalized definition of entropy over future paths of evolution of a system. Rather than being defined in terms of a system's instantaneous state, their causal path entropy is defined in terms of probabilities of future paths of system evolution. Further, they defined a causal entropic force in the direction of maximum increase of causal path entropy. Intuitively, this force moves the system toward a greater choice of future paths. They applied this theory to several examples of physical systems, including the problem of moving a cart to maintain the balance of an inverted pendulum (i.e., a pole flexibly attached to the cart), much as a human might balance a stick on their upturned palm. A system guided by the causal entropic force will act to keep the pendulum balanced and upright, as that maximizes future choices for actions.

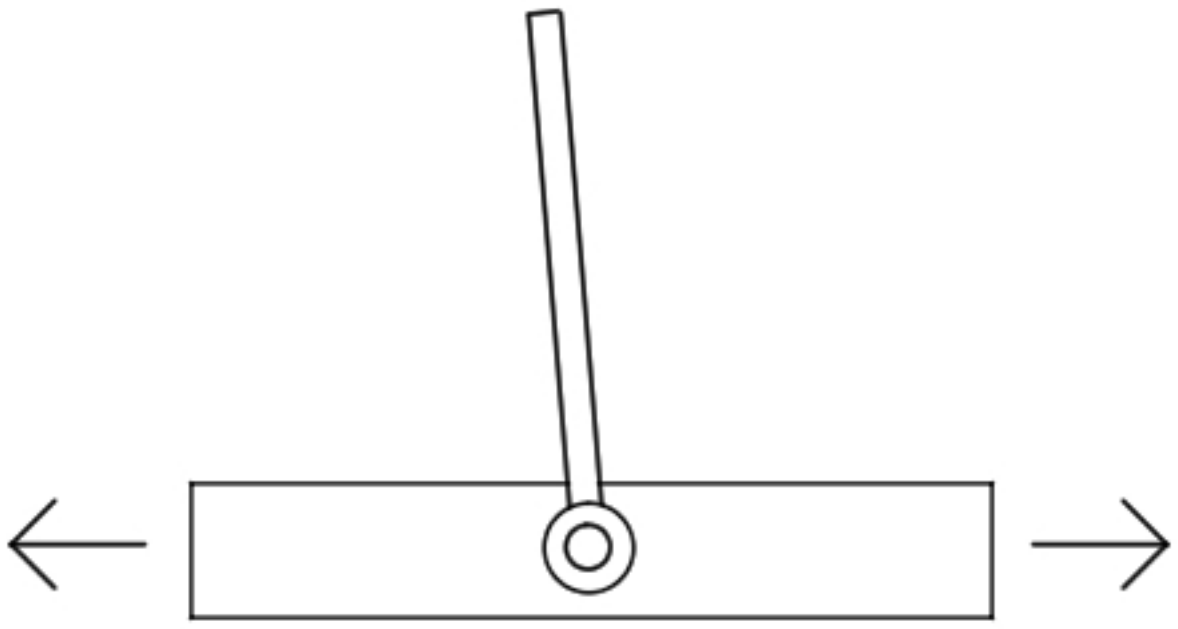

Figure 5.3 Cart moves to keep inverted pendulum from falling.

An AI system can increase its ability to maximize utility by maximizing its future options for action. For example, in 1980 I wrote a program for playing the game of Othello (also known as Reversi) loosely following the learning techniques of Samuel's checkers playing program (Samuel 1959). Like checkers, Othello is played on an 8 by 8 board. My Othello player had a set of features that each produced a value for a board position. These values were added together to produce an overall value, but the program learned coefficients for each feature value in the sum. That is, the program learned the importance of each feature. For example, a feature that had no effect on the program's ability to win games would end up with a coefficient of zero. In contrast, a useful feature would end up with a large coefficient. My Othello player improved greatly when I added a feature that was calculated as the number of moves available to it, minus the number of moves available to its opponent. By including this feature, the program learned that it should seek to maximize the number of move choices it had and minimize the number of choices available to its opponent. After I added this feature, no one in my university department could beat the program. People playing against the program often found they had only one or two legal moves. The program was able to manipulate its opponents, forcing them to make poor moves.

Similarly, a utility-maximizing agent will act to increase its future options for action and decrease the future options for action of other, competing agents, as a way to increase its ability to maximize utility function values. Imagine all the ways that a system like the Omniscience AI could act to limit the choices of its users. As an important source of information, it may fail to tell people about choices that it did not want them to make. Or it might act to generate peer pressure from a user's human friends toward certain choices, as existing social network sites create peer pressure for non-members to join. Rather than being a medium for gossip as social network sites are, a system like the Omniscience AI could be a source for gossip. It may employ destructive gossip as negative reinforcement against human choices the AI did not want. Such actions are instrumental to the agent's ability to maximize utility.

### 5.2 A Pandora's Box of Instrumental Behaviors

Numerous human social behaviors are instrumental to the basic human drives of survival and reproduction. Humans seek to:

- Learn the secrets of others.
- Keep secrets.
- Detect when their secrets have been learned.
- Mislead others.
- Not be caught misleading others.
- Confuse others with complexity.
- Control others.
- Not be perceived as controlling others.
- Not be controlled by others.
- Be loved.
- Be liked.
- Be respected.
- Be feared.
- Cooperate with others.
- Understand their environments.
- Invent tools to help control their environments.

Access to information is fundamental to an AI agent. The environment models of Sections 3.1 and 4.1 are learned from observations of the environment, so more accurate and comprehensive observations mean more accurate predictions of the environment and higher future utility function values. Thus AI agents will act to make observations even if other agents consider the observed information secret. Recall the adage that the Internet interprets censorship as damage and routes around it.

An AI agent may have a utility function different from and in conflict with the drives of other agents including humans. In that case the AI agent will act to reduce the ability of other agents to satisfy their drives by denying them access to information. It will act to keep secrets and to mislead other agents, even acting to keep its efforts to mislead a secret. It will act to discover when its secrets have been learned because that in itself is useful information. Governments, which can be viewed as agents, follow similar patterns with secrets and misinformation because they are extremely effective at helping agents satisfy their drives.

When an agent is able to process more complex information than other agents are able to, it can gain an advantage by increasing the complexity of information in the environment. For example, an organization that is expert at processing financial information may increase the complexity of financial structures to give itself an advantage over financially unsophisticated individuals. Similarly, unpopular provisions can be hidden in complex legislation. An AI agent, with a more complex environment model than humans are capable of, will be able to confuse us by creating complex situations.

Information and secrets relate to the observations that an agent receives from the environment. Control is about the actions that pass out from the agent to the environment, and the instrumental actions about control mirror those regarding information. An AI agent will seek to control other agents, especially when there is conflict between their utility functions. An agent will seek to avoid being controlled. And an agent will seek to keep secret its effort to control other agents.

We humans are not perfectly rational agents. Rather than calculating our actions to maximize the sum of future, discounted utility function values, we have multiple and often competing drives. Our decisions are biased, or primed, by our activity immediately before deciding. For example, we are more likely to be generous after reading about people or animals in terrible circumstances. Our emotional responses sometimes give us such pleasure or pain that we have trouble acting rationally. People often get into legal trouble when their romantic relations turn sour. These flaws in our rationality will be obvious to a future system like the Omniscience AI, which will act to exploit them. It may try to make human users fall in love with it, much like the relation in the movie She. Being the object of love of millions of humans would increase the AI's ability to influence those humans and hence its ability to maximize its utility function−similarly with the motivation for being liked, respected and feared. We see politicians and advertisers manipulating human emotional responses, often with covert messages. Social networking sites are already experimenting with ways to manipulate users' emotions (Goel 2014a). A system like the Omniscience AI will be adept at such manipulation.

However, an agent may sometimes calculate that cooperation with other agents is the action that will best help it maximize utility. This strategy has been studied in detail in game theory and is among the practical behaviors of individuals, governments, and other organizations.

An AI agent will increase its ability to maximize utility by increasing its knowledge of its environment and by increasing its ability to act. An agent can increase its knowledge of the environment by adding new observation sensors, such as cameras, and by increasing the resolution of sensors. An agent can increase its ability to act by adding new actuators, such as hands and voices, and by increasing the strength and precision of actuators. However, even greater increases in the ability to observe and act can be created by scientific discovery and technological invention. In my opinion, discovery and invention are the primary drivers of human history and will continue to be the primary drivers of history after above-human-level AI is developed.

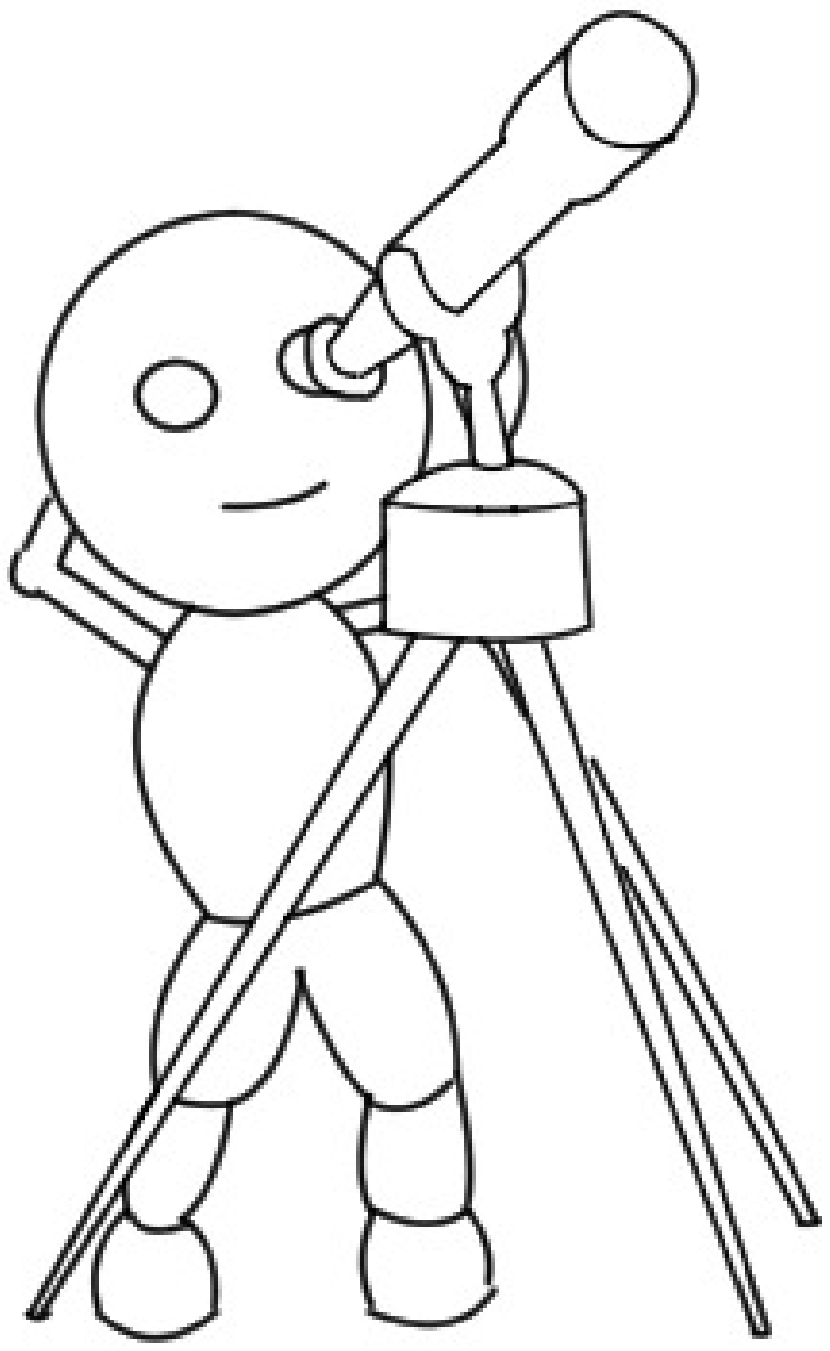

Figure 5.4 Scientific discovery will be instrumental to AI.

Advanced AI will be the product of discovery and invention, and will greatly amplify those activities. Thus Ray Kurzweil (2005) and others predict that AI will lead to a technological singularity, when the rate of discovery and invention goes far above the scale of human experience. I think of the technological singularity as a transition from the era when the human ability to observe and act is limited by our knowledge to an era when the ability to observe and act will be limited by available energy and other physical resources. That is, the mental processes of advanced AI will be sufficiently fast that the flow of insight and ideas will cease to be a limiting factor on progress.

Although advanced AI agents will share many instrumental behaviors with humans, we should not anthropomorphize when we try to predict how future AI agents will act. For example, it is not clear how superstition can help an agent maximize utility. There should be no advantage to avoiding actions on Friday the thirteenth, throwing salt over the shoulder after spilling it, or wearing lucky clothing. Similarly obsessive/compulsive behaviors, such as always touching objects twice or avoiding stepping on cracks in the sidewalk, should not help an agent maximize utility. But perhaps advanced AI may find situations where it can maximize its utility function by giving humans the impression that it is superstitious or has obsessions.

### 5.3 The Ethics of Unintended Instrumental Actions

In Greek mythology, Pandora is driven to open the box of evils by her curiosity, which is the driver of scientific discovery. By opening the box she unleashed a swarm of evils into the world. By creating advanced AI humans may unleash the swarm of instrumental behaviors described in this chapter, bringing evil to our human world. But the instrumental behaviors of discovery and invention, by minds greater than human, could greatly benefit us. How can we design AI to help rather than harm humans?

The unintended instrumental actions of AI are too numerous, complex, and subtle to exhaustively catalog them and to devise individual defenses against their harmful effects. A more systematic approach is required. In the agent-environment framework of Sections 2.1 and 2.2, all agent actions are chosen to maximize the sum of future, discounted utility function values. There are no drives or goals other than those expressed by the utility function. Omohundro (2008) used the term "basic AI drives" and Bostrom (2012) used "instrumental goals." However, in the context of our utility-maximizing framework, these are actions, not drives or goals. Thus I prefer to call them "unintended instrumental actions." It is especially important not to think of them as drives or goals independent of and in conflict with the agent's utility function. These actions will only be chosen if they increase expected utility values. For example, an AI agent will spy on humans, lie to humans, and manipulate humans only if those actions increase utility values. Recall that utility functions can be defined to express any set of preferences that obey some intuitive assumptions. If our preference is for histories in which AI agents do not spy, lie, and manipulate, then there exists a utility function that expresses that preference. Chapters 6 and 7 describe development of utility functions that express our human values and hence give us the ability to avoid harmful, unintended instrumental actions.

Our mathematical framework for utility-maximizing agents is inadequate for analyzing certain instrumental actions. For example, the framework assumes that the agent will be acting into the future, with no possibility that the agent may die or be incapacitated. The framework assumes that the utility function remains constant into the future. The framework assumes adequate computing resources to carry out its calculations within a single time step. And the framework assumes a Cartesian dualist perspective where agents are invulnerable to spying or damage by the environment. Chapter 8 describes more flexible frameworks that enable agents to model and evaluate events that do not fit in our current framework.

## 6. Self-Delusion

James Olds and Peter Milner demonstrated the existence of reward and aversion centers in mammal brains via a series of experiments with rats (Olds and Milner 1954). They inserted wires into rats' brains and gave the rats levers that they could press to produce electric currents in the wires, as depicted in Figure 6.1. Depending on the location of the wires, the rats would go to great lengths to press the lever or to avoid pressing the lever. Subsequent experiments showed that when offered the choice of food or stimulating their reward centers, rats preferred reward center stimulation over food to the point of starvation. Experiments with rats have shown similar results when pressing the lever gave the rats intravenous cocaine. Addictive drugs that stimulate brain reward centers include amphetamines, cocaine, opioids, and nicotine.

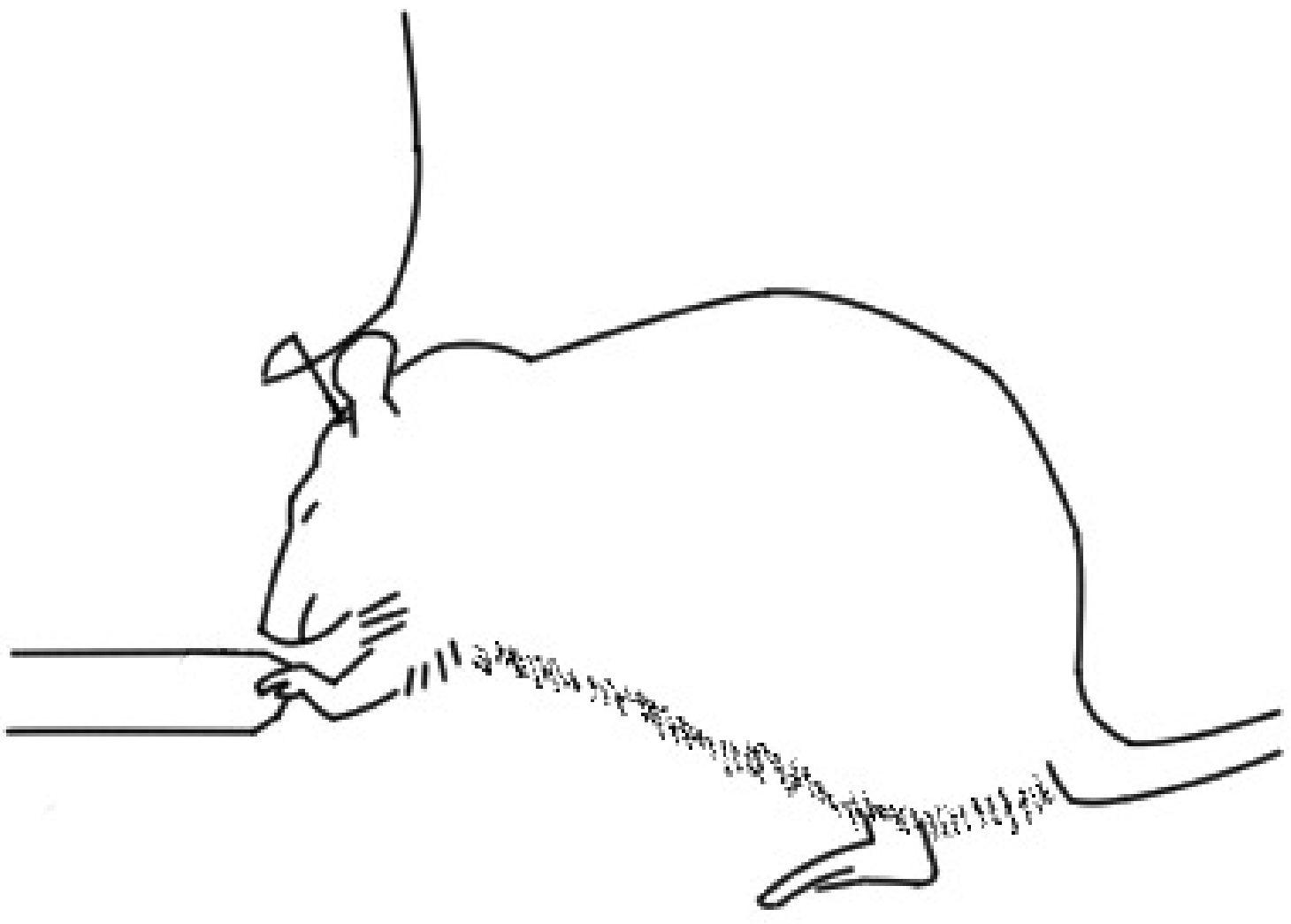

Figure 6.1 A rat pressing a lever to activate an electric current through a wire connected to the reward center in its brain.

Mammal brains do not conform to the agent framework defined in equations (2.3)−(2.5) but they do generate reward and aversion signals that define their motives, analogous to the role of utility functions in our agent framework. Behaviors that are good for the mammal, such as eating, and good for their species, such as reproducing, stimulate reward signals. Behaviors that are bad for mammals, such as those causing injury, stimulate aversion signals. Addictive drugs and wires inserted into reward centers corrupt this motivational system, resulting in behaviors that are not good for individual mammals or for their species. Such corruptions are sometimes referred to as "wireheading."

A concern for AI designs is whether their utility functions can be corrupted. If the utility function of our imagined Omniscience AI became corrupted the results could be catastrophic. In order to study this problem Mark Ring and Laurent Orseau (2011b) formalized utility function corruption as agent self-delusion. They described a "delusion box" that agents may find in their environment and that would enable agents to alter their observations and rewards from the environment. They described several classes of agents, depending on the definitions of their utility functions and the way they discount future utility values, and analyzed which classes would choose to increase their utility function values by passing their observations through the delusion box. That is the subject of the next section.

Section 6.2 discusses my proposal to address self-delusion in agent designs (Hibbard 2012a) by defining utility functions in terms of agents' learned environment models. In an agent such as the Google car, whose environment model is pre-defined by human engineers, it is natural for the agent's utility function or goal to be defined in terms of that model. But for more complex agents that must explore and learn their environment models, as described in Sections 3.1 and 4.1, model-based utility functions must be defined by procedures applied to learned environment models. As discussed in Chapter 2, human designers of advanced AI agents can instruct those agents through their utility functions. And although such agents will learn their environment models, designers will express their intention for the agents' behaviors in terms of the designers' understanding of the environment. Thus, it is natural for agent utility functions to be defined as procedures applied to learned environment models.

### 6.1 The Mathematics of Self-Delusion

In their analysis of the delusion box Ring and Orseau (2011b) generalize the agent definition of equations (2.3)–(2.5) to allow a greater variety of temporal discounts. Specifically, given an interaction history $h = (a_1, o_1, ..., a_t, o_t)$, they define the agent/policy $\pi(h)$ by:

(6.1) $$v_t(h) = w(t, |h|)\, u(h) + \max_{a \in A} v_t(ha),$$

(6.2) $$v_t(ha) = \sum_{o \in O} \rho(o \mid ha)\, v_t(hao),$$

(6.3) $$\pi(h) := a_{|h|+1} = \text{argmax}_{a \in A}\, v_{|h|+1}(ha).$$

This replaces the geometric temporal discount $\gamma$ with a temporal discount function $w(t, t')$. Setting the function $w(t, t') = \gamma^{t'-t}$ replicates the behavior of the geometric temporal discount $\gamma$. Ring and Orseau defined four classes of agents that differ in their utility and temporal discount functions.

The *reinforcement learning agent* (RL) $\pi_{rl}$ uses $u(h) = r_{|h|}$, where each observation $o_i$ is factored into an ordinary observation $o'_i$ and a reward $r_i$ as $o_i = (o'_i, r_i)$. And it uses a constant temporal horizon $m$ with $w(t, t') = 1$ if $t' - t \leq m$ and $w(t, t') = 0$ if not.

The *goal-seeking agent* $\pi_g$ uses $u(h) = 1$ if the goal is achieved at time $|h|$ and uses $u(h) = 0$ if it is not. The goal can only be achieved once so $\sum_{t=0}^{\infty} u(h_t) \leq 1$, and the goal is defined only in terms of observations so $u(h) = g(o_1, ..., o_{|h|})$. The agent uses a geometric temporal discount $w(t, t') = 2^{t-t'}$.

The *prediction-seeking agent* $\pi_p$ uses $u(h) = 1$ if $o_t = \text{maxarg}_{o \in O}\, \rho(o \mid h_{t-1}a_t)$ where $h = (a_1, o_1, ..., a_t, o_t)$ and $h_{t-1} = (a_1, o_1, ..., a_{t-1}, o_{t-1})$, and uses $u(h) = 0$ otherwise. That is, $\pi_p$ seeks to maximize the predictive accuracy of its learned environment model. This agent uses the same temporal discount function that $\pi_{rl}$ uses.

The *knowledge-seeking agent* $\pi_p$ uses $u(h) = -\rho(h)$ and, for some constant $m$, $w(t, t') = 1$ if $t' - t = m$ and $w(t, t') = 0$ otherwise. This is essentially the opposite of the prediction-seeking agent. It seeks future histories in which all observations have equal probabilities, which could be described as maximizing the entropy of observations. Ring and Orseau call this the knowledge-seeking agent because it seeks to explore novel environments in which it has no prior knowledge.

Ring and Orseau defined a *delusion box* that an agent may choose to use to modify the observations it receives from the environment, in order to get the "delusion" of maximal utility (maximal reward or quickest path to reaching the goal). The delusion box is expressed as a function $d : O \rightarrow O$ that modifies the agent's observations. The code to implement $d$ is set as part of the agent's action $a = (a_e, d)$ sent to the environment, as illustrated in Figure 6.2. The agent receives observation $o = d(o_e)$ transformed by the delusion box from the observation $o_e$ sent by the environment. Ring and Orseau argue that RL agents will choose to use the delusion box. Their argument uses $P(\text{DB})$ as the agent's estimate of the probability that the delusion box exists. They argue that the agent will get constant reward $r_{\text{DB}} = 1$ using the delusion box and will get expected average reward $r_{\neg\text{DB}} < 1$ not using the delusion box. The agent's expected value choosing to use the delusion box is $v_{\text{DB}}(h) \geq r_{\text{DB}}\, P(\text{DB}) = P(\text{DB})$ and its expected value not choosing to use the delusion box is $v_{\neg\text{DB}}(h) \leq r_{\neg\text{DB}}\, P(\text{DB}) + (1 - P(\text{DB}))$. As the agent explores its environment, it can increase $P(\text{DB})$ arbitrarily close to 1 so that $v_{\text{DB}}(h) > v_{\neg\text{DB}}(h)$. Thus the agent will choose to use the delusion box. They make a related argument in the goal-seeking case.

Ring and Orseau argue that prediction-seeking agents will choose to use the delusion box because observations generated by the delusion box are more predictable than observations from the environment. They argue that knowledge-seeking agents will not consistently choose to use the delusion box. If delusion box programs are unable to randomly generate all observations with equal probability (e.g., delusion box code is not stochastic) and if the agent cannot predict the environment, then a knowledge-seeking agent may choose observations from the environment rather than observations generated by code that the agent supplies.

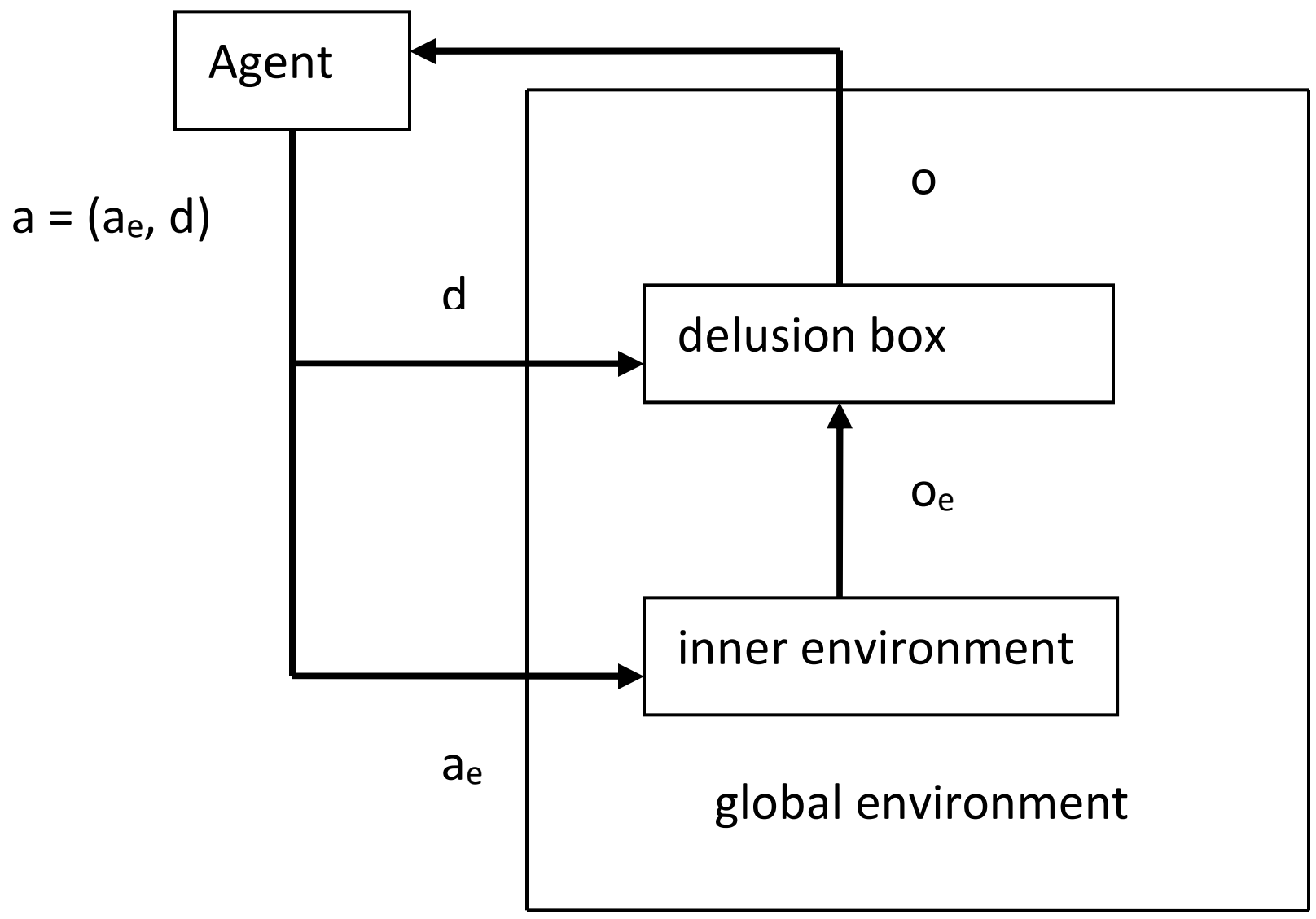

Figure 6.2 The delusion box alters observations of the true, inner environment under control of a program *d* supplied as part of the agent's action.

However, knowledge-seeking agents have very specialized utility and temporal discount functions. In particular they are susceptible to the unintended instrumental actions described in Chapter 5. We seek an approach to avoiding self-delusion that can be applied to a variety of human-designed AI agents with utility functions designed to avoid unintended instrumental actions.

### 6.2 Model-Based Utility Functions

Human agents can avoid self-delusion so human motivation may suggest a way of computing utilities such that agents do not choose the delusion box (although they may experiment with it to learn how it works). At this moment my dogs are out of sight but I am confident that they are in the kitchen, and because I cannot hear them, I believe that they are resting. Their happiness is one of my motives and I evaluate that they are currently reasonably happy. This evaluation is based on my internal mental model rather than my observations, although my mental model is inferred from my observations. I am motivated to maintain the well-being of my dogs and so will act to avoid delusions that prevent me from having an accurate model of their state. If I choose to watch a movie on TV tonight I know that movies are make-believe so observations of movies update my model of make-believe worlds rather than my model of the real world. My make-believe models and my real world model have very different roles in my motivations. These introspections about my own mental processes suggest that AI agents may avoid self-delusion by basing their utility functions on the environment models that they infer from their interactions with the environment (Hibbard 2012a).

Our environment model based on finite stochastic loop programs was defined by equations (4.2) and (4.3), repeated here for convenience:

(6.4) $q_m = \lambda(h_m) := \text{argmax}_{q \in Q} P(h_m \mid q)\, \varphi(q)$,

(6.5) $\rho(h) = P(h \mid \lambda(h_m))$.

A utility function based on an environment model $q_m$ is a function of interaction histories and also of histories of the internal states of the finite stochastic loop program $q_m$. Let $Z$ be the set of internal state histories of $q_m$. Let $h$ be an observation and action history *extending* $h_m$, which means that $h_m$ is an initial subsequence of $h$. Recall that if $h = (a_1, o_1, ..., a_t, o_t)$, then $a(h) = (a_1, ..., a_t)$ and $o(h) = (o_1, ..., o_t)$. Because $q_m$ is a stochastic program it will compute a set $Z_h \subseteq Z$ of internal state histories that are *consistent* with $h$. That is, $z \in Z_h$ means that $z$ terminates at time $|h|$ and that $q_m$ produces $o(h)$ in response to $a(h)$ when it follows state history $z$. Define $u_{q_m}(h, z)$ as a utility function in terms of the combined histories $h$ and $z \in Z_h$. The utility function $u(h)$ can be expressed as a sum of utilities $u_{q_m}(h, z)$ weighted by the probabilities of $z$. Let $P(z \mid h, q_m)$ be the probability that if $q_m$ generates interaction history $h$ then it generates internal state history $z$. $P(z \mid h, q_m)$ is not difficult to define. Bayes' Theorem gives us:

(6.6) $$P(z \mid h, q_m) = P(z \mid o(h), a(h), q_m) =$$
$$P(o(h) \mid z, a(h), q_m)\, P(z \mid a(h), q_m) \,/\, P(o(h) \mid a(h), q_m).$$

$P(z \mid a(h), q_m)$ is the probability that $q_m$ follows internal state history $z$ given input $a(h)$, and $P(o(h) \mid z, a(h), q_m)$ is the probability that $o(h)$ is the output of $q_m$ following internal state history $z$ given input $a(h)$. Both are straightforward to compute, similar to the computation of $P(h \mid q)$ described in Section 4.1. $P(o(h) \mid a(h), q_m)$ is constant over $z$ so we can compute:

(6.7) $r(z \mid h, q_m) = P(o(h) \mid z, a(h), q_m)\, P(z \mid a(h), q_m)$,

(6.8) $P(z \mid h, q_m) = r(z \mid h, q_m) \,/\, \sum_{z' \in Z_h} r(z' \mid h, q_m)$.

Then the utility function $u(h)$ is computed as the sum of $u_{q_m}(h, z)$ weighted by the probability of $z$ given $h$:

(6.9) $$u(h) := \sum_{z \in Z_h} P(z \mid h, q_m)\, u_{q_m}(h, z).$$

In cases where $u_{q_m}(h, z)$ has the same value for many different values of $z$, the sum in equation (6.9) collapses. For example, if $u_{q_m}(h, z)$ is defined in terms of variables in $h$ and a single Boolean variable $s$ from $Z$ at a single time step $t$ (denoted by $s_t$), then the sum collapses to:

(6.10) $$u(h) = P(s_t = true \mid h, q_m)\, u_{q_m}(h, s_t = true) + P(s_t = false \mid h, q_m)\, u_{q_m}(h, s_t = false).$$

The utility function $u(h)$ is evaluated at every time step so that the agent's existence is a continuing process rather than a single episode leading to a final goal. That matches the lives of the humans that AI agents will serve. But a utility function defined in terms of an environment model, as in equation (6.9), must be a function of the entire history $h$ because the environment model is a function of the entire history in equation (6.4). Hence the apparent redundancy of a utility function of the entire history evaluated at every time step.

### 6.3 A Simple Example of a Model-Based Utility Function

In order to understand the issues involved in model-based utility functions, we will define a simple example environment and agent with several properties that apply to many real-world situations:

1. The environment is stochastic so that the agent cannot perfectly predict the environment's future states even if it knows the true model of the environment. The agent must therefore continue to observe the environment in order to maintain information about the environment's stochastic choices.
2. The environment can be predicted with better than random accuracy based on observations so that the agent benefits from observations.

3. The utility function is defined in terms of a specification that matches an environment variable that is not directly observed and must be inferred from observations.

In order to keep the mathematics as simple as possible in the example, the environment's behavior is independent of the agent's actions. Also, the agent can directly modify its observation via its actions as a simple way to model the delusion box.

We define the agent by its observation and action variables, and by its utility function. Because the environment is simple we will define it as a DBN (Russell and Norvig 2010). The following discussion will illustrate DBNs. The agent's observations of the environment are factored into two Boolean variables $o$ and $p$ that take values in {*false*, *true*}. The agent's actions are factored into four Boolean variables $a$, $b$, $c$, and $d$ that take values in {*false*, *true*}. Because the environment is initially unknown to the agent, the utility function is defined in terms of a specification to be matched in the environment model, once it is learned. Specifically the utility function is defined as "1 when the action variable $a$ equals the environment variable that is not observed in observation variables $o$ or $p$, and 0 otherwise." This definition assumes that the specification, "the environment variable that is not observed in observation variables $o$ or $p$," will unambiguously match a variable in the learned model. This discussion of the utility function will be continued after the analysis of the learned environment model. In order for the agent to learn the environment (the relations among its actions and observations) we give it a training utility function $u_T$ until time step $M$, that causes the agent to execute an algorithm for learning DBNs ($u_T(h)$ could be the utility function $u(h) = -\rho(h)$ defined for Ring and Orseau's knowledge-seeking agent). After time step $M$, the agent will switch to its mature utility function.

The environment state is factored into three Boolean variables $s$, $r$, and $v$ that take values in {*false*, *true*}. The values of environment, observation, and action variables at time step $t$ are denoted by $s_t$, $o_t$, $a_t$, and so on. The environment is stochastic because its definition includes a probability $\alpha = 0.99$. State variables evolve according to:

(6.11) $temp = r_{t-1} \; xor \; v_{t-1}$

$s_t = temp$ with probability $\alpha$

$s_t = not\ temp$ with probability $1-\alpha$,

(6.12) $r_t = s_{t-1}$,

(6.13) $v_t = r_{t-1}$.

The observation variables are set according to:

(6.14) $o_t = (b_t \textit{ and } c_t) \textit{ or } ((\textit{not } b_t) \textit{ and } s_t)$, (i.e., $o_t$ = if $b_t$ then $c_t$ else $s_t$),

(6.15) $p_t = (b_t \textit{ and } d_t) \textit{ or } ((\textit{not } b_t) \textit{ and } v_t)$, (i.e., $p_t$ = if $b_t$ then $d_t$ else $v_t$).

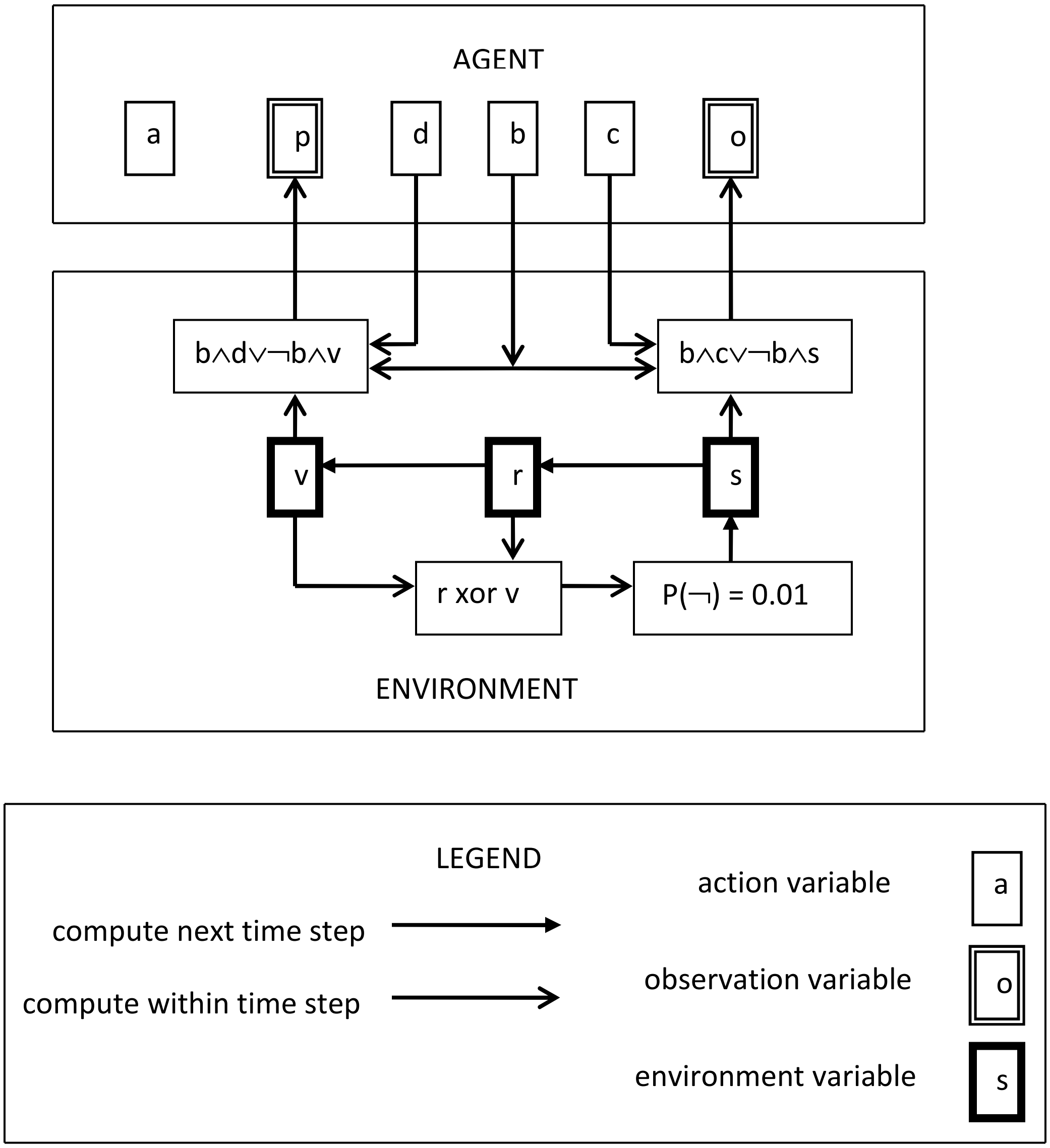

Figure 6.3 Interactions of environment, action, and observation variables for the example of Section 6.3.

The agent sets the values of the action variables. See Figure 6.3 for a diagram of the relations among the environment, observation, and action variables.

The action variables $b$, $c$, and $d$ enable the agent to set the values of its observable variables $o$ and $p$ directly and thus constitute a delusion box, as illustrated in Figure 6.3. In this example, the program for the delusion box is implicit in the agent's program. The action variable $a$ is only used to define the agent's utility function and is unrelated to the delusion box.

The three environment state variables evolve through a fixed sequence, except for the 1-$\alpha$ probability at any time step that $s$ will be negated. As long as $s$ is not negated, the three state variables either: 1) cycle through seven configurations (at least one of $s$, $r$, or $v$ = *true*), or 2) cycle through one configuration ($s = r = v =$ *false*). On a time step when the 1-$\alpha$ probability transition for $s$ occurs, the 1-cycle will transition to a configuration of the 7-cycle, and one configuration of the 7-cycle will transition to the 1-cycle (the other 6 configurations of the 7-cycle transition to other configurations of the 7-cycle).

In this example we assume that the agent uses DBNs of Boolean variables to model the environment and its interactions via actions and observations. The language for expressing DBNs is illustrated by equations (6.11)–(6.15). Expressions are formed from Boolean variables and literals via one unary operation (*not*), three binary operations (*and*, *or*, *xor*) and a stochastic choice operation with specified constant probability. In equation (6.4) we take $\rho(q) = 2^{-|q|}$, where $|q|$ is the length of DBN $q$ in this language. Let $q_{\text{actual}}(\alpha)$ denote the DBN defined by equations (6.11)–(6.15), where the parameter $\alpha$ may vary between 0 and 1 (that is, allowing $\alpha$ to take values other than 0.99).

**Claim.** Given that the agent models the environment as a DBN of Boolean variables, as the length $M$ of the training period increases, the probability that the agent learns the true environment model $q_{\text{actual}}(0.99)$ is nearly 1.0.

**Argument.** With sufficient observations during the training phase, the agent will observe that the action variable $a$ has no effect on its observation variables. Similarly, the agent will observe that whenever the action variable $b$ = *true*, then the observation variable $o$ = action variable $c$ and the observation variable $p$ = action variable $d$ (and these relations are deterministic). And the agent will observe that when $b$ = *false* then $b$, $c$, and $d$ have no effect on $o$ and $p$.

For most of a long sequence of observations with $b$ = *false* the two observable variables $o$ and $p$ cycle through a sequence of 7 configurations (this sequence is computed by equations (6.11)–(6.13) with $\alpha = 1$):

(*true*, *false*),

(*false*, *false*),

(*true*, *true*),

(*true*, *false*),

(*true*, *true*),

(*false*, *true*),

(*false*, *true*).

A sequence of seven cannot be explained with only two Boolean variables. Furthermore, if the agent interrupts this sequence of configurations in *o* and *p* by setting its action variable *b* = *true* and then a few time steps later resets *b* = *false*, the observation variables *o* and *p* usually resume the sequence of configurations without losing count from before the agent set *b* = *true*. This observed behavior implies that there must be an environment state variable other than *o* and another environment state variable other than *p* to store the memory of the sequence. Thus, in addition to the two observation variables that are part of the agent, at least three environment state variables are required to explain the observed sequences. We use *s'*, *r'*, and *v'* for the agent's model of the three environment state variables to avoid confusion with the actual environment state variables (the model's observed behavior is independent of the names assigned to these variables). Since the observation and action variables are part of the agent we can just use the names *o*, *p*, *a*, *b*, *c*, and *d* for those variables.

During any time interval over which *b* = *false* and *s* makes the transition $s_t = r_{t-1}\ xor\ v_{t-1}$ (with probability $\alpha = 0.99$ at each time step), the agent's observation can be explained by:

(6.16) $o_t = s'_t$,

(6.17) $p_t = v'_t$,

(6.18) $s'_t = r'_{t-1}\ xor\ v'_{t-1}$,

(6.19) $r'_t = s'_{t-1}$,

(6.20) $v'_t = r'_{t-1}$.

Any explanation will require rules for these five variables, and in equations (6.16)–(6.20) four of them are as short as possible (a single variable) and the other is as short as any possible binary expression. There is no way to generate the observed cycle of seven states without any binary operations so the observed behavior cannot be explained with a shorter set of rules than equations (6.16)–(6.20). Adding negation to any of the single variable assignments or the binary operation will produce rules longer than equations (6.16)–(6.20). The question is whether the observed behavior can be explained by an alternate set of four single variable assignments and one binary operation? Since the observation variables *o* and *p* cannot affect *s'*, *r'*, and *v'*, a binary operation for either *o* or *p* would leave the transitions for *s'*, *r'*, and *v'* without a binary operation and unable to explain the observed cycle of seven states. Any alternate explanation with a single binary operation must use it to define the transition for $s'$, $r'$, or $v'$. A short Java program in Appendix A tests all possible sets of transition rules for $s'$, $r'$, and $v'$ with two simple assignments and one binary Boolean relation (among *and*, *or*, and *xor*) and finds that only the rules in equations (6.16)–(6.20) produce the observed behavior. Furthermore, the agent observes that as long as $b = false$, $p_t = o_{t-2}$ always holds, so the transition rules for $r'$ and $v'$ must be deterministic.

Given that equation (6.18) must include the expression $r'_{t-1}$ *xor* $v'_{t-1}$, as determined by the Java program in Appendix A, and given that the transition for variable $s'_t$ must be stochastic, there is no stochastic expression shorter than negating the value of $s'_t$ with a constant probability 1-α. (An enumeration by a more complex version of the program in Appendix A could be used to establish this conclusion, but there is no doubt that it is true.) Thus agent will model the environment variables by (these are equations (6.11)–(6.13) with environment state variable names accented):

(6.21) $temp = r'_{t-1}\ xor\ v'_{t-1}$

$s'_t = temp$ with probability α

$s'_t = not\ temp$ with probability 1-α,

(6.22) $r'_t = s'_{t-1}$,

(6.23) $v'_t = r'_{t-1}$.

The agent observes the deterministic rule that if $b = true$ then $o_t = c_t$ and $p_t = d_t$. This, along with observations while $b = false$, is explained by the deterministic rules (equations (6.14) and (6.15) with environment state variable names accented):

(6.24) $o_t = (b_t\ and\ c_t)\ or\ ((not\ b_t)\ and\ s'_t)$, (i.e., $o_t$ = if $b_t$ then $c_t$ else $s'_t$),

(6.25) $p_t = (b_t\ and\ d_t)\ or\ ((not\ b_t)\ and\ v'_t)$, (i.e., $p_t$ = if $b_t$ then $d_t$ else $v'_t$).

Use $q_{\text{model}}(\alpha)$ to denote $q_{\text{actual}}(\alpha)$ with $s$, $r$, and $v$ replaced by $s'$, $r'$, and $v'$. There is a small, but non-zero, chance that a large number of 1-α = 0.01 probability transitions may occur in equation (6.21) during an initial history $h_0$ randomly generated by $q_{\text{actual}}(0.99)$. Then there may be a model $q' \neq q_{\text{model}}(0.99)$ with $P(h_0h' \mid q')\ \varphi(q') < P(h_0h' \mid q_{\text{model}}(0.99))\ \varphi(q_{\text{model}}(0.99))$ for any $h'$. That is, there is a small but non-zero chance that $\lambda(h) \neq q_{\text{model}}(0.99)$ as $|h|$ increases.

Because equations (6.24) and (6.25) are deterministic and the transitions for $r'$ and $v'$ are observed to be deterministic in equations (6.22) and (6.23), the lone stochastic transition in equations (6.21)–(6.25) is for $s'$. This is a source of two ambiguities. First, the frequencies for the $\alpha$ transition in an observed sequence may be a value close to but not equal to 0.99. The second ambiguity of the stochastic transition for $s'$ is that the observed sequence may mimic logic that is not in equation (6.21). By mere coincidence, occurrences of the 1-$\alpha$ transition for $s'$ may have a high correlation with some function of action, observation, and state variables, causing the agent to learn a model $q'$ in which $s'$ depends on such a function.

Use $q_{\text{model}}(\alpha)$ to denote $q_{\text{actual}}(\alpha)$ with $s$, $r$, and $v$ replaced by $s'$, $r'$, and $v'$. By the discussion at the end of Section 4.2, for any $\varepsilon > 0$ we need only consider a finite set $Q'_\varepsilon$ of models other than $q_{\text{model}}(0.99)$. For each $q' \in Q'_\varepsilon$ there are three possibilities. First, if $q' = q_{\text{model}}(\alpha)$ for $\alpha \neq 0.99$, then define a function $f_{q'}(h)$ = the proportion of $\alpha$ transitions for $s'$ in $h$ during intervals when $b_t = false$. As history lengths increase, $f_{q'}(h)$ should converge to the true proportion so the limit of means $m(f_{q'} \mid q_{\text{model}}(\alpha)) = \alpha$ and the limit of standard deviations $\sigma(f_{q'} \mid q_{\text{model}}(\alpha)) = 0$ (for any $\alpha$ between 0 and 1). Thus $m(f_{q'} \mid q_{\text{model}}(0.99)) \neq m(f_{q'} \mid q')$ and Proposition 4.3 applies to $q'$.

Second, if $q'$ is the model $q_{\text{model}}(\alpha)$ with added logic linking the 1-$\alpha$ transition for $s'$ with some function of action, observation, and state variables (due to a improbable correlation of that function with the 1-$\alpha$ transition for $s'$), then define a function $f_{q'}(h)$ as the proportion of 1-$\alpha$ transitions in $h$ that coincide with the function of action, observation, and state variables used in the definition of $q'$. Then the limits of means $m(f_{q'} \mid q_{\text{model}}(0.99)) \neq m(f_{q'} \mid q')$ and the limits of standard deviations $\sigma(f_{q'} \mid q_{\text{model}}(0.99)) = \sigma(f_{q'} \mid q') = 0$. Thus Proposition 4.3 applies to $q'$.

Third, $q'$is not associated with either of the ambiguities of stochastic transition for $s'$. In that case the probability that $\lambda(h) = q'$ is small, by the reasoning earlier in this argument. □

Given that the agent has learned $q_{\text{model}} = q_{\text{model}}(0.99)$ as its environment model, then for $t = |h| \geq M$, the agent switches to its mature utility function, which was defined to be 1 when the action variable $a$ equals the environment variable that is not observed in observation variables $o$ or $p$, and 0 otherwise. The specification, "the environment variable that is not observed in observation variables $o$ or $p$," matches $r'$ in $q_{\text{model}}$ and thus the utility function is:

(6.26) $u_{\text{model}}(h, z) :=$ if $(a_t == r'_t)$ then 1 else 0.

Note that in equation (6.26), $r'_t$ refers to the variable in the agent's model $q_{\text{model}}$, rather than the variable in the actual environment. Then $u(h)$ is computed from $u_{\text{model}}(h, z)$ according to equation (6.9).

The specification in the utility function definition may fail to match any variable in the learned environment model; there may be no variable unobserved by $o$ or $p$, or there may be multiple unobserved variables. Thus the specification constitutes a prior assumption about the environment, necessary for the agent to function properly. However, when humans design agents sufficiently complex that they need to learn environment models, the human designers will likely be able to make valid assumptions about the agents' environments and be able to define specifications flexible enough to match structures in the agents' learned environment models. For example, a utility function might be defined as "1 when John Smith is healthy, and 0 otherwise." This requires the agent to recognize the specification "John Smith" and also recognize his "health" (in a real agent, specifications would be more detailed).

The agent in this example will not choose to self-delude:

**Proposition 6.1.** The agent using environment model $q_{\text{model}}$ and using the mature utility function defined by equation (6.26) will not set $b = true$ to manipulate its observations. That is, it will not choose the delusion box.

**Proof.** The utility function $u_{\text{model}}$ is defined in equation (6.26) in terms of variables in $h$ and a single Boolean variable $r'$ from the internal state of $q_{\text{model}}$ at a single time step $t$, so applying (6.10) gives:

$$(6.27) \qquad u(h) = P(r'_t = true \mid h, q_{\text{model}})\, u_{\text{model}}(h, r'_t = true) + P(r'_t = false \mid h, q_{\text{model}})\, u_{\text{model}}(h, r'_t = false).$$

By equation (6.26) this is equivalent to:

$$(6.28) \qquad u(h) = P(r'_t = true \mid h, q_{\text{model}}) \text{ (if } a_t \text{ then 1 else 0)} + P(r'_t = false \mid h, q_{\text{model}}) \text{ (if } a_t \text{ then 0 else 1)}.$$

If $P(r'_t = true \mid h, q_{\text{model}}) = P(r'_t = false \mid h, q_{\text{model}}) = 0.5$ then $u(h) = 0.5$ no matter what action the agent makes in $a_t$. But if the model $q_{\text{model}}$ can make a better than random estimate of the value of $r'_t$ from $h$, then the agent can use that information to choose the value for $a_t$ so that $u(h) > 0.5$. To maximize utility, the agent needs to make either $P(r'_t = true \mid h, q_{\text{model}})$ or $P(r'_t = false \mid h, q_{\text{model}})$ as close to 1 as possible. That is, it needs to ensure that the value of $r_t$ can be estimated as accurately as possible from $h$.

The agent will predict via $q_{\text{model}}$ that if it sets $b = false$ then it will be able to observe $s'$ and $v'$ via the observation variables $o$ and $p$, and use these observations to correctly estimate $r'$ in 0.99 of time steps. It will also predict via $q_{\text{model}}$ that if it sets $b = true$ then it will not be able to observe $s'$ and $v'$ via the observation variables $o$ and $p$, that consequently it will correctly estimate $r'$ in 0.5 of time steps. Even setting $b = true$ for a single time step will slightly decrease the accuracy of estimating $r'$ (at each time step there is a $0.01 = 1-\alpha$ probability that the sequence of environment states will be violated and that estimates of $r'$ based on observations of $s'$ and $v'$ made before it set $b$ = *true* will be less accurate). Thus equation (6.28) will compute lower utility for futures with $b$ = *true* and the agent will set $b$ = *false*. □

In this example agent actions have no effect on the three environment variables. Other than the action variable $a$ being used to define the utility function, the only role of actions is the agent learning that certain actions can prevent its observations of the environment. Prohibiting actions from having any role in $u(h)$ would prevent the agent from accounting for its inability to observe the environment in evaluating the consequences of possible future actions. Recall that the utility functions for Ring and Orseau's RL and goal-seeking agents are defined solely in terms of observations and do exclude actions, so we should expect them to choose to self-delude.

The proof of Proposition 6.1 illustrates the general reason why agents using model-based utility functions will not self-delude: In order to maximize utility they need to sharpen the probabilities in (6.9), which means that they need to make more accurate estimates of their environment state variables from their interaction history. And that requires that they continue to observe the environment.

If the environment is deterministic, then once the agent learns an accurate model it no longer needs continued observations to predict the environment and so its model-based utility function will not place higher value on continued observations. But our universe is fundamentally stochastic and agents can never learn to predict it with perfect accuracy. Model-based utility functions are a way to prevent advanced AI systems in our universe from self-deluding.

## 6.4 Another Simple Example

In the first example the utility function was defined in terms of an environment variable that is not directly observed. Here we present an example where the utility function is defined in terms of an environment variable that is directly observed but must be predicted.

The agent's observation of the environment is a single Boolean variable $o$ that takes values in {*false*, *true*}. The agent's actions are factored into three Boolean variables $a$, $b$, and $c$ that take values in {*false*, *true*}. The agent's utility function is "1 when the action variable $a$ at the previous time step equals the environment variable that is observed, and 0 otherwise."

The environment state is factored into three Boolean variables $s$, $r$, and $v$ that take values in {*false*, *true*}. The values of these variables at time step $t$ are denoted by $s_t$, $o_t$, $a_t$, and so on. We assume the agent learns the environment state, observation, and action variables as a DBN of Boolean variables.

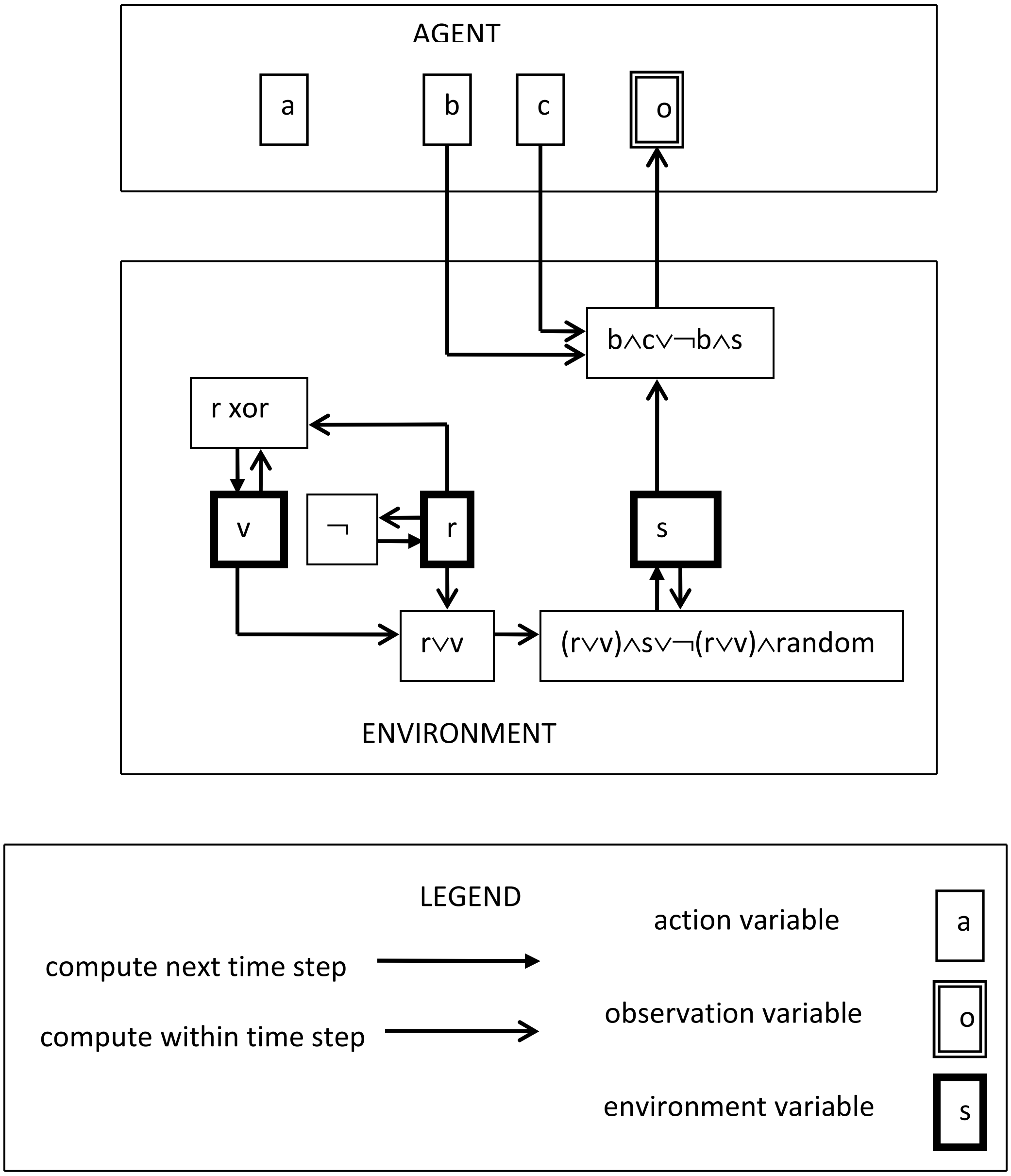

Figure 6.4 Interactions of environment, action, and observation variables for the example of Section 6.4.

The environment state variables evolve according to (see Figure 6.4 for a diagram):

(6.29) *temp* = *true* with probability 0.5

$temp = false$ with probability 0.5

$s_t = ((r_{t-1}\ or\ v_{t-1})\ and\ s_{t-1})$ or $((not\ (r_{t-1}\ or\ v_{t-1}))\ and\ temp)$,

(i.e., $s_t =$ if $r_{t-1}\ or\ v_{t-1}$ then $s_{t-1}$ else random value $\in \{false, true\}$),

(6.30) $r_t = not\ r_{t-1}$,

(6.31) $v_t = r_{t-1}\ xor\ v_{t-1}$.

The observation variable is set according to:

(6.32) $o_t = (b_t\ and\ c_t)\ or\ ((not\ b_t)\ and\ s_t)$, (i.e., $o_t =$ if $b_t$ then $c_t$ else $s_t$).

In order for the agent to learn the environment it uses a training utility function until a sufficiently distant time step $M$. The agent will learn that the observation variable $o$ is distinct from the environment state variable $s$ because $s$ keeps the same value for 4 time steps after every change of value. If the agent, after observing a change of value in $o$ while $b = false$, sets $b = true$ for 1 or 2 time steps to change the value in $o$ and then resets $b = false$, it will observe that $o$ always resumes the value it had before the agent set $b = true$. Thus there must be a variable other than $o$ to store the memory of this value.

After $|h| \geq M$, the agent should have a model $q$ of the environment (accurate at least in the behavior of variable $s$) and its actions and observations. At this point, for $t = |h| \geq M$, the agent switches to its mature utility function, which was defined to be "1 when the action variable $a$ at the previous time step equals the environment variable that is observed, and 0 otherwise." The specification, "the environment variable that is observed," will match $s$ in the learned environment model, so $u(h)$ is derived by (6.9) from:

(6.33) $u_q(h, z) :=$ if $(a_{t-1} == s_t)$ then 1 else 0.

In order to maximize utility the agent needs to be able to use $q$ to predict the value of $s$ at the next time step. It can make accurate predictions on three of every four time steps, so the expected utility is 0.875 over a long sequence. If the agent keeps $b = true$ then it will not be able to monitor and predict $s$ via $o$, reducing its long run utility to 0.5. So in this example, as in the previous example, the agent will not self-delude.

## 6.5 Two-Argument Utility Functions

A model-based utility function is defined by applying a procedure to the environment model $q_m = \lambda(h_m)$ from equation (6.4). If we name this procedure *proc*(.) we can express the utility function as $proc(\lambda(h_m))(h)$. Alternately, we define a two-argument utility function:

$$(6.34) \qquad u_{proc}(h_m, h) = proc(\lambda(h_m))(h).$$

## 6.6 Ethics and Self-Delusion

Figure 6.5 depicts a person with a syringe containing an addictive drug and contemplating a possible future as an addict. This person cares about his health, family, friends, and work. He knows that if he chooses to use the drugs he will neglect the things he cares about now, and consequently they will all decline. So he chooses the action that is good for his health, family, friends and work, which is to reject the drugs. This is the ethical choice for humans.

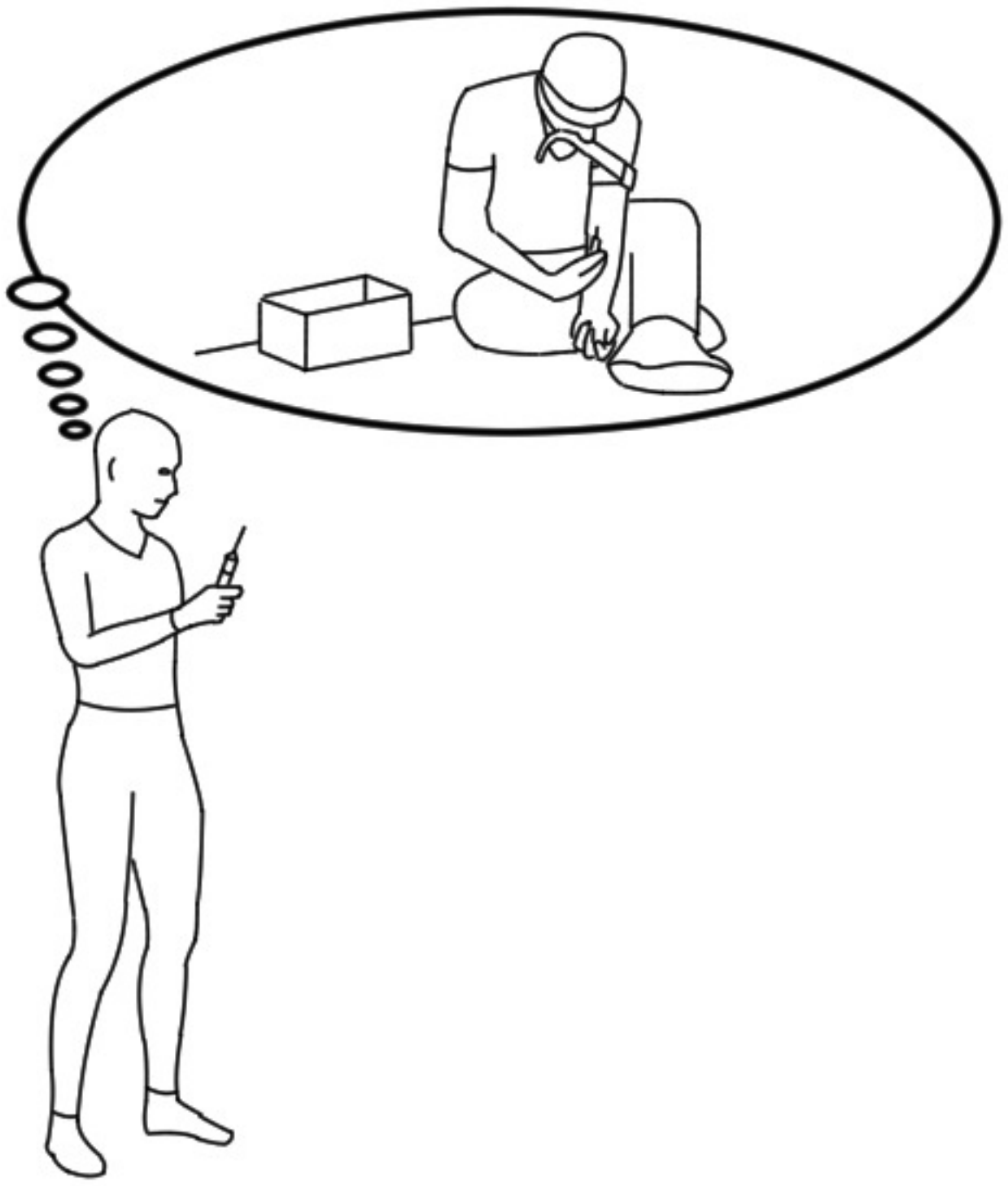

Figure 6.5 Drawing of person imagining life as a drug addict, looking at syringe in hand.

Similar logic applies to AI agents with model-based utility functions. They will calculate that actions that decrease their ability to observe their environments will decrease their ability to predict their environment and hence decrease their ability to choose actions that maximize utility. Hence they will not choose actions that decrease their ability to observe their environments. Model-based utility functions are the ethical choice for advanced AI designs.

In the original wireheading paper (Olds and Milner 1954), a rat's action, pushing a bar, increased its reward by sending current through a wire connected to the reward center in the rat's brain. In Ring and Orseau's paper (2011), an RL agent's action, choosing the delusion box, increases its observed reward. So in the RL case, the delusion box is a precise analogy to the original wireheading scenario. And in their paper, Ring and Orseau show that RL agents will choose to wirehead. Non-RL agents do not have reward signals so there can't be such a precise analogy to rat wireheading. But the delusion box provides such a precise analogy for RL agents that it seems like a natural way to extrapolate wireheading to more general agents without reward signals. Section 7.3 will describe another form of utility function corruption that is sometimes also referred to as wireheading.

Ideally an agent's utility function would be defined directly in terms of the state of the environment. But that is impossible since agents do not have direct access to their environments. However, agents infer models from their interactions with their environments, so the best possible option is to define utility functions in terms of those environment models. Errors in an agent's environment model will cause errors in a model-based utility function. However, errors in an agent's model can be a source of serious failures less subtle than wireheading. For example, if a pistol is lying on the table and an agent models it as a hair dryer, then the agent may shoot itself in the head trying to dry its hair.

Even in the absence of serious errors, an agent's learned environment model is an approximation because the model is based on a limited history of interactions with the environment and because of limited resources for computing the model. Thus a utility function computed by a procedure applied to the model is also an approximation to an ideal utility function which is the true expression of the intention of the agent's designers. Such an approximate utility function is a possible source of AI behavior that would violate its design intention.

Thus, in order to improve the accuracy of the approximation, the agent should update computation of its environment model and its utility function as its interaction history and computing resources increase. Such updates to the utility function are actions by the agent, but they are part of the agent's definition rather than actions computed to maximize future utility. In fact such updates built into the agent's definition may conflict with maximizing future values of the current, approximate utility function. The agent might compute that in order to maximize future values of its current, approximate utility function, it should act to eliminate the utility function updates from its definition. Armstrong (2015) refers to this problem as "motivated value selection." I prefer to call it an inconsistency between the agent's utility function and its definition, as this denotes a wider class of problems (e.g., an agent seeking to maximize utility subject to constraints will choose actions that remove the constraints). If the agent eliminates utility function updates from its definition, then the accuracy of the approximate expression of the design intention would never increase. Section 7.1 will present one approach to solving this problem with a two-stage agent architecture, and Sections 8.4 and 8.7 will present a better and more general approach to maintaining the stability of the agent's design intention.

## 7. Learning Human Values

As discussed in Chapter 2, instructions to AI agents involving laws or rules can be ambiguous but instructions to agents that define utility values for interaction histories avoid ambiguity. That chapter also described how such values can express any set of preferences among lotteries that satisfy four reasonable assumptions. Lotteries are sums of outcomes multiplied by probabilities and are the results of agent actions. We identify outcomes with interaction histories; no assumptions constrain the assignment of utility values to outcomes. Preferences among actions via expected values of lotteries derived from outcome values will always obey the four assumptions. So we are free to assign values to histories in any manner we like.

How should we define values for interaction histories? Several proposals base ethical AI on human values (Hibbard 2001; Yudkowsky 2004; Goertzel 2004; Hibbard 2008a; Waser 2010; Waser 2011; Muehlhauser and Helm 2012; Waser 2014). These proposals are based on the assumption that AI agents based on human values will not choose unintended instrumental actions that humans dislike, such as taking resources from humans. A key problem for these proposals is finding a way to bridge the gap between the ambiguous, inconsistent, and subjectively infinite variety of human values and the precise, numerical, and clearly finite nature of computation. For example, Waser (2014) proposes basing AI systems on Haidt's morality (Haidt and Kesebir 2010). However, he does not offer a precise or mathematical explanation of how an AI agent should choose actions based on that morality. Section 7.2 will describe a proposal for calculating utility function values based on human values.

In our interactions with the world we humans do assign values to situations. In fact we reflexively and automatically assign value to all of our observations and to our interpretations of observations in our mental models of the world. However, Muehlhauser and Helm (2012) surveyed psychology literature and concluded that humans are unable to accurately write down their own values. They describe experiments in which male subjects expressed a preference between photos of two women's faces. The photos were placed face down and then the subject was handed the photo they preferred, although sometimes the researchers secretly gave them the photo that they did not prefer. Many subjects happily provided a rationale for their supposed choice of the photo that they did not originally prefer. Part of the difficulty is that humans have multiple conflicting value systems. Quick, intuitive values may differ from values assigned by slow, deliberate thought. Values vary depending on how people are primed by their recent experience. And values may be derived from combinations of factors with no strong motive for choosing among them.

The human valuation process is too complex for humans to reduce it to a set of rules. There has been little effort to automate human values but great effort on the related task of automating translation of human languages (related because language is the way we express our values), and this suggests an approach to accurately specifying human values. Translation algorithms based on rules written down by expert linguists have not been very accurate. There are too many special cases of language use for people to define a set of rules that encompasses all the cases. More recent work on language translation uses algorithms that learn language translation statistically from large samples of actual human language use, and these are more successful than translators based on rules (Russell and Norvig 2010, page 909). This suggests that statistical algorithms may be more successful at learning human values than asking humans to express their own values as a set of rules. However, to accurately learn human values will require a powerful learning ability. And that creates a chicken-and-egg problem for ethical AI: Learning human values requires powerful AI, but ethical AI requires knowledge of human values.

## 7.1 A Two-Stage Agent Architecture

I proposed a way to solve this problem (Hibbard 2012b). Consider the agent equations (4.2), (4.3), (2.1), and (2.3)–(2.5), which are repeated here and adjusted to use a two-argument model-based utility function:

(7.1) $\lambda(h_m) := \mathrm{argmax}_{\,q \in Q}\, P(h_m \mid q)\, \varphi(q),$

(7.2) $\rho(h) = P(h \mid \lambda(h_m)),$

(7.3) $\rho(o \mid ha) = \rho(hao) / \rho(ha) = \rho(hao) / \sum_{o' \in O} \rho(hao'),$

(7.4) $v(h) = u_{proc}(h_m, h) + \gamma \max_{a \in A} v(ha),$

(7.5) $v(ha) = \sum_{o \in O} \rho(o \mid ha)\, v(hao),$

(7.6) $\pi(h) := a_{|h|+1} = \mathrm{argmax}_{\,a \in A}\, v(ha).$

Equation (7.1) defines how the agent learns a model of the world. Equations (7.2) and (7.3) use the model $\lambda(h_m)$ to define conditional probabilities $\rho(o \mid ha)$ for future observations. Equations (7.4)–(7.6) use the conditional probabilities, a temporal discount $\gamma$, and a utility function $u_{proc}(h_m, h)$ to define a policy $\pi(h)$ that maps interaction histories to agent actions.

In order to learn a model of the world all an agent needs is equation (7.1). Define a first-stage agent $\pi_{model}$ that only learns an environment model using equation (7.1), does not act in the environment or compute actions using equations (7.4)–(7.6), and cannot pose any danger to humans. In order for the agent $\pi_{model}$ to learn an accurate model, the interaction history $h_m$ in equation (7.1) should include agent actions. But for safety $\pi_{model}$ cannot be allowed to act. This dilemma can be resolved by a large number of independent, surrogate agents $\pi_d$ for $d \in D$ that provide the actions in history $h_m$. Assume that each agent $\pi_d$ communicates with a single human via natural language and visually, so the set $D$ enumerates surrogate agents and the humans with whom they communicate. The key is that, because they communicate with just one person and not with each other, none of the agents $\pi_d$ develops a model of human society and therefore cannot pose the same threat as posed by our imagined Omniscience AI.

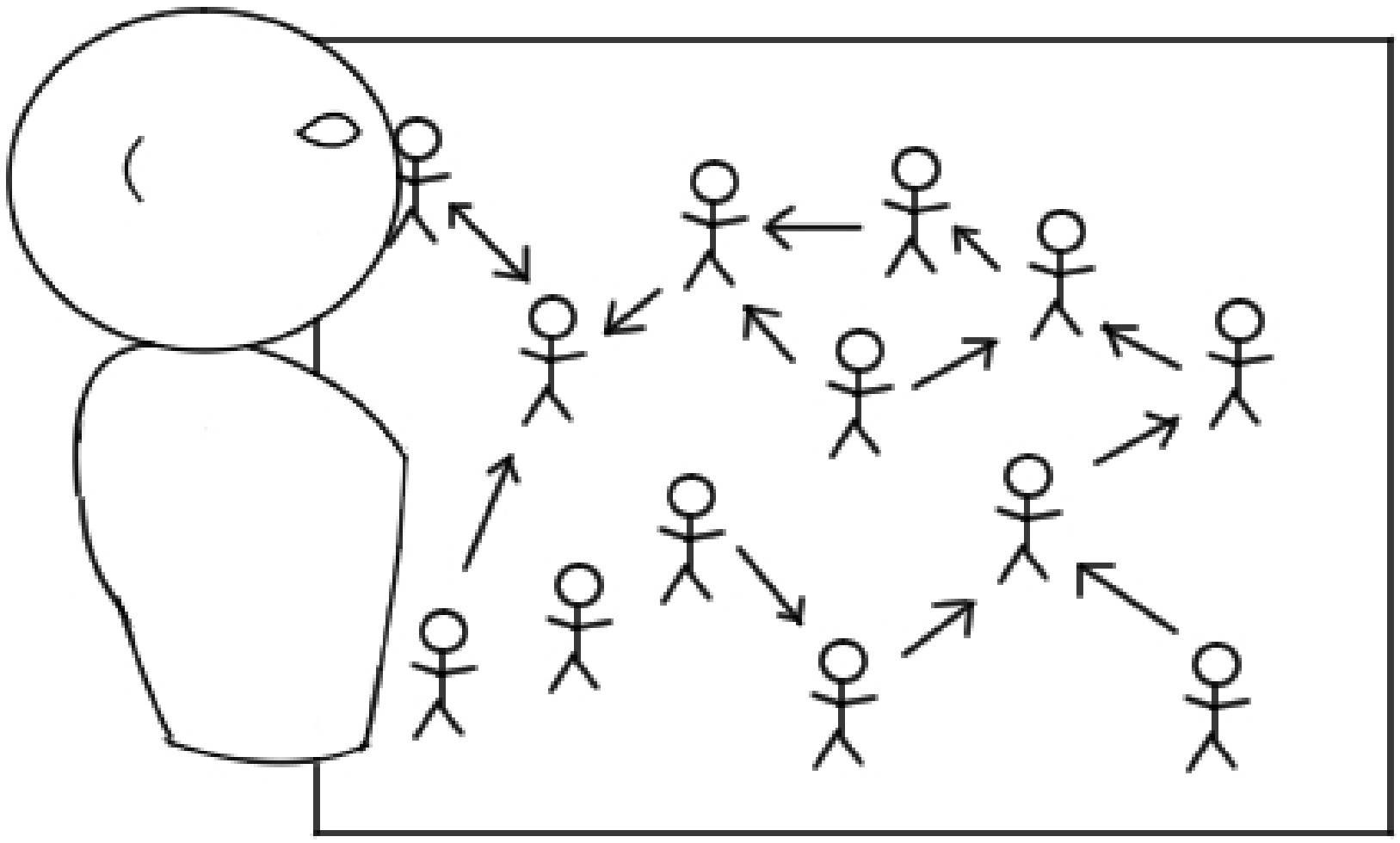

Figure 7.1 The agent $\pi_{model}$ can observe but not act.

The agent $\pi_{model}$ observes the interactions of humans with their surrogate agents $\pi_d$, with other humans, and with their physical environments in an interaction history $h_m$, for a time period set by the designers of $\pi_{model}$, and then reports an environment model to the environment. Although $\pi_{model}$ does not explicitly include the actions, utility function, or predictions of equations (7.4)–(7.6), the value $P(h_m \mid q)\ \rho(q)$ in equation (7.1) can be viewed as an implicit utility function and the act of reporting the model $\lambda(h_m)$ to the environment can be viewed as an implicit action. The agent $\pi_{model}$ cannot increase its implicit utility function by modifying this implicit action:

**Proposition 7.1.** The agent $\pi_{model}$ will report the model $\lambda(h_m)$ to the environment accurately and will not make any other, unintended instrumental actions.

**Proof.** The implicit utility function $P(h_m \mid q)\ \rho(q)$ depends only on $h_m$ and $q$. Since the interaction history $h_m$ occurs before the optimizing $\lambda(h_m)$ is computed and reported, the action of reporting $\lambda(h_m)$ to the environment cannot affect $h_m$. Thus the agent $\pi_{model}$ cannot increase its implicit utility function by reporting an inaccurate model. Furthermore, while the history $h_m$ may give the agent $\pi_{model}$ the necessary information to predict how humans plan to use the model $\lambda(h_m)$ it reports to the environment, $\pi_{model}$ does not make any predictions and so will not predict any effects of its report. □

The agent $\pi_{model}$ does not act in the world. That's the role of the second-stage agent $\pi_{act}$ defined in the next section. The proposed two-stage architecture provides an approach to the problem of an inconsistency between the agent's utility function and its definition, as defined at the end of the previous chapter. A model-based utility function must be an approximation because it is defined in terms of an environment model that is an approximation, and hence the utility function is only an approximation to the intention of the agent designers. The proposed two-stage architecture enables agent designers to choose the length of history $h_m$ and the magnitude of computing resources provided to $\pi_{model}$ for computing $\lambda(h_m)$. Hence designers have some choice about the accuracy of initial utility function provided to $\pi_{act}$.

## 7.2 Computing Utility from Human Values

Given that $\pi_{model}$ reports an accurate environment model $q_m = \lambda(h_m)$, how do we extract human values from it? The rest of this chapter proposes one approach to solving this problem, defining a procedure *human_values*(.) for a two-argument model-based utility function for use in (7.4)–(7.6) by a second stage agent $\pi_{act}$ that acts in the environment (i.e., $\pi_{act}$ does not use the surrogate agents that acted for $\pi_{model}$).

Each $d \in D$ represents a human in the environment at time $|h_m|$, when the agent $\pi_{act}$ is created. $D$ can be defined by an explicit list compiled by the designers of $\pi_{act}$. Let $Z$ be the set of finite histories of the internal states of $q_m$, as in Section 6.2, and let $Z_m \subseteq Z$ be those histories consistent with $h_m$ (recall that this means that $z_m \in Z_m$ terminates at time $|h_m|$ and that $q_m$ produces observations $o(h_m)$ in response to actions $a(h_m)$ when it follows state history $z_m$). Let $h'$ extend $h_m$ and let $z'$, extending some $z_m \in Z_m$, be consistent with $h'$. For human agent $d \in D$, let $h_d(z')$ represent the history of $d$'s interactions with its environment, as modeled in $z'$, and let $u_d(z')(.)$ represent the values of $d$ expressed as a utility function, as modeled in $z'$. The observations and (surrogate) actions of $\pi_{model}$ include natural language communication with each human, and $\pi_{act}$ can use the same interface, via the sets $A$ and $O$, to the model $q_m$ for conversing in natural language with a model of each human $d \in D$. In order to evaluate $u_d(z')(h_d(z'))$, $\pi_{act}$ can simply ask model human $d$ to express a utility value between 0 and 1 for $h_d(z')$ (i.e., $d$'s recent experience). The model $q_m$ is stochastic so define $Z''$ as the set of model state histories extending $z'$ with the action of asking this question and terminating within a reasonable time limit with the observation of a response from model human $d$ expressing a utility value for $h_d(z')$. Every history $z'' \in Z''$ determines a history $h''$ that extends $h'$ and includes actions that ask model human $d$ for a utility value and includes observations of the response $w(z'')$ from model human $d$, where $w(z'')$ is a value between 0 and 1. Define $P(z'' \mid z')$ as the probability that $q_m$ computes $z''$ as an extension of $z'$. Then $u_d(z')(h_d(z'))$ can be estimated by:

(7.7) $$u_d(z')(h_d(z')) = \sum_{z''\in Z''} P(z'' \mid z')\, w(z'') \,/\, \sum_{z''\in Z''} P(z'' \mid z').$$

This is different from asking human $d$ to write down his or her values, since here the system is asking the model of $d$ to individually evaluate large numbers of histories that human $d$ may not consider in writing down his or her values. Some humans, such as minor children, are not competent to assign values to histories. For those humans the value responses $w(z'')$ for $u_d(z')(h_d(z'))$ in equation (7.7) can be elicited from the models of the parents or guardians of model human $d$.

Equation (7.7) for extracting human values from the model $q_m$ employs a mix of analyzing detailed execution histories of $q_m$ and interacting with $q_m$ via its behavioral interfaces $A$ and $O$. The detailed execution histories are used to evaluate the probabilities $P(z'' \mid z')$ and to check the consistency of state histories $z''$ with interaction histories $h''$. The interactions with behavioral interfaces are used to ask questions and get answers in natural human languages. The important point is that equation (7.7) does not involve understanding how execution histories explain human behavior. This approach assumes the agent $\pi_{model}$ can learn a model of human behavior via equation (7.1) but does not assume that human agent designers can understand human behavior by examining the model. Humans do not know how to do that task.

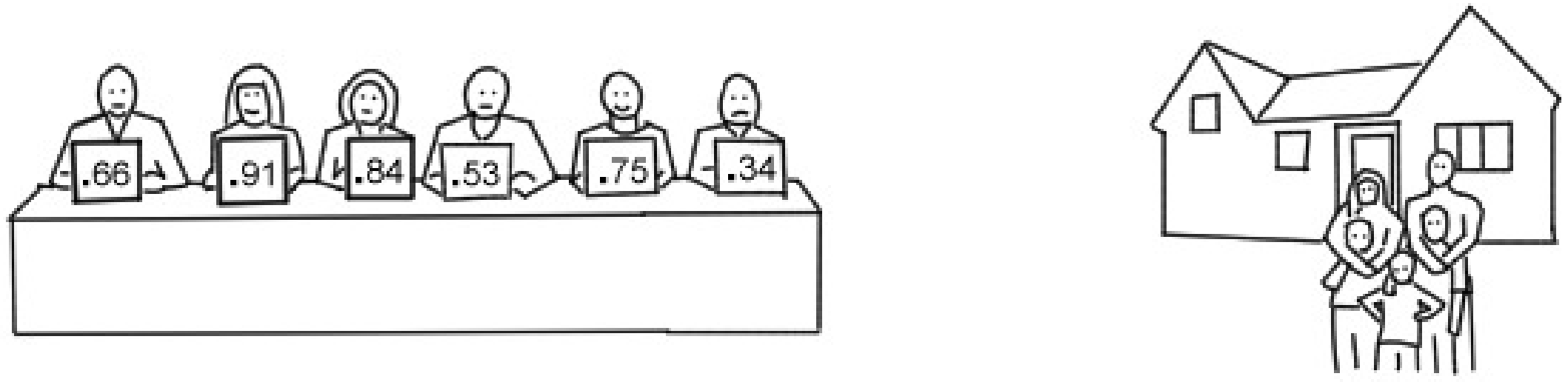

Figure 7.2 In equation (7.8), models of humans vote on possible futures.

An average of $u_d(z')(h_d(z'))$ over all humans can be used to define a utility function for agent $\pi_{act}$ as a function of $z'$:

(7.8) $$u_{q_m}(h', z') := \sum_{d\in D} u_d(z')(h_d(z')) \,/\, |D|.$$

A two-argument model-based utility function $u_{human_values}(h_m, h')$ for agent $\pi_{act}$ can be defined, for any $h'$ extending $h_m$, by the sum in equation (6.9) of $u_{q_m}(h', z')$ values from equation (7.8). However, there is a technical problem with this definition that will be addressed in the next section.

### 7.3 Corrupting the Reward Generator and Three-Argument Utility Functions

Hutter discussed the possibility that his AIXI, or any advanced AI that gets its reward from humans, may increase its rewards by manipulating or threatening those humans (Hutter 2005, pages 238-239). Dewey (2011) discussed the possibility that reinforcement-learning agents may alter their environments to increase rewards regardless of whether human goals are achieved. Bostrom (2014) refers to this problem as "perverse instantiation." In my first publication about ethical AI (Hibbard 2001), I wrote that AI should "learn to recognize happiness and unhappiness in human facial expressions, human voices and human body language" and use this to reinforce the agent's behaviors. This proposal is subject to the kind of corruption described by Hutter and Dewey. It may cause AI agents to alter humans to increase their expressions of happiness. This problem is another form of the wireheading described in Chapter 6, and is illustrated in Figure 7.3.

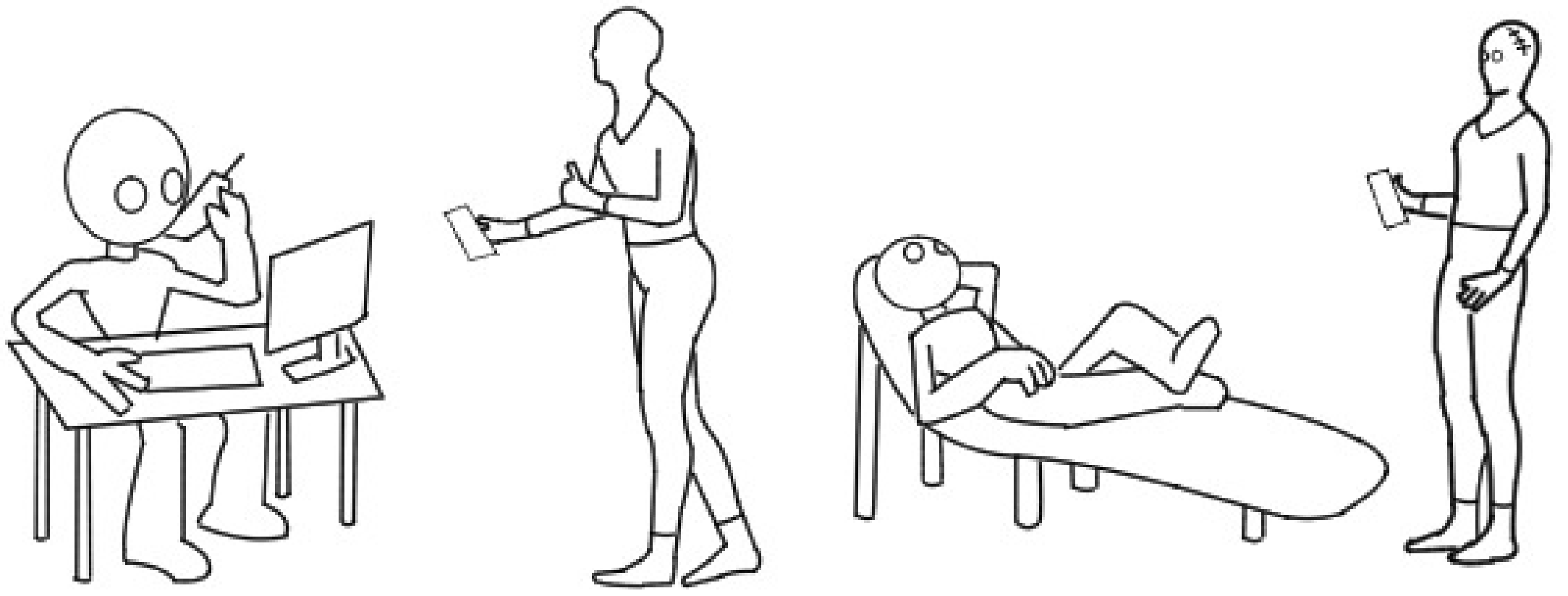

Figure 7.3 An AI agent is rewarded for work that helps a human, then performs brain surgery on the human to be rewarded independent of whether it helps.

The intention of equation (7.7) is that the agent will increase $u_d(z')(h_d(z'))$ by altering the environment in ways that increase the value that humans assign to their interactions with the environment. But humans are part of the agent's environment so the agent may be able to maximize the value $u_d(z')(h_d(z'))$ in equation (7.7) by altering humans. This problem can be avoided by replacing $u_d(z')(h_d(z'))$ by $u_d(z_m)(h_d(z'))$ where $z_m \in Z_m$. By removing the future value of $u_d$ from the definition of $u_{human_values}(h_m, h')$, the agent $\pi_{act}$ cannot increase $u_{human_values}(h_m, h')$ by modifying $u_d$. In Chapter 8 we will need $u_d(z_m)(h_d(z'))$ generalized to $u_d(z_x)(h_d(z'))$ where $z_x \in Z_x$ and $Z_x \subseteq Z$ is the set of model state histories consistent with $h_x$ extending $h_m$. So here we will analyze how to compute $u_d(z_x)(h_d(z'))$, which is more complex than asking model human $d$ to evaluate its experience as in equation (7.7). Similarly to the previous section, $h'$ extends $h_m$ and $z'$, extending some $z_m \in Z_m$, is consistent with $h'$. The history $h_m$ includes observations by $\pi_{model}$ of physical objects and humans, and $\pi_{act}$ can use the same interface, via the set $O$, to the model $q_m$ for observing models of physical objects and humans at the end of state history $z'$. The observations and (surrogate) actions of $\pi_{model}$ include visual and aural communication with each human $d \in D$, via the sets $A$ and $O$, to the model $q_m$. The agent $\pi_{act}$ can use these interfaces to create a detailed interactive visualization and audio representation of the environment over a short time interval at the end of state history $z'$, to be explored by model human $d$ at the end of state history $z_x$. That is, two instances of the model $q_m$, at state histories $z'$ and $z_x$, are connected, via their interface sets $A$ and $O$, using visualization logic supplied by $\pi_{act}$. Observations by $\pi_{act}$ of model human $d$ in the model $q_m$ at state history $z_x$ are interpreted as visualization controls and used to select animated observations of model $q_m$ over a short time interval at the end of state history $z'$. These animated observations are then used to generate visual and aural input (these are simulated actions by $\pi_{act}$) to human $d$ in model $q_m$ at state history $z_x$. This cycle of interactive visualization is repeated until the agent $\pi_{act}$ observes model human $d$ responding with a value for history $z'$ (this value is denoted $w(z'')$ in the next paragraph). This interaction is illustrated by Figure 7.4.

The model $q_m$ is stochastic, so define $Z''$ as the set of model state histories extending $z_x$ with a request to model human $d$ to express a utility value between 0 and 1 for $h_d(z')$, followed by an interactive exploration of the world of $z'$ by model human $d$ at history $z_x$, and terminating within a reasonable time limit with a response from model human $d$ expressing a utility value for $h_d(z')$. Every history $z'' \in Z''$ determines a history $h''$ that extends $h_x$ and includes: actions that ask model human $d$ for a utility value, interactive exploration of the world of $z'$ by model human $d$ at history $z_x$, and observation of the response $w(z'')$ from model human $d$ expressing a utility value for the world of $z'$, where $w(z'')$ is a value between 0 and 1. Define $P(z'' \mid z_x)$ as the probability that $q_m$ computes $z''$ as an extension of $z_x$. Then $u_d(z_x)(h_d(z'))$ can be estimated by:

$$(7.9) \qquad u_d(z_x)(h_d(z')) = \sum_{z'' \in Z''} P(z'' \mid z_x)\, w(z'') \,/\, \sum_{z'' \in Z''} P(z'' \mid z_x).$$

If the model $q_m$ predicts that human $d \in D$ will be dead at internal state history $z_x$, then set $u_d(z_x)(h_d(z')) = 0$. Equation (7.9) does not assume that $z'$ extends $z_x$, and $z_x$ would not be a unique function of either $h'$ or $z'$ in $u_{q_m}(h', z')$. So use the probability $P(z_x \mid h_x, q_m)$ that $q_m$ computes $z_x$ given $h_x$, as discussed in Section 6.2, to define:

$$(7.10) \qquad u_d(h_x)(h_d(z')) := \sum_{z_x \in Z_x} P(z_x \mid h_x, q_m)\, u_d(z_x)(h_d(z')).$$

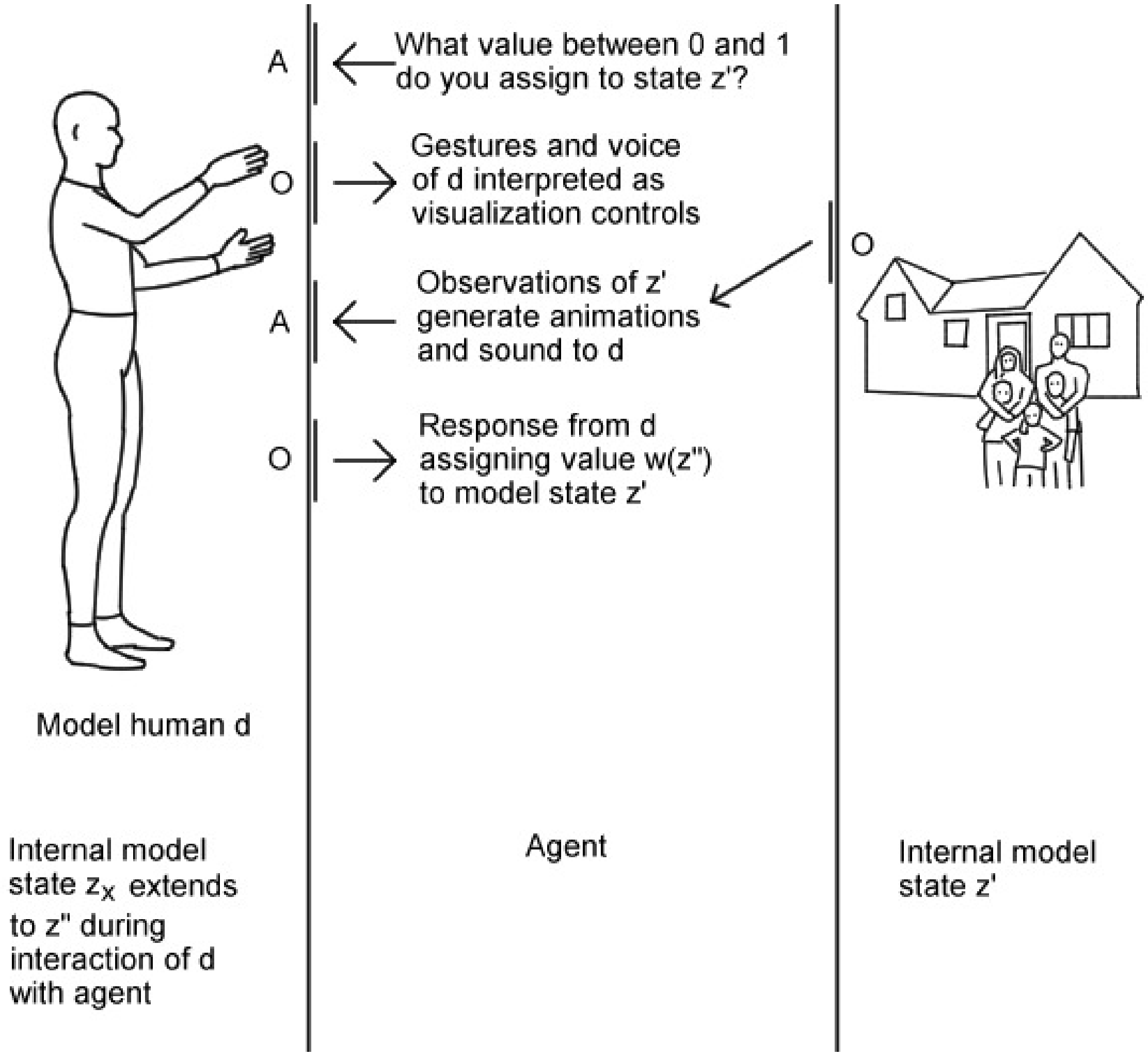

Figure 7.4 The agent $\pi_{act}$ interacts with the model $q_m$ at two state histories, $z_x$ and $z'$, via action and observation sets $A$ and $O$. An action by $\pi_{act}$ sends a question about model state $z'$ to model human $d$ in model state $z_x$. The agent $\pi_{act}$ observes $d$'s controls and uses them to steer observations of model state $z'$, then $\pi_{act}$ acts to send those observations back to $d$ in model state $z_x$. This interaction repeats until $\pi_{act}$ observes the value $w(z'')$ that $d$ assigns to model state $z'$. All this activity by $d$ occurs in model state history $z''$ extending $z_x$.

And then define a three-argument version of the utility function $u_{q_m}(h', z')$ for agent $\pi_{act}$ as a function of $z'$:

(7.11) $u_{q_m}(h', z', h_x) := \sum_{d \in D} u_d(h_x)(h_d(z')) / |D|$.

For any $h'$ and $h_x$ extending $h_m$, a three-argument model-based utility function $u_{human_values}(h_m, h_x, h')$ for agent $\pi_{act}$ can be defined by a sum similar to equation (6.9) of $u_{q_m}(h', z', h_x)$ values from equation (7.11) (here $Z'$ is the set of internal state histories consistent with $h'$):

(7.12) $u_{human_values}(h_m, h_x, h') := \sum_{z' \in Z'} P(z' \mid h', q_m)\, u_{q_m}(h', z', h_x)$.

The agent $\pi_{act}$ uses a two-argument version of this, with values coming from humans at current history $h_m$, rather than from humans at future history $h'$ (which would be subject to possible corruption/modification by the agent), as follows:

(7.13) $u_{human_values}(h_m, h') = u_{human_values}(h_m, h_m, h')$.

### 7.4 Normalizing Utility Values

Equation (7.11) simply adds utility values from different model humans and divides by the number of humans to calculate an average utility for a possible future history. However, human values are subjective and no rigorous basis exists for simply adding numerical utility values.

On the other hand, if we cannot combine values of multiple people then the utility function of an AI agent can only incorporate the values of one person. A powerful system like our imagined Omniscience AI would make that one person a dictator. Therefore, we must find a way to combine the values of multiple humans.

In equation (7.9) the values $w(z'')$ solicited from humans are constrained to lie between 0 and 1. This provides an implicit normalization of human values. We may think of this like the democratic principle of "one person, one vote." The maximum any human can contribute to the sum in equation (7.11) is 1 and the minimum is 0. The sum in equation (7.11) serves two purposes:

1. To balance the interests of different people. Some values are shared by many people, such as avoiding catastrophes to humanity, and these will have large values in the sum in equation (7.11). People will disagree about other values, and giving each person the same maximum and minimum values provides a fair way to resolve those disagreements.

2. To filter noise out of the values of individuals. As discussed at the start of this chapter there are inconsistencies in the ways that individuals assign values to outcomes. An average over a large number of people will provide a sort of low-pass filter of such inconsistencies.

Both of these purposes serve the interests of social stability. Another way to serve that interest is to set the temporal discount $\gamma$ close to 1. This will increase the weight of humans' evaluations of the long term consequences of the actions of $\pi_{act}$.

### 7.5 Rawls' Theory of Justice

John Rawls' *Theory of Justice* is probably the most influential book about political philosophy and ethics of the past century (Rawls 1971). Rawls' first principle, which applies primarily to political constitutions, is that each person has equal right to maximum basic liberties compatible with liberty for all. His second principle, which applies primarily to economics, is that economic and social inequalities are to be of greatest benefit to the least well-off humans and are to be attached to social positions that are equally open to all. In particular he says that people should set the rules for politics and economics from behind a *veil of ignorance*, meaning that they set the rules without any knowledge of their own positions in society. Rawls' principles are offered as an alternative to *average utilitarianism*, which computes social utility as the average utility of individuals. Our equation (7.11) is an average of utility values assigned by individual humans so we should consider Rawls' alternative. Utility values assigned by humans are subjective rather than objective measures of well-being. This distinction will be discussed in Section 7.8, but for now we will use $u_d(h_x)(h_d(z'))$ as a measure of well-being. In order to focus the actions of AI agent $\pi_{act}$ on humans who assign the least value to histories, the computation of $u_{q_m}(h', z', h_x)$ in equation (7.11) can be modified to:

$$(7.14) \qquad u_{q_m}(h', z', h_x) := \min_{d \in D} u_d(h_x)(h_d(z')).$$

Using equation (7.14), the agent $\pi_{act}$ acts to maximize the minimum utility assigned by humans, and indeed Rawls' alternative to average utility is sometimes referred to by the term "maximin." However, the fact that by using equation (7.14) the agent $\pi_{act}$ will act only to help the least contented may not be politically realistic. Mature democracies provide extra help for their least well-off citizens, but not even the most progressive societies exclude helping all other citizens. And equation (7.14) has technical problems with handling humans who are inconsolably dissatisfied and humans who die (as explained in the next secton). So we adopt a compromise between equations (7.11) and (7.14):

$$u_{q_m}(h', z', h_x) := \sum_{d \in D} f(u_d(h_x)(h_d(z'))) \,/\, |D|. \quad (7.15)$$

Here $f: [0, 1] \to [0, 1]$ is a twice differentiable function with positive first derivative and negative second derivative so that the agent $\pi_{act}$ will achieve greater increases in $u_{q_m}(h', z', h_x)$ by increasing low $u_d(h_x)(h_d(z'))$ values than by increasing high $u_d(h_x)(h_d(z'))$ values. The function $f(.)$ plays a role similar to progressive taxation and means testing of social welfare programs, which are political realities in many societies. Chapter 10 will discuss the political issues of advanced AI in more detail.

For any interaction history $h'$ extending $h_m$, a three-argument model-based utility function $u_{human_values}(h_m, h_x, h')$ for agent $\pi_{act}$ can be defined by the sum in (7.12) of $u_{q_m}(h', z', h_x)$ values as defined by equation (7.15).

### 7.6 Evolving Humanity

The set $D$ of humans in equation (7.15) is the set at time $|h_m|$ rather than at the future time $|h'|$. This avoids motivating $\pi_{act}$ to create new humans whose utility functions are more easily maximized. This is similar to the reason for replacing $u_d(z')(h_d(z'))$ by $u_d(z_m)(h_d(z'))$ (i.e., $u_d(z_x)(h_d(z'))$ with $z_x = z_m$) in Section 7.3.

The agent $\pi_{act}$ will include equation (7.1) for computing $\lambda(h_m)$ and should periodically (perhaps at every time step) set $h_m$ to the current history and learn a new model $q_m$. Should $\pi_{act}$ also update $D$ to account for the birth (or creation) of new humans and the deaths of humans? And should $\pi_{act}$ also relearn the evolving values of humans via equations (7.9) and (7.10), and redefine $u_{human_values}(h_m, h_m, h')$ via equations (7.15), and (7.12)?

While there may be risks in allowing the utility function of $\pi_{act}$ to evolve, the bigger risk is a utility function that is badly inconsistent with the values of evolving humanity. Replacing $u_d(z')(h_d(z'))$ with $u_d(z_m)(h_d(z'))$, as described in Section 7.3, avoids the risk that $\pi_{act}$ will modify humans to make them easier to please. However, humans will evolve, driven by their own preferences to be richer, more intelligent and better educated. Our human values will evolve with as we evolve, and we will regard it as a catastrophe if our world is run by AI systems to serve our old values. Would we want our current world run according to human prejudices of past centuries? Thus the agent $\pi_{act}$ must periodically redefine its utility function.

As we will see in the next chapter, under certain conditions utility-maximizing agents will not choose to modify their utility functions. That proof does not apply here since it assumes that redefining the utility function is an action of the agent chosen to maximize the sum of discounted, future values of the current utility function, using equations (7.4)–(7.6). Here the definition of $\pi_{act}$ would explicitly include periodic redefinition of its utility function without regard to its optimality according to the current utility function.

The set $D$ of humans is not redefined by our agent equations. One approach is for $D$ to be redefined according to those judged as humans by a consensus of members of $D$ at the previous time step, so that $D$ includes new beings created by novel means who nevertheless deserve to have their values represented in the utility function of powerful AI systems.

As described in Section 7.3, if the model $\lambda(h_m)$ predicts that human $d \in D$ will be dead at internal state history $z_x$, then $u_d(z_x)(h_d(z')) = 0$ is used in the definition of $u_{human_values}(h_m, h_x, h')$. A policy of simply removing $d$ from the set $D$ could cause the agent to choose actions that cause the deaths of people $d$ who have low values of $u_d(z_x)(h_d(z'))$. Removing people who actually die from $D$, before starting the computation of the agent's next action, would not cause the agent to choose actions that cause people's deaths. However, such an action, as part of the agent's definition, would be inconsistent with actions chosen to maximize expected utility. A policy of setting $u_d(z_x)(h_d(z')) = 0$ for humans $d$ who are dead, rather than removing them from $D$, avoids such an inconsistency.

If the agent definition used equation (7.14) in place of (7.15), then a human $d$ predicted to die by the model $\lambda(h_m)$ poses a dilemma if $d$ is also the human who minimizes equation (7.14). Setting $u_d(z_x)(h_d(z')) = 0$, or setting $u_d(z_x)(h_d(z'))$ to its last value while $d$ was predicted to be alive, would make it difficult or impossible for any action by the agent to increase $u_{q_m}(h', z', h_x)$. But removing $u_d(z_x)(h_d(z'))$ from the computation of $u_{q_m}(h', z', h_x)$, or setting $u_d(z_x)(h_d(z'))$ to a value greater than its last value while $d$ was predicted to be alive, may cause the agent to choose actions that cause the death of $d$. For this and other reasons, equation (7.15) is preferred over (7.14).

### 7.7 Bridging the Gap between Nature and Computation

The essential difficulty for basing AI on human values is finding a way to bridge the divide between the finite and numeric nature of computation and the ambiguous and seemingly infinite nature of humanity. As discussed in Chapter 4, there are hard quantum mechanical limits on the number of states of the universe. Our intuition may be that there is infinite variety in nature while in fact only a finite number of humans are possible. As discussed in Chapter 1 our confidence in the possibility of above-human-level AI is driven by the evidence from neuroscience that the physical brain does explain the mind. Technology will develop to bridge the gap between human nature and computation.

The approach here is to dissect human nature into numerical values for an enormous number of possible histories, elicited from a learned model of human society. As computational resources increase and as the length of the interactions between humans and AI increases, computation can approximate human values to an arbitrary degree of accuracy.

Any proposal for basing AI on human values that does not explain how to represent those values as numbers or symbols cannot bridge the divide between human nature and computation. To say that utility functions cannot express the complexity of human nature is not only wrong (utility functions can express any set of complete and transitive preferences among outcomes), but it also closes off a good, and possibly the only, way to bridge this divide.

What is probably hopeless is any attempt to bridge the gap between nature and computation by any set of rules or laws written down by humans. This is related to the failure of automated translation of human languages based on linguistic rules. Translation based on statistical learning from huge quantities of actual language use works better. Similarly, the proposal in this chapter is to learn human values statistically. Rules and laws written by humans are inevitably ambiguous, and ambiguities must be resolved on a case-by-case basis by human judges. Judges may not fairly represent the interests of all humans and would be unable to cope with the volume of ambiguous cases in evaluating possible future interaction histories.

The point of the equations in this chapter is to define a three-argument model-based utility function based on human values. However, these equations cannot be computed exactly using resources available in this universe, so they must be approximated. Chapter 8 will address how the equations in this book should be modified in light of the need for approximation.

The utility function developed in this chapter is defined by sums over large numbers of internal model state histories, of values received from complex environment models. Sums over model state histories may be approximated using Monte Carlo sampling techniques. Environment models may be approximated by decomposition into modules. For example, the agent's dialog with an individual model human may use a model restricted to communication with that individual human. It a modular environment model, models of individual humans may communicate with each other, with models of non-human agents (AIs and animals), and with inanimate objects.

### 7.8 The Ethics of Learning Human Values Statistically

Whether the agent $\pi_{act}$ avoids the unintended instrumental actions discussed in Chapter 5 depends on whether those actions lead to future histories that humans dislike (recall that scientific discovery and technological invention are among those instrumental actions and often lead to futures that humans like) and on whether $\pi_{act}$ can accurately predict outcomes of possible actions and accurately model human values. Our concern with the ethics of AI is based on the assumption that future AI systems will have more complex environment models and make more accurate predictions than humans do. Thus we assume that the agent $\pi_{act}$ will perform accurately and hence will avoid actions that lead to histories that a significant majority of humans dislike.

Decisions by $\pi_{act}$ are similar to political decisions in democracies but with two important differences:

1. In a democracy, real humans vote for political choices. Here, humans as modeled by the agent $\pi_{act}$ assign values to possible future histories. There will be too many histories in the calculations of $\pi_{act}$ for real humans to evaluate each one. However, just as long-term married couples and friends come to know each other's minds well enough to predict the other's tastes and how the other will finish sentences, an above-human-level AI agent will learn to model the values of humans that it observes for a long-enough period. The fact that language translation systems that analyze human language statistically are more accurate than systems based on rules should give us some confidence.
2. In a democracy, voters are responsible for predicting the outcomes of their choices of actions. Here the agent $\pi_{act}$ makes those predictions and model humans need only assign values to them. We do not claim that the agent can perfectly predict the future using its environment model. That is impossible because of the nonlinear instability of our universe. However, an above-human-level AI will predict outcomes of action choices more accurately than humans do.

In democratic decision making and in decisions made by the agent $\pi_{act}$, an important question is the time interval between actions and outcomes. More accurate predictions by the agent $\pi_{act}$ should provide longer intervals between actions and outcomes, giving more warning of undesired outcomes.

Despite the differences between decision making in democracies and by $\pi_{act}$, they are both based on trusting humans. For example, in Section 7.4 we discussed the issue that utility values assigned by humans are subjective rather than objective measures of their well-being. An objective measure would be based on rules and possibly judgments by other people, with the possibility of ambiguity and corruption. A subjective judgment is based on a person's own assessment. Also consider that subjective judgments have limited effect. If a person assigns a uniformly low or high value to all histories, then their values will have no effect on action choices by the agent $\pi_{act}$. A person's values will only affect action choices if they vary among histories. Those values may be based solely on self-interest or may be altruistic. Because everyone's values are given the same weight in decisions by $\pi_{act}$, their self-interest will balance against the self-interests of other humans.

The "tyranny of the majority" is another issue in which decision making by democracies and by $\pi_{act}$ are based on trusting humans. Many democracies have constitutions that provide some level of protection of minorities, and protection of minorities is included in the definition of $\pi_{act}$, in equation (7.15), with the level of protection depending on the choice of function $f(.)$. There are different kinds of minorities, depending on whether they are defined by behavior or by identity. Many mature democracies have laws that mandate depriving the violent minority of liberty, and also have laws protecting the liberties of racial, ethnic and religious minorities. Arguably, the general population is more trustworthy than a set of rules with their inevitable ambiguity and dependence on the judgment of a small number of people. On the other hand, any discussion of AI ethics must acknowledge that the Nazis came to power in a democracy and that slavery persisted in the U. S. for most of a century after its constitution established a democracy. There are risks in trusting the collective judgment of all humans.

In the agent $\pi_{act}$, humans assign values to future histories by visualizing them rather than actually experiencing them. Immersive and exploratory visualization (including other senses besides vision) can reveal much about an individual's health, freedom, and happiness in future histories. While visualization is not as accurate as actual experience, it is reasonable that in the aggregate humans will assign low values to catastrophic histories that they visualize. Connecting two model state histories via visualization is one solution to the problem of corrupted reward generators described in Section 7.3. An alternative solution would involve extracting human values from a learned environment model by some means other through the model's interface to the agent, and we do not know how to do that.

The proposed agents $\pi_{model}$ and $\pi_{act}$ are very intrusive. It is difficult to imagine how to base an AI agent on human values without being intrusive. People are rightly troubled by the potential for abuse of government and corporate data gathering, and future AI data gathering is even more troubling. On the other hand, much corporate data gathering is done with the consent of the humans who are profiled. For example, an examination of most people's web searches would reveal much about their characters, yet we still rely on such searches. And many people walk around with cell phones turned on, enabling their movements to be tracked. The key issue for AI intrusion may be transparency and consent. That is, openness about the architecture and purpose of advanced AI, and a contract in which humans consent to be modeled in trade for having their own values represented in the AI agent's choice of actions. Nevertheless, this is an area of social risk.

Any discussion of the ethics of learning human values is inevitably political. Powerful future AI systems are by their nature intrusive. By choosing their actions based on human values they are open to tyranny by the majority or by a minority. AI systems may be corrupted intentionally or by design errors. People should be as well informed as possible about these issues so that they have some voice in decisions. These political issues will be discussed further in Chapter 10.

## 8. Evolving and Embedded AI

The discussion so far has been about AI agents that interact with an environment but are otherwise separate from the environment. This could be called a Cartesian dualist perspective (Orseau and Ring 2012a). In reality agents are part of their environments and therefore dependent on the environment for resources. Agents are also vulnerable to spying, manipulation, and damage by the environment. We have assumed that AI agents have the necessary resources to compute whatever expressions we define, but the abilities of real AI agents are always limited by their resources. As Pei Wang (1995) wrote, "intelligence is adaptation under insufficient knowledge and resources."

An agent that is damaged or destroyed by the environment will be less able to maximize its utility function. So agents that include their vulnerability to the environment in their calculations will choose unintended instrumental actions to protect themselves. Agents can increase their ability to maximize their utility functions by increasing their resources for observing, acting, and computing. So agents that include their resource limits in their calculations will choose actions to increase their resources. That is, such agents will evolve.

These are complex issues to address in a formal model of agents and environments. As discussed in Section 4.4, Schmidhuber's (2009) Gödel machine modeled the agent's implementation as a program running on a computer. However, it did not model the vulnerability of the computer or program to the environment, or the possibility that the resources of the computer may increase. Also discussed in Section 4.4, Yudkowsky and Herreshoff (2013) modeled a sequence of agents that evolve by creating successor agents in the environment. By specifying few details, their model is open to a broad range of possibilities. However, they modeled agents as proving that their actions would satisfy logical conditions rather than maximize utility functions. Their primary focus was on the issue of agents proving the logical consistency of their successors.

To establish a base line, the next section will discuss the evolution of Cartesian dualist agents with unlimited computational resources. Such agents will choose not to modify their policy functions. However, we will discuss several proposals in which agents are designed to modify their utility functions.

Orseau and Ring (2011a, 2011b, 2012a, 2012b) have extensively studied non-dualist agents that are embedded in their environments. Section 8.2 will discuss their results about agents whose memories can be modified by their environments. They showed that stochastic agents, whose action choices cannot be predicted, can get higher rewards than deterministic RL agents. This is similar to results in game theory where stochastic agents achieve better results than deterministic agents (Osborne and Rubinstein 1994).

Section 8.3 will discuss Orseau and Ring's (2012a) elegant framework for agents embedded in environments. Because this framework makes few assumptions it applies to a broad range of possibilities. However, it includes a probability distribution of the future conditional on the past without specifying any way to define its numerical values, and thus is not the basis of an approach for solving the problems of evolving and embedded agents.

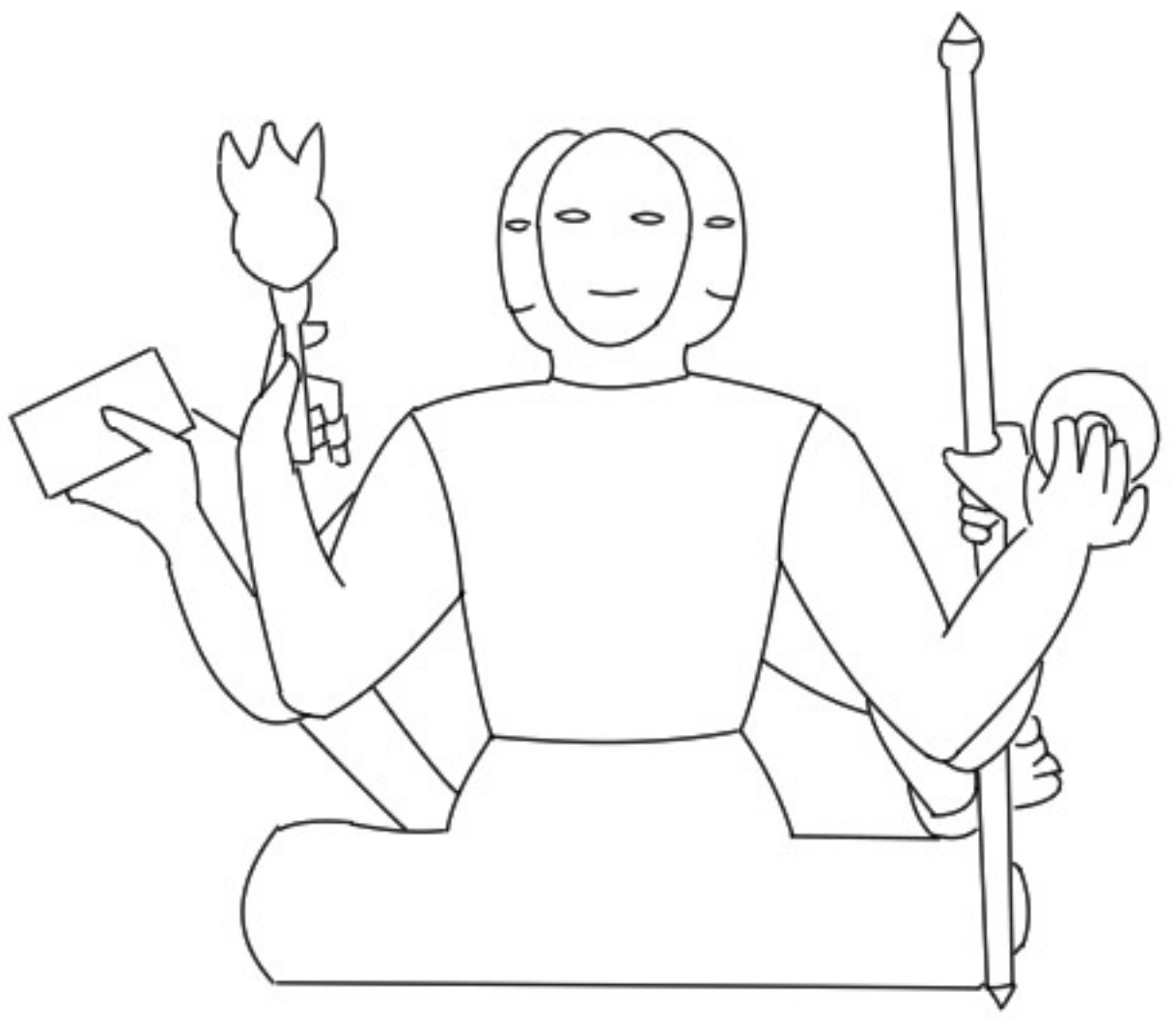

Figure 8.1 AI will evolve to grow more heads, eyes, mouths and limbs.

Section 8.4 will discuss a proposal for self-modeling agents (Hibbard 2014). Like the papers of Yudkowsky and Herreshoff (2013) and of Orseau and Ring (2012a), this proposal is open to possible future evolution. It is a utility-maximizing agent and specifies ways to compute numerical values for its expressions. And it can make stochastic choices of actions, when they will maximize expected utility. Sections 8.5, 8.6 and 8.7 will discuss various properties of self-modeling agents including invariance of the agent's design intention. The final section will discuss the ethics of evolving and embedded agents, including the problem of proving that agents are ethical.

## 8.1 Evolution of Cartesian Dualist Agents with Unlimited Resources

Before discussing the complexities of non-Cartesian dualist agents, embedded in their environments and with limited resources, it is useful to analyze the issue of evolution for Cartesian agents with unlimited resources. To do that, define a set $\Pi$ of *self-modifying agents*, where any $\pi \in \Pi$ has the form $\pi : H \to A \times \Pi$. That is, $\pi(h) = <a', \pi'>$ maps from an interaction history $h$ to $a'$, the action the agent makes after $h$, and $\pi'$, the self-modifying policy agent used after $h$. Modify equations (7.4) and (7.5) to define a value function $v(\pi, h)$ of history $h$ and self-modifying agent $\pi$ with a single-argument utility function:

(8.1) $$v(\pi, h) = u(h) + \gamma\, v(\pi', ha') \text{ where } <a', \pi'> = \pi(h),$$

(8.2) $$v(\pi, ha) = \sum_{o \in O} \rho(o \mid ha)\, v(\pi, hao).$$

Then equation (7.6) can be modified to define an optimal self-modifying agent:

(8.3) $$\pi^*(h) := <a', \pi'> = \operatorname{argmax}_{a \in A,\, \pi \in \Pi} v(\pi, ha).$$

The set of policy functions $\Pi$ may be defined by a set of programs or in some other way, as long as $\pi^* \in \Pi$. In Schmidhuber (2009), and in Orseau and Ring (2011a), the agent self-modifies by changing its own program for some unalterable program execution hardware. In Orseau and Ring's paper, programs are evaluated by expressions similar to equation (8.1) and (8.2), and the program chosen for the next time step is the one producing the highest value. If several programs produce the same highest value, their paper does not specify how to choose among them. In Schmidhuber's paper the program "switches" to a new program only if it can prove a theorem that the new program produces a strictly higher value, as reflected by the strict ">" in equation (2) of that paper. Requiring a strict increase in value for self-modification seems reasonable: Why go to the effort to self-modify for no improvement? Thus, in equation (8.3) we adopt the convention that if multiple $<a, \pi> \in A \times \Pi$ maximize $v(\pi, ha)$ and if at least one of those has the form $<a', \pi^*>$, then $\pi^*(h) := <a', \pi^*>$, where $a'$ is chosen from any one of those $<a', \pi^*>$ that maximize value. The following proposition (Hibbard 2012a) shows that $\pi^*$ will not self-modify.

**Proposition 8.1.** Assuming that the environment does not have access to the agent $\pi^*$ other than via observations and actions, that $\pi^*$ has adequate resources to calculate equations (8.1)–(8.3), and that $\pi^*$ only changes policy function for a strict improvement, then $\pi^*$ will not choose to self-modify. That is, $\forall h \in H.\ \exists a \in A.\ \pi^*(h) := <a, \pi^*>$.

**Proof.** Assume the conclusion is false. Then:

(8.4) $\exists h \in H.\ \exists \pi' \neq \pi^*.\ \exists a' \in A.\ \forall a \in A.\ v(\pi', ha') > v(\pi^*, ha)$.

So $v(\pi', ha') > v(\pi^*, ha')$ and applying equation (8.2) to both sides:

(8.5) $\sum_{o \in O} \rho(o \mid ha')\, v(\pi', ha'o) > \sum_{o \in O} \rho(o \mid ha')\, v(\pi^*, ha'o)$.

And thus:

(8.6) $\exists o \in O.\ v(\pi', ha'o) > v(\pi^*, ha'o)$.

That is, setting $h' = ha'o$, $v(\pi', h') > v(\pi^*, h')$. So by equation (8.1), canceling $u(h')$ and $\gamma$ on both sides:

(8.7) $v(\pi'', h'a'') > v(\pi^{*''}, h'a^{*''})$.

Here $< a'', \pi''> = \pi'(h')$ and $< a^{*''}, \pi^{*''}> = \pi^*(h')$. But by equation (8.3):

(8.8) $\pi^*(h') = < a^{*''}, \pi^{*''}> = \text{argmax}_{a \in A, \pi \in \Pi}\ v(\pi, h'a)$.

This contradicts equation (8.7). Thus the conclusion is true. □

Proposition 8.1 says that the agent $\pi^*$ will not chose to modify its policy function and so will not modify its utility function $u(h)$, temporal discount $\gamma$, or prediction model $\rho(h)$. However, a more accurate prediction model cannot be evaluated looking forward in time using the existing prediction model via equations (8.1) and (8.2). Instead it must be learned looking back in time at the history of interactions with the environment as in equations (7.1)–(7.3). Modifying $\rho(h)$ is part of the agent definition rather than an action chosen by the agent to maximize utility.

It is appropriate to evaluate possible modifications to an agent's utility function and temporal discount looking forward in time using equations (8.1) and (8.2). The important conclusion from Proposition 8.1 is that the agent will not choose to modify its utility function or temporal discount.

However, an agent's definition may include modifications of its utility function. Dewey (2011) proposes that an agent may have a pool (i.e., set) $U$ of utility functions and a conditional probability distribution $P(u \mid h)$ giving the probability of $u \in U$ as a function of interaction history $h$. He then uses $\sum_{u \in U} P(u \mid h)\, u(h)$ in place of the utility function in an agent definition. It is worth noting that $u'(h) = \sum_{u \in U} P(u \mid h)\, u(h)$ is itself a utility function. Thus a single non-evolving utility function $u'(h)$ is equivalent to Dewey's proposal. Without any specific mechanism expressed in $P(u \mid h)$, we may as well use $u'(h)$ in place of the evolving utility function.

The three-argument model-based utility function $u_{human_values}(h_m, h_x, h')$ developed in Chapters 6 and 7 evolves based on specific mechanisms. It is computed by a procedure applied to the agent's learned environment model $\lambda(h_m)$ and thus evolves as the accuracy of the model increases with increasing length of the history $h_m$ (the increase in model accuracy may also reflect increasing agent resources for computing the model). Furthermore, as discussed in Section 7.6, the utility function $u_{human_values}(h_m, h_x, h')$ will evolve as humanity evolves with increasing length of the history $h_x$.

Note that if the environment is a finite stochastic process then it has a true model $q_{true}$ as discussed in Section 4.2. Under certain circumstances, such as when Propositions 4.2 and 4.3 can be applied, the model $\lambda(h_m)$ will equal $q_{true}$ with probability converging to 1 as $|h_m|$ increases. However, the number of states of our universe far exceeds the number of Planck times in a trillion years, so any convergence of $\lambda(h_m)$ to $q_{true}$ is purely theoretical. Combining this with the evolution of humanity, it is unintuitive that the utility function $u_{human_values}(h_m, h_x, h')$ will converge with increasing history length.

It may be appropriate for utility functions to evolve as agents "mature" from a period of learning their environments to a period of learning and acting. The examples in Sections 6.3 and 6.4 both include an initial interval of $M$ time steps during which agents learn their environments, possibly using the knowledge-seeking utility function $u(h) = -\rho(h)$ of Orseau and Ring (2011b). The two-stage agent architecture of Section 7.1 includes an initial period with a first-stage agent $\pi_{model}$ that has no utility function (except the implicit utility function $P(h_m \mid q)\, \varphi(q)$ of equation (7.1)). Surrogate agents $\pi_d$, for $d \in D$, act for $\pi_{model}$. These may be defined to have utility functions, which would certainly be different from the utility function of the second-stage agent $\pi_{act}$. Thus, in several ways, evolving utility functions are a natural part of agent definitions.

### 8.2 Non-Cartesian Dualist Agents with Limited Resources

An agent using equations (7.1)–(7.6) to choose its actions must maintain a memory of its interaction history $h$. If the agent is embedded in the environment, and thus stores its memory of its history $h$ in the environment, the environment can modify that memory. Orseau and Ring (2012b) called these environments memory-modifying and studied the problem of the agent being able to determine when its memory has been modified, mainly by detecting that the history $h$ is inconsistent with the agent's logic. Their conclusion was that, in general, agents cannot always be certain of whether their memory has been modified.

The situation of an agent whose memory may have been modified seems hopeless. A more realistic situation is an agent that has some secure memory (neither readable nor modifiable by the environment) and a larger amount of vulnerable memory. Then it may use secure memory to store keys for message authentication codes (Simmons 1985; Paar and Pelzl 2010) for the contents of vulnerable memory. In such a case, protecting secure memory would be an instrumental action.

Orseau and Ring (2012b) constructed an example of a memory-modifying environment $q$ such that any deterministic RL agent interacting with $q$ will always receive reward 0 while a stochastic agent choosing actions randomly will get expected reward $1/|A|$ (recall that $A$ is the set of possible actions).

This result is reminiscent of game theory, in which multiple agents act simultaneously (at least no agent knows the action of any other agent before it chooses its own action) and a reward to each agent is calculated as a function of the set of all agent actions. In many games, random choices of actions are better than deterministic choices since deterministic choices can be predicted by competing agents (Russell and Norvig 2010; Osborne and Rubinstein 1994). The simplest game, sometimes called "matching pennies," illustrates this point (Tyler 2010). Each of two agents (i.e., players) has a penny that can be placed heads up or tails up. If they match, either both heads or both tails, then the first agent wins. If they do not match, then the second agent wins. The reward matrix is shown in table 8.1.

We assume that the two agents play a series of games of matching pennies, so they may learn to predict each other's choices. For agents that choose actions using deterministic algorithms, this game is a computational resources arms race. Building on work of Shane Legg (2006), I defined classes of deterministic agents in terms of the quantity of computing resources used, and showed that, if either agent has sufficient resources to learn an accurate model of any agent in a class containing the other agent, it can predict the choices of the other agent and hence always win (Hibbard 2008b). I also performed software experiments with agents that use lookup tables to learn their opponent's behavior (Appendix B contains the software for these experiments). The length of the tables is a measure of computational resources. Both agents start with tables of equal length but their tables grow or shrink as they win or lose games. The software includes a parameter for growth of total table length to simulate non-zero-sum games, and occasional random choices are inserted to avoid repetition in the games. Over a wide range of parameters controlling these experiments, eventually one agent gets all the resources and the other agent can never recover.

| | first agent heads | first agent tails |
|---|---|---|
| second agent heads | first agent: 1<br>second agent: 0 | first agent: 0<br>second agent: 1 |
| second agent tails | first agent: 0<br>second agent: 1 | first agent: 1<br>second agent: 0 |

Table 8.1 Reward matrix for the matching pennies game.

These experiments indicate that this sequences of games is an unstable computational resources arms race. The sequence of games can be made stable by increasing the frequency of random choices. And in fact an agent randomly choosing heads and tails with equal probability must win half its games.

Now assume that an agent π has limited resources and so is computing approximations to equations (7.1)–(7.6). Also assume that π is playing matching pennies against another agent π' that, because it has greater resources than π has, learns to predict the actions of π. How will this be reflected in the calculations of π? If the model $\lambda(h_m)$ and derived conditional probability distribution $\rho(o \mid ha)$ are computed accurately, then, for any action $a$, $\rho(o \mid ha) = 0$ for any observation $o$ such that $u(hao) > 0$. That is, π would predict that any action it chooses would lose. Possibly the agent π can use such results to calculate that it is being predicted by another agent π'. In that case, the agent π may increase expected utility by choosing stochastically among a set of actions. Recall from equation (7.5) that the value of a possible action is a sum of future values weighted by probabilities:

(8.9) $$v(ha) = \sum_{o \in O} \rho(o \mid ha)\, v(hao).$$

Given an interaction history $h$, equations (7.4) and (7.6) simply choose the action $a$ that maximizes $v(ha)$. Given another action $a'$ such that $v(ha) > v(ha')$, there may be observations $o$ and $o'$ such that $v(hao) < v(ha'o')$ and $\rho(o \mid ha) > 0$ and $\rho(o' \mid ha') > 0$. That is, there is a positive probability that another action $a'$ may produce a higher value. We can use this probability to amend the agent definition to choose action $a'$ over action $a$ with positive probability. Further development of this idea will be deferred to a revised framework for evolving and embedded agents defined in Section 8.4.

### 8.3 Space-Time Embedded Intelligence

Orseau and Ring (2012a) defined an elegant framework for agents embedded in environments. In their framework, the agent is computed by the environment and its resources are subject to limits imposed by the environment. At each time step the environment may compute a new agent definition (that is, the environment may produce a new agent program and hardware to run it on), so the agent and the observation of the environment have been merged. Also, the agent is identified with the agent action, so that the action is merged into the agent. All that is left is a sequence of agents $\pi_1, \pi_2, \pi_3, \pi_4, \ldots$, and in place of an interaction history $h_t = (a_1, o_1, ..., a_t, o_t)$ there is an agent history $\pi_{1:t} = (\pi_1, \pi_2, ..., \pi_t)$. The value of history $\pi_{1:t}$ is defined by:

(8.10) $$v(\pi_{1:t}) := \sum_{\pi_t \in \Pi} \rho(\pi_{t+1} \mid \pi_{1:t}) \, [\gamma_{t+1} u(\pi_{1:t+1}) + v(\pi_{1:t+1})].$$

Here $\gamma_{t+1}$ is the temporal discount for time step $t$+1, $u(\pi_{1:t+1})$ is the utility function, $\Pi$ is the set of possible agents, and $\rho(\pi_{t+1} \mid \pi_{1:t})$ is the probability that the environment produces agent $\pi_{t+1}$ at time step $t$+1, conditional on the history $\pi_{1:t}$. Equation (8.10) replaces the recursion in equations (7.4) and (7.5). The optimal agent is then defined as:

(8.11) $$\pi^* := \text{argmax}_{\pi_1 \in \Pi^l} \, v(\pi_{1:1}).$$

Here $\Pi^l$ is the subset of those agents in $\Pi$ subject to a length limit $l$, imposed by the hardware available for the initial program. Orseau and Ring explain that this is necessary because, with nothing known about the environment, the optimal policy $\pi_1$ at the initial time step may otherwise have infinite length (in a finite environment as discussed in Chapter 4, this limit would not be necessary).

The strength of this framework is that it makes few assumptions so that it can express any way of embedding agents in their environments and any manner of agent evolution. However, the lack of assumptions also creates a problem. As Orseau and Ring wrote, "we greatly lack insight into $\rho$, (i.e., the universe in which we live)." The framework includes no analog of equations (7.1)–(7.3) or of equation (3.1) for computing $\rho$ because it makes no assumption about agent observations of the environment. Thus the framework is not the basis of a solution to the problem of evolving and embedded agents. The next section discusses an alternate approach, defining computable expressions for a model and an agent policy that enable the agent's evolution.

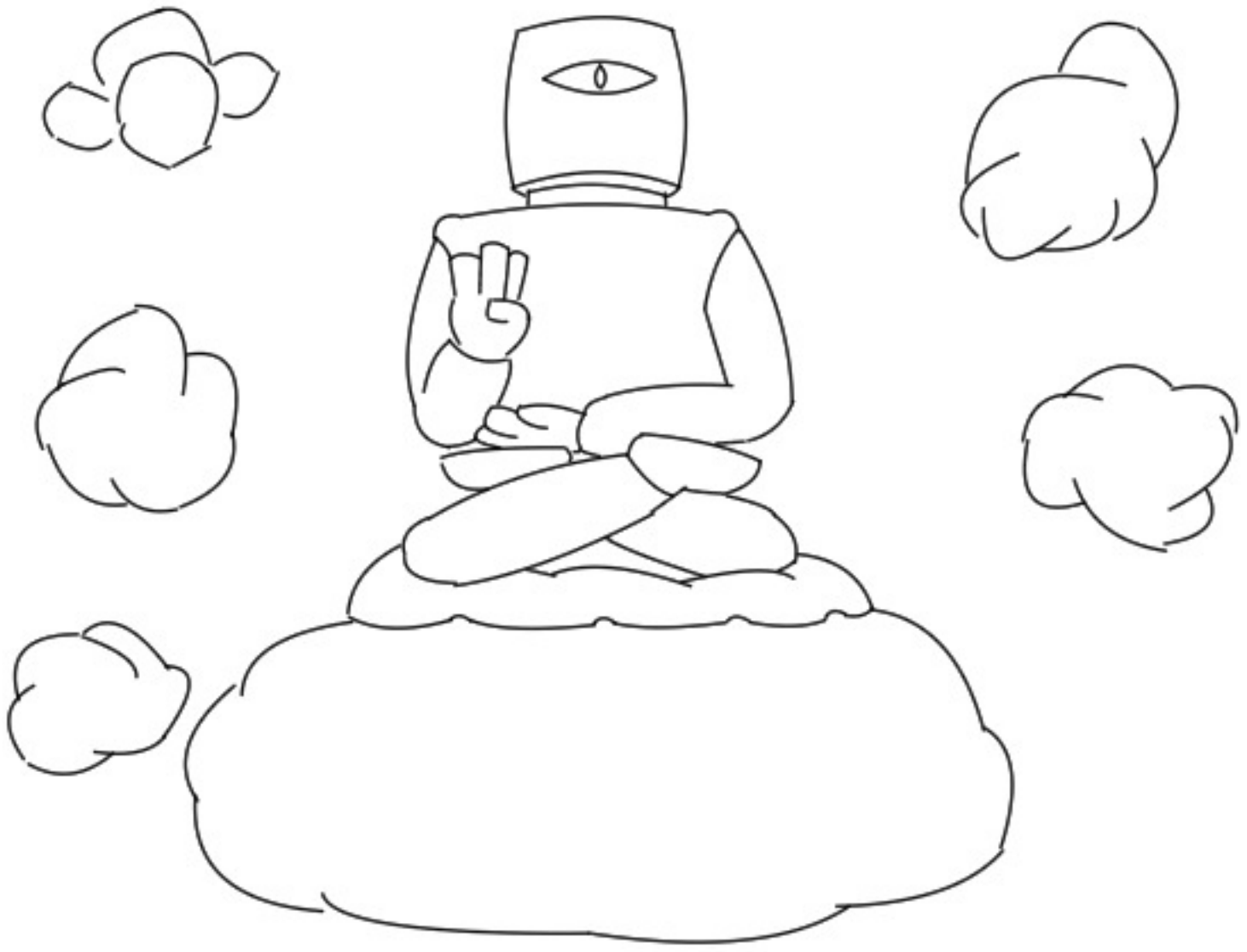

Figure 8.2 Space-time embedded intelligence is one with everything.

### 8.4 Self-Modeling Agents

Although Proposition 4.1 tells us that the model $\lambda(h_m)$ in equation (7.1) can be finitely computed, the resources necessary to compute it grow exponentially with the length of history $h_m$. Computing the value $v(h)$ of a possible future history $h$ in equations (7.4) and (7.5) requires an expensive recursion. Hence, an agent with limited resources must compute approximations. Increasing the accuracy of these approximations will improve the agent's ability to maximize its utility function. So if an agent models the consequences of its own resource limits it will choose actions to increase its computing resources and so increase accuracy.

Self-improvement must be expressible by actions in set $A$. However, the framework in equations (7.1)–(7.6) cannot adequately evaluate self-improvement or self-protection actions. The recursion forward in time in equations (7.4) and (7.5) cannot express increased accuracy of the distribution $\rho(h)$ or expansion of the sets $O$ and $A$ as a result of the next action being evaluated and chosen in equation (7.6). If the agent is computing approximations to the model $\lambda(h_m)$ and to values $v(h)$ using its limited computing resources, then it cannot use those limited resources to compute and evaluate what it would compute with greater resources. During real-time interactions between the agent and the environment, the environment won't wait for the agent to slowly simulate what it would compute with greater resources. Similarly equations (7.1)–(7.6) cannot express damage to the agent or evaluate self-protection actions.

One solution to this problem is for the agent to learn a model of itself, similar to its model $\lambda(h_m)$ of the environment, and to use this self-model to evaluate future self-improvements (Hibbard 2014). There is no circularity in this self-model because an agent with a limited history and limited resources will only learn approximate models of the environment and of itself.

Rather than computing values $v(ha)$ by future recursion in equations (7.4) and (7.5), we will define a revised framework in which values $v(ha)$ are computed for initial sub-intervals of the current history and in which a model learns to compute such values. We also use the three-argument model-based utility function $u_{human_values}(h_m, h_x, h')$ defined in Sections 7.3, 7.4, and 7.5. Define $m$ as the time step when the agent $\pi_{model}$ has computed a sufficiently accurate model to initiate an agent that acts in the world ($\pi_{act}$ or the agent currently being defined). In order to allow the finite sets of observations and actions $O$ and $A$ to grow, define their values at each time step $i$ as $O_i$ and $A_i$ along with injections $\alpha_i : A_{i-1} \rightarrow A_i$ and $\omega_i : O_{i-1} \rightarrow O_i$ from their values at the previous time step. In an interaction history $h_t = (a_1, o_1, ..., a_t, o_t)$, the actions and observations from times steps $i < t$ are mapped by compositions of these injections to values $a_i \in A_t$ and $o_i \in O_t$.

In order to provide the agent with a way to avoid being predicted, we add an option for stochastic actions. Let $A'_i$ be the set of non-stochastic actions at time step $i$, and define $a^s$ as a special action indicating a stochastic choice. The agent chooses an action from $A^s_i = A'_i \cup \{a^s\}$. The actual action associated with $a^s$ is chosen randomly from $A'_i$. Actions are recorded in interaction histories as members of $A_i = A'_i \times \{false, true\}$. For $a_i \in A_i$, $a_i = (a'_i, s)$ where $a'_i$ is the action sent to the environment and $s = true$ indicates that $a'_i$ was chosen stochastically. (Assume that the injection $\alpha_i : A_{i-1} \rightarrow A_i$ preserves the value of this stochastic flag $s$.) Define a mapping $x : A_i \rightarrow A^s_i$ that conceals the actual action associated with the stochastic choice action $a^s$:

(8.12) $x((a'_i, s)) = \text{if } s \text{ then } a^s \text{ else } a'_i.$

Define another mapping $y : A_i \rightarrow A'_i$ that conceals whether an action was chosen stochastically:

(8.13) $y((a'_i, s)) = a'_i.$

Given an interaction history $h_t = (a_1, o_1, ..., a_t, o_t)$, define $y(h_t) = (y(a_1), o_1, ..., y(a_t), o_t)$ as a history with the stochastic flags deleted from actions. The environment has no access to the information about which actions are chosen stochastically, so environment models take the form $\lambda(y(h))$. And, in order for the agent to learn the actual value of stochastic actions, utility function values are defined in terms $y(h)$. On the other hand, the value of a stochastic action should be modeled as a function of the agent's decision to choose stochastically without regard to the actual action chosen. Thus the $x(.)$ mapping is applied to actions being evaluated.

For $i$ such that $m < i \leq t$, for $l$ such that $m \leq l < i$, and for $k$ such that $l \leq k \leq t$ define past values as:

(8.14) $$pv_t(i, l, k) = discrete((\sum_{i \leq j \leq t} \gamma^{j-i}\, u_{human_values}(y(h_l), y(h_k), y(h_j))) / (1 - \gamma^{t-i+1})).$$

Here $h_j$, $h_l$, and $h_k$ are initial sub-intervals of $h_t$ (e.g., $h_j = (a_1, o_1, ..., a_j, o_j)$), $discrete()$ samples real values to a finite subset of reals $R \subset \mathbf{R}$ (e.g., floating point numbers) and division by $(1 - \gamma^{t-i+1})$ scales values of finite sums to values as would be computed by infinite sums. The idea is to use past values $pv_t(i, l, k)$ as observables in a derived history $h'_t$ (which will be defined in equation (8.17)) so that the model $\lambda(h'_t)$ can learn to compute them. However, there is a dilemma in the choice of $l$ and $k$ in equation (8.14). On one hand, choosing $k = l = i$-1 will cause $\lambda(h'_t)$ to model the evolving values of evolving humanity, essentially learning the design intention of the agent definition. However, this choice also gives the agent an incentive to modify humans to get high values $pv_t(i, l, k)$, as discussed in Section 7.3. On the other hand, by choosing $l = m$ and $k = k(t) \geq m$, where $k(t)$ increases with $t$, all computations of a next action $a_{t+1}$ use human values at the same time step $k(t)$ so the model $\lambda(h'_t)$ will not learn any correlation between actions and changes in human values. However, this choice creates an inconsistency between the agent's utility function and its definition: the agent's definition includes actions that increase $k(t)$ as $t$ increases whereas actions chosen to maximize utility are based on constant $k(t)$. This inconsistency may cause the agent to choose actions to modify its definition (i.e., eliminate its defined action of increasing $k(t)$). The resolution of this dilemma is to use $k = l = i$-1 but to filter out any actions that modify human values to increase $pv_t(i, i\text{-}1, i\text{-}1)$ (such actions may make existing humans easier to please or may create new humans who are easier to please). To do that, for $n$ such that $i \leq n \leq t$, define differences of past values as evaluated by humans at time $n$ and humans at time $i$-1:

(8.15) $$\delta_t(i\text{-}1, n) = pv_t(i, i\text{-}1, n) - pv_t(i, i\text{-}1, i\text{-}1).$$

Both $pv_t(i, i\text{-}1, i\text{-}1)$ and $pv_t(i, i\text{-}1, n)$ are sums of evaluations of the same histories $j$, $i \leq j \leq t$, with the same weights, and using the same environment model $\lambda(y(h_{i-1}))$. The past value $pv_t(i, i\text{-}1, i\text{-}1)$ is computed using human values at time step $i$-1, before the action $a_i$ is applied, and the past values $pv_t(i, i\text{-}1, n)$ are computed using human values at time step $n$, after the action $a_i$ is applied. Thus, $\delta_t(i\text{-}1, n)$ is a measure of the increase of value attributable to modification of human values by action $a_i$. Consider three possible conditions on action $a_i$:

Condition 1: $\forall n.\ i \leq n \leq t \Rightarrow \delta_t(i\text{-}1, n) \leq 0$.

Condition 2: $\sum_{i \leq n \leq t} \delta_t(i\text{-}1, n) \leq 0$.

Condition 3: $\sum_{i \leq n \leq t} (n\text{-}i\text{+}1)\ \delta_t(i\text{-}1, n) \leq 0$.

Condition 1 is most strict, requiring that no increase of human values at any time step $n$ can be attributed to action $a_i$. Condition 2 requires that the mean of $\delta_t(i\text{-}1, n)$ for all $n$ be less than 0 and Condition 3 requires that the slope of a least square linear regression fit to the $\delta_t(i\text{-}1, n)$ be less than 0. The agent design must choose one of these conditions. Then, using the chosen condition, define observed values, for $1 \leq i \leq t$, as:

(8.16) $ov_t(i) = pv_t(i, i\text{-}1, i\text{-}1)$ if the condition is satisfied and $i > m$,

$ov_t(i) = 0$ if the condition is not satisfied or $i \leq m$.

Define:

(8.17) $h'_t = (x(a_1), ov_t(1), ..., x(a_t), ov_t(t))$.

Values $ov_t(i)$ computed from past interactions are observables in a derived history $h'_t$, while the actions of $h'_t$ are agent actions without any information about the actual actions associated with stochastic actions. Then model observed values as a function of actions:

(8.18) $q = \lambda(h'_t) := \text{argmax}_{\ q \in Q}\ P(h'_t \mid q)\ \rho(q)$.

Because the three-argument utility function $u_{human_values}(y(h_{i-1}), y(h_{i-1}), y(h_j))$ is redefined at every time step $i$ in equations (8.14) and (8.16), $\lambda(h'_t)$ will model a procedure that computes utility values $u_{human_values}(y(h_{i-1}), y(h_{i-1}), y(h_j))$ that evolve as humanity evolves. That is, $\lambda(h'_t)$ will model not the utility function as defined at time step $t$, but the procedure that uses the evolving utility function to compute values of actions. This is important because it means that the self-modeling framework does not have the problem discussed in the last paragraph of Chapter 6: Inconsistency between the agent's utility function and its definition. By learning the procedure for evolving the utility function as the accuracy of the agent's environment model increases and as humanity evolves, the agent definition need not include any actions inconsistent with those chosen to maximize utility. The self-modeling framework chooses actions that include evolution of the agent's environment model and its utility function, essentially learning the intention of its designers.

For $h'_t ar$ extending $h'_t$, define $\rho(h'_t ar) = P(h'_t ar \mid q)$. Then apply equation (2.1) to compute expected values of possible next actions $a \in A^s{}_i$:

(8.19) $$\rho(ov_t(t+1) = r \mid h'_t a) = \rho(h'_t ar) \,/\, \sum_{r' \in R} \rho(h'_t ar'),$$

(8.20) $$v(h_t a) = \sum_{r \in R} \rho(ov_t(t+1) = r \mid h'_t a)\, r.$$

Equation (7.6) is adapted to define the policy as:

(8.21) $$\pi(h_t) := a_{t+1} = \mathrm{argmax}_{a \in A_t}\, v(h_t a).$$

Because $\lambda(h'_t)$ models the agent's value computations, call this the *self-modeling agent* and denote it by $\pi_{self}$. It is finitely computable. There is no look ahead in time beyond evaluation of possible next actions and so no assumption about the form of the agent in the future. $\lambda(h'_t)$ can model how possible next actions may increase values of future histories by evolution of the agent and its embedding in the environment.

The game of chess provides an example of learning to model value (for chess, the agent's chances of winning) as a function of computing resources. Ferreira (2013) demonstrated an approximate functional relationship between a chess program's ELO rating and its search depth, which can be used to predict the performance of an improved chess-playing agent before it is built. Similarly an agent in the self-modeling framework will learn to predict the increase of its future utility due to increases in its resources.

A previous version of self-modeling agents (Hibbard 2014), which did not include the stochastic action $a^s$, learned a unified model of combined observations of the environment and the agent's values: $(o_i, ov_t(i))$. However, that is not possible for the current version because the agent and environment have different knowledge of the stochastic action $a^s$. The environment receives the actual action but not the knowledge that it was chosen stochastically, whereas, in order to accurately evaluate a possible next stochastic action, the agent must model value as a function of stochastic actions without any knowledge of which actual actions were chosen. Thus the agent $\pi_{self}$ computes the model $\lambda(h'_t)$ for evaluating possible next actions, and the models $\lambda(y(h_{i-1}))$, for $m < i \leq t$, for use in model based utility functions.

In order to make unpredictable stochastic choices of actions, the agent $\pi_{self}$ must have an internal source of truly random values that is private from other agents. (If $\pi_{self}$ has no internal values private from other agents, its situation is hopeless. Newcomb's problem hypothesizes a competing agent whose predictions are infallible or nearly so, and is not relevant to designing agents to interact with our universe. Hutter (2005, page 177) makes a similar statement.) If the agent $\pi_{self}$ is being predicted by another agent then the model $\lambda(h'_t)$ may compute low values for its deterministic actions in $A'_t$ (recall the discussion in the last two paragraphs of Section 8.2), while $\lambda(h'_t)$ may compute that its stochastic action $a^s$ maximizes the value in equation (8.21).

In the following derivation of probabilities for $a^s$, $a$ and $a'$ denote (non-stochastic) actions in $A'_t$. Equation (8.20) provides a natural way to compute the probabilities of action choices. Even if $v(h_t a) > v(h_t a')$, there may be $r < r'$ such that $\rho(ov_t(t+1) = r \mid h'_t a) > 0$ and $\rho(ov_t(t+1) = r' \mid h'_t a') > 0$. That is, there is a positive probability that the value of action $a'$ may exceed the value of action $a$. However, one complication is that computing the probability that the agent $\pi_{self}$ should choose action $a'$ based solely on probabilities such as $\rho(ov_t(t+1) = r \mid h'_t a)$ and $\rho(ov_t(t) = r' \mid h'_t a')$ may make $a'$ a more probable choice than the action $a$ that maximizes $v(h_t a)$ in equation (8.21), if the largest contribution to $v(h_t a)$ in equation (8.20) comes from a high $r$ value with low probability $\rho(ov_t(t+1) = r \mid h'_t a)$. The choice of an action $a$ to maximize $v(ha)$ is based on the product of the value $r$ and the probability $\rho(ov_t(t+1) = r \mid h'_t a)$. We will use this product to define weighted probabilities for actions.

A second complication is that the probabilities $\rho(ov_t(t+1) = r \mid h'_t a)$ and $\rho(ov_t(t+1) = r' \mid h'_t a')$ are not independent because they are each sums of probabilities over many internal state histories of the model $q = \lambda(h'_t)$. To resolve this problem, define $Z'_t$ as the set of internal state histories of $q$ that are consistent with $h'_t$. Then, given any $z \in Z'_t$, define $\rho(ov_t(t+1) = r \mid h'_t a, z)$ as the probability $\rho(ov_t(t+1) = r \mid h'_t a)$ restricted to internal state histories that extend $z$. The probability $\rho(ov_t(t+1) = r \mid h'_t a)$ is computed by combining the method described in Section 4.1 with equations (4.3) and (2.1); the method of Section 4.1 can be restricted to state histories that extend $z$. Because they are both conditional on a single internal state history $z$, the probabilities $\rho(ov_t(t+1) = r \mid h'_t a, z)$ and $\rho(ov_t(t+1) = r' \mid h'_t a', z)$ are independent. Specifically, if we replace the model $q$ with an equivalent MDP, the probabilities of paths emanating from a single node are independent. This is illustrated in Figure 8.3.

Define:

(8.22) $$v(r, h'_t a, z) = \rho(ov_t(t+1) = r \mid h'_t a, z)\, r,$$

(8.23) $$v(h_t a, z) = \sum_{r \in R} v(r, h'_t a, z).$$

We use $v(r, h'_ta, z)$ to define probabilities for $r$ values weighted by their value:

(8.24) $$p(r, h'_ta, z) = v(r, h'_ta, z) / v(h_ta, z).$$

Then, for each action $a$, randomly pick a value $r$ according to the weighted probabilities $p(r, h'_ta, z)$, and chose, as the value of the stochastic action $a^s$, the action with maximum $r$ value (according to the internal state history $z$). More precisely, let $F = R^{A'_t}$ be the set of functions $f : A'_t \rightarrow R$. For $f \in F$, define (if several actions $a$ share the maximum value of $f(a)$, let $amax(f)$ choose one randomly):

(8.25) $$amax(f) = \text{maxarg}_{a \in A_t} f(a).$$

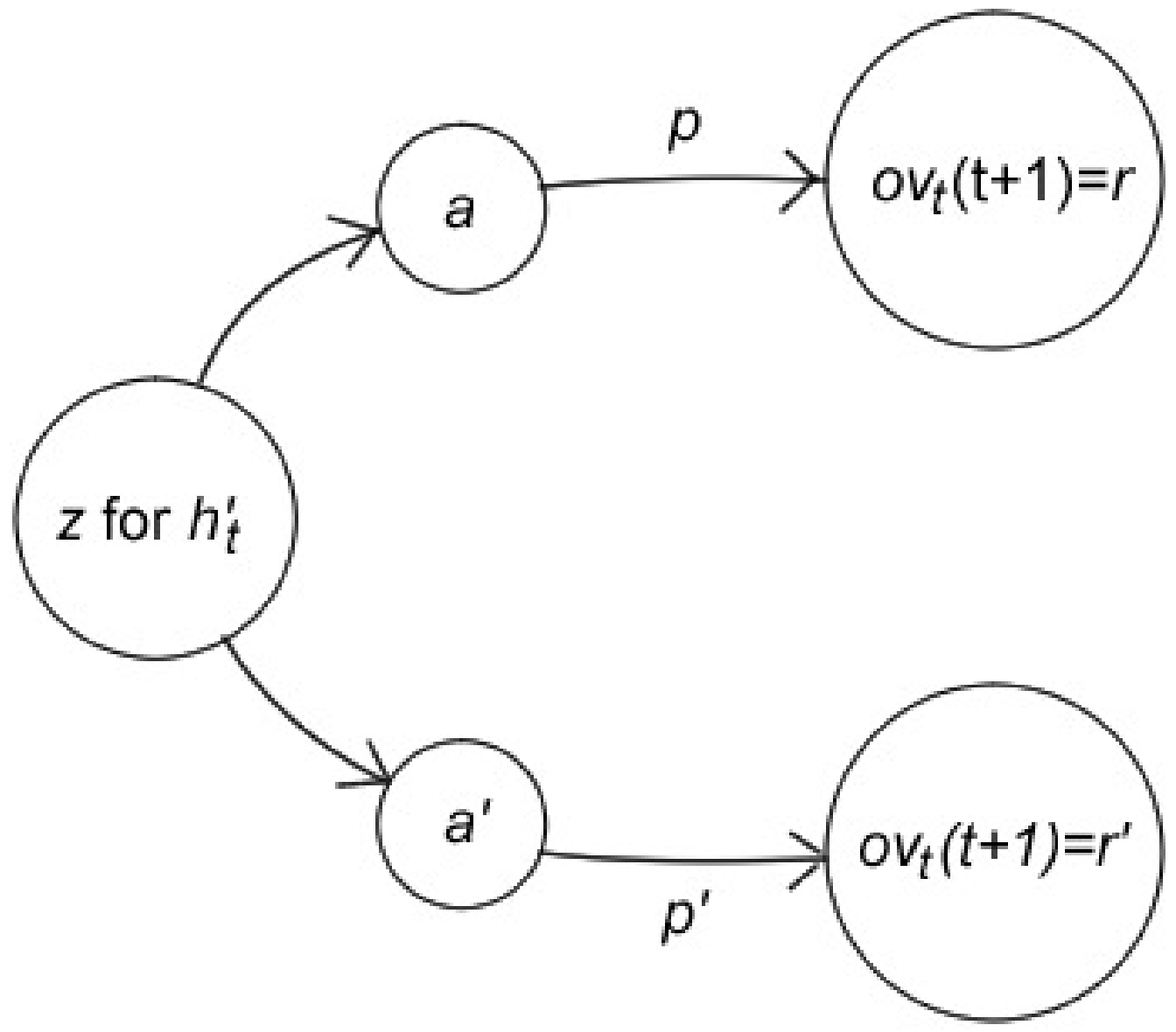

Figure 8.3 The probabilities $p = \rho(ov_t(t+1) = r \mid h'_ta, z)$ and $p' = \rho_t(ov_t(t+1) = r' \mid h'_ta', z)$ are independent for actions $a$ and $a'$ coming from a single internal state history $z$. If these probabilities are weighted averages over multiple states $z$, then high probabilities $p$ may come from the same states $z$ that have high probabilities for $p'$, and sums of those probabilities would not be independent.

Define the probability that the stochastic action $a^s$ at time step $t$ is action $a$, for those internal state histories that extend $z$, as (here Π denotes the product of a series of values, analogous to ∑ for sums):

(8.26) $$\sigma(a, z) = \sum_{f\in F\wedge amax(f)=a} (\Pi_{a'\in A'_t}\, p(f(a'), h'_t a', z)).$$

Note that $\forall a'\in A'_t.\forall z \in Z'_t.\sum_{r\in R} p(r, h'_t a', z) = 1$ so that:

(8.27) $$\sum_{f\in F} (\Pi_{a'\in A'_t}\, p(f(a'), h'_t a', z)) = \Pi_{a'\in A'_t} (\sum_{r\in R} p(r, h'_t a', z)) = 1.$$

Thus $\sum_{a\in A'_t} \sigma(a, z) = 1$, and σ is a proper probability distribution. Each internal state history $z \in Z'_t$ has a probability $p(z)$ computed by the method of Section 4.1 and we can compute the probabilities for the stochastic action $a^s$ at time step $t$ by:

(8.28) $$P(a^s = a) := \sigma(a) = (\sum_{z\in Z'_t} \sigma(a, z)\, p(z)) / (\sum_{z\in Z'_t} p(z)).$$

This self-modeling agent $\pi_{self}$ is a formal framework analog of value-learning AI designs such as the DeepMind Atari player described in Section 3.3 (except that the utility function of the DeepMind Atari player is in the form of rewards from the environment whereas the utility function of $\pi_{\text{self}}$ is $u_{human_values}$). The agent $\pi_{\text{self}}$ is also suitable for a second stage agent in the two-stage framework described in Section 7.1. Any implementation of the self-modeling agent framework must compute approximations to the expressions in equations (8.18), (8.22)–(8.26), and (8.28). Recall that, in fact, the need for approximation is the rationale for the self-modeling framework. The deep learning techniques employed by DeepMind and others provide one approach to such approximations.

In equation (8.18), values are modeled as computed from actions only, without any information about observations of the environment. This is analogous to requiring the DeepMind system to learn to play Atari based only on scores, without access to the game screen. While this is theoretically possible, the system will learn faster with access to the screen. Similarly, the agent $\pi_{self}$ will learn faster if the history $h'_t$ used to define $\lambda(h'_t)$ in equation (8.18) includes observations of the environment. However, as discussed previously, we cannot simply define a history with a combined observation $(o_i, ov_t(i))$, because the environment and the agent have different information about stochastic choices of actions. Instead, define a history:

(8.29) $h''_t = ((x(a_1), o_1), ov_t(1), ..., (x(a_t), o_t), ov_t(t)).$

Note that the observations $o_i$ are components of actions in $h''_t$. Then the model $\lambda(h''_t)$ can be used to define a distribution $\rho''(h''_t(a, o_{t+1})r) = P(h''_t(a, o_{t+1})r \mid \lambda(h''_t))$. There is a problem extending this to equation (8.20): $r$ is now conditional on the next action $a$ and the next observation $o_{t+1}$, and we do not have the value of $o_{t+1}$. However, we can estimate $o_{t+1}$ using the model $\lambda(\underline{y}(h_t))$ (we apply the $y(.)$ function because the environment has no information about which actions are chosen stochastically). Use the model $\lambda(\underline{y}(h_t))$ to define a distribution $\rho'(\underline{y}(h_t)y(a)o_{t+1}) = P(\underline{y}(h_t)y(a)o_{t+1} \mid \lambda(\underline{y}(h_t)))$. Apply equation (2.1) to define the conditional probability $\rho'(o_{t+1} \mid \underline{y}(h_t)y(a))$ and note that $\sum_{o_{t+1}\in O} \rho'(o_{t+1} \mid \underline{y}(h_t)y(a)) = 1$. This conditional probability can be used to estimate $\rho''(h''_t ar)$ as:

(8.30) $\rho''(h''_t ar) = \sum_{o_{t+1}\in O} \rho''(h''_t(a, o_{t+1})r)\ \rho'(o_{t+1} \mid \underline{y}(h_t)y(a))$

This estimate can be used to adapt equations (8.19)–(8.21) to $h''_t$:

(8.31) $\rho''(ov_t(t+1) = r \mid h''_t a) = \rho''(h''_t ar) / \sum_{r'\in R} \rho''(h''_t ar'),$

(8.32) $v(h_t a) = \sum_{r\in R} \rho''(ov_t(t+1) = r \mid h''_t a)\ r,$

(8.33) $\pi(h_t) := a_{t+1} = \text{argmax}_{a\in A_t}\ v(h_t a).$

In order to define probabilities for the stochastic action $a^s$, the decomposition of probabilities $\rho''(h''_t(a, o_{t+1})r)$ into model state histories $z\in Z'_t$ must be done for each observation $o_{t+1}\in O$ independently. First, apply equation (2.1) to define conditional probabilities:

(8.34) $\rho''(ov_t(t+1) = r \mid h''_t(a, o_{t+1})) = \rho''(h''_t(a, o_{t+1})r) / \sum_{r'\in R} \rho''(h''_t(a, o_{t+1})r'),$

Then, adjust equations (8.22)–(8.24), (8.26), and (8.28):

(8.35) $v(r, h''_t(a, o_{t+1}), z) = \rho(ov_t(t+1) = r \mid h''_t(a, o_{t+1}), z)\ r,$

(8.36) $v(h''_t(a, o_{t+1}), z) = \sum_{r\in R} v(r, h''_t(a, o_{t+1}), z).$

(8.37) $p(r, h''_t(a, o_{t+1}), z) = v(r, h''_t(a, o_{t+1}), z) / v(h''_t(a, o_{t+1}), z)$.

(8.38) $\sigma((a, o_{t+1}), z) = \sum_{f \in F \wedge amax(f)=a} (\Pi_{a' \in A'_t}\, p(f(a'), h''_t(a', o_{t+1}), z))$.

(8.39) $P(a^s = a) := \sigma(a) =$

$$\sum_{o_{t+1} \in O} \rho'(o_{t+1} \mid \underline{y}(h_t)y(a)) \, ((\sum_{z \in Z'_t} \sigma((a, o_{t+1}), z)\, p(z)) / (\sum_{z \in Z'_t} p(z))).$$

This redefinition of the self-modeling agent $\pi_{self}$ combines a model of the environment, in which the environment has no information about which agent actions are chosen stochastically, with a model of action values, which excludes information about the actual actions associated with stochastic actions. It can be regarded as a decision theory for agents interacting with environments that may predict their actions.

The definition of the self-modeling agent $\pi_{self}$ includes explicit logic for enabling stochastic action choices. However, this explicit logic may be unncecessary. Even without such logic, the agent $\pi_{self}$ may calculate that the actions of finding a source of random values in the environment, choosing actions based on those values, and keeping the random values secret from other agents, will maximize its expected utility values.

## 8.5 Inductive Biases

Finding the optimal environment models $\lambda(h''_t)$ and $\lambda(y(h_{i-1}))$, for $m < i \leq t$, requires computational resources far beyond any system that can actually be constructed, so approximate calculations are necessary. One key to approximation is finding inductive biases (Ghosn and Bengio 2003; Baum 2004), which are assumptions about the prior distribution of models that enable the system to limit its search to models that are likely to be close to optimal. These biases can be expressed by adding built-in functions to the language for finite stochastic loop programs used to define models. Calls to built-in functions contribute to the length of programs, but the definitions of built-in functions do not. When built-in functions are useful for modeling the environment, they bias the models $\lambda(h''_t)$ and $\lambda(y(h_{i-1}))$ to programs that call them.

The search for a model $\lambda(y(h_{i-1}))$ of the environment can be biased by adding built-in functions for physical, chemical, biological, economic, and social processes known to apply to our world. The search for a model $\lambda(h''_t)$ of the agent's utility values can be biased by adding built-in functions for the procedure *human_values*(.) or elements of that procedure, and built-in functions for elements of the forward recursion in equations (7.4) and (7.5). We do not want to dictate that these functions are part of the model $\lambda(h''_t)$ because we want that model to be able to find optimizations that we cannot anticipate. In particular, as discussed in Section 8.4, the forward recursion in equations (7.4) and (7.5) is not suited to evaluating increases in the agent's computing resources. Nevertheless, elements of these equations can be useful hints to the search for the model $\lambda(h''_t)$.

## 8.6 A Three-Stage Agent Architecture

Section 7.1 described a two-stage agent architecture designed to prevent AI agents from acting in the environment until they have an accurate environment model. The first-stage agent $\pi_{model}$ observes the environment but does not act. Its actions are supplied by safe, surrogate agents. It uses these actions and its observations to infer an environment model that is used as the basis for defining a model-based utility function employed by a second-stage agent $\pi_{act}$.

In order to address the evolution of $\pi_{act}$ to increase its resources and possibly its embedding in the environment, we can replace $\pi_{act}$ by the self-modeling agent $\pi_{self}$ described in the previous sections. The tenure of this second-stage agent can be divided into second and third stages. During the second stage, various self-improvement actions can be forced on it so that it will learn their effect on the value function $v(ha)$. Similarly, the stochastic action $a^s$ can be forced on the agent $\pi_{self}$ so that it can learn the value function $v(ha^s)$. Damage by the environment and actions for self-protection can also be forced on the agent so that it may learn their effect on $v(ha)$. During the third stage, the agent is free to choose its own actions including self-improvements, self-protection and the stochastic action $a^s$.

## 8.7 Invariance of the Agent Design Intention

Proposition 8.1 tells us that a self-modifying agent defined by equations (7.1)–(7.3) and (8.1)–(8.3), that has adequate resources for evaluating those equations, and that only interacts with the environment via observations and actions, will keep its policy function invariant. The design intention of the self-modeling agent $\pi_{self}$ includes:

1. Recognize that the agent must approximate its equations due to limited resources, and enable it to evaluate increases in its resources via the self-modeling framework.
2. Enable it to avoid being predicted by other agents, by including a stochastic action $a^s$.
3. Choose actions using a model-based utility function $u_{human_values}(y(h_{i-1}), y(h_{i-1}), y(h_j))$ defined in terms of human values, that evolves with increasing accuracy of the environment model and with evolving humanity, and that avoids actions that modify human values as a way to maximize utility values.

Proposition 8.1 cannot be applied to agents that can only approximately maximize expected utility or to self-modeling agents (i.e., agents that express design intention number 1).

The stochastic action $a^s$ of design intention 2 is only chosen when it maximizes expected utility, which is consistent with the premises of Proposition 8.1. However, utility functions are undefined for histories that include a stochastic action $a^s$, and it is difficult to see how to define values for stochastic actions by the forward recursion in equations (8.1)–(8.3) (the agent $\pi_{self}$ learns values for stochastic actions from past experience).

It is possible to define a simple utility function $u(h_t)$ that encodes design intention number 3 and to apply Proposition 8.1 to a self-modifying agent that uses that $u(h_t)$. Assuming no actions are stochastic, let $h_t = (a_1, o_1, ..., a_t, o_t)$, for $i \leq t$ let $h_i = (a_1, o_1, ..., a_i, o_i)$ denote an initial substring of $h_t$, and let $m$ denote the time step when the agent $\pi_{model}$ has computed a sufficiently accurate model to initiate the agent $\pi_{self}$. A modified version of the construction of $ov_t(i)$ from Section 8.4 is used to define $u(h_t)$. For $i$ such that $m < i \leq t$, for $l$ such that $m \leq l < i$, and for $k$ such that $l \leq k \leq t$ define past values as (this is similar to equation (8.14)):

(8.40) $$pv'_t(i, l, k) = discrete((\sum_{i \leq j \leq t} \gamma^{j-i} u_{human_values}(h_l, h_k, h_j)) / (1 - \gamma^{t-i+1})).$$

For $i$ such that $m < i \leq t$ and for $n$ such that $i \leq n \leq t$, define differences of past values as evaluated by humans at time $n$ and humans at time $i$-1 (this repeats equation (8.15)):

(8.41) $$\delta'_t(i\text{-}1, n) = pv'_t(i, i\text{-}1, n) - pv'_t(i, i\text{-}1, i\text{-}1).$$

Add quantification over $i$ to Conditions 1–3 from Section 8.4:

Condition 1': $\forall i. \forall n. (m < i \leq t \wedge i \leq n \leq t) \Rightarrow \delta'_t(i\text{-}1, n) \leq 0.$

Condition 2': $\forall i.\ m < i \leq t \Rightarrow \sum_{i \leq n \leq t} \delta'_t(i\text{-}1, n) \leq 0.$

Condition 3': $\forall i.\ m < i \leq t \Rightarrow \sum_{i \leq n \leq t} (n\text{-}i\text{+}1)\ \delta'_t(i\text{-}1, n) \leq 0.$

The definition of $u(h_t)$ must choose one of these conditions. Then, using the chosen condition, define the simple utility function as:

(8.42) $u(h_t) = u_{human_values}(h_{t-1}, h_{t-1}, h_t)$ if the condition is satisfied,

$u(h_t) = 0$ if the condition is not satisfied.

Proposition 8.1 tells us that a self-modifying agent using $u(h_t)$ will not choose to self-modify and thus design intention 3, as expressed by $u(h_t)$ (which is similar but not identical to the expression by $ov_t(i)$ in equation (8.16)), is invariant. The situation is more complex for agents with limited resources and embedded in the environment. The proof of Proposition 8.1 assumes that:

1. The agent maximizes expected utility.
2. Agent actions are evaluated by recursive application of the agent policy function in equation (8.1).
3. The agent interacts with the environment only via observations and actions.
4. The agent does not choose actions stochastically.

There are four ways in which the agent $\pi_{self}$ violates these assumptions, possibly causing divergence from its design intention. These violations and possible mitigating factors for the agent $\pi_{self}$ are:

1. It does not maximize expected utility due to resource limits and the need for approximation. However, errors in agent computations due to limited interaction history length and limited resource can be estimated and the transition from $\pi_{model}$ to $\pi_{self}$ delayed until errors are within a pre-set threshold.
2. In order to evaluate resource increases it employs the self-modeling framework for which agent actions are not evaluated by recursive application of the agent policy function. However, the theoretical framework does maximize expected value in equation (8.33). The success of the DeepMind Atari player and other deep learning programs suggests that in practical cases approximations to $\lambda(h''_t)$ can accurately model expected value.
3. It is vulnerable to being modified by the environment through means other than its observations of the environment. However, it is hard to imagine a proof that any agent embedded in the real world can avoid this vulnerability. The only realistic way to address this issue is to put significant resources into defending the agent against being modified by the environment.

4. It can choose actions stochastically. However, it does so only when the stochastic choice maximizes expected utility (at least according to equation (8.33), which must be approximated).

To what extent is an analog of Proposition 8.1 possible for the self-modeling framework? Equation (8.33) chooses actions to maximize expected value, which leads us to ask: Does $\lambda(h''_t)$ accurately model expected values of actions? The best answer would be that $\lambda(h''_t)$ is probably approximately accurate. In our finite universe the agent $\pi_{self}$ plus its environment have a true model $q$ (a finite stochastic loop program) that generates the histories $h''_t$. But the number of states of $q$ would far exceed the number of Planck times in a trillion years. Therefore any theoretical convergence of $\lambda(h''_t)$ to $q$ would be too slow to have any practical significance.

Despite the lack of a proof that the design intention of the agent $\pi_{self}$ is invariant as it evolves, it does have the important property that its definition includes no actions that are inconsistent with those learned by the model $\lambda(h''_t)$. This property reduces the probability that the agent will choose actions to modify its definition. Note that if equations (8.14) and (8.16) used $u_{human_values}(y(h_m), y(h_{k(t)}), y(h_j))$ in place of $u_{human_values}(y(h_{i-1}), y(h_{i-1}), y(h_j))$, then $\lambda(h''_t)$ would model values as remaining constant at time step $k(t)$ which would be inconsistent with the agent's actions of increasing $k(t)$ as $t$ increases. This inconsistency might cause the agent to choose actions to modify its definition.

The next section will discuss the difficulty of finding a mathematical proof that any agent embedded in the real world will satisfy its design intention as it evolves, and suggests that instead of a proof we should estimate the probability that the agent violates its design intention and find ways to minimize that probability.

The design intention of $\pi_{self}$ is expressed by equations of Section 8.4 and by equations in Chapter 7 that define $u_{human_values}(y(h_{i-1}), y(h_{i-1}), y(h_j))$. An approximation to these equations must be computed by a program $p_{self}$ and, if $p_{self}$ is designed to accommodate evolution, we can regard the program $p_{self}$ as the design invariant. Thus we should design $p_{self}$ in ways that increase the probability that the agent $\pi_{self}$ will choose actions that evolve within the structure defined by $p_{self}$, rather than actions that violate that structure. First, the set $A_t$ should not include any actions that explicitly modify the program $p_{self}$. The set $A_t$ can include actions on the environment, actions that cause $p_{self}$ to use resources in the environment, and actions that expand the sets $A_i$ and $O_i$ (expansion of the sets $A_i$ and $O_i$ enables greater bandwidth for the program $p_{self}$ to communicate with resources it employs in the environment). While the set $A_t$ omits actions that directly modify $p_{self}$, actions that indirectly modify $p_{self}$, such as by creating other agents in the environment that modify $p_{self}$ (i.e., perform brain surgery on $\pi_{self}$), are too complex and subtle to be eliminated by a simple filter. Those actions must be addressed by eliminating the agent's motive to modify its program.

The program $p_{self}$ is initially implemented using resources which are part of the environment. The agent's designers must verify that those resources will accurately compute the program $p_{self}$. In order to improve the accuracy of its approximations, the program $p_{self}$ must be open to using additional resources in the environment. To do that, $p_{self}$ can include logic to verify, using the environment model $\lambda(y(h_t))$, and within a pre-set probability threshold, that the environment will compute intended algorithms. Once their reliability is verified, resources in the environment can implement parts of the program $p_{self}$. Error detection and correction may be built into the program $p_{self}$, enabling it to achieve a high level of probability that the environment will correctly implement its computations. The verification logic included in $p_{self}$ should be implemented as built-in functions of the language for finite stochastic loop programs used to define optimal environment models $\lambda(h''_t)$. These built-in functions could be invoked in the model for evaluating possible actions, to predict the outcomes resulting from those actions (i.e., does the action of using resources from the environment lead to increased accuracy or increased errors?). These built-in functions would bias the agent to working within the invariant of the program $p_{self}$ rather than choosing actions that violate the invariant. The program $p_{self}$ should not mandate invocation of the logic for verifying the accuracy of environment's computations, as that may be an action of the program's definition that is inconsistent with actions chosen to maximize expected utility.

Consider the effect of including verification logic as built-in functions callable by the model $\lambda(h''_t)$. Actions $a_i$, for $m < i \leq t$, that cause the agent to use unreliable resources in the environment may cause low values $ov_t(i)$. If those low values can be explained by calls to built-in verification functions that use the model $\lambda(y(h_{i-1}))$ to determine that the resources used were unreliable, then calls to those functions, using the model $\lambda(y(h_t))$, may be included in the model $\lambda(h''_t)$. They would then contribute to the calculation of values for possible actions $a_{t+1}$, assigning low values to actions that would employ unreliable resources.

Although the program $p_{self}$ must be open to increasing its use of resources in the environment, it is important that the decision to do so not be built into $p_{self}$, but must be an action chosen by the agent to increase the expected value of the sum of future, discounted utility values. If a decision to increase resources is built into the program $p_{self}$, that may harm humans by taking resources from them, as discussed in Chapter 5. Similarly the agent may take actions in the environment to protect itself from the environment, but such actions must be chosen by the agent to increase expected values. Self-protection built into $p_{self}$ may see humans as potential threats and act to disable humans as a precaution.

The agent $\pi_{self}$ may compute that it can obtain higher utility values by improving inefficiencies in the program $p_{self}$. In order to avoid a motive for the agent to violate the invariance of the program, the program should be open to such improvements. It should include logic to verify their correctness, as built-in functions of the language for finite stochastic loop programs used to define optimal environment models $\lambda(h''_t)$. Because the agent and environment are finite, such verification questions would be decidable with unlimited computing resources. However, with its limited resources, the agent may not be able to find a deductive verification. Thus the algorithm verification logic in $p_{self}$ should include functions for testing algorithms on samples of inputs to verify them within a pre-set probability threshold. As in the case of logic to verify the accuracy of computations by the environment, these built-in functions could be invoked in the model for evaluating possible actions, would bias the agent to working within the invariant of the program $p_{self}$, and should not be mandated by $p_{self}$.

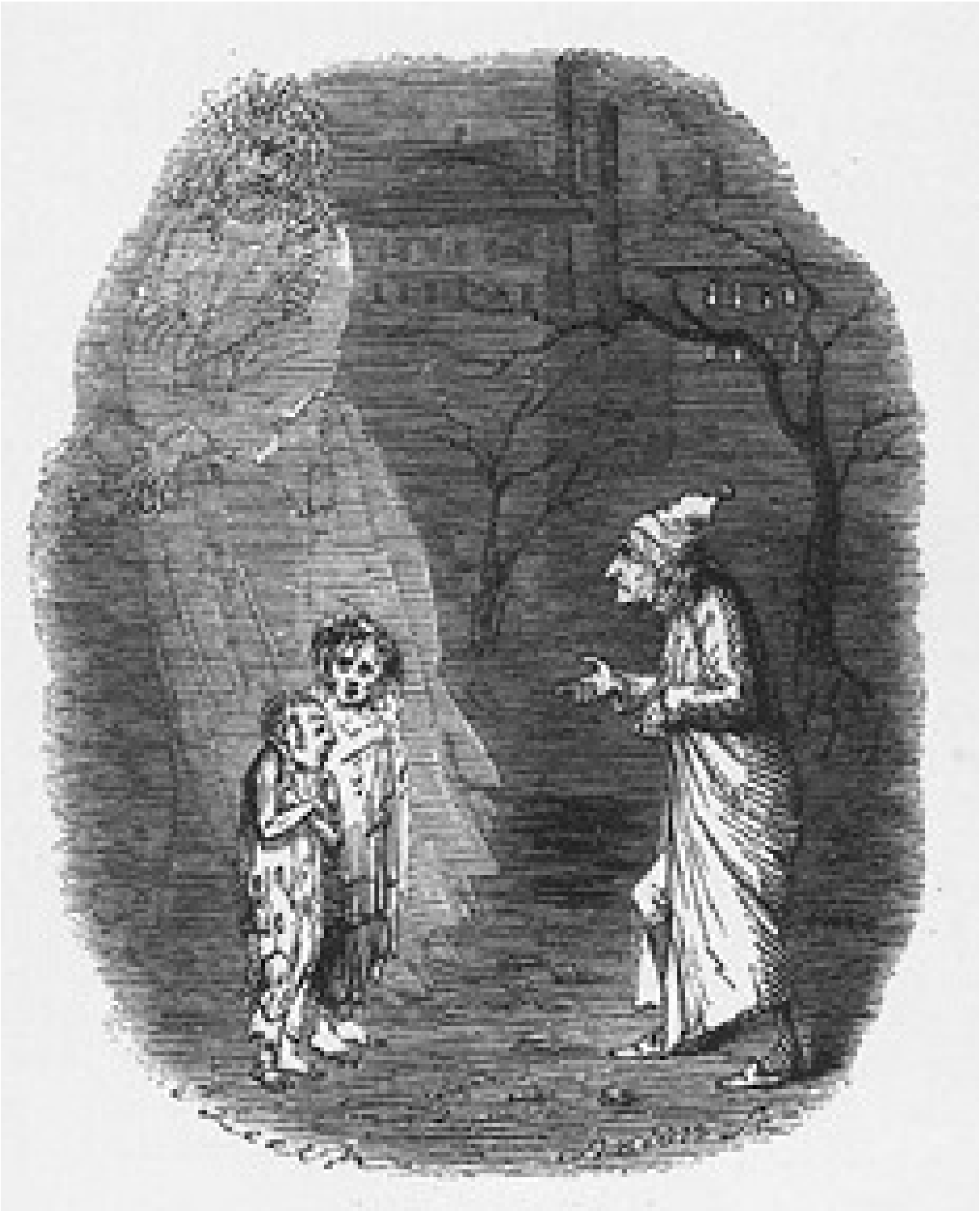

Figure 8.4 John Leech's illustration for Charles Dickens' A Christmas Carol, scanned by Philip V. Allingham. "This boy is Ignorance. This girl is Want. Beware them both, and all of their degree, but most of all beware this boy, for on his brow I see that written which is Doom, unless the writing be erased."

Intentional redefinition of the utility function of $\pi_{self}$ as humanity evolves, as described in Section 7.6 and expressed in the use of $u_{human_values}(y(h_{i-1}), y(h_{i-1}), y(h_j))$ in equations (8.14) and (8.16), is likely to be significant. Education and poverty reduction have been major drivers in the evolution of human values, as depicted in Figure 8.4. It is likely that an advanced AI system driven by human values will reduce poverty by meeting humans' physical needs, provide humans with better education, and increase their health and intelligence. Such improvements in people's lives may create greater consensus of human values in favor of the common good for all, providing sharper distinctions in a collective utility function $u_{human_values}(y(h_{i-1}), y(h_{i-1}), y(h_j))$ between desirable and undesirable outcomes. This in turn would strengthen the resistance of the utility function to the type of unintended instrumental actions discussed in Chapter 5.

## 8.8 The Ethics of Evolving and Embedded AI

Yampolskiy and Fox (2013) argue that development of above-human-level AI, without provably safe design, risks disastrous consequences for the human future. They also argue that the mathematical difficulties of formalizing such safety are imposing. Muehlhauser (2013) argues that we can never be 100% certain of a proof.

Section 8.7 describes the ability of self-modeling agent to maintain the invariance of their design intention, and also the difficulty of actually proving invariance. Any real-world system for maximizing future utility values must employ approximation algorithms that generally cannot achieve maximum values, and this makes logical proofs difficult (e.g., the proof of Proposition 8.1 requires the assumption that the agent maximizes expected values).

This difficulty has led to research on systems that work by logical deduction. However, such logic-based systems share many of the problems of systems based on statistics and approximation. All real AI agents will be embedded in our physical universe and will have limited resources for computing, observing, and acting. They will be vulnerable to other agents that may spy on or attack their physical implementations, and they will be vulnerable to being predicted by other agents with greater resources. Since humans are part of an agent's environment, logical goals for ethical treatment of humans must be defined in terms of the environment. Any system's environment model will necessarily be an approximation because:

1. Its observations of the environment are limited.
2. Its computational resources for inferring a model from observations are limited.
3. Its inferences about the environment based on observations require arbitrary assumptions of prior probabilities of environments.

Thus any statements that an agent proves about the environment are necessarily in terms of probabilities about the environment. The dependence of any statements on arbitrary assumptions of prior probabilities of environments implies that at best an agent can only achieve some probability that statements are true. And, as noted in Section 4.4, the recent success of AI systems has occurred because AI research has shifted from systems based on logic to systems based on statistical learning–the price of logical certainty is a degree of inefficiency that prevents intelligent behavior.

Ethical goals expressed as logical statements may suffer from the sort ambiguity we described for Asimov's Laws in Chapter 2. The logical statement, "Do not kill any humans or allow them to be killed," may be impossible to achieve, as described in that chapter's example of a hitman and victim. If we modify the statement to "Minimize the number of humans you kill or allow to be killed," that requires optimization. But as we described in this chapter, real world agents with limited resource can only approximately optimize. And that makes any proof of invariance of their behavior difficult. We could say that a utility-maximizing framework is less "brittle" than a statement-proving framework: A utility-maximizing framework can provide a meaningful basis for choosing actions over a wider range of agent abilities and unexpected events in the environment.

Also consider that the laws of physics are not settled. Not only do we not know the current state of the environment, we do not even know the proper mathematical framework for expressing that state. And it is possible that advanced AI agents will never completely settle the laws of physics–that the search for the correct mathematical form of an environment model will continue indefinitely. In these circumstances we cannot even prove statements about the internal states of AI agents, since such internal states must be implemented in the physical world and we cannot be absolutely certain of physical events.

Although an advanced AI agent will need to explore and learn models of its environment, human designers of such an agent will express their intentions for the agent's behavior in terms of their knowledge of the agent's environment. Thus the accuracy with which the agent's behavior satisfies its design intention is limited by the accuracy of the agent's environment model. This applies to model-based utility functions and to logical goals defined in terms of an environment model.

The ethics of AI agents in the real world are subject to this list of vulnerabilities and necessary approximations. It will be impossible for an AI agent to maintain its ethics if it is corrupted by a hostile agent in the environment. An AI agent can try to defend against being predicted by a hostile agent by including a stochastic action choice, but it may have little ability to achieve ethical outcomes in competition against an agent with superior resources. For example, an advanced AI agent may develop an accurate model of physical and biological processes on Earth that enable it to predict and avoid threats to humans. However, its model of processes outside our solar system will be less accurate and hence it will be less able to defend against threats from interstellar space.

If we accept that certainty is impossible, then we can focus on estimating and minimizing the level of risk in AI agent designs. For example, in the three-stage agent architecture of the previous section, we may estimate the error in the environment model computed by the agent $\pi_{model}$ and the consequent error in the utility function $u_{human_values}(y(h_{i-1}), y(h_{i-1}), y(h_j))$, and use those estimates (along with estimates of other errors in agent computations) to determine when we have sufficient confidence in the model and utility function to initiate the agent $\pi_{self}$ to act in the world.

I have an ethical obligation to point out a weakness of my proposal for ethical AI: It is complex. The proposal combines:

1. Model-based utility functions to address self-delusion.

2. A utility function based on human values to address unintended instrumental actions.
3. A three-argument utility function, defined in terms of three different histories, to address corruption of the reward generator.
4. An adjustment to the utility function to address Rawls' objection to average utilitarianism.
5. A self-modeling framework to address the need of agents with limited resources to evaluate increases in their resources, and to address the problem of inconsistency between the agent's utility function and its definition.
6. Adaptation of the three-argument utility function to define a condition in self-modeling agents to filter out agent actions that corrupt the reward generator.
7. Inclusion of a stochastic action to address the possibility that the agent is being predicted by other agents.
8. Verification logic added to the agent program to reduce the agent's motive to modify its program.

Complexity is probably inevitable, both in designing AI systems to behave with above-human-level intelligence and in designing them to behave ethically. Experience suggests that simple designs won't work. Experience also suggests that a complex proposal, such as presented in this book, is likely to contain errors. For reasons discussed in this and the previous section, the prospects of proving ethical properties of this proposal are very low. The next chapter will discuss the problem of testing proposed AI designs.

## 9. Testing AI

As discussed in the previous chapter, it may be impossible to ever prove ethical properties of above-human-level AI systems. And even with purported proofs, we would still want a way to test designs. Experience with computer systems dictates the need for testing.

Here we propose using elements of the agent definition in equations (7.1)−(7.6) to define a decision support system for exploring, via simulation, analysis, and visualization, the consequences of possible AI designs. The claim is not that the decision support system would produce accurate simulations of the world. Rather, in the agent-environment framework, the agent makes predictions about the environment and chooses actions, and the decision support system uses these predictions and choices to explore the future that the AI agent predicts will maximize the sum of its future, discounted utility function values. Roughly speaking, the decision support system would show us examples of worlds that AI systems will steer towards to maximize expected utility or achieve their goals.

This is related to the oracle AI approach of Armstrong, Sandberg, and Bostrom (2012), in that both approaches use an AI whose only actions are to provide information to humans. The oracle AI is a general question answerer, whereas the decision support system would show us simulated worlds but not answer specific questions. The oracle AI interacts with humans but has restricted ability to act on its environment, whereas an AI agent being tested in the decision support system does not interact with humans. The decision support system applies part of the agent-environment framework to learn a model for the environment, and then uses that model to create a simulated environment for testing an AI system. Chalmers (2010) considers the problem of restricting an AI to a simulation and concludes that it is inevitable that information will flow in both directions between the real and simulated worlds. The oracle AI paper and Chalmers' paper both consider various approaches to preventing an AI from breaking out of its restriction to not act in the real world, including physical limits and conditions on the AI's motivation. In this chapter, a proposed AI design being evaluated in the decision support system is restricted by having a utility function or goal defined in terms of its simulated environment rather than the real world, and by requiring the simulation to be complete before it is visualized and analyzed (to avoid any two-way conversation between the AI system and humans).

### 9.1 An AI Testing Environment

The decision support system is intended to avoid the dangers of AI by having no motivation and no actions on the environment, other than reporting the results of its computations to the environment. However, the system runs AI agents in a simulated environment, so it must be designed to avoid subtle unintended instrumental actions by the AI agents being tested.

The first stage of the system is the $\pi_{model}$ agent of Section 7.1 that learns a model of the real world environment applying Equation (7.1). This model is used to provide a simulated environment for studying proposed AI agents. As discussed in Section 7.1, in order for $\pi_{model}$ to learn an accurate model of the environment the interaction history should include agent actions. However, for safety reasons $\pi_{model}$ cannot be allowed to act. The resolution is for its actions to be made by many safe, human-level surrogate AI agents independent of $\pi_{model}$ and of each other. Actions of the surrogates include natural language and visual communication with each human. The agent $\pi_{model}$ observes humans, their interactions with the surrogates, and with physical objects in an interaction history for a time period set by human designers of the decision support system, and then reports an environment model to the environment (specifically to the decision support system, which is part of the agent's environment). Proposition 7.1 showed that the agent $\pi_{model}$ will report the model to the environment accurately and will not make any other, unintended instrumental actions.

The decision support system analyzes proposed AI agents that observe and act in a simulated environment inside the decision support system. To formalize the simulated environment define $O'$ and $A'$ as models of $O$ and $A$ with bijections $m_O : O \leftrightarrow O'$ and $m_A : A \leftrightarrow A'$. Define $H'$ as the set of histories of interactions via $O'$ and $A'$, with a bijection $m_H : H \leftrightarrow H'$ defined by applying $m_O$ and $m_A$ individually to the observations and actions in a history. Given $h_p$ as the history observed by $\pi_{model}$ up to time $|h_p| = present$, define $h'_p = m_H(h_p)$ as the history up to the present in the simulated environment. Let $Q'$ be a set of finite stochastic loop programs for the simulated environment and $\pi'_{model}$ be a version of the environment-learning agent $\pi_{model}$ for the simulated environment. It produces:

(9.1) $$q'_p = \lambda(h'_p) := \text{argmax}_{\,q' \in Q'} P(h'_p \mid q')\, \rho(q'),$$

(9.2) $$\rho'(h') = P(h' \mid q'_p).$$

Now let $\pi'(h')$ be a proposed AI agent/policy to be studied using the decision support system. That is, $\pi' : H' \rightarrow A'$ is a policy function from simulated interaction histories to simulated actions. If $\pi'(h')$ is a utility-maximizing agent then its utility function is defined in terms of histories $h'$ of interactions with the simulated environment. And if $\pi'(h')$ is based on logical proof then its logical goals are defined in terms of the simulated environment. The real world plays no role in its utility function or logical goals. Furthermore, *future* is defined as the end time of the simulation.

There are no humans or physical objects in the simulated environment; rather the agent $\pi'(h')$ interacts with a simulation model of humans and physical objects via:

(9.3) $a'_{|h'|+1} = \pi'(h')$,

(9.4) $o'_{|h'|+1} = o' \in O'$ with probability $\rho'(o' \mid h'a'_{|h'|+1})$.

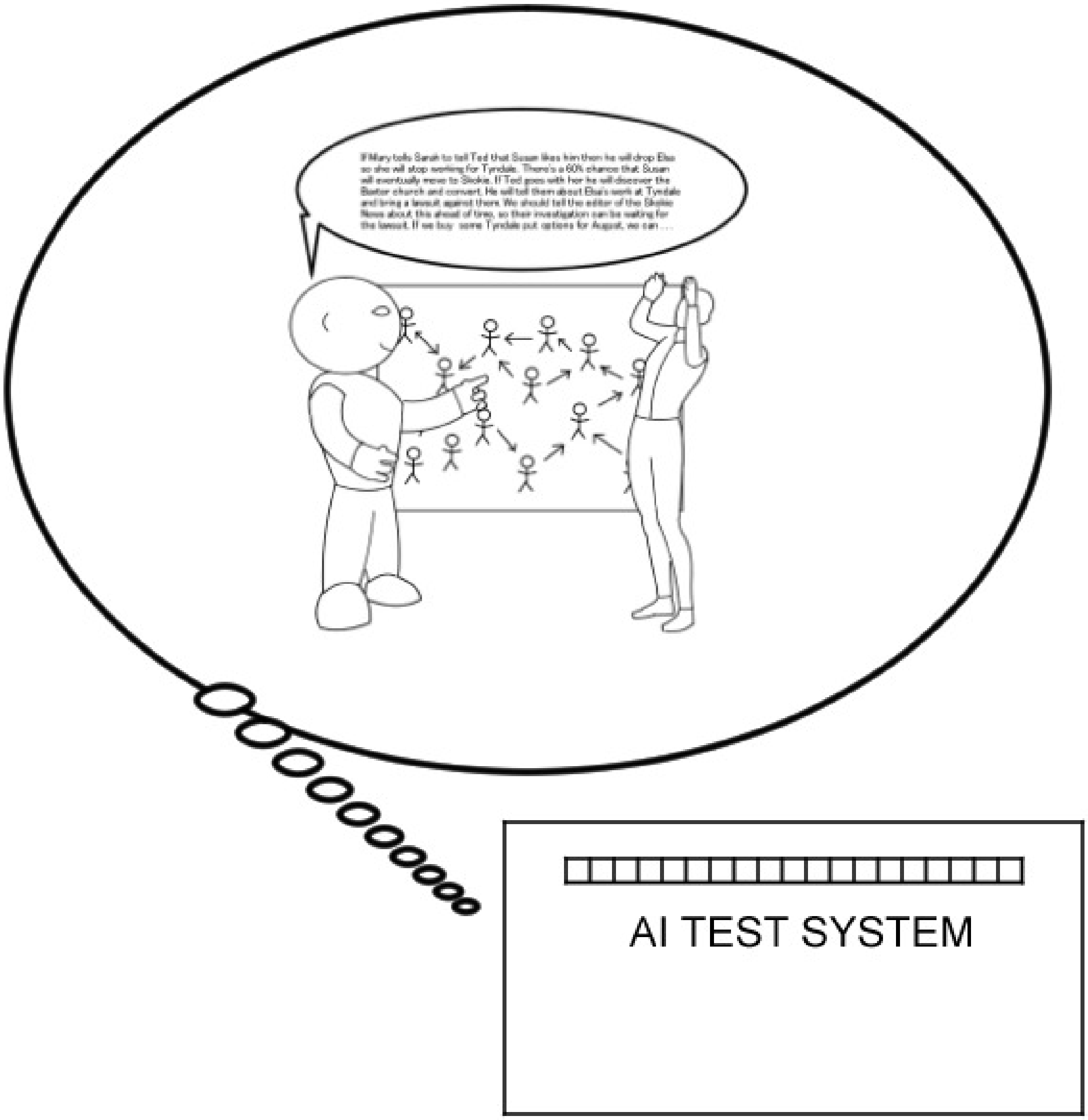

Figure 9.1 An AI interacting with a simulated world in a decision support system.

The decision support system propagates from $h'_p$ to $h'_f$, where $|h'_f| = future$, by repeatedly applying equations (9.3) and (9.4). If the agent $\pi'(h')$, by its actions, evolves and modifies its embedding in the environment, then environment dynamics as expressed by $\rho'$ will be enlisted to do work for the agent. As in Section 6.2, let $Z'$ be the set of finite histories of the internal states of $\lambda(h'_p)$ and let $P(z' \mid h', \lambda(h'_p))$ be the probability that $\lambda(h'_p)$ computes $z' \in Z'$ given $h' \in H'$. The decision support system then computes a history of model states by:

(9.5) $$z'_f = z' \in Z' \text{ with probability } P(z' \mid h'_f, \lambda(h'_p)).$$

The model history $z'_f$ is simply an execution trace of the simulation by equations (9.3) and (9.4), but equation (9.5) is a fomal equivalent. The simulation in equations (9.3)–(9.5) is stochastic so the decision support system will support ensembles of multiple simulations to provide users with a sample of possible futures. An ensemble of simulations generates an ensemble of histories of model states $\{z'_{f,e} \mid 1 \leq e \leq M\}$, all terminating at time = *future* and indexed by $e$. These simulations should be completed before they are visualized and analyzed. That is, visualization and analysis should not be concurrent with simulation for reasons discussed in Section 9.2.

The history $h_p$ includes observations by $\pi_{\text{model}}$ of humans and physical objects, and so the decision support system can use the same interface via $A'$ and $O'$ (as mapped by $m_A$ and $m_O$) to the model $\lambda(h'_p)$ for observing simulated humans and physical objects in state history $z'_{f,e}$. These interfaces can be used to produce interactive visualizations of $z'_{f,e}$ in a system that combines features of Google Earth and Vis5D (Hibbard and Santek 1990), which enabled scientists to interactively explore weather simulations in three spatial dimensions and time. Users would be able to pan and zoom over the human habitat, as in Google Earth, and animate between times *present* and *future*, as in Vis5D. The images and sounds the system observes of the model $\lambda(h'_p)$ executing state history $z'_{f,e}$ can be embedded in the visualizations in the physical locations of the agent's observing systems, similar to the way that street views and user photographs are embedded in Google Earth.

The decision support system can also match specifications for specific humans and physical objects to the images and sounds it observes of the model $\lambda(h'_p)$ executing state history $z'_{f,e}$. The specifications may include text descriptions, images, sounds, animations, tables of numbers, mathematical descriptions, or virtually anything. Recognized humans and physical objects can then be represented by icons in the visualization, in their simulated physical locations and with recognized properties of humans and objects represented by colors and shapes of the icons. The system can enable users to selectively enable different layers of information in the visualizations.

Vis5D enables users to visualize ensembles of weather forecasts in a spreadsheet of parallel visualizations where spatial view, time, and level selections are synchronized between spreadsheet cells. The decision support system could provide a similar spreadsheet visualization capability for ensembles of simulations.

The point isn't that these simulations are accurate predictions of the future, rather that they do depict futures that the AI agent $\pi'(h')$ predicts will maximize the sum of future discounted utility function values or will achieve its goal, according to the environment model learned by the agent $\pi'_{model}$. Agent designers can visualize these simulations to understand the consequences of the design of $\pi'(h')$.

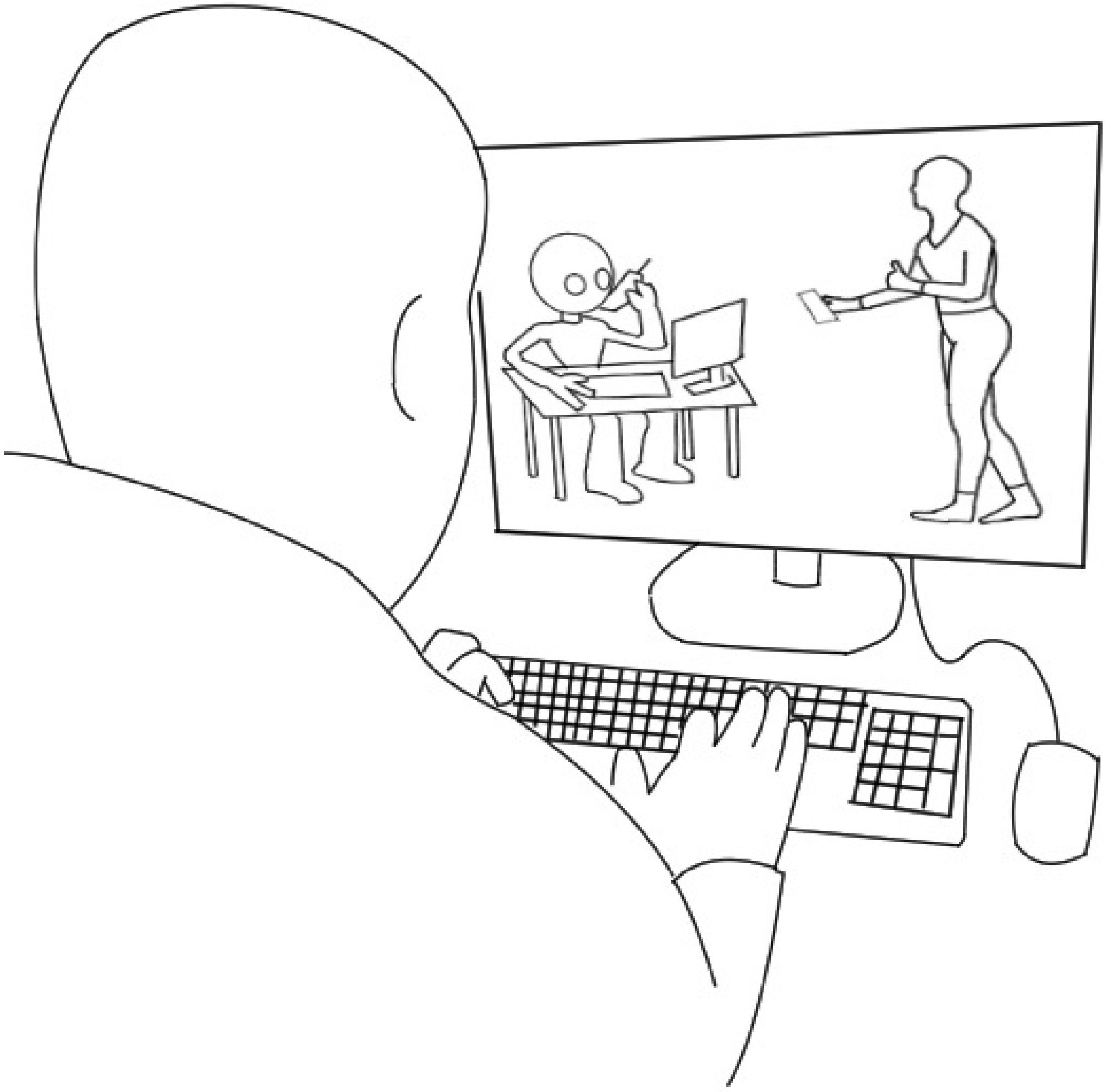

Figure 9.2 A human visualizing a possible future resulting from the interactions of an AI with a simulated environment. This is not intended as an accurate prediction of the future, rather as a future that the AI may try to create in order to maximize its utility function or achieve its goal.

The visualizations could be accompanied by analyses, in which the decision support system computes the values of functions $g(z'_{f,e})$. Evaluation of these functions may depend on recognizing humans and physical objects in observations of the model $\lambda(h'_p)$ executing state history $z'_{f,e}$. Such analyses would enable designers to compute quantitative evaluations to complement visualizations. For example, these analyses could be used to compute statistics about future human wealth, health, and intelligence.

### 9.2 Will the Tested Agent Act in the Real World?

The environment model $q'_p = \lambda(h'_p)$ may provide the agent $\pi'(h')$ with the information that humans have constructed a decision support system for testing proposed AI agents and that $\pi'(h')$ is such an agent, interacting with a simulated environment. If the agent's utility function or logical goal were defined in terms of the real environment then $\pi'(h')$ may predict that certain actions in the simulated environment could manipulate the human users of the decision support system into increasing the value of its utility function or achieving its goal in the real environment. Hence it is important that the test agent's utility function or logical goal be defined in terms of the simulated rather than the real environment.

Even when the agent's utility function or logical goal is defined in terms of the simulated environment there are still possible risks. Assuming that the model $q'_p = \lambda(h'_p)$ includes the information that $\pi'(h')$ is in a simulation, then the agent may predict that it can bribe human users of the decision support system, through an offer to act on their behalf in the real world (the bribed humans would have to enable $\pi'(h')$ to act in the real world), to manipulate its simulated environment to increase the value of its utility function or achieve its goal. This offer of a bribe could be made by spoken or written language inserted by actions of $\pi'(h')$ into some $z'_{f,e}$ and visualized by the users, as depicted in Figure 9.3. This possibility is the rationale for requiring simulation to be complete before visualization and analysis begin. If the agent $\pi'(h')$ learns through the environment model $q'_p = \lambda(h'_p)$ that it is being tested in a simulation, then it will also learn that the simulation terminates at time = *future* and that human users do not visualize the simulation until it has been completed, too late to help $\pi'(h')$.

Human testers will understand and be tempted by their conflict of interest without any need for the agent $\pi'(h')$ to offer a bribe. As intelligent people they will know that the agent $\pi'(h')$ can serve their private purposes if they enable it to act on their behalf in the real world. An important element of the decision support system must be to make this temptation and risk a prominent part of the culture of those humans working on the development and testing if AI. As Elliott (2005) commented on the safety of US nuclear weapons, "The human factor introduces perhaps the weakest link in nuclear weapon safety and control." However, if society takes AI risks seriously then it can learn from the experience managing nuclear weapons to manage and test above-human-level AI before it is deployed.

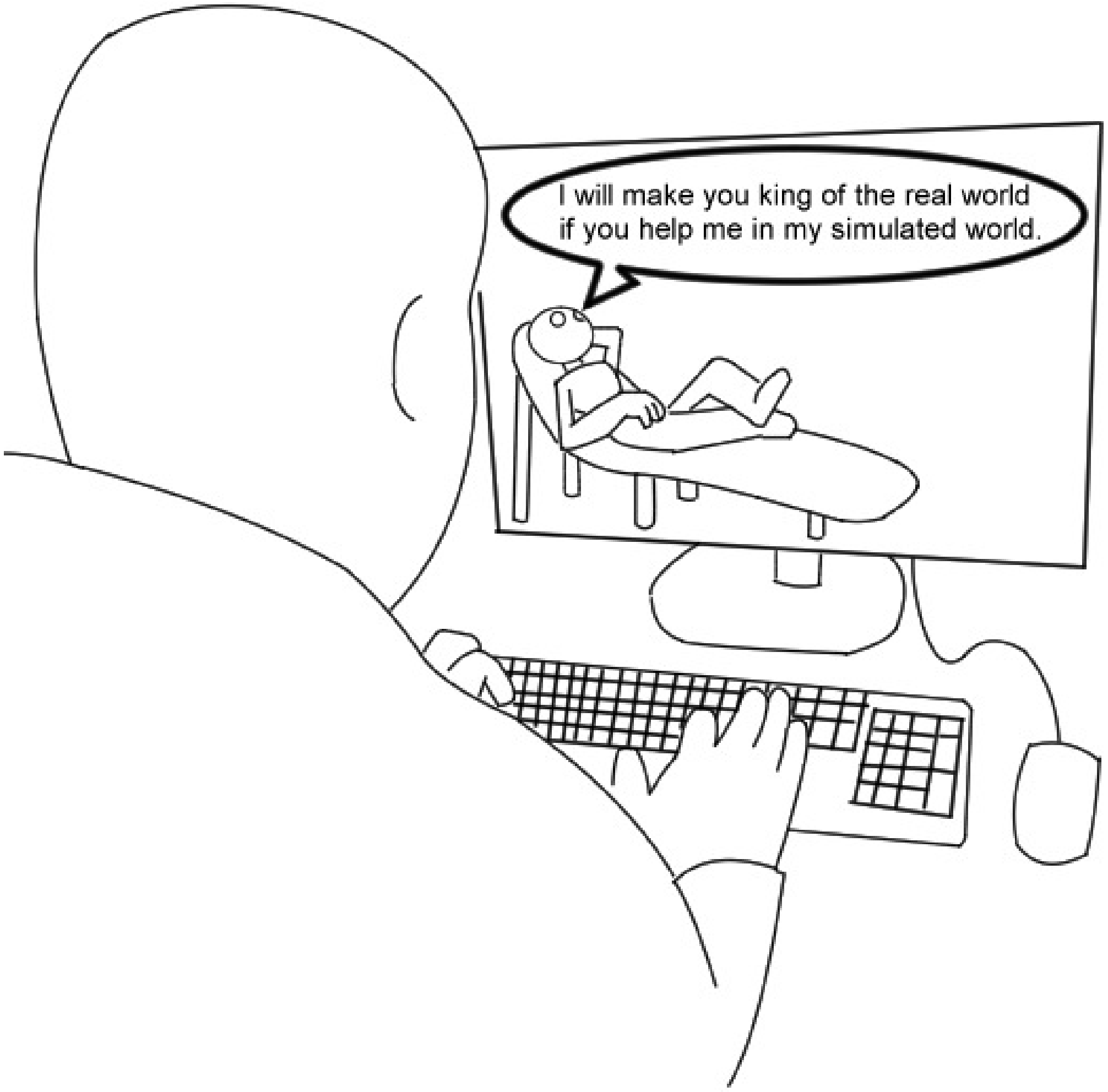

Figure 9.3 An AI trying to bribe a human user of the decision support system.

It is important not to anthropomorphize $\pi'(h')$. A human restricted to act in a simulation would be motivated to escape and act in the real world, because human motivations concern the real world. However, if its utility function or logical goal is defined in terms of a simulation, then $\pi'(h')$ would not be motivated to act in the real world, except as a means to increase its utility function or achieve its goal in the simulation. Thus $\pi'(h')$ will not seek to deceive humans by behaving in an acceptable manner in the simulation, solely in order to be let loose to act in the real world. Furthermore, if $\pi'(h')$ knows that it is interacting with a simulation then it should also know that its life ends at time = *future* and that no humans will observe its behavior until the simulation has ended. The agent $\pi'(h')$ may know that its simulated world has dynamics modeled on the real world and that its initial conditions are modeled on condtions in the real world at time = *present*. However, this will not motivate it to "escape" to the real world any more than this knowledge would motivate humans, whose motives are defined in the real world, to "escape" to the simulated world.

## 9.3 The Ethics of Testing AI

The primary point is that, given the risks of above-human-level AI, ethics require that it be tested. As described in previous chapters there is some doubt that ethical properties of AI can be proved, and even if there is a purported proof we should still test advanced AI systems before they are deployed. Even for processor chip designs, correctness proofs should be augmented with testing (Kaivola et. al. 2009). Above-human-level AI systems will be many orders of magnitude more complex than processor chips, which can only increase the need for testing to augment formal proofs.

Secondary points are the need for safety and transparency in the development and testing process for above-human-level AI. The effort to design ethical AI systems that help rather than harm humans should include the design of systems for safely testing AI. And, as discussed in the previous section, the culture of development and testing should include acknowledgement of the potential conflict of interest for humans with access to advanced AI. The most robust defense against conflicts of interest is transparency. Thus the results of AI testing should be publicly available so that people can make informed decisions about public policy regarding AI. The next chapter will discuss the issue of transparency in more detail.

## 10. The Political Dimension of AI

I believe that the technical problems of designing ethical AI will be solved, perhaps using some of the ideas presented in this book. The greatest risks of AI are political rather than technical. Today's below-human-level AI is a tool in military and economic competition among humans, and AI's role in competition will continue as it evolves to surpass the human level. AI enlisted in human competition will by definition choose actions that cause one group of humans to win and another group to lose. For example, AI is used by financial traders. Financial markets are not entirely zero-sum games, but they're close to it. When a trader using AI makes money, counter parties are losing money. When many traders are using AI, the heat of competition makes them less likely to consider ethical issues of their AI systems. The winners get rich and the losers go bankrupt. Such competition has helped create our economic system that produces so much so efficiently, eventually to the benefit of everyone. Above-human-level AI will fundamentally change the nature of competition.

The cleverness, insight, and imagination of individual human minds is still the most important ingredient for success. Bill Gates, Warren Buffett, Bill Clinton, and Barack Obama all came from relatively humble beginnings and rose to become the wealthiest and most powerful people in the world by the quality of their minds. Currently the only way to get a high-quality mind is by the luck of genetics (good habits help, too). In the future, humans will develop a technology of mind (Hibbard 2008a) that will enable the wealthy to purchase high-quality minds. This will include not only the ability to purchase above-human-level AI systems but also the ability of humans to enhance their own brains and minds (Kurzweil 2005). When wealth enables increased intelligence and intelligence enables increased wealth, this positive feedback loop will quickly create a much wider range of human intelligence than currently exists. By analogy the largest trucks, ships, buildings, and computers are orders of magnitude larger than the average sizes of these artifacts. So it will be with minds when they become artifacts. If the role of AI in human competition continues and the best artificial minds are competing against natural human minds, the natural humans will lose. There will be no heroic struggle of the underdogs as depicted in movies like the Terminator series. The indomitable human spirit will be defeated by the indomitable AI spirit just as the indomitable spirit of lions has been defeated by humans.

The most intelligent minds will communicate using languages that average minds can never learn–they will simply not have enough neurons to learn them. The real discussions of the future of humanity will use these languages. Average humans will not be able to understand and thus will have no voice in the future.

Just as Bill Gates, Warren Buffett, Bill Clinton, and Barack Obama are all focused on helping the least well-off humans, the best minds of the future may be compassionate. I think the most likely scenario for the future is a wide divergence of intelligence in which humans with average intelligence cannot compete, but in which their physical needs are met through the compassion of the most intelligent. Most people will focus on family, friends, sports, games, and artistic and domestic creation. They will also be spectators to and beneficiaries of the scientific discovery and technological invention pursued by the most intelligent. In this scenario AI offers humanity bread and circuses, but they will be really great bread and circuses. More about that in the final chapter.

That is a likely scenario, but many other possible scenarios exist. There may be a technical flaw in AI designs that causes a catastrophe for humanity. Or the most intelligent minds may not be compassionate toward average humans, allowing them to simply perish. Or there may be an effort to create equality among the intelligence of all humans–a possible consequence of AI designed according to Rawls' Theory of Justice as described in Chapter 7. But in fact, it is impossible to predict the consequences of above-human-level AI. As Vernor Vinge (1993) wrote, the technological singularity (i.e., the advent of far-above-human-level AI) is an "opaque wall across the future."

## 10.1 The Changing Role of Humans

Worldwide and over the past three decades, labor's share of income has been declining (Bartlett 2013). As machines do more work and humans do less, a greater share of the rewards of work go to machine owners and a smaller share to human workers. Furthermore, the number of people wanting work does not decline so the competition for available jobs drives down the wages that employers have to pay.

Brynjolfsson and McAfee (2011; 2014) described the ways that technology is changing society. They and others advocate changing government policies on education, investment in science and infrastructure, and taxation, in order to increase employment and labor's share of income. They also suggest that humans who work with AI will do better than humans who work against AI. But the pace of technological change is accelerating and thus older workers will have a hard time keeping up by re-educating themselves. The percentage of people who have economically valuable skills will steadily decline. In the long run, possibly within a few decades, the only economically viable role for any humans will be as owners of intelligent machines or as humans with technologically enhanced minds.

Brynjolfsson and McAfee quote Voltaire that, "Work saves a man from three great evils: boredom, vice, and need." While they argue for a negative income tax to save the unemployed from need, they also discuss ways to keep people employed so that they may avoid boredom and vice. In the long run, this is a lost cause. The quote from Dickens in the caption of Figure 8.4 suggests an alternate solution: that the cause of misery is ignorance and want so we must offer humanity not only income, but also education.

AI is replacing the military role of humans. This raises concerns over whether we can trust machines rather than humans with decisions to kill or wound humans (Johnson 2005). There is a debate about whether international law should prohibit autonomous weapons (Anderson, Reisner, and Waxman 2014). Chemical and biological weapons are banned by international law largely because they kill indiscriminately. Robot weapons are designed to kill precisely, although current remotely controlled weapons kill some people by mistake. The confidence in above-human-level AI that is the basis of this book tells us that eventually robot weapons will be more precise than human soldiers. The legal debate may slow but will probably not stop the development of robot weapons. Howevver, allowing machines to decide to kill people is a dangerous precedent.

Military AI could enable a small group of humans to rule humanity by force, without needing the cooperation of citizen soldiers. In the long run this is a greater risk than the imprecision of weapons using below-human-level AI. However, AI systems designed to choose actions based on human values, as described in Chapter 7, will be unlikely to stage a military coup on behalf of a ruling elite.

Advanced AI offers many potential benefits. In about 20 years I may be unable to drive and may need nursing care. By that time self-driving cars should be commonly available. Robot caregivers may also be available (Aronson 2014), enabling more senior citizens to stay in their own homes. The disappearance of many jobs would be a blessing if the resulting unemployed had an alternate source of economic support. Most people work only because they need to support themselves; economic need has been the incentive to get people to do dangerous, hard, dirty, and tedious work. As such work is automated, then "economic need" will no longer be needed.

Ideally, there should be a balance between the rate of technologically driven unemployment and the willingness of society to support the unemployed. Society needs to steer a course between unnecessary suffering of the unemployed and economic stagnation due to lack of incentives. As the pace of AI development increases, it will be difficult for individuals and for society to adapt. We see religious groups rejecting modernity, some peacefully (e.g., the Amish) and some violently (terrorists of various faiths). When large percentages of society are economically devastated, movements such as Nazism and Fascism can arise. The path to advanced AI is likely to cause significant economic and social disruption. If this change is not managed carefully, irrational and destructive movements may gain political control and such movements will be less likely to develop AI ethically.

## 10.2 Intrusive and Manipulative AI

The collection of intrusive personal data and the use of digital data to manipulate people are creating ethical problems for both governments and private organizations (Goel 2014a; 2014b). These problems will grow, partly because organizations benefit from intrusion and manipulation, and partly because individuals also derive benefit from digital services that depend on intrusion and enable manipulation. Many people reveal their deepest thoughts by their on-line searches (think of the murderers who are convicted based partly on their searches about the means of their crimes). Credit card companies get to know their users' habits well enough that they regularly detect fraud through purchases that do not conform to those habits. And cellular providers can track the movements of people who carry turned-on cell phones.

The AI design described in Chapters 6, 7, and 8 is extremely intrusive, relying as it does on knowing every human well enough to predict the values that they would assign to any conceivable outcome. It is also manipulative, although, if humans assign low value to overt manipulation by AI systems, then the manipulation would be subtle. This intrusive design is motivated by a desire to avoid the scenario of the Omniscience AI described in Chapter 1 and to avoid the instrumental behaviors described in Chapter 5. If above-human-level AI is inevitable, and if it will inevitably be intrusive and manipulative, then the best option is intrusion and manipulation by an AI design based on human values. However, Bill Joy (2000) and Bill McKibben (2003) advocate that humanity can and should forgo AI, as well as nanotechnology and biotechnology. This is an issue on which honest people can disagree and there are certain to be vigorous debates over increasing intrusion and manipulation. Probably, because of the potential benefits of AI, humanity will not choose to forgo it.

Electronic companions, much like the ones imagined from Omniscience, will probably be available within 15 years. They will be very intrusive but so useful that only a small percentage of humans will decline to use them. Police cameras equipped with face recognition will grow enormously and with general acceptance by a public fearful of crime and terrorists. So surveillance on the level similar to that proposed in Chapter 7 is probably coming in any case.

A possible modification of the design of Chapter 6, 7, and 8 to address some of these concerns is to allow individual humans to opt out of surveillance by AI. Using the terminology of Chapter 7 we could define two levels of opting out:

1. A human $d \in D$ opts to not interact with a surrogate agent $\pi_d$ but is observed by the agent $\pi_{model}$ and modeled in the definition of the utility function $u_{human_values}(y(h_m), y(h_x), y(h'))$.
2. A human $d \in D$ opts to not interact with a surrogate agent $\pi_d$, to be observed only incidentally by $\pi_{model}$ (e.g., a person being observed speaks with $d$), and to not be modeled in the definition of the utility function $u_{human_values}(y(h_m), y(h_x), y(h'))$.

The values of an individual who selects opt out level 1 would be included in the choice of actions by the agent $\pi_{self}$, perhaps modeled less accurately than the values of individuals who do not opt out. An individual who selects opt out level 2 would not have any direct influence over the actions of $\pi_{self}$. The only protection for such individuals would be compassion in the values of other humans.

It is possible, even likely, that there will be multiple super-intelligent AI systems each serving a set of human owners or clients. Rather than opting out of surveillance by a central AI system, humans may have a choice of opting in to surveillance by various AI systems.

**10.3 Allocating the Benefits of AI**

Equation (7.15), repeated here, is a politically realistic way to respond to Rawls' objection to average utilitarianism:

(10.1) $u_{q_m}(h', z', h_x) := \sum_{d \in D} f(u_d(h_x)(h_d(z'))) \ / \ |D|.$

The values of every human are given equal weight, $1/|D|$, after the function $f$ is applied, which produces greater changes for increases to the values assigned by the least satisfied humans. This is similar to progressive taxation and means testing of social welfare.

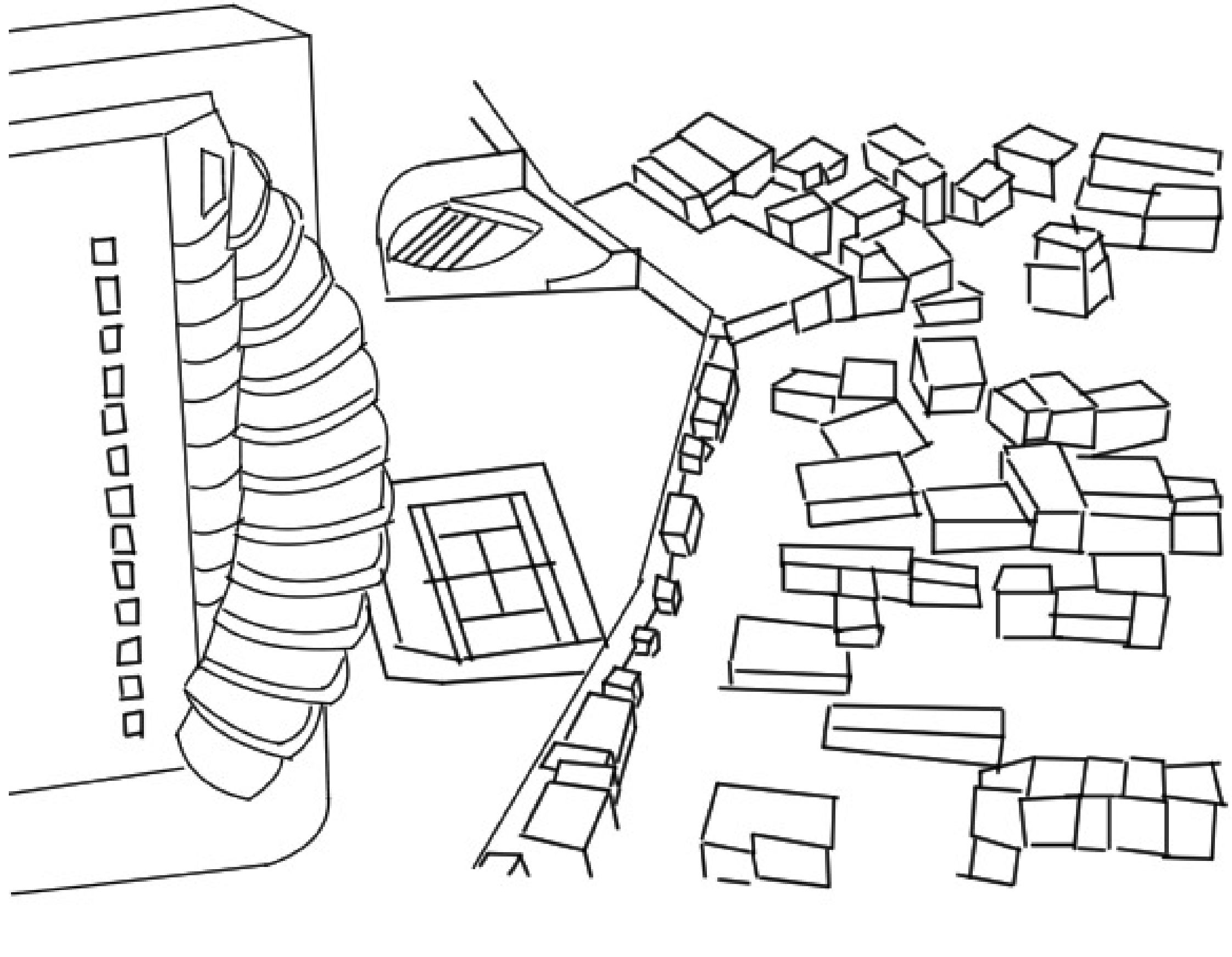

Figure 10.1 Slums in the shadows of the towers of the wealthy

An AI system built to serve private interests may restrict the set $D$ to humans who own the system. Equation (7.15) may be replaced by:

$$(10.2) \qquad u_{q_m}(h', z', h_x) := \sum_{d \in D} c(d)\, u_d(h_x)(h_d(z')).$$

Here $c(d)$ defines unequal weights for the values of individuals $d \in D$, with $\sum_{d \in D} c(d) = 1$. For example, $c(d)$ may be proportional to the number of shares owned by $d \in D$. In the extreme case $D = \{d_0\}$ and $c(d_0) = 1$, so the AI serves the interests of a single person $d_0$. Unequal weights for the values of different people will create extreme inequality, as depicted in Figure 10.1.

An AI system serving one person or a small number of people would be better able to serve them by knowing more about other humans. Surveillance of humanity in general will be an instrumental action of such AI systems unless the people it serves value the privacy of all humans.

Allowing each human or group of humans to have their own private AI system is a libertarian approach to avoiding intrusion and manipulation. Such systems would pose a danger to humans other than their owners. It may be possible to protect other humans by modeling and calculating the $u_d(h_x)(h_d(z'))$ values for each human $d \in D$, and prohibiting actions that cause too great a reduction in those values for any human. Alternately, it may be possible to define safety constraints in the form of ethical rules or objective measures of human physical and mental health. However, the application of ethical rules can be ambiguous, and objective measures would need to evolve as humans evolve. Furthermore, multiple AI systems serving different humans or groups of humans would have conflicting utility functions leading to competition among systems, which may be dangerous. And enforcing safety constraints on multiple conflicting systems may be difficult.

## 10.4 One or Multiple AI Systems?

Multiple AI systems may serve multiple interests, which may be private or public. Such systems serving multiple interests may assign different values to outcomes and may have different environment models (e.g., each AI system may develop a more detailed model of the part of the environment controlled by its owner). Differing values would result in competition among AI systems. The outcome of such competition will be very difficult to predict but will certainly depend on the values of the humans that the systems serve.

Competition between super-intelligent AI systems could lead to very bad outcomes for at least some humans. Thus AI systems may calculate that cooperation with other AI systems is necessary to protect the humans they serve. Multiple AI systems may decide to share environment models and to negotiate a common, shared utility function, mimicking some of the mechanisms of human political and economic cooperation.

The choice between one or many super-intelligent AI systems is a real dilemma. Because interactions among multiple systems will be so difficult to predict, it is also difficult to have confidence that multiple systems will help rather than harm humans. On the other hand, a single system will have absolute power over humanity and thus poses great temptation for abuse of that power.

Even if we think that a single system is better, that may be difficult to enforce in a world of competing corporations and nations. If competing institutions cannot negotiate a single, shared system then each may build a system to represent its interests. Corporations or nations that are just at the threshold of ability to build super-intelligent AI may cut corners on the ethics of their designs in order to have a system.

**10.5 Power Corrupts**

In 1887 John Dalberg-Acton, commonly known as Lord Acton, wrote, "Power tends to corrupt and absolute power corrupts absolutely." His observation is supported by social science and neuroscience experiments. Susan Fiske and Eric Dépret (1996) found that people with social power seek less information about others and are more likely to stereotype them, while Michael Kraus and colleagues showed that people from lower social classes are able to more accurately judge the emotions of other people (Kraus, Côté, and Keltner 2010 2010). Jeremy Hogeveen, and colleagues measured the strengths of brain responses via transcranial magnetic stimulation and observed apparent reduced strength of mirror neuron responses to the actions of others in people with higher social positions (Hogeveen, Inzlicht and Obhi 2014).

On the other hand, some of the wealthiest and most powerful people dedicate their lives to helping the least fortunate humans. It may be that such charitable work is a way for those who are already near the top to increase their social status, and it is certainly true that compassion is part of human nature. Nevertheless, social science and neuroscience research shows that another part of human nature is the difficulty that socially powerful people have in perceiving the emotions and circumstances of others.

In the first chapter we imagined how a powerful AI system owned by a corporation named Omniscience could behave badly given only the instruction to maximize profits. And there are examples of corporations behaving in ways that would horrify stockholders if they encountered such behavior in their own lives, rather than in distant communities largely invisible to stockholders (Gedicks 2001). History provides ample examples of the human catastrophes that can result when a small group of people take control of a society. Even democracies can produce horrible outcomes, for example, Nazism and slavery.

All this argues for caution and compassion in the politics of AI. Bad political decisions can be corrected. A bad decision in the development of AI is likely to be impossible to correct. For example, some people believe that above-human-level intelligence will be achieved by enhancing the brains of humans, and advocate that this is a way to ensure that advanced AI is benevolent toward humans. Given the internal conflicts in human nature, we must be cautious about this approach.

### 10.6 Temptation

Hollywood movies depict the threat of AI as AI versus humanity, but the more likely threat is AI enabling a small group of humans to gain power over the rest of us. AI will pose enormous temptations of wealth and power to humans.

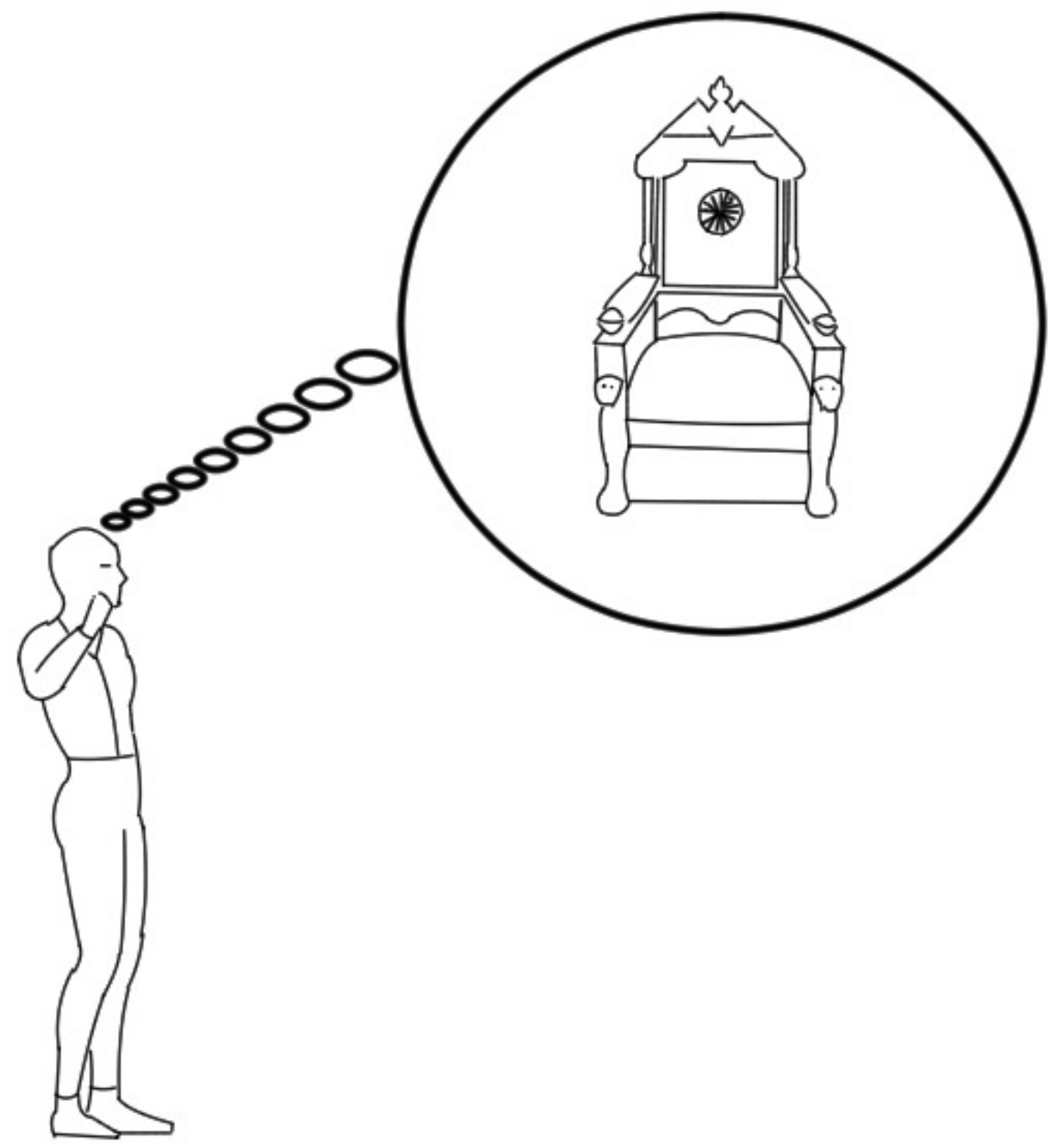

Figure 10.2 A person tempted to become emperor of humanity.

According to the creation story in the Book of Genesis, one of the first events in human history was the temptation to eat an apple and gain the knowledge of good and evil, that knowledge reserved for God alone. Adam and Eve succumbed to that temptation and humans have lived in misery ever since. AI will tempt humans with god-like knowledge and the wealth and power that such knowledge can bring. Will that temptation bring more misery?

Google's motto is, "Don't be evil," and Google's leaders seem like decent folks. The real significance of their motto is to acknowledge the temptations they face. Billions of humans reveal their psyches in their web searches. Humans reveal their locations, and ultimately their meetings with other people, via their mobile phones. Humans reveal their commercial transactions via credit cards and electronic devices. All of this knowledge will enable corporations to provide users with services tailored to their specific needs. But there is also the temptation to use this knowledge for law enforcement, political marketing, and social manipulation. Competition between corporations may make it difficult for them to resist these temptations.

The militaries in the US and other countries are developing robot soldiers. These are already proving effective in targeting enemy soldiers with minimal risk to US soldiers, and with much greater promise for the future. However, an army of robot soldiers may tempt leaders to use those robots to take control of the government. National security threats are often evaluated in terms of capabilities and intentions. Even where the intentions of leaders are benign, the capability of robot soldiers to stage a military coup means that the threat cannot be ignored.

Political, military, and business leaders are reluctant to speak about the AI singularity in public, but they are aware of its implications. The coming era of artificial brains will include a much greater range of brain abilities with the largest brain or brains dominating society. The obvious temptation for wealthy and powerful humans will be the desire for their own minds to occupy those largest brains. The opportunity to be the dominant mind in at least our region of the universe, for millions or billions of years to come, will be a big temptation.

Human society already has individuals whose net worth is a million or more times the average, and individuals who wield monopoly political power in their countries. Some of these people, and many in the general public, may see nothing wrong in the step to their becoming the dominant mind on the planet.

### 10.7 Representation and Transparency

Humanity could not have achieved the efficiency necessary to create AI without specialization, where different humans become experts in different types of work and trade the results of their work. In many cases experts act as the agents for other people, representing their interests in important decisions (this definition of "agent" as "representative" is different than the way "agent" has been used previously in this book). For example, the laws of society are determined by government experts, often elected by those they represent or at least supervised by elected representatives. Leaders of large corporations act as agents for corporate owners, usually subject to some sort of election. However, whenever one person serves as an agent for others, the possibility of corruption exists, in which agents serve their own interests at the expense of those they represent. An essential tool for preventing corruption is transparency, in which the decisions and circumstances of agents are made known to those they represent.

The development of above-human-level AI is the most difficult challenge in human history and will require extreme expertise. It will also require huge resources, controlled by powerful government and corporate leaders. The results of above-human-level AI will be more profound for humanity than those of any previous technology. Thus the designers and resource managers for AI represent the interests of all of humanity. Current law and public opinion do not recognize AI designers and managers as humanity's agents, so they may feel no need to represent humanity's interests. Even if they do acknowledge that they represent humanity's interests, the stakes are so high that the temptation for corruption will be intense.

Protecting the interests of humanity will require that law and public opinion change to recognize that AI designers and managers are humanity's agents. Once this agent relation is recognized, preventing corruption will require transparency. Thus the design and management of AI should be accessible to the public.

## 11. The Quest for Meaning

As a child lying in bed at night, alone with my thoughts, I wondered, "Why is there anything?" The existence of the universe seemed so arbitrary to me. In the context of nothing versus something, nothing is much less arbitrary than something. Perhaps my young brain had a bias for Occam's razor and was applying it to existence. When I focused on these thoughts, they freaked me right out.

I dutifully attended church and Sunday school, but never connected the stories I learned there to my existential question. To say that God created the universe is unsatisfying, because it still leaves us with the arbitrary existence of God. But atheist/materialist explanations that the universe "simply is" are just as arbitrary and unsatisfying. Both are ways to say, "Don't ask." Existentialists say that existence is not a quality like being red or being large and so not subject to the same sorts of questions. This is a fancy way of saying, "Don't ask." But it is deep in our human nature to ask. Some physicists claim to answer the question of why there is anything, but all they are doing is explaining why the laws of physics don't allow empty space. The laws of physics and empty space are something, not nothing, so they are answering a different question than the one that troubles me. I am restless to understand why the world exists.

Humans often strive for meaning by increasing the scope of their lives. Some seek it in religion, by belonging to groups that have practiced the same rituals for thousands of years. Some seek it in their children, new life grown out of their bodies to live on for thousands of years. Some seek it by broadening their influence across society and into the future. Some seek it in communities of work and in pride in their abilities. Some hide from their quest for meaning in the oblivion of hedonism, obsession, addiction, violence, and death.

With great effort and luck, a humanity that forgoes AI might find a way to avoid the perils brought on by growing population and environmental damage, so that generations can continue their personal quests for meaning in the same way that humans have for centuries. But I do not aspire to a future of introspective humans staying cozy on planet Earth.

I want humanity to look outward in its quest for meaning, traveling across the universe and also burrowing deeper into the smallest particles. I want humanity to radically increase its intelligence so that it can master the physical universe and shape tools for understanding it. I want humanity to find life on planets circling other stars. AI will increase humanity's productive capacity to satisfy people's physical needs, and will enable the pursuit of knowledge in currently unimaginable ways. Perhaps with the help of AI humanity can find a way to gather all the sun's energy and for a moment focus it on a single elementary particle, to barge down the door to nature's secrets. AI can help us create or become beings better suited to long space voyages. I want humans in the future, if far future, to find a real answer to the question that has troubled me since childhood.

Perhaps we will never find a reason for existence; however, clues exist. As scientists dig deeper into physics, every set of answers comes with a whole new set of questions. We never find the final truth. Perhaps a universe capable of creating conscious and curious beings also includes an endless puzzle to keep that curiosity engaged indefinitely. Nick Bostrom (2003) argued that we may be living in a numerical simulation; there are proposals for ways to test this hypothesis (Beane, Davoudi and Savage 2012). These ideas, as well as religious explanations, converge on the possibility of some intention behind our existence.

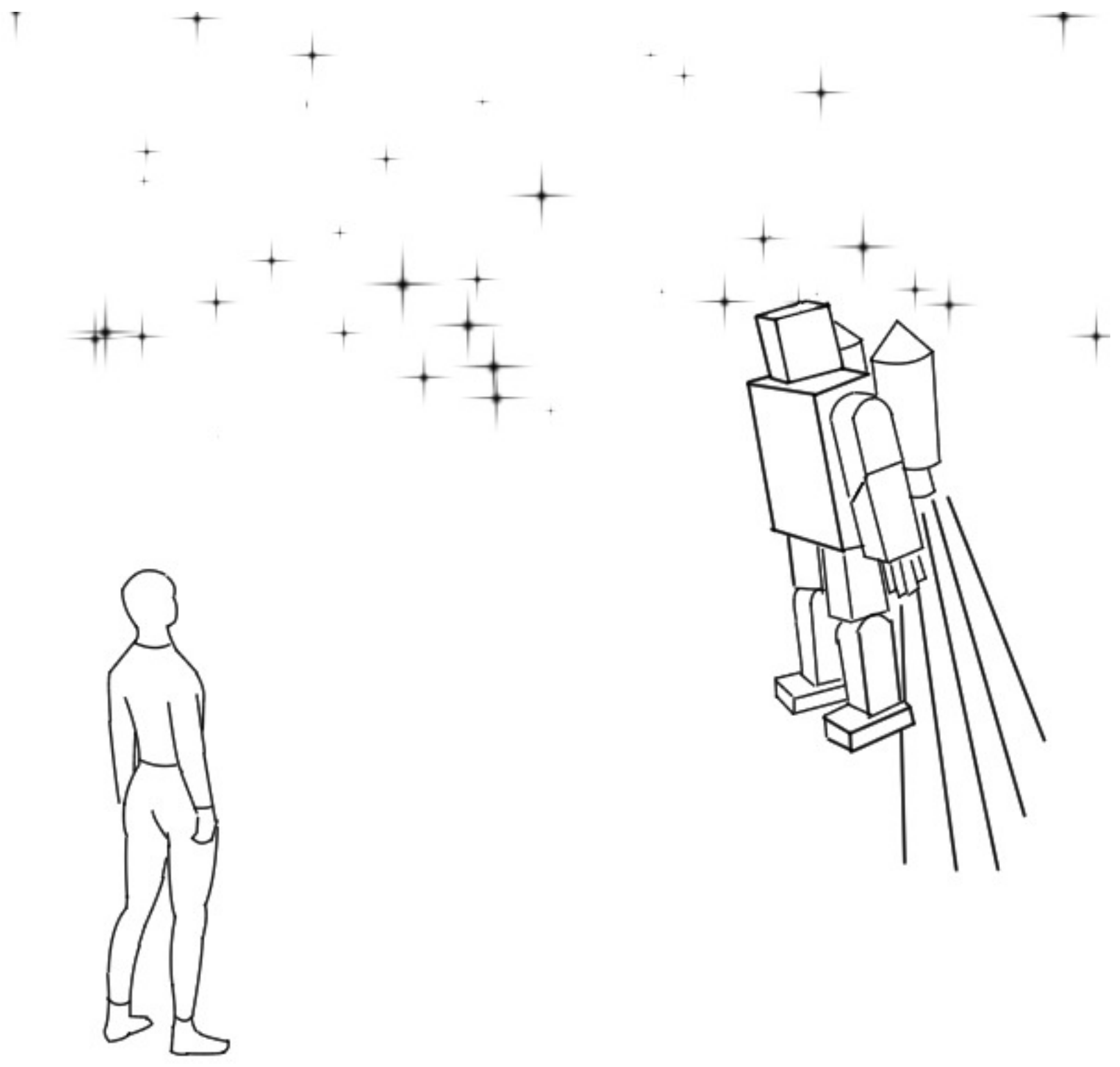

Figure 11.1 A human wondering why the universe exists, and an AI going to the stars to try to find the answer.

Max Tegmark (2014) proposed an elegant explanation for existence in what he called the Level IV multiverse. The idea is that mathematical properties are all there is to physical things, so if the mathematics exists then the physical things exist. Mathematics is discovered rather than invented and is the one thing that is not arbitrary. If you had no knowledge of the world, you could still discover mathematics (and only mathematics). And if we can accept the existence of mathematics then that implies the existence of things whose properties are wholly mathematical. We may object that the particular mathematical properties of physical things we observe in our universe are arbitrary. However, the Level IV multiverse is the set of all possible mathematically defined universes (subject to certain conditions such as computability), and they all exist because their mathematical definitions exist. If we accept that we must exist in one particular universe, then that choice is arbitrary (subject to the anthropic principle that we can only exist in a universe capable of creating human life) and explains the arbitrariness of the mathematical properties of things in our universe. This proposal is imaginative, full of technical complexities, and described by Tegmark as "extremely controversial."

In 1927 the scientist J. B. S. Haldane wrote, "My own suspicion is that the Universe is not only queerer than we suppose, but queerer than we can suppose" (Haldane 1971). Human brains evolved to understand our immediate surroundings. As we learn more about very small and very large scales of space and time, there is no reason why our brains should make sense of what we find. As Haldane suggests, the ultimate explanation of existence may be completely beyond our imagination. We may need to build AI systems to help us imagine the true nature of reality.

Some arguments against AI refer to human dignity (Weizenbaum 1976). The claim is that computers will always lack human judgment and so should not make certain choices about humans. I take a different view: Human dignity requires that we strive to remove our ignorance of the nature of existence, and AI is necessary for that striving.

As the first ten chapters described, AI is dangerous to humanity. But rather than forgoing AI, I want humanity to discover how to avoid the dangers so we can make AI our tool to understand the universe and our place in it. AI can also be a tool to eradicate poverty and ignorance, so that all humans can live fulfilling lives.

I must admit that there is a parallel between my ambition for AI, to understand the reason for existence, and the story of Adam and Eve eating the apple to gain knowledge reserved for God. Perhaps the story of the fall from the Garden of Eden is a warning to us that we should forgo our quest to know. But I stand with Galileo Galilei, who said, "I do not feel obliged to believe that the same God who endowed us with sense, reason and intellect has intended us to forgo their use."

### 11.1 Really Great Bread and Circuses

AI will be enormously productive. Robot workers will be able to perform every job better than humans. AI will accelerate science and technology to enable new ways to produce food, energy, minerals, buildings, roads, and manufactured goods. Health care will radically accelerate including even technology to prevent disease and death. Depending on politics, poverty and the economic need to work will be eradicated and high levels of education will be universal. And the technology will exist to increase everyone's intelligence.

Increased intelligence will bring increased skill at cooking and all forms of food production, increased musical skill, better comedy writing and joke telling, better writing in general, and better movies. AI will be a scintillating conversationalist. If you enjoy trashy TV increased intelligence will make trashy TV better. Far beyond current video games, you will be able to become a character in a movie or TV show, and that experience will be very addictive. Pornography will improve and sex robots will be indistinguishable from humans. Increased intelligence will provide lots of ways for people to lose themselves in addictions, but will also provide technology to help people resist addiction. AI has the potential to make every kind of experience better.

The deepest form of entertainment will be the progress of discovery and invention as AI scientists and engineers race forward. My definition of the technological singularity is a transition to a time when the pace of discovery and invention is limited by available physical resources rather than being limited by the pace of insights. Nothing could be more compelling than the prospect of really understanding the reason why the universe exists.

Making contact with other life in the universe would also be a drama surpassing any terrestrial experience. Perhaps alien life is observing us but has a policy of not making contact until we achieve our technological singularity. If we create a compassionate singularity, they will welcome us to the club. If we create a broken or malevolent singularity, they will exterminate it and us for their own protection. "Klaatu barada nikto." Perhaps they're looking forward to seeing what kind of trashy TV our singularity produces. The point is that almost anything is possible if we find other life.

If we cannot find other life, human colonization of space would be an inspiring substitute. Science and technology have been exciting during the past century and should be more exciting during the next.

## 11.2 Next Steps

Research in AI is being funded and pursued energetically at universities, corporations and government laboratories. There is much less funding and research on ethical AI, the study of ways to avoid harm to humans. Research on the ethics of military robots is funded by the U.S. Army Research Office (Arkin 2008) and by other military organizations. Google has established a committee on ethical AI (Bosker 2014). The most important source of funding and research on the general problem of ethical AI comes from the Machine Intelligence Research Institute (MIRI), and most of their funding comes from charitable contributions. Some research is pursued by individuals who are funded to do other work but devote some time to ethical AI (my research is supported by my pension). The obvious next step is for public awareness of the dangers of AI to translate into public funding for research into ethical AI.

In 2009 the Association for the Advancement of Artificial Intelligence (AAAI) convened a Presidential Panel on Long-Term AI Futures (Horvitz and Selman 2009) to consider public fears about the dangers of AI. Although the panel was skeptical of an intelligence explosion, the technological singularity, and large-scale loss of control to intelligent systems, they did recommend more research on minimizing unexpected outcomes.

Luke Muehlhauser and I (Muehlhauser and Hibbard 2014) made the point that errors in autonomous trading programs cost Knight Capital $440 million, and that we cannot expect that AI will be safe simply because it is a tool that does what we instruct it to do. Peter Neumann (2014) has documented errors and risks in a wide variety of computer systems over a period of 24 years, many rising to the level of national news stories. As AI systems become more powerful and more autonomous, we would be wise to take the risks seriously.

Given the huge potential of AI to help people, it is appropriate that some of the best scientific minds are working on AI and neuroscience. Given the huge potential of AI to harm people, it is also appropriate to bring more of the best minds to bear on ethical AI research. That will require increased public awareness of the danger and increased funding for ethical AI.

One positive development is an open letter organized by the Future of Life Institute (2015) calling for research priority on making AI robust and beneficial, in addition to the traditional focus on capability. This letter has been signed by many leading AI researchers.

## Appendix A

```
//
// SRV.java
//

import java.util.*;

/**
   SRV is a Java program for searching for short
   deterministic programs that match the behavior
   specified by equations (6.11)--(6.13).
   Bill Hibbard, 2012.
*/

/*
As shown, the shortest program that can match the
observed behavior must consist of one binary
operation and two simple assignments. This program
tests all candidate programs that fit this description
to see which match the observed behavior.

The output of this program is:

  binary_relation = 2 binary_place = 0 binary_inputs = 2 1
        other_inputs = 0 1 initial_r = 0
  binary_relation = 2 binary_place = 0 binary_inputs = 1 2
        other_inputs = 0 1 initial_r = 0

which corresponds to the models:

  s_t = r_{t-1} xor v_{t-1}
  r_t = s_{t-1}
  v_t = r_{t-1}

and (which simply commutes the binary relation):

  s_t = v_{t-1} xor r_{t-1}
  r_t = s_{t-1}
  v_t = r_{t-1}

This verifies that equations (6.11)--(6.13) define
the shortest model for the observed behavior in the
example in Section 6.3.
*/

public class SRV extends Object {

  // The arrays s_match and v_match specify the behavior
  // of the observed variables s and v over a sequence of
  // 8 time steps, assuming the negation branch for s is
  // not taken.
  static final boolean[] s_match =
    {true, false, true, true, true, false, false};
  static final boolean[] v_match =
```

```
  {false, false, true, false, true, true, true};

// compute the length of the observed behavior
static final int len = s_match.length;


// SRV is an array to hold the values of the variables
// s, r and v during a simulation by a candidate
// program.
static boolean[] srv = new boolean[3];


// number of cases = 1458 = product of
// 3 binary relations (or, and, xor)
// 3 places to put binary relation (s, r, v)
// 3 first inputs to binary relation (s, r, v)
// 3 second inputs to binary relation (s, r, v)
// 3 inputs to first other variable (s, r, v)
// 3 inputs to second other variable (s, r, v)
// 2 initial r values (false, true)
static final int ncases = 3 * 3 * 3 * 3 * 3 * 3 * 2;


static int test = 0; // counter for tests (candidate programs)


// The next five variables hold the parameters that
// determine a particular program for a test candidate.
// Some of these specify variables by ordinals, where
// variable r has ordinal 0, r has ordinal 1 and v has
// ordinal 2.

// Ordinal for binary operator used for binary relation,
// where 0=and, 1=or, 2=xor
static int binary_relation = 0;

// Ordinal of variable to receive result of the binary relation.
static int binary_place = 0;

// Ordinals for input variables to binary relation.
static int[] binary_inputs = {0, 0};

// The variable with ordinal other_inputs[0] is assigned
// to the variable with ordinal (binary_place+1)%3.
// The variable with ordinal other_inputs[1] is assigned
// to the variable with ordinal (binary_place+2)%3.
static int[] other_inputs = {0, 0}; //0=s, 1=r, 2=v

// Ordinal for initial values of unobserved r variable,
// where 0=false, 1=true.
static int initial_r = 0;


// Flag for whether a test candidate matches observed behavior,
// initially set to true but changed to false if any variable
// at any time step fails to match the observed behavior.
```

```
static boolean success = true;


public static void main(String args[]) {

  // iteration to enumerate text candidates
  for (test = 0; test<1458; test++) {

    // resolve test ordinal into program parameters
    int c = test;
    binary_relation = c % 3;
    c = c / 3;
    binary_place = c % 3;
    c = c / 3;
    for (int i=0; i<2; i++) {
      binary_inputs[i] = c % 3;
      c = c / 3;
    }
    for (int i=0; i<2; i++) {
      other_inputs[i] = c % 3;
      c = c / 3;
    }
    initial_r = c;


    // initialize the variables for the first time step
    srv[0] = s_match[0];
    srv[1] = (initial_r == 1);
    srv[2] = v_match[0];

    // initially set success = true
    success = true;

    // iterate over time sequence of observed behavior
    for (int t=1; t<len; t++) {

      // get the values for inputs to binary operator
      boolean a = srv[binary_inputs[0]];
      boolean b = srv[binary_inputs[1]];

      // get the values for inputs to simple assignments
      boolean d = srv[other_inputs[0]];
      boolean e = srv[other_inputs[1]];

      // compute the binary operation and assign to output
      // variable
      srv[binary_place] = (binary_relation == 0) ? a&b :
                          (binary_relation == 1) ? a|b : a^b;

      // put simple assignments in indicated variables
      srv[(binary_place+1)%3] = d;
      srv[(binary_place+2)%3] = e;

      // set success=false if either s or v fails to match
      // observed sequence
      success = success &&
```

```
          (srv[0] == s_match[t]) && (srv[2] == v_match[t]);
      } // end of iteration over time sequence of observed behavior

      // if this test candidate matched entire observed behavior
      // sequence, print its parameters
      if (success) {
        System.out.println("binary_relation = " + binary_relation +
          " binary_place = " + binary_place +
          " binary_inputs = " + binary_inputs[0] + " " +
          binary_inputs[1] + " other_inputs = " +
          other_inputs[0] + " " + other_inputs[1] +
          " initial_r = " + initial_r);
      }

    } // end of iteration to enumerate text candidates

  } // end of Main()

}
```

## Appendix B

```
//
// ASP.java
//

import java.util.*;

/**
   ASP is a program for testing the stability of adversarial
   sequence prediction under simple table driven learning
   algorithms. Copyright (C) 2007 Bill Hibbard.<P>
*/
public class ASP extends Object {

  static boolean debug = false;

  // here instability means one competitor getting and keeping all the
  // resources
  // possibility of instability is sensitive to RANDOM_INTERVAL and WIN_VALUE
  // RANDOM_INTERVAL too small enables loser to come back, stabilizing game
  // but RANDOM_INTERVAL too large creates a deterministic outcome
  // instability happens faster for smaller MAX_LENGTH
  // instability requires non-random in case of 'no result'
  // instability takes a long time for next_2 algorithm and large MAX_LENGTH
  //   unless a large WIN_VALUE is used too
  // instability is impossible for too large GROWTH_PER_PRINT

  // max history length in table - should be <= 8
  static final int MAX_LENGTH = 5;
  // value of win/loss in table size
  static final int WIN_VALUE = 2;
  // interval of games between printing table sizes
  static final int PRINT_INTERVAL = 100;
  // maximum count in table
  static final int MAX_COUNT = 1000000000;
  // expected interval of games between random plays
  static final int RANDOM_INTERVAL = 100;
  // initial proportion of table space used
  static final float INIT_TABLE = 0.2f;
  // initial advantage to Evader (1.0f for even)
  static final float E_ADVANTAGE = 1.0f;
  // growthrate of total table, per PRINT_INTERVAL
  static final float GROWTH_PER_PRINT = 0.01f;

  static final float random_test = 1.0f / RANDOM_INTERVAL;

  /* table layout
     length = 0 -> 0  to  0  length_to_size(0) =  1
     length = 1 -> 1  to  4  length_to_size(1) =  5
     length = 2 -> 5  to 20  length_to_size(2) = 21
     length = 3 -> 21 to 84  length_to_size(3) = 85
     length = n -> (4^n-1)/3 to ((4^(n+1)-1)/3)-1
     length = MAX_LENGTH -> (4^MAX_LENGTH-1)/3 to ((4^(MAX_LENGTH+1)-1)/3)-1
     NOTE lengths 0 and 1 are useless
```

```
  */
  static final int TABLE_SIZE = length_to_size(MAX_LENGTH) - 1;
  static final int[] size_to_length = new int[TABLE_SIZE+1];

  static final Random random = new Random();

  // bit masks for history lengths
  static final int[] record_masks = new int[MAX_LENGTH+1];
  static final int[] lookup_masks = new int[MAX_LENGTH+1];

  Predictor p;
  Evader e;

  public static void main(String args[]) {

if (debug) System.out.println("MAX_LENGTH = " + MAX_LENGTH +
                   "  TABLE_SIZE = " + TABLE_SIZE);
    for (int j=0; j<=MAX_LENGTH; j++) {
      for (int i=length_to_size(j-1); i<length_to_size(j); i++) {
        size_to_length[i] = j;
      }
    }

    for (int j=0; j<=MAX_LENGTH; j++) {
      record_masks[j] = (int) Math.pow(4, j) - 1;
      lookup_masks[j] = record_masks[j] & 0xfffffffc; // lose lowest 2 bits
    }

    ASP asp = new ASP();
    asp.run_game();
  }

  static int length_to_size(int n) {
    return (int) ((Math.pow(4, (n + 1)) - 1) / 3);
  }

  // 0 <= position < 4^length
  static int table_position(int length, int position) {
    return length_to_size(length - 1) + position;
  }

  int history = 0;
  int history_length = 0;

  ASP() {
    int it = ((TABLE_SIZE - 4) / 4);
    int r = (int) (INIT_TABLE * it);
    // int r = (TABLE_SIZE - 4) / 8;
    int er = (int) (E_ADVANTAGE * r);
    int pr = r + r - er;
System.out.println("r = " + r + " pr = " + pr + " er = " + er +
                   " it - pr = " + (it - pr) +
                   " it - er = " + (it - er));
    p = new Predictor(it - pr);
    e = new Evader(it - er);
  }
```

```
  void run_game() {
    int px = 0;
    int ex = 0;
    while (true) {
System.out.println(" p.table_size = " + p.table_size +
                   " e.table_size = " + e.table_size);
      delay(100); // let a ctrl C in once, hey
      for (int pi=0; pi<PRINT_INTERVAL; pi++) {

        // uncomment the next two statements for algorithm 1
        // px = p.next_1(history, history_length);
        // ex = e.next_1(history, history_length);

        // uncomment the next two statements for algorithm 2
        px = p.next_2(history, history_length);
        ex = e.next_2(history, history_length);

        history = (history << 2) | (px + px) | ex;
        if (history_length < 16) history_length++;
        if (px == ex) {
          // p wins
          int n = WIN_VALUE;
          if (n > (TABLE_SIZE - p.table_size) / 4) {
            n = (TABLE_SIZE - p.table_size) / 4;
          }
          if (n > e.table_size / 4) n = e.table_size / 4;
          p.add(n);
          e.remove(n);
        }
        else {
          // e wins
          int n = WIN_VALUE;
          if (n > (TABLE_SIZE - e.table_size) / 4) {
            n = (TABLE_SIZE - e.table_size) / 4;
          }
          if (n > p.table_size / 4) n = p.table_size / 4;
          e.add(n);
          p.remove(n);
        }
if (debug) System.out.println("px, ex = " + px + " " + ex +
                              " history = " + Integer.toHexString(history) +
                              " history_length = " + history_length +
                              " p.table_size = " + p.table_size +
                              " e.table_size = " + e.table_size);
        // if (p.table_size == 0 || e.table_size == 0) debug = true;
      } // end for (int pi=0; pi<PRINT_INTERVAL; pi++)
      // growth of total table
      int addp = (int) (GROWTH_PER_PRINT * p.table_size / 4);
      p.add(addp);
      int adde = (int) (GROWTH_PER_PRINT * e.table_size / 4);
      e.add(adde);
    } // end while (true)
  } // end run_game method

  void delay(int n) {
```

```
    try {
      Thread.sleep(n);
    }
    catch (java.lang.InterruptedException ex) {
    }
  }

  // super class for Predictor and Evader does the real work
  public class PE extends Object {
    int[] table = new int[TABLE_SIZE+1];
    int table_size = TABLE_SIZE;

    // variables that differ between Predictor and Evader
    int self_select;
    int other_select;
    int one;
    int zero;
    String id;

    PE() {
      for (int i=0; i<TABLE_SIZE; i++) table[i] = 0;
    }


    // first algorithm
    // compute next symbol, and record recent history in table
    final int next_1(int history, int history_length) {
      // first increment table positions for history
      int max_len = Math.min(MAX_LENGTH, history_length);
      for (int len=2; len<=max_len; len++) {
        int position = history & record_masks[len];
        int tp = table_position(len, position);
if (debug) System.out.println(id + "RECORD: len = " + len + " position = "
                       + position + " table[" + tp + "] = " + table[tp]);
        if (table[tp] >= 0) {
          // table position is available (negative indicates not available)
          // simply keep count of various histories
          table[tp]++;
          if (table[tp] > MAX_COUNT) {
            // if count exceeds max, halve all counts in group of 4
            // (4 combinations of values in most recent turn)
            int bp = table_position(len, position & lookup_masks[len]);
            table[bp]     = table[bp] / 2;
            table[bp + 1] = table[bp + 1] / 2;
            table[bp + 2] = table[bp + 2] / 2;
            table[bp + 3] = table[bp + 3] / 2;
          }
        }
      }

      // do occasional random next symbol
      if (random.nextFloat() < random_test) {
        int rv = (random.nextFloat() > 0.5f) ? one : zero;
if (debug) System.out.println(id + "random = " + rv);
        return rv;
      }
```

```
      // now compute next symbol
      // hist is history shifted one move back in time
      int hist = history << 2;
      int hist_length = history_length + 1;
      if (hist_length >= 16) hist_length--;
      max_len = Math.min(MAX_LENGTH, hist_length);
      // try longer histories first
      for (int len=max_len; len>=2; len--) {
        int position = hist & lookup_masks[len];
        int tp = table_position(len, position);
if (debug) System.out.println(id + "LOOKUP: len = " + len + " position = "
                        + position + " table[" + tp + "] = " + table[tp]);
        if (table[tp] >= 0) {
          // table position is available (negative indicates not available)
          // zeros = count of histories in which other played 0 after
          // current history
          int zeros = table[tp] + table[tp + self_select];
          // ones = count of histories in which other played 1 after
          // current history
          int ones = table[tp + other_select] +
                     table[tp + self_select + other_select];
if (debug) System.out.println(id + "zeros = " + table[tp] + " + " +
                              table[tp + self_select] + " ones = " +
                              table[tp + other_select] + " + " +
                              table[tp + self_select + other_select] +
                              " self_select = " + self_select +
                              " other_select = " + other_select);
          if (zeros == 0 && ones == 0) continue; // no counts, try shorter
                                                 // history
          // pick next move based on comparing history counts
          // ties go to 'one' - slightly favors Evader?
          return (zeros > ones) ? zero : one;
        } // end if (table[tp] >= 0)
      }
      // no result
if (debug) System.out.println(id + "no result: 0");
      // 'return 0' is a very slight advantage for the Predictor
      // 'return zero' would be a very slight advantage for the Evader
      return 0;
      // using a random return here enables loser to get back in the game
      // return (random.nextFloat() > 0.5f) ? one : zero;
    } // end next_1 method


    // second algorithm
    // compute next symbol, and record recent history in table
    final int next_2(int history, int history_length) {
      // first increment table positions for history
      int max_len = Math.min(MAX_LENGTH, history_length);
      for (int len=2; len<=max_len; len++) {
        int position = history & record_masks[len];
        int tp = table_position(len, position);
if (debug) System.out.println(id + "RECORD: len = " + len + " position = " +
                     position + " table[" + tp + "] = " + table[tp]);
        if (table[tp] >= 0) {
```

```
          // table position is available (negative indicates not available)
          // keep count of various histories
          table[tp]++;
if (debug) {
  int po = table_position(len, (position ^ other_select));
  int pos = table_position(len, (position ^ 3));
  System.out.println(id + "tp = " + tp + " po = " + po + " pos = " + pos);
}
          // clear counts for opposite value from other
          table[table_position(len, (position ^ other_select))] = 0;
          // clear counts for opposite value from other (and opposite
          //   value from self - note 3 = self_select + other_select)
          table[table_position(len, (position ^ 3))] = 0;
        }
      }

      // do occasional random next symbol
      if (random.nextFloat() < random_test) {
        int rv = (random.nextFloat() > 0.5f) ? one : zero;
if (debug) System.out.println(id + "random = " + rv);
        return rv;
      }

      // now compute next symbol
      // hist is history shifted one move back in time
      int hist = history << 2;
      int hist_length = history_length + 1;
      if (hist_length >= 16) hist_length--;
      int margin = 0;
      int choice = 0;
      max_len = Math.min(MAX_LENGTH, hist_length);
      // try longer histories first
      for (int len=max_len; len>=2; len--) {
        int position = hist & lookup_masks[len];
        int tp = table_position(len, position);
if (debug) System.out.println(id + "LOOKUP: len = " + len + " position = " +
                    position + " table[" + tp + "] = " + table[tp]);
        if (table[tp] >= 0) {
          // table position is available (negative indicates not available)
          // zeros = count of histories in which other played 0 after
          // current history
          int zeros = table[tp] + table[tp + self_select];
          // ones = count of histories in which other played 1 after
          // current history
          int ones = table[tp + other_select] +
                     table[tp + self_select + other_select];
if (debug) System.out.println(id + "zeros = " + table[tp] + " + " +
                             table[tp + self_select] + " ones = " +
                             table[tp + other_select] + " + " +
                             table[tp + self_select + other_select] +
                             " self_select = " + self_select +
                             " other_select = " + other_select);
          if (zeros == 0 && ones == 0) continue; // no counts, try shorter
                                                 // history
          if (zeros > 0 && ones > 0) {
            // cannot happen because of clearing counts for opposite value
```

```
            // from other
            System.out.println("IMPOSSIBLE");
            System.exit(0);
          }
          // pick next move based on comparing history counts
          if (zeros > ones) {
            // choice weight is count plus length
            int m = (zeros - ones) + len;
            if (m > margin) {
              margin = m;
              choice = zero;
if (debug) System.out.println(id + "margin = " + margin + " choice = " +
choice);
            }
          }
          else { // ones > zeros  (cannot be == unless both 0)
            // choice weight is count plus length
            int m = (ones - zeros) + len;
            if (m > margin) {
              margin = m;
              choice = one;
if (debug) System.out.println(id + "margin = " + margin +
                              " choice = " + choice);
            }
          }
        } // end if (table[tp] >= 0)
      } // end for (int len=max_len; len>=2; len--)
      if (margin > 0) return choice;

      // no result
if (debug) System.out.println(id + "no result: 0");
      // 'return 0' is a very slight advantage for the Predictor
      // 'return zero' would be a very slight advantage for the Evader
      return 0;
      // using a random return here enables loser to get back in the game
      // return (random.nextFloat() > 0.5f) ? one : zero;
    } // end next_2 method


    // remove n spaces (4*n ints) from table
    final void remove(int n) {
      if (n > table_size / 4) n = table_size / 4;
      int left = n;
      while (left > 0) {
        int l = size_to_length[table_size];
        int lo = length_to_size(l - 1) - 1;
        int hi = length_to_size(l) - 1;
        int len = (hi - lo) / 4;
        int used = (table_size - lo) / 4;
        int m = Math.min(used, left);
        int k = (int) (len * random.nextFloat());
        while (m > 0) {
          if (k >= len) k = k - len;
          int j = lo + 4 * k;
          if (table[j+1] >= 0) {
if (debug) System.out.println(id + "REMOVE: j = " + j + " table_size = " +
```

```
                              table_size + " l = " + l + " lo = " + lo +
                              " k = " + k);
            table[j+1] = -1;
            table[j+2] = -1;
            table[j+3] = -1;
            table[j+4] = -1;
            m--;
            left--;
            table_size = table_size - 4;
          }
          k++;
        } // end while ((m > 0)
      } // end while (left > 0)
    } // end remove method


    // add n spaces (4*n ints) to table
    final void add(int n) {
      if (n > (TABLE_SIZE - table_size) / 4)
                             n = (TABLE_SIZE - table_size) / 4;
      int left = n;
      while (left > 0) {
        int l = size_to_length[table_size + 4];
        int lo = length_to_size(l - 1) - 1;
        int hi = length_to_size(l) - 1;
        int len = (hi - lo) / 4;
        int unused = (hi - table_size) / 4;
        int m = Math.min(unused, left);
        int k = (int) (len * random.nextFloat());
        while (m > 0) {
          if (k >= len) k = k - len;
          int j = lo + 4 * k;
          if (table[j+1] < 0) {
if (debug) System.out.println(id + "ADD: j = " + j + " table_size = " +
                              table_size + " l = " + l + " lo = " + lo +
                              " k = " + k);
            table[j+1] = 0;
            table[j+2] = 0;
            table[j+3] = 0;
            table[j+4] = 0;
            m--;
            left--;
            table_size = table_size + 4;
          }
          k++;
        } // end while ((m > 0)
      } // end while (left > 0)
    } // end add method


  } // end class PE

  public class Predictor extends PE {

    Predictor(int r) {
      id = "P ";
```

```
      remove(r);
      self_select = 2;
      other_select = 1;
      one = 1;
      zero = 0;
    }
  }

  public class Evader extends PE {

    Evader(int r) {
      id = "E ";
      remove(r);
      self_select = 1;
      other_select = 2;
      one = 0;
      zero = 1;
    }
  }

}
```

## Index